\PassOptionsToPackage{table,xcdraw}{xcolor}
\PassOptionsToPackage{usenames, dvipsnames}{color}
\documentclass[compsoc]{IEEEtran}
\makeatletter
\newcommand\footnoteref[1]{\protected@xdef\@thefnmark{\ref{#1}}\@footnotemark}
\makeatother

\usepackage[utf8]{inputenc} % allow utf-8 input
\usepackage[T1]{fontenc}    % use 8-bit T1 fonts
\usepackage{hyperref}       % hyperlinks
\usepackage{url}            % simple URL typesetting
\usepackage{booktabs}       % professional-quality tables
\usepackage{amsfonts}       % blackboard math symbols
\usepackage{nicefrac}       % compact symbols for 1/2, etc.
\usepackage{microtype}      % microtypography

\usepackage{subfigure}
\usepackage{graphicx}
\usepackage{caption} % Add this to the preamble

\usepackage{cite} % Add this to the preamble

\usepackage{subfigure}
\usepackage{color}

\usepackage{comment}
\usepackage{algorithm,algpseudocode}
\usepackage{rotating}
\usepackage{multirow}
\usepackage{tabularray}
\usepackage[table,xcdraw]{xcolor}

\usepackage{pifont} % for \ding symbols like \ding{55}
\usepackage{amssymb} % for \checkmark symbol

\usepackage{enumitem}
\usepackage{url}
\usepackage{xspace}
\usepackage{subfigure}
\usepackage{amsmath}
\usepackage{tabularx}  
\usepackage{graphicx}
\usepackage{array}
\usepackage{balance}
\usepackage{newtxmath}

\usepackage{tikz}
\usepackage{forest}

\usepackage{multicol}

\definecolor{lavender}{RGB}{190, 184, 220} % Light lavender color
\definecolor{beige}{RGB}{231, 218, 210} % Light beige color
\definecolor{teal}{RGB}{153, 153, 153} % Light teal color
\definecolor{bittersweet}{rgb}{1.0, 0.44, 0.37}% You can adjust the RGB values as needed

\definecolor{c0}{HTML}{7d9bbc} % Soft mint green
\definecolor{c1}{HTML}{b3d2f4} % Soft sky blue98b0a8
\definecolor{c2}{HTML}{92d9da} % Warm coral orange
\definecolor{c3}{HTML}{c1f6e5}  % Dusty rose

\definecolor{colorn1}{HTML}{00c3a9}
\definecolor{colorn2}{HTML}{00aeb6}
\definecolor{colorn3}{HTML}{0097bb}
\definecolor{colorn4}{HTML}{007fb6}
\definecolor{colorn5}{HTML}{0067a8}

\definecolor{root}{HTML}{B2B2B2}
\usepackage{pgfmath}
\usetikzlibrary{decorations.text, arrows.meta,calc,shadows.blur,shadings}
\def\arcarrow[#1](#2)(#3)[#4]{
    \draw[thick,-,>=latex,color=#1,line width=1pt,shorten >=-2pt, shorten <=-2pt] 
        let \p1 = (#2), \p2 = (#3), % To access cartesian coordinates x, and y.
            \n1 = {veclen(\x1,\y1)}, % Distance from the origin
            \n2 = {veclen(\x2,\y2)}, % Distance from the origin
            \n3 = {atan2(\y1,\x1)} % Angle where arctext starts.
        in (\n3-#4: \n1) -- (\n3-#4: \n2); % Draw the arrow.
}

\pgfmathsetmacro{\nodeHeight}{15}    % Height of main arcs
\pgfmathsetmacro{\nodeRadius}{1.5}  % Radius for main nodes
\pgfmathsetmacro{\subnodeHeight}{13} % Height of sub-arcs
\pgfmathsetmacro{\subnodeRadius}{2.5} % Radius for subnodes
\pgfmathsetmacro{\subsubnodeHeight}{8} % Height of sub-arcs
\pgfmathsetmacro{\subsubnodeRadius}{3.5} % Radius for subnodes
\pgfmathsetmacro{\subsubsubnodeHeight}{10} % Height of sub-arcs
\pgfmathsetmacro{\subsubnodeRadius}{3.3} % Radius for subnodes

\pgfmathsetmacro{\offet}{-15} % Radius for subnodes

\newcommand{\levelOneFontSize}{\fontsize{5pt}{7pt}\selectfont}
\newcommand{\levelTwoFontSize}{\fontsize{4pt}{4pt}\selectfont}
\newcommand{\levelThreeFontSize}{\fontsize{3.5pt}{3pt}\selectfont}

\def\arctextnode[#1][#2][#3](#4)(#5)(#6)[#7]#8{
    \draw[
        color=white,
        thick,
        line width=1.3pt,
        fill=#2
    ]
    (#5:#4cm+#3) coordinate (above #1) arc (#5:#6:#4cm+#3)
    -- (#6:#4) coordinate (right #1) -- (#6:#4cm-#3) coordinate (below right #1)
    arc (#6:#5:#4cm-#3) coordinate (below #1)
    -- (#5:#4) coordinate (left #1) -- cycle;
    
    \pgfmathsetmacro{\midangle}{(#5+#6)/2}
    \pgfmathparse{\midangle < 0 ? \midangle + 360 : \midangle}
    \let\positiveangle\pgfmathresult

    \pgfmathparse{\positiveangle > 180 ? \positiveangle + 90 : \positiveangle - 90}
    \let\calculatedangle\pgfmathresult

    {%
    \node at (\midangle:#4) [
        rotate=\calculatedangle,
        transform shape, 
        align=center,          % Allows for line breaks
        inner sep=0pt,
        font=#7\bf\color{white}, % Apply the font size variable
        text width=2cm         % Adjust as needed
    ] {#8};
    }
}

\usepackage{listings}

\ifCLASSINFOpdf
\else
\fi
\newcommand{\eg}{\textit{e.g.},\xspace}

\newcommand{\etal}{\textit{et al.},\xspace}
\newcommand{\etc}{\textit{etc}\xspace}

\usepackage[usenames,dvipsnames]{color}
\usepackage{etoolbox}

\begin{document}

\title{Large Language Model-Brained GUI Agents: \\A Survey}

\author{Chaoyun~Zhang,
        Shilin~He,
        Jiaxu~Qian,
        Bowen~Li,
        Liqun~Li,
        Si~Qin,
        Yu~Kang,
        Minghua~Ma,
        Guyue~Liu,
        Qingwei~Lin,
        Saravan~Rajmohan,
        Dongmei~Zhang,
        Qi~Zhang
\thanks{Version: v8 (major update on May 2, 2025)}
\thanks{Chaoyun Zhang, Shilin He, Jiaxu Qian, Liqun Li, Si Qin, Yu Kang, Minghua Ma, Qingwei Lin,  Saravan Rajmohan, Dongmei Zhang and Qi Zhang are with Microsoft. e-mail: \{chaoyun.zhang, shilin.he, v-jiaxuqian, liqun.li, si.qin, yu.kang, minghuama, qlin, saravan.rajmohan, dongmeiz, zhang.qi\}@microsoft.com.}% <-this % stops a space
% \thanks{Minghua Ma and Saravan Rajmohan are with M365 Research, Microsoft, USA. e-mail: \{minghuama, saravan.rajmohan\}@microsoft.com.}% <-this % stops a space
\thanks{Bowen Li is with Shanghai Artificial Intelligence Laboratory, China. e-mail: libowen.ne@gmail.com.}% <-this % stops a space
\thanks{Guyue Liu is with Peking University, China. e-mail: guyue.liu@gmail.com.}
\thanks{For any inquiries or discussions, please contact Chaoyun Zhang and Shilin He.}
}

\newcounter{figurecounter}
\newcounter{tablecounter}

% 定义 \fg 命令，自动编号并使用紫色显示

\newcommand{\fg}[1]{%
    \refstepcounter{figurecounter}%
    \textcolor{purple}{[Figure \thefigurecounter: #1]}%
}

\newcommand{\tb}[1]{%
    \refstepcounter{tablecounter}%
    \textcolor{orange}{[Table \thetablecounter: #1]}%
}

\markboth{Journal of \LaTeX\ Class Files, December~2024}%
{Zhang \MakeLowercase{\textit{et al.}}: Large Language Model-Brained GUI Agents: A Survey}





\IEEEtitleabstractindextext{%
\begin{abstract}
    Graphical User Interfaces (GUIs) have long been central to human-computer interaction, providing an intuitive and visually-driven way to access and interact with digital systems. Traditionally, automating GUI interactions relied on script-based or rule-based approaches, which, while effective for fixed workflows, lacked the flexibility and adaptability required for dynamic, real-world applications. The advent of Large Language Models (LLMs), particularly multimodal models, has ushered in a new era of GUI automation. They have demonstrated exceptional capabilities in natural language understanding, code generation, task generalization, and visual processing. This has paved the way for a new generation of ``LLM-brained'' GUI agents capable of interpreting complex GUI elements and autonomously executing actions based on natural language instructions. These agents represent a paradigm shift, enabling users to perform intricate, multi-step tasks through simple conversational commands. Their applications span across web navigation, mobile app interactions, and desktop automation, offering a transformative user experience that revolutionizes how individuals interact with software. This emerging field is rapidly advancing, with significant progress in both research and industry.
    
    To provide a structured understanding of this trend, this paper presents a comprehensive survey of LLM-brained GUI agents, exploring their historical evolution, core components, and advanced techniques. We address critical research questions such as existing GUI agent frameworks, the collection and utilization of data for training specialized GUI agents, the development of large action models tailored for GUI tasks, and the evaluation metrics and benchmarks necessary to assess their effectiveness. Additionally, we examine emerging applications powered by these agents. Through a detailed analysis, this survey identifies key research gaps and outlines a roadmap for future advancements in the field. By consolidating foundational knowledge and state-of-the-art developments, this work aims to guide both researchers and practitioners in overcoming challenges and unlocking the full potential of LLM-brained GUI agents. We anticipate that this survey will serve both as a practical cookbook for constructing LLM-powered GUI agents, and as a definitive reference for advancing research in this rapidly evolving domain.

    The collection of papers reviewed in this survey will be hosted and regularly updated on the GitHub repository: {\color{bittersweet}\url{https://github.com/vyokky/LLM-Brained-GUI-Agents-Survey}}. Additionally, a searchable webpage is available at 
    
    {\color{bittersweet}\url{https://aka.ms/gui-agent}} for easier access and exploration.
\end{abstract}
\begin{IEEEkeywords}
Large Language Model, Graphical User Interface, AI Agent, Automation, Human-Computer Interaction
\end{IEEEkeywords}

}
\maketitle

% \fi

\IEEEpeerreviewmaketitle

\section{Introduction\label{sec:introduction}}
\begin{figure*}[!h]
    \centering
    \includegraphics[width=\textwidth]{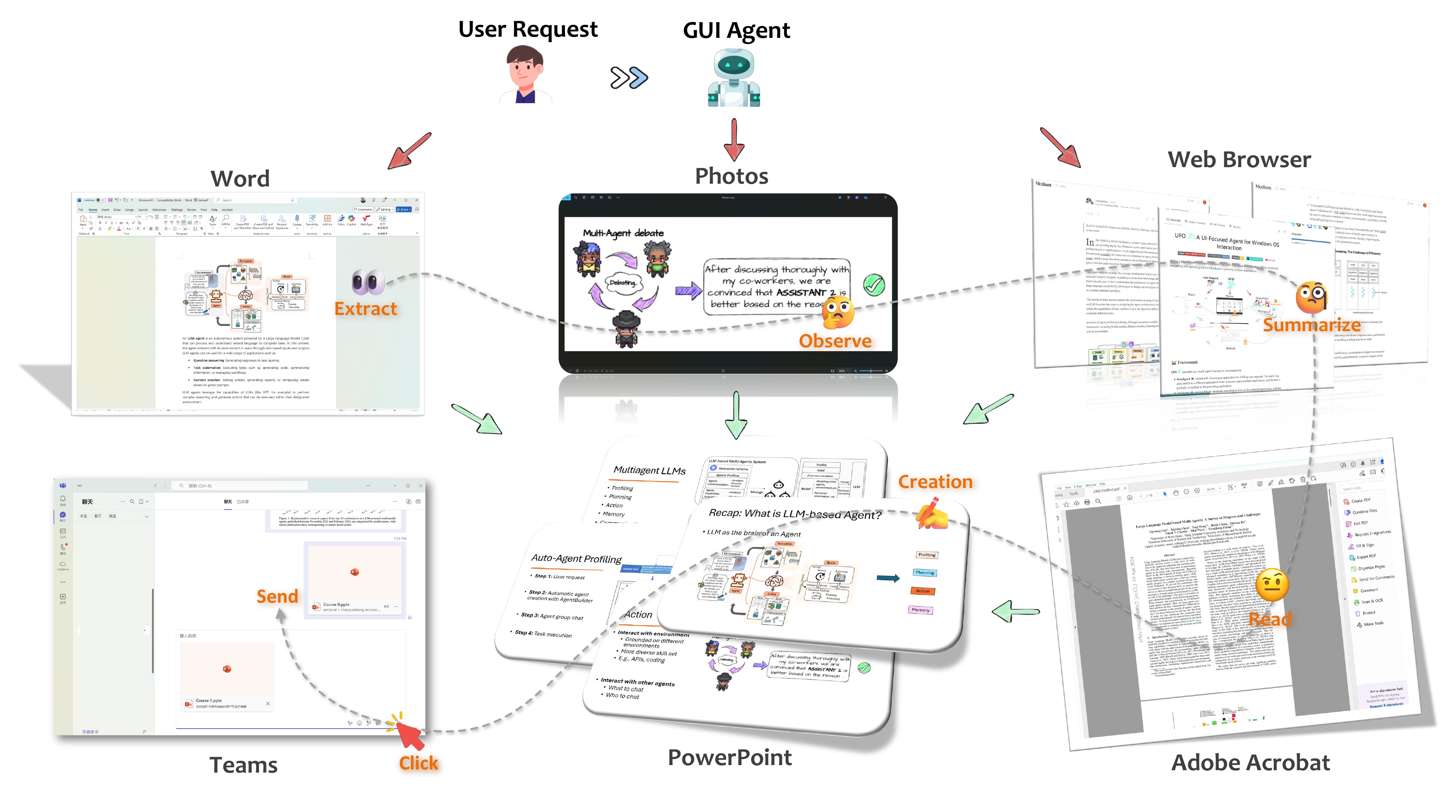}
    \vspace{-2em}
    \caption{Illustration of the high-level concept of an LLM-powered GUI agent. The agent receives a user's natural language request and orchestrates actions seamlessly across multiple applications. It extracts information from Word documents, observes content in Photos, summarizes web pages in the browser, reads PDFs in Adobe Acrobat, and creates slides in PowerPoint before sending them through Teams.}
    \label{fig:gui_agent}
    % \vspace{-2em}
\end{figure*}

Graphical User Interfaces (GUIs) have been a cornerstone of human-computer interaction, fundamentally transforming how users navigate and operate within digital systems \cite{Jansen1998TheGU}. Designed to make computing more intuitive and accessible, GUIs replaced command-line interfaces (CLIs) \cite{Sampath2021AccessibilityOC} with visually driven, user-friendly environments. Through the use of icons, buttons, windows, and menus, GUIs empowered a broader range of users to interact with computers using simple actions such as clicks, typing, and gestures. This shift democratized access to computing, allowing even non-technical users to effectively engage with complex systems. However, GUIs often sacrifice efficiency for usability, particularly in workflows requiring repetitive or multi-step interactions, where CLIs can remain more streamlined \cite{Michalski2006TheEO}.

While GUIs revolutionized usability, their design, primarily tailored for human visual interaction, poses significant challenges for automation. The diversity, dynamism, and platform-specific nature of GUI layouts make it difficult to develop flexible and intelligent automation tools capable of adapting to various environments. Early efforts to automate GUI interactions predominantly relied on script-based or rule-based methods \cite{hellmann2011rule, jRapture}. Although effective for predefined workflows, these methods were inherently narrow in scope, focusing primarily on tasks such as software testing and robotic process automation (RPA) \cite{ivanvcic2019robotic}. Their rigidity required frequent manual updates to accommodate new tasks, changes in GUI layouts, or evolving workflows, limiting their scalability and versatility. Moreover, these approaches lacked the sophistication needed to support dynamic, human-like interactions, thereby constraining their applicability in complex or unpredictable scenarios.

The rise of Large Language Models (LLMs)\footnote{By LLMs, we refer to the general concept of foundation models capable of accepting various input modalities (\eg visual language models (VLMs), multimodal LLMs (MLLMs)) while producing output exclusively in textual sequences \cite{wikipedia_llm}.} \cite{zhao2023survey, naveed2023comprehensive}, especially those augmented with multimodal capabilities \cite{yin2023survey}, has emerged as a game changer for GUI automation,  redefining the the way agents interact with graphical user interfaces. Beginning with models like ChatGPT~\cite{wu2023brief}, LLMs have demonstrated extraordinary proficiency in natural language understanding, code generation, and generalization across diverse tasks \cite{liu2024large, shen2024llm, fengfar, zhao2023survey}. The integration of visual language models (VLMs) has further extended these capabilities, enabling these models to process visual data, such as the intricate layouts of GUIs \cite{hong2023cogagentvisuallanguagemodel}. This evolution bridges the gap between linguistic and visual comprehension, empowering intelligent agents to interact with GUIs in a more human-like and adaptive manner. By leveraging these advancements, LLMs and VLMs offer transformative potential, enabling agents to navigate complex digital environments, execute tasks dynamically, and revolutionize the field of GUI automation.

\subsection{Motivation for LLM-Brained GUI agents\label{sec:introduction:motivation}}
With an LLM serving as its \textbf{``brain''}, LLM-powered GUI automation introduces a new class of intelligent agents capable of interpreting a user's natural language requests, analyzing GUI screens and their elements, and autonomously executing appropriate actions. Importantly, these capabilities are achieved without reliance on complex, platform-specific scripts or predefined workflows. These agents, referred to as \textbf{``LLM-brained GUI agents''}, can be formally defined as:
\begin{quote} 
    \textit{Intelligent agents that operate within GUI environments, leveraging LLMs as their core inference and cognitive engine to generate, plan, and execute actions in a flexible and adaptive manner.} 
\end{quote}
This paradigm represents a transformative leap in GUI automation, fostering dynamic, human-like interactions across diverse platforms. It enables the creation of intelligent, adaptive systems that can reason, make decisions in real-time, and respond flexibly to evolving tasks and environments. We illustrate this high-level concept in Figure~\ref{fig:gui_agent}. 

Traditional GUI automation are often limited by predefined rules or narrowly focused on specific tasks, constraining their ability to adapt to dynamic environments and diverse applications. In contrast, LLM-powered GUI agents bring a paradigm shift by integrating natural language understanding, visual recognition, and decision-making into a unified framework. This enables them to generalize across a wide range of use cases, transforming task automation and significantly enhancing the intuitiveness and efficiency of human-computer interaction. Moreover, unlike the emerging trend of pure Application Programming Interface (API)-based agents—which depend on APIs that may not always be exposed or accessible—GUI agents leverage the universal nature of graphical interfaces. GUIs offer a general mechanism to control most software applications, enabling agents to operate in a non-intrusive manner without requiring internal API access. This capability not only broadens the applicability of GUI agents but also empowers external developers to build advanced functionality on top of existing software across diverse platforms and ecosystems. Together, these innovations position GUI agents as a versatile and transformative technology for the future of intelligent automation.

This new paradigm enables users to control general software systems with conversational commands \cite{xu2025every}. By reducing the cognitive load of multi-step GUI operations, LLM-powered agents make complex systems accessible to non-technical users and streamline workflows across diverse domains. Notable examples include SeeAct~\cite{zheng2024gpt4visiongeneralistwebagent} for web navigation, AppAgent~\cite{zhang2023appagentmultimodalagentssmartphone} for mobile interactions, and UFO~\cite{zhang2024ufouifocusedagentwindows} for Windows OS applications. These agents resemble a ``virtual assistant'' \cite{Guan2023IntelligentVA} akin to J.A.R.V.I.S. from Iron Man—an intuitive, adaptive system capable of understanding user goals and autonomously performing actions across applications. The futuristic concept of an AI-powered operating system that executes cross-application tasks with fluidity and precision is rapidly becoming a reality \cite{zhang2024operating, mei2024aios}.

Real-world applications of LLM-powered GUI agents are already emerging. For example, Microsoft Power Automate utilizes LLMs to streamline low-code/no-code automation\footnote{\url{https://www.microsoft.com/en-us/power-platform/blog/power-automate/revolutionize-the-way-you-work-with-automation-and-ai/}}, allowing users to design workflows across Microsoft applications with minimal technical expertise. Integrated AI assistants in productivity software, like Microsoft Copilot\footnote{\url{https://copilot.microsoft.com/}}, are bridging the gap between natural language instructions and operations on application. Additionally, LLM-powered agents show promise for enhancing accessibility \cite{Aljedaani2024DoesCG}, potentially allowing visually impaired users to navigate GUIs more effectively by converting natural language commands into executable steps. These developments underscore the timeliness and transformative potential of LLM-powered GUI agents across diverse applications.

The convergence of LLMs and GUI automation addresses longstanding challenges in human-computer interaction and introduces new opportunities for intelligent GUI control \cite{Chin2024HumanCenteredLU}. This integration has catalyzed a surge in research activity, spanning application frameworks \cite{zhang2024ufouifocusedagentwindows}, data collection \cite{cheng2024seeclickharnessingguigrounding}, model optimization \cite{hong2023cogagentvisuallanguagemodel}, and evaluation benchmarks \cite{Zhuge2024AgentasaJudgeEA}. Despite these advancements, key challenges and limitations persist, and many foundational questions remain unexplored. However, a systematic review of this rapidly evolving area is notably absent, leaving a critical gap in understanding.

\subsection{Scope of the Survey\label{sec:introduction:scope}}
To address this gap, this paper provides a pioneering, comprehensive survey of LLM-brained GUI agents. We cover the historical evolution of GUI agents, provide a step-by-step guide to building these agents, summarize essential and advanced techniques, review notable tools and research related to frameworks, data and models, showcase representative applications, and outline future directions. Specifically, this survey aims to answer the following research questions (RQs):
\begin{enumerate} 
    \item \textbf{RQ1:} What is the historical development trajectory of LLM-powered GUI agents? (Section~\ref{sec:evolution})
    \item \textbf{RQ2:} What are the essential components and advanced technologies that form the foundation of LLM-brained GUI agents? (Section~\ref{sec:agent_foundation})
    \item \textbf{RQ3:} What are the principal frameworks for LLM GUI agents, and what are their defining characteristics? (Section~\ref{sec:framework}) 
    \item \textbf{RQ4:} What are the existing datasets, and how can comprehensive datasets be collected to train optimized LLMs for GUI agents? (Section~\ref{sec:data})
    \item \textbf{RQ5:} How can the collected data be used to train purpose-built Large Action Models (LAMs) for GUI agents, and what are the current leading models in the field? (Section~\ref{sec:model})
    \item \textbf{RQ6:} What metrics and benchmarks are used to evaluate the capability and performance of GUI agents? (Section~\ref{sec:evaluation})
    \item \textbf{RQ7:} What are the most significant real-world applications of LLM-powered GUI agents, and how have they been adapted for practical use? (Section~\ref{sec:applications})
    \item \textbf{RQ8:} What are the major challenges, limitations, and future research directions for developing robust and intelligent GUI agents? 
    (Section~\ref{sec:limitation})
\end{enumerate}
Through these questions, this survey aims to provide a comprehensive overview of the current state of the field, offer a guide for building LLM-brained GUI agents, identify key research gaps, and propose directions for future work. This survey is one of the pioneers to systematically examine the domain of LLM-brained GUI agents, integrating perspectives from LLM advancements, GUI automation, and human-computer interaction.

\subsection{Survey Structure\label{sec:introduction:structure}}

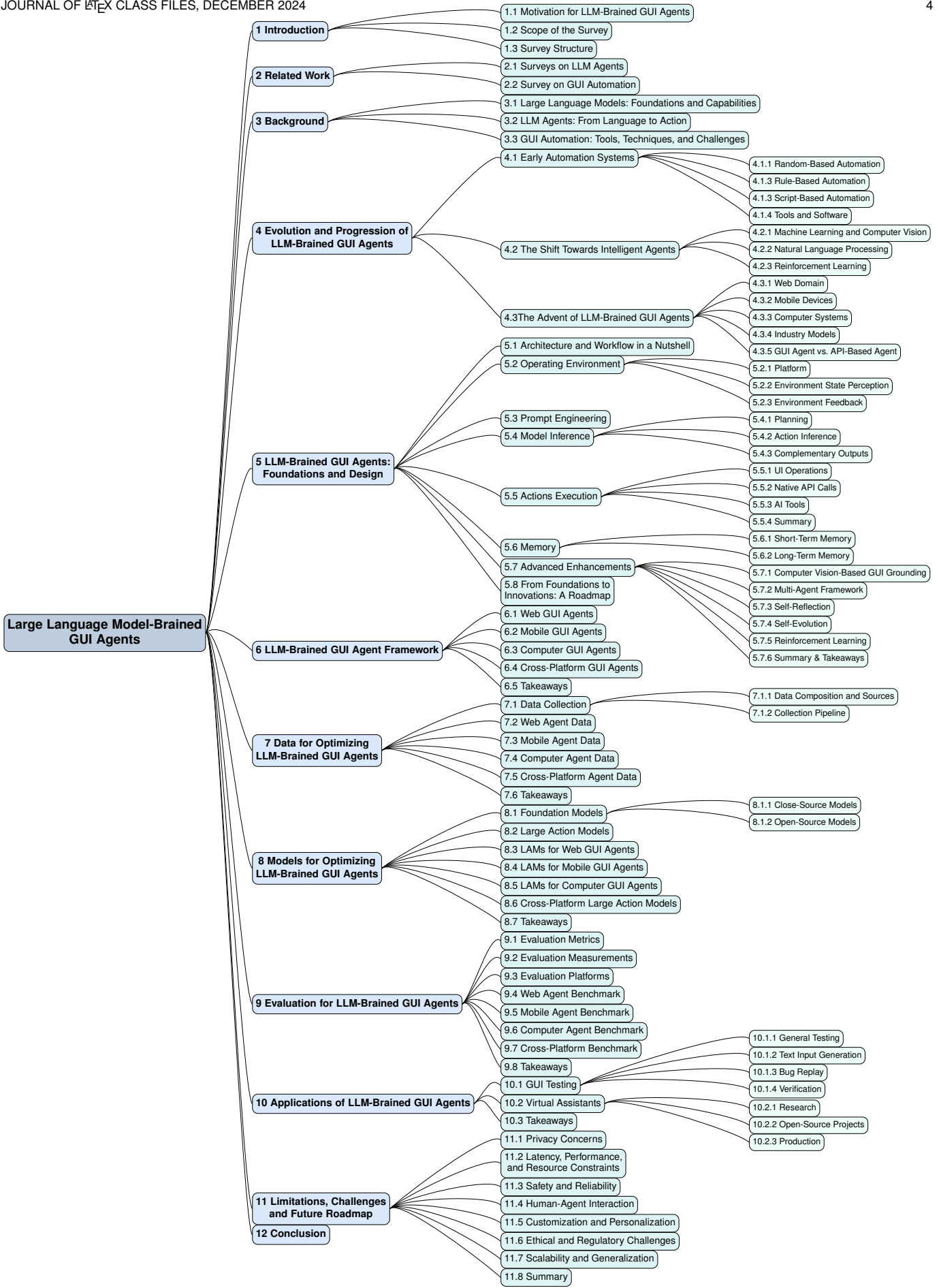
\begin{figure*}[!h]
    \centering
    \vspace*{-1.5em}
    \resizebox{\textwidth}{!}{ % Resize the entire figure to fit \textwidth
    \begin{forest}
    for tree={
        grow=east,
        parent anchor=east,
        child anchor=west,
        rounded corners,
        draw,
        edge path={
            \noexpand\path [draw, \forestoption{edge}]
            (!u.parent anchor) .. controls +(10pt,10pt) and +(-10pt,10pt) .. (.child anchor)\forestoption{edge label}; % Creates an arc
        },
        anchor=west,
        l=8.1cm, % Adjusts level distance (reduce to make tree flatter)
        s sep=0pt, % Adjusts sibling separation (reduce to make tree flatter)
        % Styles for each level
        tier/.wrap pgfmath arg={tier#1}{level()},
        tier0/.style={fill=c0!50, text=black, font=\large\bfseries, rounded corners=5pt, align=center}, % Soft teal for root
        tier1/.style={fill=c1!50, text=black, font=\normalsize\bfseries, rounded corners=5pt, align=center}, % Light lavender for first level
        tier2/.style={fill=c2!30, text=black, font=\small, rounded corners=5pt, align=center}, % Light beige for second level
        tier3/.style={fill=c3!30, text=black, font=\footnotesize, rounded corners=5pt, align=center} % Light gray for third level
    }
    [
    {Large Language Model-Brained\\ GUI Agents}, tier0
        [\ref{sec:conclusion} Conclusion, tier1]
        [\ref{sec:limitation} {Limitations, Challenges} \\ and Future Roadmap, tier1
            [\ref{sec:limitation:summary} Summary, tier2]
            [\ref{sec:limitation:scalability} Scalability and Generalization, tier2]
            [\ref{sec:limitation:ethical} Ethical and Regulatory Challenges, tier2]
            [\ref{sec:limitation:customization} Customization and Personalization, tier2]
            [\ref{sec:limitation:hai} Human-Agent Interaction, tier2]
            [\ref{sec:limitation:safety} Safety and Reliability, tier2]
            [\ref{sec:limitation:latency} {Latency, Performance,} \\ and Resource Constraints, tier2]
            [\ref{sec:limitation:privacy} Privacy Concerns, tier2]
        ]
        [\ref{sec:applications} Applications of LLM-Brained GUI Agents, tier1
            [\ref{sec:applications:takeaways} Takeaways, tier2]
            [\ref{sec:applications:assistants} Virtual Assistants, tier2
                [\ref{sec:applications:assistants:production} Production, tier3]
                [\ref{sec:applications:assistants:projects} Open-Source Projects, tier3]
                [\ref{sec:applications:assistants:research} Research, tier3, yshift=-40pt]
            ]
            [\ref{sec:applications:testing} GUI Testing, tier2
                [\ref{sec:applications:testing:verification} Verification, tier3, yshift=40pt]
                [\ref{sec:applications:testing:bug} Bug Replay, tier3]
                [\ref{sec:applications:testing:text} Text Input Generation, tier3]
                [\ref{sec:applications:testing:general} General Testing, tier3]                                        
            ]
        ]
        [\ref{sec:evaluation} Evaluation for LLM-Brained GUI Agents, tier1
            [\ref{sec:evaluation:takeaways} Takeaways, tier2]
            [\ref{sec:evaluation:cross} Cross-Platform Benchmark, tier2]
            [\ref{sec:evaluation:computer} Computer Agent Benchmark, tier2]
            [\ref{sec:evaluation:mobile} Mobile Agent Benchmark, tier2]
            [\ref{sec:evaluation:web} Web Agent Benchmark, tier2]
            [\ref{sec:evaluation:platform} Evaluation Platforms, tier2]
            [\ref{sec:evaluation:measurement} Evaluation Measurements, tier2]
            [\ref{sec:evaluation:metric} Evaluation Metrics, tier2]
        ]
        [\ref{sec:model} Models for Optimizing \\LLM-Brained GUI Agents, tier1
            [\ref{sec:model:takeaways} Takeaways, tier2]
            [\ref{sec:model:cross} Cross-Platform Large Action Models, tier2]
            [\ref{sec:model:computer} LAMs for Computer GUI Agents, tier2]
            [\ref{sec:model:mobile} LAMs for Mobile GUI Agents, tier2]
            [\ref{sec:model:web} LAMs for Web GUI Agents, tier2]
            [\ref{sec:model:lam} Large Action Models, tier2]
            [\ref{sec:model:foundation} Foundation Models, tier2
                [\ref{sec:model:foundation:open} Open-Source Models, tier3]
                [\ref{sec:model:foundation:close} Close-Source Models, tier3]
            ]
        ]
        [\ref{sec:data} Data for Optimizing \\LLM-Brained GUI Agents, tier1
            [\ref{sec:data:takeaways} Takeaways, tier2]
            [\ref{sec:data:cross} Cross-Platform Agent Data, tier2]
            [\ref{sec:data:computer} Computer Agent Data, tier2]
            [\ref{sec:data:mobile} Mobile Agent Data, tier2]
            [\ref{sec:data:web} Web Agent Data, tier2]
            [\ref{sec:data:collection} Data Collection, tier2
                [\ref{sec:data:collection:pipeline} Collection Pipeline, tier3]
                [\ref{sec:data:collection:source} Data Composition and Sources, tier3]
            ]
        ]
        [\ref{sec:framework} LLM-Brained GUI Agent Framework, tier1
            [\ref{sec:framework:takeaways} Takeaways, tier2]
            [\ref{sec:framework:cross} Cross-Platform GUI Agents, tier2]
            [\ref{sec:framework:computer} Computer GUI Agents, tier2]
            [\ref{sec:framework:mobile} Mobile GUI Agents, tier2]
            [\ref{sec:framework:web} Web GUI Agents, tier2]
        ]
        [\ref{sec:agent_foundation} LLM-Brained GUI Agents:\\ Foundations and Design, tier1
            [\ref{sec:agent_foundation:roadmap} From Foundations to \\Innovations: A Roadmap, tier2]
            [\ref{sec:agent_foundation:enhancement} Advanced Enhancements, tier2
                [\ref{sec:agent_foundation:enhancement:summary} Summary \& Takeaways, tier3]
                [\ref{sec:agent_foundation:enhancement:rl} Reinforcement Learning, tier3]
                [\ref{sec:agent_foundation:enhancement:self_evolution} Self-Evolution, tier3]
                [\ref{sec:agent_foundation:enhancement:self_reflection} Self-Reflection, tier3]
                [\ref{sec:agent_foundation:enhancement:multiagent} Multi-Agent Framework, tier3]
                [\ref{sec:agent_foundation:enhancement:cv} Computer Vision-Based GUI Grounding, tier3, yshift=-90pt]
            ]
            [\ref{sec:agent_foundation:memory} Memory, tier2
                [\ref{sec:agent_foundation:memory:long} Long-Term Memory, tier3]
                [\ref{sec:agent_foundation:memory:short} Short-Term Memory, tier3]
            ]
            [\ref{sec:agent_foundation:action_exe} Actions Execution, tier2
                [\ref{sec:agent_foundation:action_exe:summary} Summary, tier3]
                [\ref{sec:agent_foundation:action_exe:ai} AI Tools, tier3]
                [\ref{sec:agent_foundation:action_exe:api} Native API Calls, tier3]
                [\ref{sec:agent_foundation:action_exe:ui} UI Operations, tier3]
            ]
            [\ref{sec:agent_foundation:inference} Model Inference, tier2
                [\ref{sec:agent_foundation:inference:complementary} Complementary Outputs, tier3]
                [\ref{sec:agent_foundation:inference:action} Action Inference, tier3]
                [\ref{sec:agent_foundation:inference:plan} Planning, tier3]
            ]
            [\ref{sec:agent_foundation:prompt} Prompt Engineering, tier2
                % [\ref{sec:agent_foundation:prompt:request} Complementary Information, tier3]
                % [\ref{sec:agent_foundation:prompt:example} Demonstrated Examples, tier3]
                % [\ref{sec:agent_foundation:prompt:action} Action Information, tier3]
                % [\ref{sec:agent_foundation:prompt:state} Environment States, tier3]
                % [\ref{sec:agent_foundation:prompt:instruction} Agent Instruction, tier3]
                % [\ref{sec:agent_foundation:prompt:action} User Request, tier3]
            ]
            [\ref{sec:agent_foundation:env} Operating Environment, tier2
                % [\ref{sec:agent_foundation:env:challenge} Challenges, tier3]
                [\ref{sec:agent_foundation:env:feedback} Environment Feedback, tier3]
                [\ref{sec:agent_foundation:env:state} Environment State Perception, tier3]
                [\ref{sec:agent_foundation:env:platform} Platform, tier3, yshift=-40pt]
            ]
            [\ref{sec:agent_foundation:architecture} Architecture and Workflow in a Nutshell, tier2]
        ]
        [\ref{sec:evolution} Evolution and Progression of\\ LLM-Brained GUI Agents, tier1
            [\ref{sec:evolution:vs} GUI Agent vs. API-Based Agent, tier2]
            [\ref{sec:evolution:agent}The Advent of LLM-Brained GUI Agents, tier2
                [\ref{sec:evolution:agent:model} Industry Models, tier3]
                [\ref{sec:evolution:agent:computer} Computer Systems, tier3]
                [\ref{sec:evolution:agent:mobile} Mobile Devices, tier3]
                [\ref{sec:evolution:agent:web} Web Domain, tier3]
            ]
            [\ref{sec:evolution:intellgent} The Shift Towards Intelligent Agents, tier2
                [\ref{sec:evolution:intellgent:rl} Reinforcement Learning, tier3]
                [\ref{sec:evolution:intellgent:nlp} Natural Language Processing, tier3]
                [\ref{sec:evolution:intellgent:ml} Machine Learning and Computer Vision, tier3]
            ]
            [\ref{sec:evolution:early} Early Automation Systems, tier2
                [\ref{sec:evolution:early:tool} Tools and Software, tier3]
                [\ref{sec:evolution:early:script} Script-Based Automation, tier3]
                [\ref{sec:evolution:early:script} Rule-Based Automation, tier3]
                [\ref{sec:evolution:early:random} Random-Based Automation, tier3, yshift=-60pt]
            ]
        ]
        [\ref{sec:background} Background, tier1
            [\ref{sec:background:gui_automation} {GUI Automation: Tools,} {Techniques, and Challenges}, tier2]
            [\ref{sec:background:llm_agent} LLM Agents: From Language to Action, tier2]
            [\ref{sec:background:llm} Large Language Models: Foundations and Capabilities, tier2]
        ]
        [\ref{sec:related_work} Related Work, tier1
            [\ref{sec:related_work:llm_agent} Survey on GUI Automation, tier2]
            [\ref{sec:related_work:gui_auto} Surveys on LLM Agents, tier2]
        ]
        [\ref{sec:introduction} Introduction, tier1
            [\ref{sec:introduction:structure} Survey Structure, tier2]
            [\ref{sec:introduction:scope} Scope of the Survey, tier2]
            [\ref{sec:introduction:motivation} Motivation for LLM-Brained GUI Agents, tier2]
        ]
    ]
    \end{forest}
    }
    \captionsetup{justification=centerlast} % Center the caption
    \vspace*{-1.5em}
    \caption{The structure of the survey on LLM-brained GUI agents.\label{sec:structure}}
    \vspace*{-2em}
\end{figure*}

% [Introduction, tier1
%     [Motivation for LLM-Brained GUI Agents, tier2]
%     [Scope of the Survey, tier2]
%     [Survey Structure, tier2]
% ]

\begin{table}[t]
\centering
\caption{List of abbreviations in alphabetical order.}
\label{tab:abbreviation}
\begin{tabular}{c|l}
\hline
\textbf{Acronym} & \textbf{Explanation} \\ \hline
AI&Artificial Intelligence\\ \hline
AITW&Android in the Wild \\ \hline
AITZ&Android in The Zoo \\ \hline
API&Application Programming Interface\\ \hline
CLI&Command-Line Interface\\ \hline
CLIP & Contrastive Language-Image Pre-Training\\ \hline
CoT&Chain-of-Thought\\ \hline
CSS&Cascading Style Sheets\\ \hline
CUA&Computer-Using Agent \\ \hline
CuP&Completion under Policy\\ \hline
CV&Computer Vision\\ \hline
DOM&Document Object Model\\ \hline
DPO&Direct Preference Optimization\\ \hline
GCC&General Computer Control\\ \hline
GPT&Generative Pre-trained Transformers\\ \hline
GUI   &   Graphical User Interface\\ \hline
HCI&Human-Computer Interaction\\ \hline
HTML&Hypertext Markup Language\\ \hline
ICL&In-Context Learning\\ \hline
IoU&Intersection over Union\\ \hline
LAM & Large Action Model\\ \hline
LLM&Large Language Model\\ \hline
LSTM&Long Short-Term Memory\\ \hline
LTM&Long-Term Memory\\ \hline
MCTS&Monte Carlo Tree Search\\ \hline
MoE & Mixture of Experts\\ \hline
MDP&Markov Decision Process\\ \hline
MLLM&Multimodal Large Language Model\\ \hline
OCR&Optical Character Recognition\\ \hline
OS&Operation System\\ \hline
RAG&Retrieval-Augmented Generation\\ \hline
ReAct&Reasoning and Acting\\ \hline
RL&Reinforcement Learning\\ \hline
RLHF & Reinforcement Learning from Human Feedback\\ \hline
RNN&Recurrent Neural Network\\ \hline
RPA&Robotic Process Automation\\ \hline
UI&User Interface\\ \hline
UX&User Experience\\ \hline
VAB&VisualAgentBench\\ \hline
VLM&Visual Language Models\\ \hline
ViT & Vision Transformer\\ \hline
VQA & Visual Question Answering \\ \hline
% VR & Virtual Reality \\ \hline
SAM&Segment Anything Model\\ \hline
SoM&Set-of-Mark\\ \hline
STM&Short-Trem Memory\\ \hline
\end{tabular}
\end{table}

% Survey Structure
The survey is organized as follows, with a structural illustration provided in Figure~\ref{sec:structure}. Section~\ref{sec:related_work} reviews related survey and review literature on LLM agents and GUI automation. Section~\ref{sec:background} provides preliminary background on LLMs, LLM agents, and GUI automation. Section~\ref{sec:related_work} traces the evolution of LLM-powered GUI agents. Section~\ref{sec:agent_foundation} introduces key components and advanced technologies within LLM-powered GUI agents, serving as a comprehensive guide.  Section~\ref{sec:framework} presents representative frameworks for LLM-powered GUI agents. Section~\ref{sec:data} discusses dataset collection and related data-centric research for optimizing LLMs in GUI agent. Section~\ref{sec:model} covers foundational and optimized models for GUI agents. Section~\ref{sec:evaluation} outlines evaluation metrics and benchmarks. Section~\ref{sec:applications} explores real-world applications and use cases. Finally, Section~\ref{sec:limitation} examines current limitations, challenges, and potential future directions, and section~\ref{sec:conclusion} conclude this survey. For clarity, a list of abbreviations is provided in Table~\ref{tab:abbreviation}.
\section{Related Work\label{sec:related_work}}

% \usepackage{multirow}
\begin{table*}[t]
\centering
\caption{Summary of representative surveys and books on GUI automation and LLM agents. A \checkmark symbol indicates that a publication explicitly addresses a given domain, while an $\bigcirc$ symbol signifies that the publication does not focus on the area but offers relevant insights. Publications covering both GUI automation and LLM agents are highlighted for emphasis.}
\label{tab:survey}
\resizebox{\textwidth}{!}{ % Resize the entire figure to fit \textwidth
\begin{tabular}{|l|l|ccc|}
\hline
\multicolumn{1}{|c|}{}                              & \multicolumn{1}{c|}{}                                                                                                                                                                           & \multicolumn{3}{c|}{\textbf{Scope}}                                                                                                                                                                                                         \\ \cline{3-5} 
\multicolumn{1}{|c|}{\multirow{-2}{*}{\textbf{Survey}}}      & \multicolumn{1}{c|}{\multirow{-2}{*}{\textbf{One Sentence Summary}}}                                                                                                                                     & \multicolumn{1}{c|}{\begin{tabular}[c]{@{}c@{}}GUI \\ Automation\end{tabular}} & \multicolumn{1}{c|}{\begin{tabular}[c]{@{}c@{}}LLM \\ Agent\end{tabular}} & \begin{tabular}[c]{@{}c@{}}LLM Agent +\\  GUI Automation\end{tabular} \\ \hline
Li \etal \cite{li2006effective}                     & A book on how to develop an automated GUI testing tool.                                                                                                                                          & \multicolumn{1}{c|}{\checkmark}                                                & \multicolumn{1}{c|}{}                                                     &                                                                       \\ \hline
Rodr{\'\i}guez \etal \cite{rodriguez202130}         & A survey on automated GUI testing in 30 years.                                                                                                                                                  & \multicolumn{1}{c|}{\checkmark}                                                & \multicolumn{1}{c|}{}                                                     &                                                                       \\ \hline
Arnatovich \etal \cite{arnatovich2018systematic}    & A survey on automated techniques for mobile functional GUI testing.                                                                                                                             & \multicolumn{1}{c|}{\checkmark}                                                & \multicolumn{1}{c|}{}                                                     &                                                                       \\ \hline
Ivan{\v{c}}i{\'c} \etal \cite{ivanvcic2019robotic}  & A literature review on RPA.                                                                                                                                                                     & \multicolumn{1}{c|}{\checkmark}                                                & \multicolumn{1}{c|}{}                                                     &                                                                       \\ \hline
Said \etal \cite{said2020gui}                       & An overview on mobile GUI testing.                                                                                                                                                              & \multicolumn{1}{c|}{\checkmark}                                                & \multicolumn{1}{c|}{}                                                     &                                                                       \\ \hline
Li \cite{li2023gui}                                 & An survey on Android GUI testing.                                                                                                                                                               & \multicolumn{1}{c|}{\checkmark}                                                & \multicolumn{1}{c|}{}                                                     &                                                                       \\ \hline
Oksanen \etal \cite{oksanen2023test}                & GUI testing on Windows OS.                                                                                                                                                                      & \multicolumn{1}{c|}{\checkmark}                                                & \multicolumn{1}{c|}{}                                                     &                                                                       \\ \hline
Deshmukh \etal \cite{deshmukh2023automated}         & A survey on GUI testing for improving user experience.                                                                                                                                          & \multicolumn{1}{c|}{\checkmark}                                                & \multicolumn{1}{c|}{}                                                     &                                                                       \\ \hline
Bajammal \etal \cite{bajammal2020survey}            & A survey on the use of computer vision for software engineering.                                                                                                                                & \multicolumn{1}{c|}{\checkmark}                                                    & \multicolumn{1}{c|}{}                                                     &                                                                       \\ \hline
Yu \etal \cite{yu2023vision}                        & A survey on using computer for mobile app GUI testing.                                                                                                                                          & \multicolumn{1}{c|}{\checkmark}                                                & \multicolumn{1}{c|}{}                                                     &                                                                       \\ \hline
Syed \etal \cite{syed2020robotic}                   & A review of contemporary themes and challenges in RPA.                                                                                                                                          & \multicolumn{1}{c|}{\checkmark}                                                    & \multicolumn{1}{c|}{}                                                     &                                                                       \\ \hline
Chakraborti \etal \cite{chakraborti2020robotic}     & A review of emerging trends of intelligent process automation.                                                                                                                                  & \multicolumn{1}{c|}{\checkmark}                                                & \multicolumn{1}{c|}{}                                                     &                                                                       \\ \hline
Enriquez \etal \cite{enriquez2020robotic}           & A scientific and industrial systematic mapping study of RPA.                                                                                                                                    & \multicolumn{1}{c|}{\checkmark}                                                & \multicolumn{1}{c|}{}                                                     &                                                                       \\ \hline
Ribeiro \etal \cite{ribeiro2021robotic}             & A review of combining AI and RPA in industry 4.0.                                                                                                                                               & \multicolumn{1}{c|}{\checkmark}                                                & \multicolumn{1}{c|}{}                                                     &                                                                       \\ \hline
Nass \etal \cite{nass2021many}                      & Discuss the chanllenges of GUI testing.                                                                                                                                                         & \multicolumn{1}{c|}{\checkmark}                                                & \multicolumn{1}{c|}{}                                                     &                                                                       \\ \hline
Agostinelli \etal \cite{agostinelli2019research}    & Discuss the research challenges of intelligent RPA.                                                                                                                                             & \multicolumn{1}{c|}{\checkmark}                                                & \multicolumn{1}{c|}{}                                                     &                                                                       \\ \hline
Wali \etal \cite{wali2023task}                      & A review on task automation with intelligent agents.                                                                                                                                            & \multicolumn{1}{c|}{\checkmark}                                                & \multicolumn{1}{c|}{}                                                     &                                                                       \\ \hline
Zhao \etal. \cite{zhao2023survey}                   & A comprehensive survey of LLMs.                                                                                                                                                                        & \multicolumn{1}{c|}{}                                                          & \multicolumn{1}{c|}{\checkmark}                                           &                                                                       \\ \hline
Zhao \etal. \cite{zhao2023depth}                    & A survey of LLM-based agents.                                                                                                                                                                   & \multicolumn{1}{c|}{}                                                          & \multicolumn{1}{c|}{\checkmark}                                           &                                                                       \\ \hline
Cheng \etal \cite{cheng2024exploring}               & An overview of LLM-based AI agent.                                                                                                                                                              & \multicolumn{1}{c|}{}                                                          & \multicolumn{1}{c|}{\checkmark}                                           &                                                                       \\ \hline
Li \etal \cite{li2024personal}                      & A survey on personal LLM agents on their capability, efficiency and security.                                                                                                                   & \multicolumn{1}{c|}{}                                                          & \multicolumn{1}{c|}{\checkmark}                                           &                                                                       \\ \hline
Xie \etal. \cite{xi2023rise}                        & A comprehensive survey of LLM-based agents.                                                                                                                                                     & \multicolumn{1}{c|}{}                                                          & \multicolumn{1}{c|}{\checkmark}                                           &                                                                       \\ \hline
Wang \etal. \cite{wang2024survey}                   & A survey on LLM-based autonomous agents.                                                                                                                                                        & \multicolumn{1}{c|}{}                                                          & \multicolumn{1}{c|}{\checkmark}                                           &                                                                       \\ \hline
Guo \etal \cite{guo2024large}                       & A survey of mult-agent LLM frameworks.                                                                                                                                                          & \multicolumn{1}{c|}{}                                                          & \multicolumn{1}{c|}{\checkmark}                                           &                                                                       \\ \hline
Han \etal \cite{han2024llm}                         & A survey on LLM multi-agent systems, with their challenges and open problems.                                                                                                                   & \multicolumn{1}{c|}{}                                                          & \multicolumn{1}{c|}{\checkmark}                                           &                                                                       \\ \hline
Sun \etal \cite{sun2024llm}                         & A survey on LLM-based multi-agent reinforcement learning.                                                                                                                                       & \multicolumn{1}{c|}{}                                                          & \multicolumn{1}{c|}{\checkmark}                                           &                                                                       \\ \hline
Huang \etal \cite{huang2024understanding}           & A survey on planning in LLM agents.                                                                                                                                                             & \multicolumn{1}{c|}{}                                                          & \multicolumn{1}{c|}{\checkmark}                                           &                                                                       \\ \hline
Aghzal \etal \cite{aghzal2025survey}           & A survey on automated planning in LLMs.                                                                                                                                                             & \multicolumn{1}{c|}{}                                                          & \multicolumn{1}{c|}{\checkmark}                                           &                                                                       \\ \hline
Zheng \etal \cite{zheng2025lifelong}           & Discuss the roadmap of lifelong learning in LLM agents.                                                                                                                                                             & \multicolumn{1}{c|}{}                                                          & \multicolumn{1}{c|}{\checkmark}                                           &                                                                       \\ \hline
Zhang \etal \cite{zhang2024survey}                  & A survey on the memory of LLM-based agents.                                                                                                                                                      & \multicolumn{1}{c|}{}                                                          & \multicolumn{1}{c|}{\checkmark}                                           &                                                                       \\ \hline
Shen \cite{shen2024llm}                             & A survey of the tool usage in LLM agents.                                                                                                                                                       & \multicolumn{1}{c|}{}                                                          & \multicolumn{1}{c|}{\checkmark}                                           &                                                                       \\ \hline
Chang \etal \cite{chang2024survey}                  & A survey on evaluation of LLMs.                                                                                                                                                                 & \multicolumn{1}{c|}{}                                                          & \multicolumn{1}{c|}{\checkmark}                                               &                                                                       \\ \hline
Li \etal \cite{li2024survey}                        & A survey on benchmarks multimodal applications.                                                                                                                                                 & \multicolumn{1}{c|}{}                                                          & \multicolumn{1}{c|}{\checkmark}                                               &                                                                       \\ \hline
Li \etal \cite{li2025benchmark}                        & A survey on benchmarking evaluations, applications, and challenges of visual LLMs.                                                                                                                                                 & \multicolumn{1}{c|}{}                                                          & \multicolumn{1}{c|}{\checkmark}                                               &                                                                       \\ \hline
Huang and Zhang \cite{huang2024survey2}             & A survey on evaluation of multimodal LLMs.                                                                                                                                                      & \multicolumn{1}{c|}{}                                                          & \multicolumn{1}{c|}{\checkmark}                                               &                                                                       \\ \hline
Xie \etal. \cite{xie2024large}                      & A survey on LLM based multimodal agent.                                                                                                                                                         & \multicolumn{1}{c|}{}                                                          & \multicolumn{1}{c|}{\checkmark}                                           & $\bigcirc$                                                                \\ \hline
Durante \etal \cite{durante2024agent}               & A survey of multimodal interaction with AI agents.                                                                                                                                              & \multicolumn{1}{c|}{}                                                          & \multicolumn{1}{c|}{\checkmark}                                           & $\bigcirc$                                                                \\ \hline
\rowcolor[HTML]{C0C0C0} 
Wu \etal \cite{wu2024foundations}                   & A survey of foundations and trend on multimodal mobile agents.                                                                                                                                  & \multicolumn{1}{c|}{\cellcolor[HTML]{C0C0C0}}                                  & \multicolumn{1}{c|}{\cellcolor[HTML]{C0C0C0}\checkmark}                   & \checkmark                                                            \\ \hline
\rowcolor[HTML]{C0C0C0} 
Wang \etal \cite{wang2024guiagentsfoundationmodels} & A survey on the integration of foundation models with GUI agents.                                                                                                                               & \multicolumn{1}{c|}{\cellcolor[HTML]{C0C0C0}}                                  & \multicolumn{1}{c|}{\cellcolor[HTML]{C0C0C0}\checkmark}                   & \checkmark                                                            \\ \hline
\rowcolor[HTML]{C0C0C0}
Gao \etal \cite{gao2024generalist} & A survey on autonomous agents across digital platforms.                                                                                                                               & \multicolumn{1}{c|}{\cellcolor[HTML]{C0C0C0}}                                  & \multicolumn{1}{c|}{\cellcolor[HTML]{C0C0C0}\checkmark}                   & \checkmark                                                            \\ \hline
\rowcolor[HTML]{C0C0C0}
\rowcolor[HTML]{C0C0C0}
Dang \etal \cite{nguyen2024guiagentssurvey} & A survey on GUI agents.                                                                                                                               & \multicolumn{1}{c|}{\cellcolor[HTML]{C0C0C0}}                                  & \multicolumn{1}{c|}{\cellcolor[HTML]{C0C0C0}\checkmark}                   & \checkmark                                                            \\ \hline
\rowcolor[HTML]{C0C0C0}
\rowcolor[HTML]{C0C0C0}
Liu \etal \cite{liu2025llm} & A survey on GUI agent on phone automation.                                                                                                                              & \multicolumn{1}{c|}{\cellcolor[HTML]{C0C0C0}}                                  & \multicolumn{1}{c|}{\cellcolor[HTML]{C0C0C0}\checkmark}                   & \checkmark                                                            \\ \hline
\rowcolor[HTML]{C0C0C0}
\rowcolor[HTML]{C0C0C0}
Hu \etal \cite{hu2024agents} & A survey on MLLM based agents for OS.                                                                                                                               & \multicolumn{1}{c|}{\cellcolor[HTML]{C0C0C0}}                                  & \multicolumn{1}{c|}{\cellcolor[HTML]{C0C0C0}\checkmark}                   & \checkmark                                                            \\ \hline
\rowcolor[HTML]{C0C0C0}
Shi \etal \cite{shi2025towards} & A survey of building trustworthy GUI agents.                                                                                                                              & \multicolumn{1}{c|}{\cellcolor[HTML]{C0C0C0}}                                  & \multicolumn{1}{c|}{\cellcolor[HTML]{C0C0C0}\checkmark}                   & \checkmark                                                            \\ \hline
\rowcolor[HTML]{C0C0C0}
Ning \etal \cite{ning2025survey} & A survey of agents for Web automation.                                                                                                                              & \multicolumn{1}{c|}{\cellcolor[HTML]{C0C0C0}}                                  & \multicolumn{1}{c|}{\cellcolor[HTML]{C0C0C0}\checkmark}                   & \checkmark                                                            \\ \hline
\rowcolor[HTML]{C0C0C0}
Tang \etal \cite{tang2025surveymllmbasedguiagents} & A survey of GUI agents powered by (multimodal) LLMs.                                                                                                                              & \multicolumn{1}{c|}{\cellcolor[HTML]{C0C0C0}}                                  & \multicolumn{1}{c|}{\cellcolor[HTML]{C0C0C0}\checkmark}                   & \checkmark                                                            \\ \hline
\rowcolor[HTML]{C0C0C0}
Li and Huang \etal \cite{li2025summaryguiagentsfoundation} & A summary of GUI agents powered by foundation models and enhanced through reinforcement learning                                                                                                                               & \multicolumn{1}{c|}{\cellcolor[HTML]{C0C0C0}}                                  & \multicolumn{1}{c|}{\cellcolor[HTML]{C0C0C0}\checkmark}                   & \checkmark                                                            \\ \hline
\rowcolor[HTML]{C0C0C0}
Sager \etal \cite{sager2025ai} & A review of AI agent for computer use.                                                                                                                               & \multicolumn{1}{c|}{\cellcolor[HTML]{C0C0C0}$\bigcirc$}                                  & \multicolumn{1}{c|}{\cellcolor[HTML]{C0C0C0}\checkmark}                   & \checkmark                                                            \\ \hline
\rowcolor[HTML]{C0C0C0}
\textbf{Our work}                                   & \textbf{\begin{tabular}[c]{@{}l@{}}A comprehensive survey on LLM-brained GUI agents, on their foundations, technologies, \\ frameworks, data, models, applications, challenges and future roadmap.\end{tabular}} & \multicolumn{1}{c|}{\cellcolor[HTML]{C0C0C0}\textbf{$\bigcirc$}}                   & \multicolumn{1}{c|}{\cellcolor[HTML]{C0C0C0}\textbf{\checkmark}}          & \textbf{\checkmark}                                                   \\ \hline
\end{tabular}
}
\end{table*}

The integration of LLMs with GUI agents is an emerging and rapidly growing field of research. Several related surveys and tutorials provide foundational insights and guidance. We provide a brief review of existing overview articles on GUI automation and LLM agents, as these topics closely relate to and inform our research focus. To begin, we provide an overview of representative surveys and books on GUI automation, LLM agents, and their integration, as summarized in Table~\ref{tab:survey}. These works either directly tackle one or two core areas in GUI automation and LLM-driven agents, or provide valuable insights that, while not directly addressing the topic, contribute indirectly to advancing the field.

\subsection{Survey on GUI Automation\label{sec:related_work:gui_auto}}
GUI automation has a long history and wide applications in industry, especially in GUI testing \cite{li2006effective, rodriguez202130, arnatovich2018systematic} and RPA \cite{ivanvcic2019robotic} for task automation \cite{wali2023task}.

Said \etal \cite{said2020gui} provide an overview of GUI testing for mobile applications, covering objectives, approaches, and challenges within this domain. Focusing on Android applications, Li \cite{li2023gui} narrows the scope further, while Oksanen \etal \cite{oksanen2023test} explore automatic testing techniques for Windows GUI applications, a key platform for agent operations. Similarly, Moura \etal \cite{moura2023cytestion} review GUI testing for web applications, which involves diverse tools, inputs, and methodologies. Deshmukh \etal \cite{deshmukh2023automated} discuss automated GUI testing for enhancing user experience, an area where LLMs also bring new capabilities. A cornerstone of modern GUI testing is computer vision (CV), which is used to interpret UI elements and identify actionable controls \cite{bajammal2020survey}. Yu \etal \cite{yu2023vision} survey the application of CV in mobile GUI testing, highlighting both its significance and associated challenges. In LLM-powered GUI agents, application UI screenshots are equally essential, serving as key inputs for reliable task comprehension and execution.

On the other hand, RPA, which focuses on automating repetitive human tasks, also relies heavily on GUI automation for relevant processes. Syed \etal \cite{syed2020robotic} review this field and highlight contemporary RPA themes, identifying key challenges for future research. Chakraborti \etal \cite{chakraborti2020robotic} emphasize the importance of shifting from traditional, script-based RPA toward more intelligent, adaptive paradigms, offering a systematic overview of advancements in this direction. Given RPA's extensive industrial applications, Enriquez \etal \cite{enriquez2020robotic} and Ribeiro \etal \cite{ribeiro2021robotic} survey the field from an industrial perspective, underscoring its significance and providing a comprehensive overview of RPA methods, development trends, and practical challenges.

Both GUI testing \cite{nass2021many} and RPA \cite{agostinelli2019research} continue to face significant challenges in achieving greater intelligence and robustness. LLM-powered GUI agents are poised to play a transformative role in these fields, providing enhanced capabilities and adding substantial value to address these persistent issues.

\subsection{Surveys on LLM Agents\label{sec:related_work:llm_agent}}
The advent of LLMs has significantly enhanced the capabilities of intelligent agents \cite{zhao2023depth}, enabling them to tackle complex tasks previously out of reach, particularly those involving natural language understanding and code generation \cite{cheng2024exploring}. This advancement has spurred substantial research into LLM-based agents designed for a wide array of applications \cite{li2024personal}. 

Both Xie \etal \cite{xi2023rise} and Wang \etal \cite{wang2024survey} offer comprehensive surveys on LLM-powered agents, covering essential background information, detailed component breakdowns, taxonomies, and various applications. These surveys serve as valuable references for a foundational understanding of LLM-driven agents, laying the groundwork for further exploration into LLM-based GUI agents. Xie \etal \cite{xie2024large} provide an extensive overview of multimodal agents, which can process images, videos, and audio in addition to text. This multimodal capability significantly broadens the scope beyond traditional text-based agents \cite{durante2024agent}. Notably, most GUI agents fall under this category, as they rely on image inputs, such as screenshots, to interpret and interact with graphical interfaces effectively. Multi-agent frameworks are frequently employed in the design of GUI agents to enhance their capabilities and scalability. Surveys by Guo \etal \cite{guo2024large} and Han \etal \cite{han2024llm} provide comprehensive overviews of the current landscape, challenges, and future directions in this area. Sun \etal \cite{sun2024llm} provide an overview of recent methods that leverage reinforcement learning to strengthen multi-agent LLM systems, opening new pathways for enhancing their capabilities and adaptability. These surveys offer valuable insights and guidance for designing effective multi-agent systems within GUI agent frameworks. 

In the realm of digital environments, Wu \etal \cite{wu2024foundations} presents a survey on LLM agents operating in mobile environments, covering key aspects of mobile GUI agents.  In a boarder scope, Wang \etal \cite{wang2024guiagentsfoundationmodels} present a survey on the integration of foundation models with GUI agents. Another survey by Gao \etal provides an overview of autonomous agents operating across various digital platforms \cite{gao2024generalist}, highlighting their capabilities, challenges, and applications. All these surveys highlighting emerging trends in this area.

Regarding individual components within LLM agents, several surveys provide detailed insights that are especially relevant for GUI agents. Huang \etal \cite{huang2024understanding} examine planning mechanisms in LLM agents, which are essential for executing long-term tasks—a frequent requirement in GUI automation. Zhang \etal \cite{zhang2024survey} explore memory mechanisms, which allow agents to store critical historical information, aiding in knowledge retention and decision-making. Additionally, Shen \cite{shen2024llm} surveys the use of tools by LLMs (such as APIs and code) to interact effectively with their environments, grounding actions in ways that produce tangible impacts. Further, Chang \etal \cite{chang2024survey} provide a comprehensive survey on evaluation methods for LLMs, which is crucial for ensuring the robustness and safety of GUI agents. Two additional surveys, \cite{li2024survey} and \cite{huang2024survey2}, provide comprehensive overviews of benchmarks and evaluation methods specifically tailored to multimodal LLMs. The evaluation also facilitates a feedback loop, allowing agents to improve iteratively based on assessment results. Together, these surveys serve as valuable resources, offering guidance on essential components of LLM agents and forming a foundational basis for LLM-based GUI agents.

Compared to existing surveys, our work offers a significantly more comprehensive and up-to-date overview of the LLM-powered GUI agent landscape. We curate and synthesize over 500 references, covering a wide range of topics including foundation models, data sources, system frameworks, benchmarks, evaluation methodologies, and practical deployments. While prior surveys often concentrate on narrower aspects on selected platform (\eg web, mobile), our survey takes a holistic perspective that spans the full development and deployment lifecycle. Beyond narrative summaries, we also provide consolidated reference tables for each subdomain, enabling readers to quickly categorize and locate relevant works across platforms and research themes—serving as a practical handbook for both researchers and practitioners. Furthermore, we incorporate foundational background material and propose evaluation taxonomies that make the survey accessible to newcomers, addressing gaps in prior work that often assume a high degree of prior familiarity.
\section{Background\label{sec:background}}
The development of LLM-brained GUI agents is grounded in three major advancements: \textit{(i)} large language models (LLMs) \cite{zhao2023survey}, which bring advanced capabilities in natural language understanding and code generation, forming the core intelligence of these agents; \textit{(ii)} accompanying agent architectures and tools \cite{wang2024survey} that extend LLM capabilities, bridging the gap between language models and physical environments to enable tangible impacts; and \textit{(iii)} GUI automation \cite{yeh2009sikuli}, which has cultivated a robust set of tools, models, and methodologies essential for GUI agent functionality. Each of these components has played a critical role in the emergence of LLM-powered GUI agents. In the following subsections, we provide a brief overview of these areas to set the stage for our discussion.

\subsection{Large Language Models: Foundations and Capabilities\label{sec:background:llm}}
The study of language models has a long and rich history \cite{shannon1951prediction}, beginning with early statistical language models \cite{cavnar1994n} and smaller neural network architectures \cite{chung2014empirical}. Building on these foundational concepts, recent advancements have focused on transformer-based LLMs, such as the Generative Pre-trained Transformers (GPTs) \cite{mann2020language}. These models are pretrained on extensive text corpora and feature significantly larger model sizes, validating scaling laws and demonstrating exceptional capabilities across a wide range of natural language tasks. Beyond their sheer size, these LLMs exhibit enhanced language understanding and generation abilities, as well as emergent properties that are absent in smaller-scale language models \cite{wei2021finetuned}.

Early neural language models, based on architectures like recurrent neural networks (RNNs) \cite{medsker2001recurrent} and long short-term memory networks (LSTMs) \cite{hochreiter1997long}, were limited in both performance and generalization. The introduction of the Transformer model, built on the attention mechanism \cite{vaswani2017attention}, marked a transformative milestone, establishing the foundational architecture now prevalent across almost all subsequent LLMs. This development led to variations in model structures, including encoder-only models (\eg BERT \cite{devlin2018bert}, RoBERTa \cite{liu2019roberta}, ALBERT \cite{lan2019albert}), decoder-only models (\eg GPT-1 \cite{radford2018improving}, GPT-2 \cite{radford2019language}), and encoder-decoder models (\eg T5 \cite{raffel2020exploring}, BART \cite{lewis2019bart}). In 2022, ChatGPT \cite{wu2023brief} based on GPT-3.5 \cite{ouyang2022training} launched as a groundbreaking LLM, fundamentally shifting perceptions of what language models can achieve. Since then, numerous advanced LLMs have emerged, including GPT-4 \cite{achiam2023gpt}, LLaMA-3 \cite{dubey2024llama}, and Gemini \cite{team2023gemini}, propelling the field into rapid growth. Today's LLMs are highly versatile, with many of them are capable of processing multimodal data and performing a range of tasks, from question answering to code generation, making them indispensable tools in various applications \cite{hurst2024gpt, jiang2024xpert, zhang2024allhands, liu2024large2}.

The emergence of LLMs has also introduced significant advanced properties that invigorate their applications, making previously challenging tasks, such as natural language-driven GUI agents feasible. These advancements include:
\begin{enumerate}
    \item \textbf{Few-Shot Learning \cite{mann2020language}:} Also referred to as in-context learning \cite{dong2022survey}, LLMs can acquire new tasks from a small set of demonstrated examples presented in the prompt during inference, eliminating the need for retraining. This capability is crucial for enabling GUI agents to generalize across different environments with minimal effort.
    \item \textbf{Instruction Following \cite{zhang2023instruction}:} After undergoing instruction tuning, LLMs exhibit a remarkable ability to follow instructions for novel tasks, demonstrating strong generalization skills \cite{ouyang2022training}. This allows LLMs to effectively comprehend user requests directed at GUI agents and to follow predefined objectives accurately.
    \item \textbf{Long-Term Reasoning \cite{huang2022towards}:} LLMs possess the ability to plan and solve complex tasks by breaking them down into manageable steps, often employing techniques like chain-of-thought (CoT) reasoning \cite{wei2022chain, ding2023everything}. This capability is essential for GUI agents, as many tasks require multiple steps and a robust planning framework.
    \item \textbf{Code Generation and Tool Utilization \cite{chen2021evaluating}:} LLMs excel in generating code and utilizing various tools, such as APIs \cite{shen2024llm}. This expertise is vital, as code and tools form the essential toolkit for GUI agents to interact with their environments.
    \item \textbf{Multimodal Comprehension \cite{yin2023survey}:} Advanced LLMs can integrate additional data modalities, such as images, into their training processes, evolving into multimodal models. This ability is particularly important for GUI agents, which must interpret GUI screenshots presented as images in order to function effectively \cite{white2019improving}.
\end{enumerate}
To further enhance the specialization of LLMs for GUI agents, researchers often fine-tune these models with domain-specific data, such as user requests, GUI screenshots, and action sequences, thereby increasing their customization and effectiveness. In Section~\ref{sec:model}, we delve into these advanced, tailored models for GUI agents, discussing their unique adaptations and improved capabilities for interacting with graphical interfaces.

\subsection{LLM Agents: From Language to Action\label{sec:background:llm_agent}}
Traditional AI agents have often focused on enhancing specific capabilities, such as symbolic reasoning or excelling in particular tasks like Go or Chess. In contrast, the emergence of LLMs has transformed AI agents by providing them with a natural language interface, enabling human-like decision-making capabilities, and equipping them to perform a wide variety of tasks and take tangible actions in diverse environments \cite{wang2024survey, languagemodelssolvecomputer, liu2024large, qiao2023taskweaver}. In LLM agents, if LLMs form the ``brain'' of a GUI agent, then its accompanying components serve as its ``eyes and hands'', enabling the LLM to perceive the environment's status and translate its textual output into actionable steps that generate tangible effects \cite{xi2023rise}. These components transform LLMs from passive information sources into interactive agents that execute tasks on behalf of users, which redefine the role of LLMs from purely text-generative models to systems capable of driving actions and achieving specific goals.

In the context of GUI agents, the agent typically perceives the GUI status through screenshots and widget trees \cite{boshart2003growing}, then performs actions to mimic user operations (\eg mouse clicks, keyboard inputs, touch gestures on phones) within the environment. Since tasks can be long-term, effective planning and task decomposition are often required, posing unique challenges. Consequently, an LLM-powered GUI agent usually possess multimodal capabilities \cite{xie2024large}, a robust planning system \cite{huang2024understanding}, a memory mechanism to analyze historical interactions \cite{zhang2024survey}, and a specialized toolkit to interact with its environment \cite{li2006effective}. We will discuss these tailored designs for GUI agents in detail in Section~\ref{sec:agent_foundation}.

\subsection{GUI Automation: Tools, Techniques, and Challenges\label{sec:background:gui_automation}}
GUI automation has been a critical area of research and application since the early days of GUIs in computing. Initially developed to improve software testing efficiency, GUI automation focused on simulating user actions, such as clicks, text input, and navigation, across graphical applications to validate functionality \cite{said2020gui}. Early GUI automation tools  were designed to execute repetitive test cases on static interfaces \cite{rodriguez202130}. These approaches streamlined quality assurance processes, ensuring consistency and reducing manual testing time. As the demand for digital solutions has grown, GUI automation has expanded beyond testing to other applications, including RPA \cite{ivanvcic2019robotic} and Human-Computer Interaction (HCI) \cite{li2021artificial}. RPA leverages GUI automation to replicate human actions in business workflows, automating routine tasks to improve operational efficiency. Similarly, HCI research employs GUI automation to simulate user behaviors, enabling usability assessments and interaction studies. In both cases, automation has significantly enhanced productivity and user experience by minimizing repetitive tasks and enabling greater system adaptability \cite{abuaddous2022automated, gao2024assistgui}.

Traditional GUI automation methods have primarily depended on scripting and rule-based frameworks \cite{hellmann2011rule, roscipt}. Scripting-based automation utilizes languages such as Python, Java, and JavaScript to control GUI elements programmatically. These scripts simulate a user's actions on the interface, often using tools like Selenium \cite{bruns2009web} for web-based automation or AutoIt \cite{rupp2022establishment} and SikuliX \cite{granda2021towards} for desktop applications. Rule-based approaches, meanwhile, operate based on predefined heuristics, using rules to detect and interact with specific GUI elements based on properties such as location, color, and text labels \cite{hellmann2011rule}. While effective for predictable, static workflows \cite{Xu2024LLM4WorkflowAL}, these methods struggle to adapt to the variability of modern GUIs, where dynamic content, responsive layouts, and user-driven changes make it challenging to maintain rigid, rule-based automation \cite{gove2012machine}.

CV has become essential for interpreting the visual aspects of GUIs \cite{yu2023vision, li2021screen2vec, chang2010gui}, enabling automation tools to recognize and interact with on-screen elements even as layouts and designs change. CV techniques allow GUI automation systems to detect and classify on-screen elements, such as buttons, icons, and text fields, by analyzing screenshots and identifying regions of interest \cite{zou2023object, ye2021empirical, chen2020object}. Optical Character Recognition (OCR) further enhances this capability by extracting text content from images, making it possible for automation systems to interpret labels, error messages, and form instructions accurately \cite{qian2022accelerating}. Object detection models add robustness, allowing automation agents to locate GUI elements even when the visual layout shifts \cite{white2019improving}. By incorporating CV, GUI automation systems achieve greater resilience and adaptability in dynamic environments.

Despite advances, traditional GUI automation methods fall short in handling the complexity and variability of contemporary interfaces. Today's applications often feature dynamic, adaptive elements that cannot be reliably automated through rigid scripting or rule-based methods alone \cite{gambino2018framework, he2008adaptive}. Modern interfaces increasingly require contextual awareness \cite{stefanidi2022real}, such as processing on-screen text, interpreting user intent, and recognizing visual cues. These demands reveal the limitations of existing automation frameworks and the need for more flexible solutions capable of real-time adaptation and context-sensitive responses. 

LLMs offer a promising solution to these challenges. With their capacity to comprehend natural language, interpret context, and generate adaptive scripts, LLMs can enable more intelligent, versatile GUI automation \cite{liu2024make}. Their ability to process complex instructions and learn from context allows them to bridge the gap between static, rule-based methods and the dynamic needs of contemporary GUIs \cite{brie2023evaluating}. By integrating LLMs with GUI agents, these systems gain the ability to generate scripts on-the-fly based on the current state of the interface, providing a level of adaptability and sophistication that traditional methods cannot achieve. The combination of LLMs and GUI agents paves the way for an advanced, user-centered automation paradigm, capable of responding flexibly to user requests and interacting seamlessly with complex, evolving interfaces.
\section{Evolution and Progression of LLM-Brained GUI Agents\label{sec:evolution}}
\begin{figure*}[t]
    \centering
    \includegraphics[width=\textwidth]{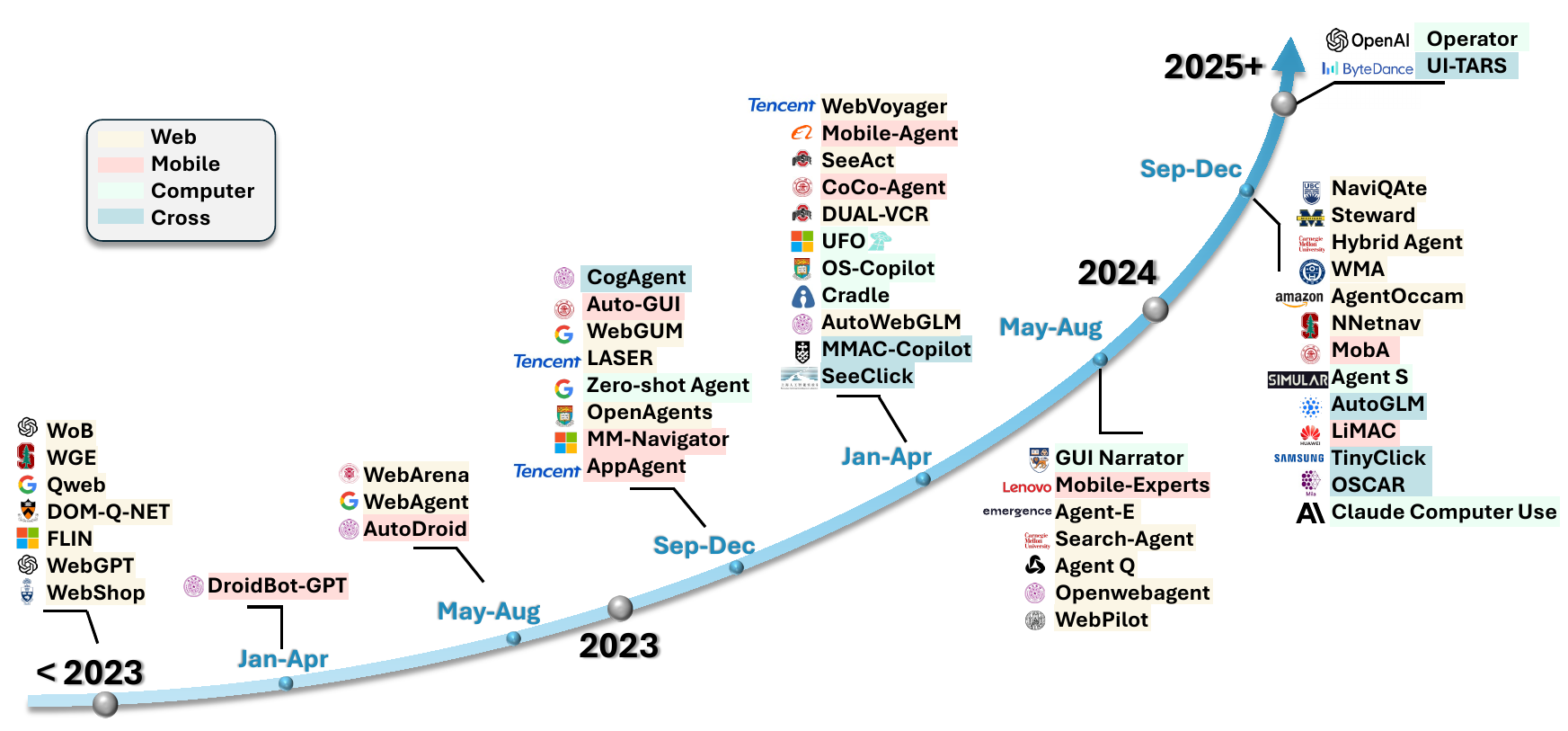}
    \caption{An overview of GUI agents evolution over years.}
    \label{fig:milestone}
    % \vspace{-2em}
\end{figure*}
``Rome wasn't built in a day.'' The development of LLM-brained GUI agents has been a gradual journey, grounded in decades of research and technical progress. Beginning with simple GUI testing scripts and rule-based automation frameworks, the field has evolved significantly through the integration of machine learning techniques, creating more intelligent and adaptive systems. The introduction of LLMs, especially multimodal models, has transformed GUI automation by enabling natural language interactions and fundamentally reshaping how users interact with software applications. 

As illustrated in Figure~\ref{fig:milestone}, prior to 2023 and the emergence of LLMs, work on GUI agents was limited in both scope and capability. Since then, the proliferation of LLM-based approaches has fostered numerous notable developments across platforms including web, mobile, and desktop environments. This surge is ongoing and continues to drive innovation in the field. This section takes you on a journey tracing the evolution of GUI agents, emphasizing key milestones that have brought the field to its present state.

\subsection{Early Automation Systems\label{sec:evolution:early}}

In the initial stages of GUI automation, researchers relied on random-based, rule-based, and script-based strategies. While foundational, these methods had notable limitations in terms of flexibility and adaptability.

\subsubsection{Random-Based Automation\label{sec:evolution:early:random}} 
Random-based automation uses random sequences of actions within the GUI without relying on specific algorithms or structured models using monkey test \cite{wetzlmaier2016framework}. This approach was widely used in GUI testing to uncover potential issues by exploring unpredictable input sequences~\cite{gui_random_testing}. While effective at identifying edge cases and bugs, random-based methods were often inefficient due to a high number of redundant or irrelevant trials.

\subsubsection{Rule-Based Automation\label{sec:evolution:early:rule}} 
Rule-based automation applies predefined rules and logic to automate tasks. In 2001, Memon \etal \cite{memon2001hierarchical} introduced a planning approach that generated GUI test cases by transforming initial states to goal states through a series of predefined operators. Hellmann \etal \cite{hellmann2011rule} (2011) demonstrated the potential of rule-based approaches in exploratory testing, enhancing bug detection. In the RPA domain, SmartRPA~\cite{agostinelli2020automated} (2020) used rule-based processing to automate routine tasks, illustrating the utility of rules for streamlining structured processes.

\subsubsection{Script-Based Automation\label{sec:evolution:early:script}} 
Script-based automation relies on detailed scripts to manage GUI interactions. Tools like jRapture~\cite{jRapture} (2000) record and replay Java-based GUI sequences using Java binaries and the JVM, enabling consistent execution by precisely reproducing input sequences. Similarly, DART~\cite{dart} (2003) automated the GUI testing lifecycle, from structural analysis to test case generation and execution, offering a comprehensive framework for regression testing.

\subsubsection{Tools and Software\label{sec:evolution:early:tool}} 
A range of software tools were developed for GUI testing and business process automation during this period. Microsoft Power Automate~\cite{powerautomate} (2019) provides a low-code/no-code environment for creating automated workflows within Microsoft applications. Selenium~\cite{selenium} (2004) supports cross-browser web testing, while Appium~\cite{appium} (2012) facilitates mobile UI automation. Commercial tools like TestComplete~\cite{testcomplete} (1999), Katalon Studio~\cite{katalon} (2015), and Ranorex~\cite{ranorex} (2007) allow users to create automated tests with cross-platform capabilities.

Although these early systems were effective for automating specific, predefined workflows, they lacked flexibility and required manual scripting or rule-based logic. Nonetheless, they established the foundations of GUI automation, upon which more intelligent systems were built.

\subsection{The Shift Towards Intelligent Agents\label{sec:evolution:intellgent}}

The incorporation of machine learning marked a major shift towards more adaptable and capable GUI agents. Early milestones in this phase included advancements in machine learning, natural language processing, computer vision, and reinforcement learning applied to GUI tasks.

\subsubsection{Machine Learning and Computer Vision\label{sec:evolution:intellgent:ml}}
RoScript~\cite{roscipt} (2020) was a pioneering system that introduced a non-intrusive robotic testing system for touchscreen applications, expanding GUI automation to diverse platforms. AppFlow~\cite{AppFlow} (2018) used machine learning to recognize common screens and UI components, enabling modular testing for broad categories of applications. Progress in computer vision also enabled significant advances in GUI testing, with frameworks~\cite{chang2010gui} (2010) automating visual interaction tasks. Humanoid~\cite{li2019humanoid} (2019) uses a deep neural network model trained on human interaction traces within the Android system to learn how users select actions based on an app's GUI. This model is then utilized to guide test input generation, resulting in improved coverage and more human-like interaction patterns during testing. Similarly, Deep GUI~\cite{yazdanibanafshedaragh2021deep} (2021) applies deep learning techniques to filter out irrelevant parts of the screen, thereby enhancing black-box testing effectiveness in GUI testing by focusing only on significant elements. These approaches demonstrate the potential of deep learning to make GUI testing more efficient and intuitive by aligning it closely with actual user behavior.

Widget detection, as demonstrated by White \etal~\cite{white2019improving} (2019), leverages computer vision to accurately identify UI elements, serving as a supporting technique that enables more intelligent and responsive UI automation. By detecting and categorizing interface components, this approach enhances the agent's ability to interact effectively with complex and dynamic GUIs \cite{xie2020uied}.

\subsubsection{Natural Language Processing\label{sec:evolution:intellgent:nlp}} 
Natural language processing capabilities introduced a new dimension to GUI automation. Systems like RUSS~\cite{xu2021groundingopendomaininstructionsautomate} (2021) and FLIN~\cite{mazumder2020flin} (2020) allowed users to control GUIs through natural language commands, bridging human language and machine actions. Datasets, such as those in~\cite{mappingnaturallanguageinstructions} (2020), further advanced the field by mapping natural language instructions to mobile UI actions, opening up broader applications in GUI control. However, these approaches are limited to handling simple natural commands and are not equipped to manage long-term tasks.

\subsubsection{Reinforcement Learning\label{sec:evolution:intellgent:rl}} 
The development of environments like World of Bits (WoB) \cite{wob} (2017) enabled the training of web-based agents using reinforcement learning (RL). Workflow-guided exploration \cite{reinforcementlearningonwebinterfaces} (2018) improved RL efficiency and task performance. DQT~\cite{lan2024deeply} (2024) applied deep reinforcement learning to automate Android GUI testing by preserving widget structures and semantics, while AndroidEnv~\cite{toyama2021androidenv} (2021) offered realistic simulations for agent training on Android. WebShop~\cite{yao2022webshop} (2022) illustrated the potential for large-scale web interaction, underscoring the growing sophistication of RL-driven GUI automation.

While these machine learning-based approaches were more adaptable than earlier rule-based systems \cite{zhang2019deep, martins2020using}, they still struggled to generalize across diverse, unforeseen tasks. Their dependence on predefined workflows and limited adaptability required retraining or customization for new environments, and natural language control was still limited.

\subsection{The Advent of LLM-Brained GUI Agents\label{sec:evolution:agent}}

The introduction of LLMs, particularly multimodal models like GPT-4o~\cite{hurst2024gpt} (2023), has radically transformed GUI automation by allowing intuitive interactions through natural language. Unlike previous approaches that required integration of separate modules, LLMs provide an end-to-end solution for GUI automation, offering advanced capabilities in natural language understanding, visual recognition, and reasoning.

LLMs present several unique advantages for GUI agents, including natural language understanding, multimodal processing, planning, and generalization. These features make LLMs and GUI agents a powerful combination. While there were earlier explorations, 2023 marked a pivotal year for LLM-powered GUI agents, with significant developments across various platforms such as web, mobile, and desktop applications.

\subsubsection{Web Domain\label{sec:evolution:agent:web}} 
The initial application of LLMs in GUI automation was within the web domain, with early studies establishing benchmark datasets and environments~\cite{yao2022webshop, wob}. A key milestone was WebAgent~\cite{gur2023real} (2023), which, alongside WebGUM~\cite{furuta2023multimodal} (2023), pioneered real-world web navigation using LLMs. These advancements paved the way for further developments~\cite{ma2023laser, zheng2024gpt4visiongeneralistwebagent, deng2024multi}, utilizing more specialized LLMs to enhance web-based interactions.

\subsubsection{Mobile Devices\label{sec:evolution:agent:mobile}} 
The integration of LLMs into mobile devices began with AutoDroid~\cite{wen2024autodroid} (2023), which combined LLMs with domain-specific knowledge for smartphone automation. Additional contributions like MM-Navigator~\cite{yan2023gpt} (2023), AppAgent~\cite{zhang2023appagentmultimodalagentssmartphone} (2023), and Mobile-Agent~\cite{wang2024mobileagentautonomousmultimodalmobile} (2023) enabled refined control over smartphone applications. Research has continued to improve accuracy for mobile GUI automation through model fine-tuning~\cite{nong2024mobileflowmultimodalllmmobile, zhang2024android} (2024).

\subsubsection{Computer Systems\label{sec:evolution:agent:computer}} 
For desktop applications, UFO~\cite{zhang2024ufouifocusedagentwindows} (2024) was one of the first systems to leverage GPT-4 with visual capabilities to fulfill user commands in Windows environments. Cradle~\cite{tan2024cradleempoweringfoundationagents} (2024) extended these capabilities to software applications and games, while Wu \etal~\cite{wu2024oscopilotgeneralistcomputeragents} (2024) provided interaction across diverse desktop applications, including web browsers, code terminals, and multimedia tools.

\subsubsection{Industry Models\label{sec:evolution:agent:model}} 
In industry, the \texttt{Claude 3.5 Sonnet} model~\cite{anthropic2024} (2024) introduced a ``computer use'' feature capable of interacting with desktop environments through UI operations \cite{hu2024dawnguiagentpreliminary}. This signifies the growing recognition of LLM-powered GUI agents as a valuable application in industry, with stakeholders increasingly investing in this technology.

OpenAI quickly followed up by releasing \texttt{Operator} \cite{cua2025} in 2025, a Computer-Using Agent (CUA) similar to Claude, achieving state-of-the-art performance across various benchmarks. This development underscores the industry's recognition of the value of GUI agents and its growing investment in the field. As interest continues to surge, GUI agent research and development are expected to become increasingly competitive, marking the beginning of a rapidly evolving landscape.

Undoubtedly, LLMs have introduced new paradigms and increased the intelligence of GUI agents in ways that were previously unattainable. As the field continues to evolve, we anticipate a wave of commercialization, leading to transformative changes in user interaction with GUI applications.

\subsection{GUI Agent vs. API-Based Agent\label{sec:evolution:vs}}
\begin{figure}[t]
    \centering
    \includegraphics[width=\columnwidth]{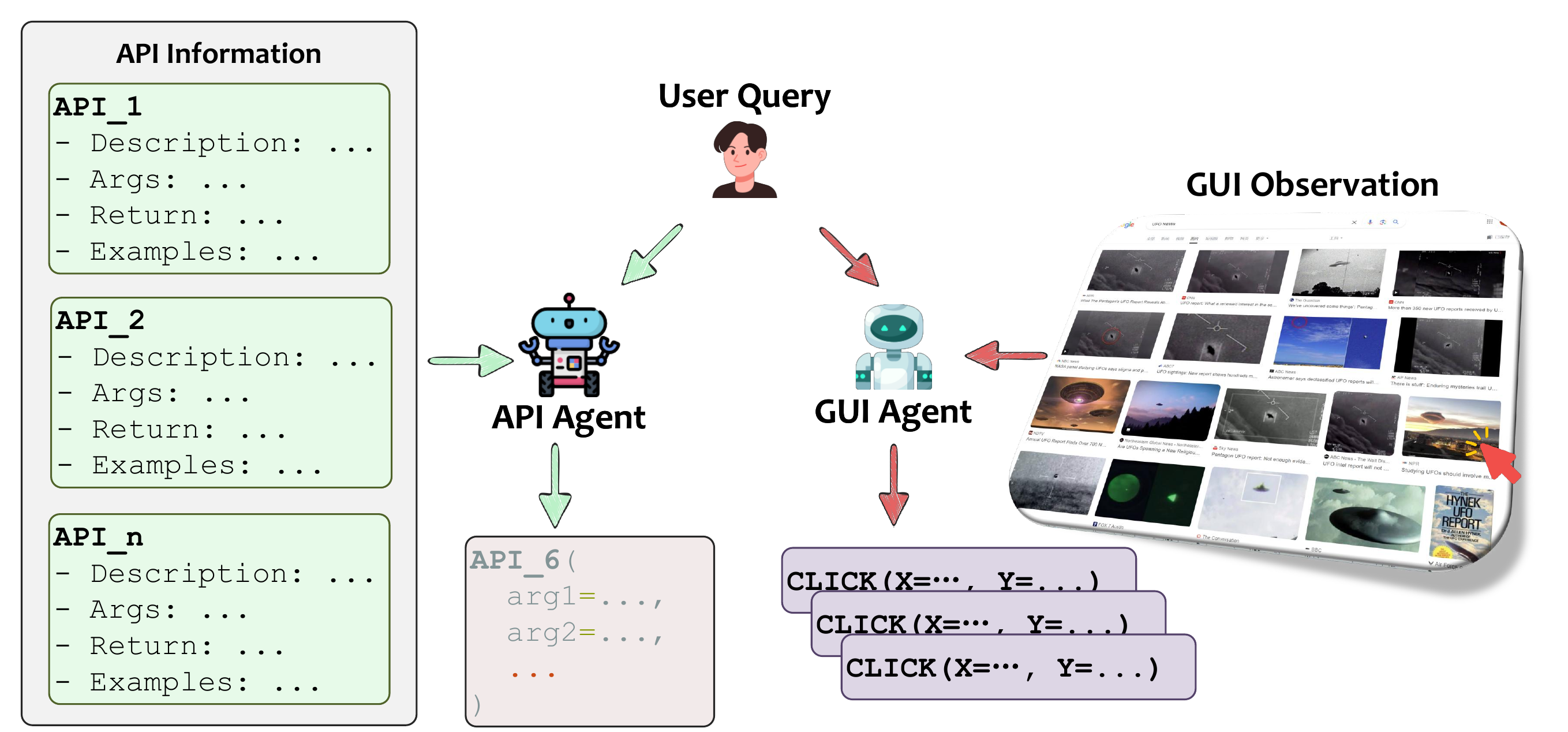}
    \caption{The comparison between API agent vs. GUI agent.}
    \label{fig:api_gui}
    % \vspace{-2em}
\end{figure}
In the field of LLM-powered agents operating within digital environments, the action space can be broadly categorized into two types: 
\begin{enumerate}
    \item \textbf{GUI Agents}, which primarily rely on GUI operations (\eg clicks, keystrokes) to complete tasks.  
    \item \textbf{API-Based Agents}, which utilize system or application-native APIs to fulfill objectives.
\end{enumerate}
We show the principle of both agent types in Figure~\ref{fig:api_gui}. Each type has distinct advantages, and a deeper understanding of these approaches is critical for designing effective agents. 

GUI operations provide a \textbf{universal control interface} that can operate across diverse applications using the same action primitives. This makes GUI agents highly generalizable, as they can interact with a wide range of software environments without requiring application-specific adaptations. However, GUI-based interactions are inherently more complex; even simple tasks may require multiple sequential steps, which can increase both the decision-making cost for the agent and the computational resources required for long-term, multi-step workflows. Another key aspect is the transparency of actions in GUI agents. Since GUI agents interact with applications in the same way a human would, by clicking, typing, and navigating through the interface, their actions are inherently more observable and interpretable to users. This transparency fosters better trust and comprehension in agent-computer interactions.

In contrast, API-based agents offer a more \textbf{efficient and direct approach} to task completion. By leveraging native APIs, tasks can often be fulfilled with a single, precise call, significantly reducing execution time and complexity. However, these native APIs are often private or restricted to specific applications, limiting accessibility and generalizability. This makes API-based agents less versatile in scenarios where API access is unavailable or insufficient. In addition, API-based agents operate behind the scenes, executing tasks through direct system calls, which, while often more efficient and reliable, can make their operations less visible and harder to debug for end users.

The most effective digital agents are likely to operate in a \textbf{hybrid manner}, combining the strengths of both approaches. Such agents can utilize GUI operations to achieve broad compatibility across software while exploiting native APIs where available to maximize efficiency and effectiveness. These hybrid agents strike a balance between generalization and task optimization, making them a \textbf{critical focus area in this survey}. For a more comprehensive comparison between GUI agents and API agents, please refer to \cite{zhang2025apiagentsvsgui}.
\section{LLM-Brained GUI Agents: Foundations and Design\label{sec:agent_foundation}}
In essence, LLM-brained GUI agents are designed to process user instructions or requests given in natural language, interpret the current state of the GUI through screenshots or UI element trees, and execute actions that simulate human interaction across various software interfaces \cite{zhang2024ufouifocusedagentwindows}. These agents harness the sophisticated natural language understanding, reasoning, and generative capabilities of LLMs to accurately comprehend user intent, assess the GUI context, and autonomously engage with applications across diverse environments, thereby enabling the completion of complex, multi-step tasks. This integration allows them to seamlessly interpret and respond to user requests, bringing adaptability and intelligence to GUI automation.

As a specialized type of LLM agent, most current GUI agents adopt a similar foundational framework, integrating core components such as planning, memory, tool usage, and advanced enhancements like multi-agent collaboration, among others \cite{wang2024survey}. However, each component must be tailored to meet the specific objectives of GUI agents to ensure adaptability and functionality across various application environments. 

In the following sections, we provide an in-depth overview of each component, offering a practical guide and tutorial on building an LLM-powered GUI agent from the ground up. This comprehensive breakdown serves as a cookbook for creating effective and intelligent GUI automation systems that leverage the capabilities of LLMs.

\subsection{Architecture and Workflow In a Nutshell\label{sec:agent_foundation:architecture}}
\begin{figure*}[t]
    \centering
    \includegraphics[width=\textwidth]{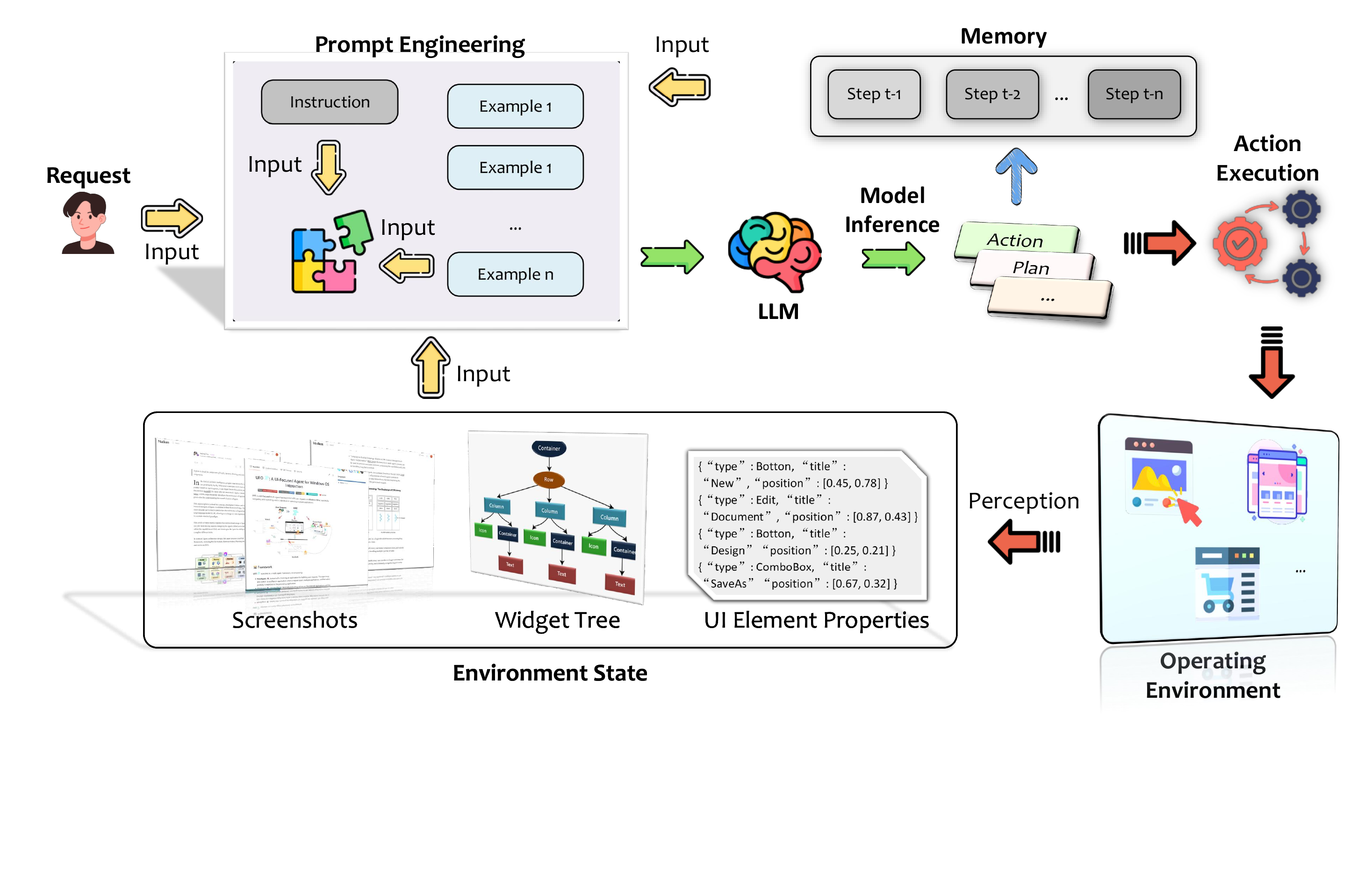}
    \vspace{-9em}
    \caption{An overview of the architecture and workflow of a basic LLM-powered GUI agent.}
    \label{fig:framework}
    % \vspace{-2em}
\end{figure*}
In Figure~\ref{fig:framework}, we present the architecture of an LLM-brained GUI agent, showcasing the sequence of operations from user input to task completion. The architecture comprises several integrated components, each contributing to the agent's ability to interpret and execute tasks based on user-provided natural language instructions. Upon receiving a user request, the agent follows a systematic workflow that includes environment perception, prompt engineering, model inference, action execution, and continuous memory utilization until the task is fully completed.

In general, it consists of the following components:
\begin{enumerate}
    \item \textbf{Operating Environment:} The environment defines the operational context for the agent, encompassing platforms such as mobile devices, web browsers, and desktop operating systems like Windows. To interact meaningfully, the agent perceives the environment's current state through screenshots, widget trees, or other methods of capturing UI structure \cite{memon2003gui}. It continuously monitors feedback on each action's impact, adjusting its strategy in real time to ensure effective task progression.
    \item \textbf{Prompt Engineering:} Following environment perception, the agent constructs a detailed prompt to guide the LLM's inference \cite{wang2023review}. This prompt incorporates user instructions, processed visual data (\eg screenshots), UI element layouts, properties, and any additional context relevant to the task. This structured input maximizes the LLM's ability to generate coherent, context-aware responses aligned with the current GUI state.
    \item \textbf{Model Inference:} The constructed prompt is passed to a LLM, the agent's inference core, which produces a sequence of plans,  actions and insights required to fulfill the user's request. This model may be a general-purpose LLM or a specialized model fine-tuned with GUI-specific data, enabling a more nuanced understanding of GUI interactions, user flows, and task requirements.
    \item \textbf{Actions Execution:} Based on the model's inference results, the agent identifies specific actions (such as mouse clicks, keyboard inputs, touchscreen gestures, or API calls) required for task execution \cite{shen2024llm}. An executor within the agent translates these high-level instructions into actionable commands that impact the GUI directly, effectively simulating human-like interactions across diverse applications and devices.
    \item \textbf{Memory:} For multi-step tasks, the agent maintains an internal memory to track prior actions, task progress, and environment states \cite{zhang2024survey}. This memory ensures coherence throughout complex workflows, as the agent can reference previous steps and adapt its actions accordingly. An external memory module may also be incorporated to enable continuous learning, access external knowledge, and enhance adaptation to new environments or requirements.
\end{enumerate}
By iteratively traversing these stages and assembling the foundational components, the LLM-powered GUI agent operates intelligently, seamlessly adapting across various software interfaces and bridging the gap between language-based instruction and concrete action. Each component is critical to the agent's robustness, responsiveness, and capability to handle complex tasks in dynamic environments. In the following subsections, we detail the design and core techniques underlying each of these components, providing a comprehensive guide for constructing LLM-powered GUI agents from the ground up.

\begin{table*}[ht]
\centering
\definecolor{HeaderBlue}{HTML}{D9EDF7} % A light blue color for header
\arrayrulecolor{black} % Table borders in black
\caption{Summary of platform-specific challenges, action spaces, and typical tasks for Web, Mobile, and Computer GUI environments.}
\label{tab:gui-platform-comparison}
\resizebox{\textwidth}{!}{ % Resize the entire figure to fit \textwidth
\begin{tabular}{p{2.7cm} p{4.7cm} p{4.7cm} p{4.7cm}}
% \rowcolor{black}
\hline
\textbf{Platform} 
& \textbf{Typical GUI Challenges} 
& \textbf{Action Space} 
& \textbf{Representative Tasks} \\
\toprule

\textbf{Mobile} 
& 
\begin{itemize}[leftmargin=*,noitemsep,topsep=0pt]
  \item Constrained screen real estate
  \item Heavy reliance on touch and gesture recognition~\cite{mitra2007gesture}
  \item App architectures (native vs.\ hybrid) 
  \item Accessibility frameworks (e.g., Android's Accessibility API, iOS VoiceOver)
  \item Platform-specific constraints (permissions, security, privacy)
\end{itemize}
& 
\begin{itemize}[leftmargin=*,noitemsep,topsep=0pt]
  \item Tap, swipe, pinch, and other touch gestures
  \item Virtual keyboard input
  \item In-app navigation (menus, tabs)
  \item Accessing hardware features (camera, GPS)
\end{itemize}
&
\begin{itemize}[leftmargin=*,noitemsep,topsep=0pt]
  \item App-based login and form filling
  \item Messaging, social media posting
  \item Location-based services and map interactions
  \item Handling push notifications and permission dialogs
\end{itemize} 
\\\hline

\textbf{Web} 
& 
\begin{itemize}[leftmargin=*,noitemsep,topsep=0pt]
  \item Dynamic and responsive layouts
  \item Asynchronous updates (AJAX, fetch APIs)
  \item HTML/DOM-based structures
  \item Cross-browser inconsistencies
\end{itemize}
& 
\begin{itemize}[leftmargin=*,noitemsep,topsep=0pt]
  \item Click, hover, scroll
  \item DOM-based form filling
  \item Link navigation and element inspection
  \item JavaScript event triggering
\end{itemize}
&
\begin{itemize}[leftmargin=*,noitemsep,topsep=0pt]
  \item Form completion (registrations, checkouts)
  \item Data extraction/web scraping
  \item Searching and filtering (e.g., e-commerce)
  \item Multi-step web navigation (redirects, pop-ups)
\end{itemize} 
\\\hline

\textbf{Computer} 
& 
\begin{itemize}[leftmargin=*,noitemsep,topsep=0pt]
  \item Full-fledged OS-level interfaces
  \item Multi-window operations and system-level shortcuts
  \item Automation APIs (\eg Windows UI Automation~\cite{oksanen2023test}) 
  \item Frequent software updates requiring adaptation
  \item Complex, multi-layered software suites
\end{itemize}
& 
\begin{itemize}[leftmargin=*,noitemsep,topsep=0pt]
  \item Mouse click, drag-and-drop
  \item Keyboard shortcuts and text input
  \item Menu navigation, toolbars
  \item Access to multiple application windows
\end{itemize}
&
\begin{itemize}[leftmargin=*,noitemsep,topsep=0pt]
  \item File management and system settings
  \item Productivity software usage (office suites, IDEs)
  \item Installing/uninstalling applications
  \item Coordinating multi-application workflows
\end{itemize}
\\
\bottomrule
\end{tabular}
}
\end{table*}

\subsection{Operating Environment\label{sec:agent_foundation:env}}
\begin{figure}[t]
    \centering
    \includegraphics[width=\columnwidth]{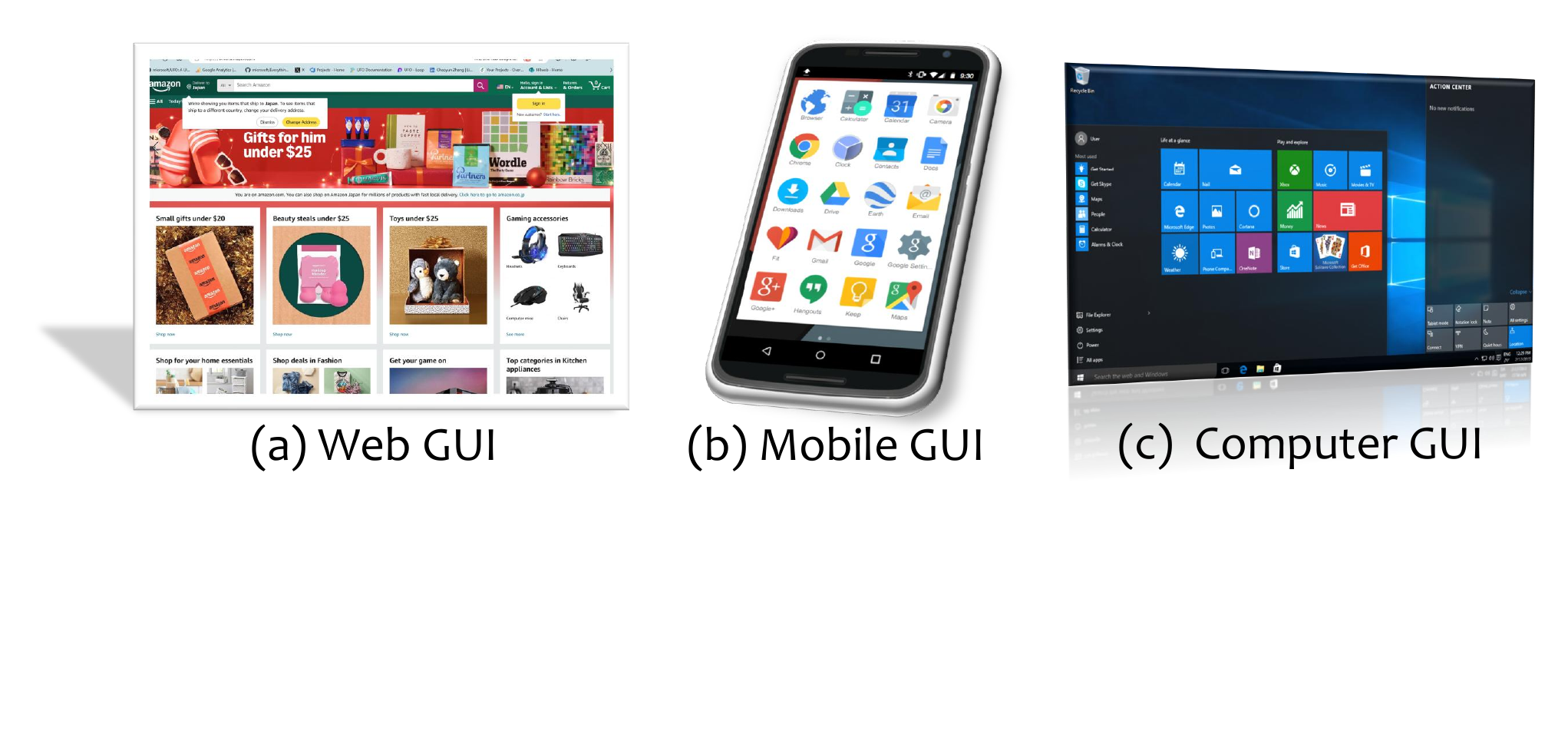}
    \vspace{-5.5em}
    \caption{Examples of GUIs from web, mobile and computer platforms.}
    \label{fig:gui}
    % \vspace{-2em}
\end{figure}

The operating environment for LLM-powered GUI agents encompasses various platforms, such as mobile, web, and desktop operating systems, where these agents can interact with graphical interfaces. Each platform has distinct characteristics that impact the way GUI agents perceive, interpret, and act within it. Examples of GUIs from each platform are shown in Figure~\ref{fig:gui}. This section details the nuances of each platform, the ways agents gather environmental information, and the challenges they face in adapting to diverse operating environments.
\begin{figure*}[t]
    \centering
    \includegraphics[width=\textwidth]{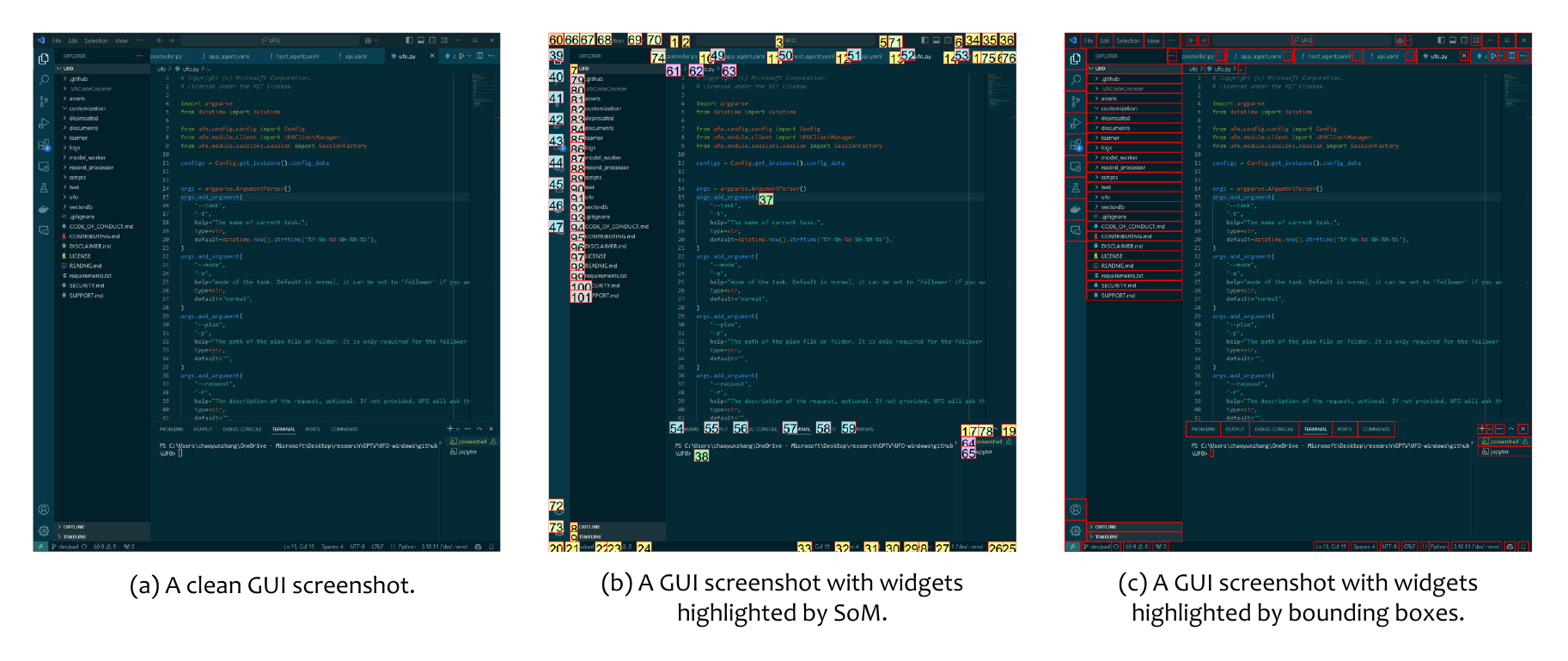}
    \vspace{-2em}
    \caption{Examples of different variants of VS Code GUI screenshots.}
    \label{fig:screenshots}
    % \vspace{-2em}
\end{figure*}

\subsubsection{Platform\label{sec:agent_foundation:env:platform}}
The operating environment for LLM-powered GUI agents encompasses various platforms, such as mobile, web, and desktop operating systems, where these agents can interact with graphical interfaces. Each platform has distinct characteristics that impact the way GUI agents perceive, interpret, and act within it. Examples of GUIs from each platform are shown in Figure~\ref{fig:gui}. This section details the nuances of each platform, the ways agents gather environmental information, and the challenges they face in adapting to diverse operating environments.
\begin{enumerate}
    \item \textbf{Mobile Platforms:} Mobile devices operate within constrained screen real estate, rely heavily on touch interactions \cite{hardy2008touch}, and offer varied app architectures (\eg native vs. hybrid apps). Mobile platforms often use accessibility frameworks, such as Android's Accessibility API\footnote{\label{Androidapi}\url{https://developer.android.com/reference/android/accessibilityservice/AccessibilityService}}\cite{lee2022systematic} and iOS's VoiceOver Accessibility Inspector\footnote{\url{https://developer.apple.com/documentation/accessibility/accessibility-inspector}}, to expose structured information about UI elements. However, GUI agents must handle additional complexities in mobile environments, such as gesture recognition \cite{mitra2007gesture}, app navigation \cite{jokinen2008user}, and platform-specific constraints (\eg security and privacy permissions) \cite{enck2011study, egele2011pios}.
    \item \textbf{Web Platforms:} Web applications provide a relatively standardized interface, typically accessible through Hypertext Markup Language (HTML) and Document Object Model (DOM) structures \cite{sierkowski2002achieving, fernandes2011web}. GUI agents can leverage HTML attributes, such as element ID, class, and tag, to identify interactive components. Web environments also present dynamic content, responsive layouts, and asynchronous updates (\eg AJAX requests) \cite{garrett2005ajax}, requiring agents to continuously assess the DOM and adapt their actions to changing interface elements.
    \item \textbf{Computer Platforms:} Computer OS platforms, such as Windows, offer full control over GUI interactions. Agents can utilize system-level automation APIs, such as Windows UI Automation\footnote{\label{uia}\url{https://learn.microsoft.com/en-us/dotnet/framework/ui-automation/ui-automation-overview}} \cite{oksanen2023test}, to obtain comprehensive UI element data, including type, label, position, and bounding box. These platforms often support a broader set of interaction types, mouse, keyboard, and complex multi-window operations. These enable GUI agents to execute intricate workflows. However, these systems also require sophisticated adaptation for diverse applications, ranging from simple UIs to complex, multi-layered software suites.
\end{enumerate}
In summary, the diversity of platforms, spanning mobile, web, and desktop environments, enable GUI agents to deliver broad automation capabilities, making them a generalized solution adaptable across a unified framework. However, each platform presents unique characteristics and constraints at both the system and application levels, necessitating a tailored approach for effective integration. By considering these platform-specific features, GUI agents can be optimized to address the distinctive requirements of each environment, thus enhancing their adaptability and reliability in varied automation scenarios.

\subsubsection{Environment State Perception\label{sec:agent_foundation:env:state}}

\begin{table*}[!h]
\small
\centering
\caption{Key toolkits for collecting GUI environment data.}
\label{table:key-toolkits}
\resizebox{\textwidth}{!}{ % Resize the entire figure to fit \textwidth
\begin{tabular}{p{2cm}|p{1.5cm}|p{3cm}|p{4cm}|p{4cm}|p{6cm}}
\hline
\multicolumn{1}{c|}{\textbf{Tool}} & \multicolumn{1}{c|}{\textbf{Platform}} & \multicolumn{1}{c|}{\textbf{Environment}} & \multicolumn{1}{c|}{\textbf{Accessible Information}}                                                           & \multicolumn{1}{c|}{\textbf{Highlight}} & \multicolumn{1}{c}{\textbf{Link}}\\ \hline

% Web Tools
Selenium & Web & Browser (Cross-platform) & DOM elements, HTML structure, CSS properties & Extensive browser support and automation capabilities &\hypertarget{url:sele}{\url{https://www.selenium.dev/}} \\ \hline

Puppeteer & Web & Browser (Chrome, Firefox) & DOM elements, HTML/CSS, network requests & Headless browser automation with rich API & \hypertarget{url:pup}{\url{https://pptr.dev/}} \\ \hline

Playwright & Web & Browser (Cross-platform) & DOM elements, HTML/CSS, network interactions & Multi-browser support with automation and testing capabilities & \url{https://playwright.dev/} \\ \hline

TestCafe & Web & Browser (Cross-platform) & DOM elements, HTML structure, CSS properties & Easy setup with JavaScript/TypeScript support & \url{https://testcafe.io/} \\ \hline

BeautifulSoup & Web & HTML Parsing & HTML content, DOM elements & Python library for parsing HTML and XML documents & \url{https://www.crummy.com/software/BeautifulSoup/} \\ \hline

Protractor & Web & Browser (Angular) & DOM elements, Angular-specific attributes & Designed for Angular applications, integrates with Selenium & \url{https://www.protractortest.org/} \\ \hline

WebDriverIO & Web & Browser (Cross-platform) & DOM elements, HTML/CSS, network interactions & Highly extensible with a vast plugin ecosystem & \url{https://webdriver.io/} \\ \hline

Ghost Inspector & Web & Browser (Cross-platform) & DOM elements, screenshots, test scripts & Cloud-based automated browser testing and monitoring & \url{https://ghostinspector.com/} \\ \hline

Cypress & Web & Browser (Cross-platform) & DOM elements, HTML/CSS, network requests & Real-time reloads and interactive debugging & \url{https://www.cypress.io/} \\ \hline\hline

% Mobile Tools
UIAutomator & Mobile & Android & UI hierarchy, widget properties, screen content & Native Android UI testing framework & \hyperlink{uia}{\url{https://developer.android.com/training/testing/ui-automator}} \\ \hline

Espresso & Mobile & Android & UI components, view hierarchy, widget properties & Google's native Android UI testing framework & \url{https://developer.android.com/training/testing/espresso} \\ \hline

Android View Hierarchy & Mobile & Android & UI hierarchy, widget properties, layout information & View hierarchy accessible via developer tools & \url{https://developer.android.com/studio/debug/layout-inspector} \\ \hline

iOS Accessibility Inspector & Mobile & iOS & Accessibility tree, UI elements, properties & Tool for inspecting iOS app UI elements & \url{https://developer.apple.com/documentation/accessibility/accessibility-inspector} \\ \hline

XCUITest & Mobile & iOS & UI elements, accessibility properties, view hierarchy & Apple's iOS UI testing framework & \hyperlink{url:xcui}{\url{https://developer.apple.com/documentation/xctest/user_interface_tests}} \\ \hline

Flutter Driver & Mobile & Flutter apps & Widget tree, properties, interactions & Automation for Flutter applications & \url{https://flutter.dev/docs/testing} \\ \hline

Android's MediaProjection API & Mobile & Android & Screenshots, screen recording & Capturing device screen content programmatically & \url{https://developer.android.com/reference/android/media/projection/MediaProjection} \\ \hline\hline

% Computer Tools
Windows UI Automation & Computer & Windows & Control properties, widget trees, accessibility tree & Native Windows support with OS integration & \hypertarget{url:uia}{\url{https://docs.microsoft.com/windows/win32/winauto/entry-uiauto-win32}}\\ \hline

Sikuli & Computer & Windows, macOS, Linux & Screenshots (image recognition), UI elements & Image-based automation using computer vision & \url{http://sikulix.com/} \\ \hline

AutoIt & Computer & Windows & Window titles, control properties, coordinates & Scripting language for Windows GUI automation & \url{https://www.autoitscript.com/site/autoit/} \\ \hline

Inspect.exe & Computer & Windows & UI elements, control properties, accessibility tree & Tool for inspecting Windows UI elements & \url{https://docs.microsoft.com/windows/win32/winauto/inspect-objects} \\ \hline

macOS Accessibility API & Computer & macOS & Accessibility tree, UI elements, control properties & macOS support for accessibility and UI automation & \url{https://developer.apple.com/accessibility/} \\ \hline

Pywinauto & Computer & Windows & Control properties, UI hierarchy, window information & Python-based Windows GUI automation & \url{https://pywinauto.readthedocs.io/} \\ \hline

Electron Inspector & Computer & Electron apps & DOM elements, HTML/CSS, JavaScript state & Tool for Electron applications & \url{https://www.electronjs.org/docs/latest/tutorial/automated-testing} \\ \hline

Windows Snipping Tool & Computer & Windows & Screenshots & Tool for capturing screenshots in Windows & \url{https://www.microsoft.com/en-us/windows/tips/snipping-tool} \\ \hline

macOS Screenshot Utility & Computer & macOS & Screenshots, screen recording & Tool for capturing screenshots and recording screen & \url{https://support.apple.com/guide/mac-help/take-a-screenshot-or\%2Dscreen-recording\%2Dmh26782/mac} \\ \hline\hline

% Cross-Platform Tools
AccessKit & Cross-Platform & Various OS & Accessibility tree, control properties, roles & Standardized APIs across platforms & \url{https://github.com/AccessKit/accesskit} \\ \hline

Appium & Cross-Platform & Android, iOS, Windows, macOS & UI elements, accessibility properties, gestures & Mobile automation framework & \hypertarget{url:appium}{\url{https://appium.io/}} \\ \hline

Robot Framework & Cross-Platform & Web, Mobile, Desktop & UI elements, DOM, screenshots & Extensible with various libraries & \url{https://robotframework.org/} \\ \hline

Cucumber & Cross-Platform & Web, Mobile, Desktop & Step definitions, UI interactions & BDD framework supporting automation tools & \url{https://cucumber.io/} \\ \hline

TestComplete & Cross-Platform & Web, Mobile, Desktop & UI elements, DOM, control properties & Tool with extensive feature set & \url{https://smartbear.com/product/testcomplete/overview/} \\ \hline

Katalon Studio & Cross-Platform & Web, Mobile, Desktop & UI elements, DOM, screenshots & All-in-one automation solution & \url{https://www.katalon.com/} \\ \hline

Ranorex & Cross-Platform & Web, Mobile, Desktop & UI elements, DOM, control properties & Tool with strong reporting features & \url{https://www.ranorex.com/} \\ \hline

Applitools & Cross-Platform & Web, Mobile, Desktop & Screenshots, visual checkpoints, DOM elements & AI-powered visual testing & \url{https://applitools.com/} \\ \hline

\end{tabular}
}
\end{table*}

\begin{figure}[t]
    \centering
    \includegraphics[width=\columnwidth]{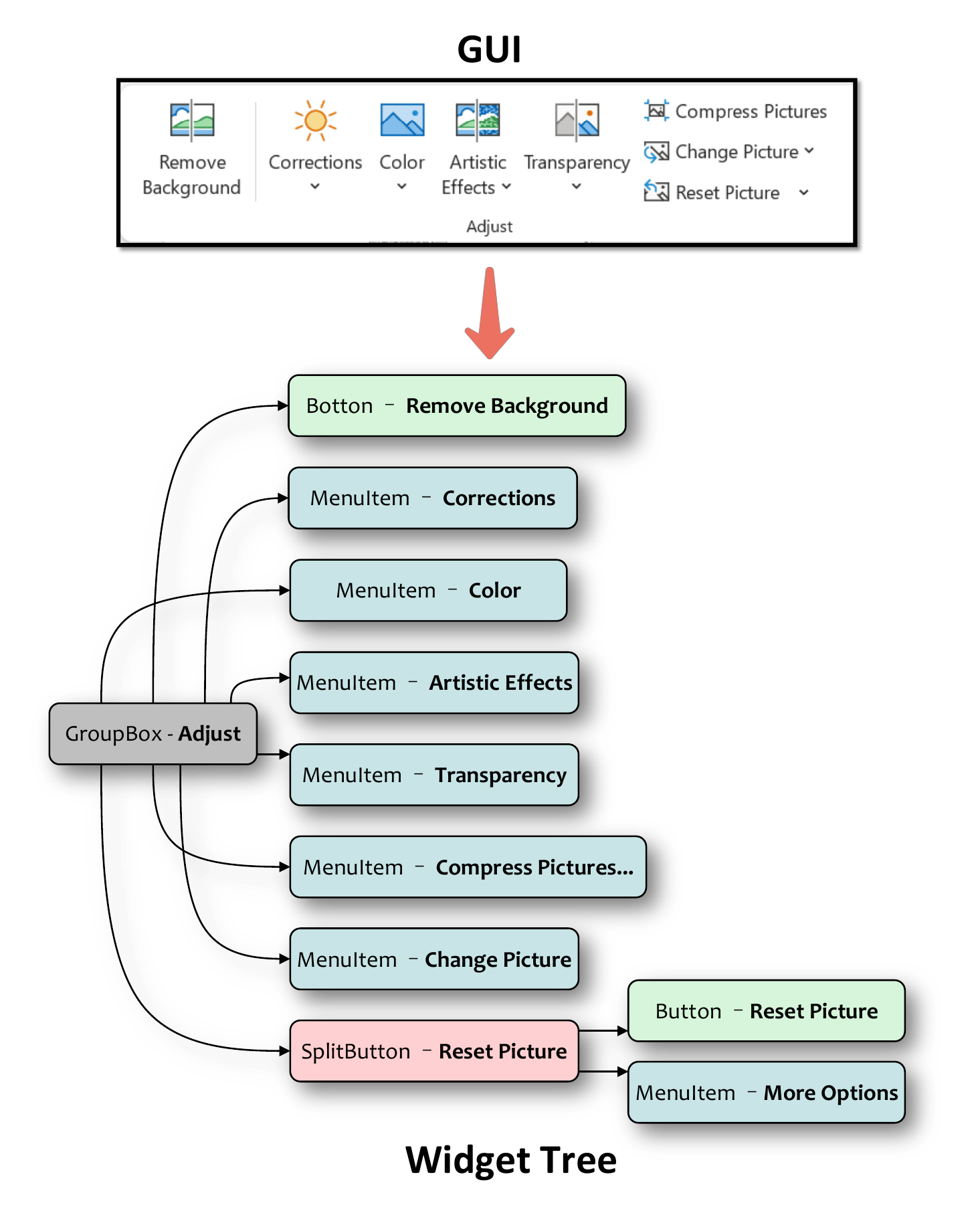}
    \vspace{-2em}
    \caption{An example of a GUI and its widget tree.}
    \label{fig:widget_tree}
    % \vspace{-2em}
\end{figure}

\begin{figure}[t]
    \centering
    \includegraphics[width=\columnwidth]{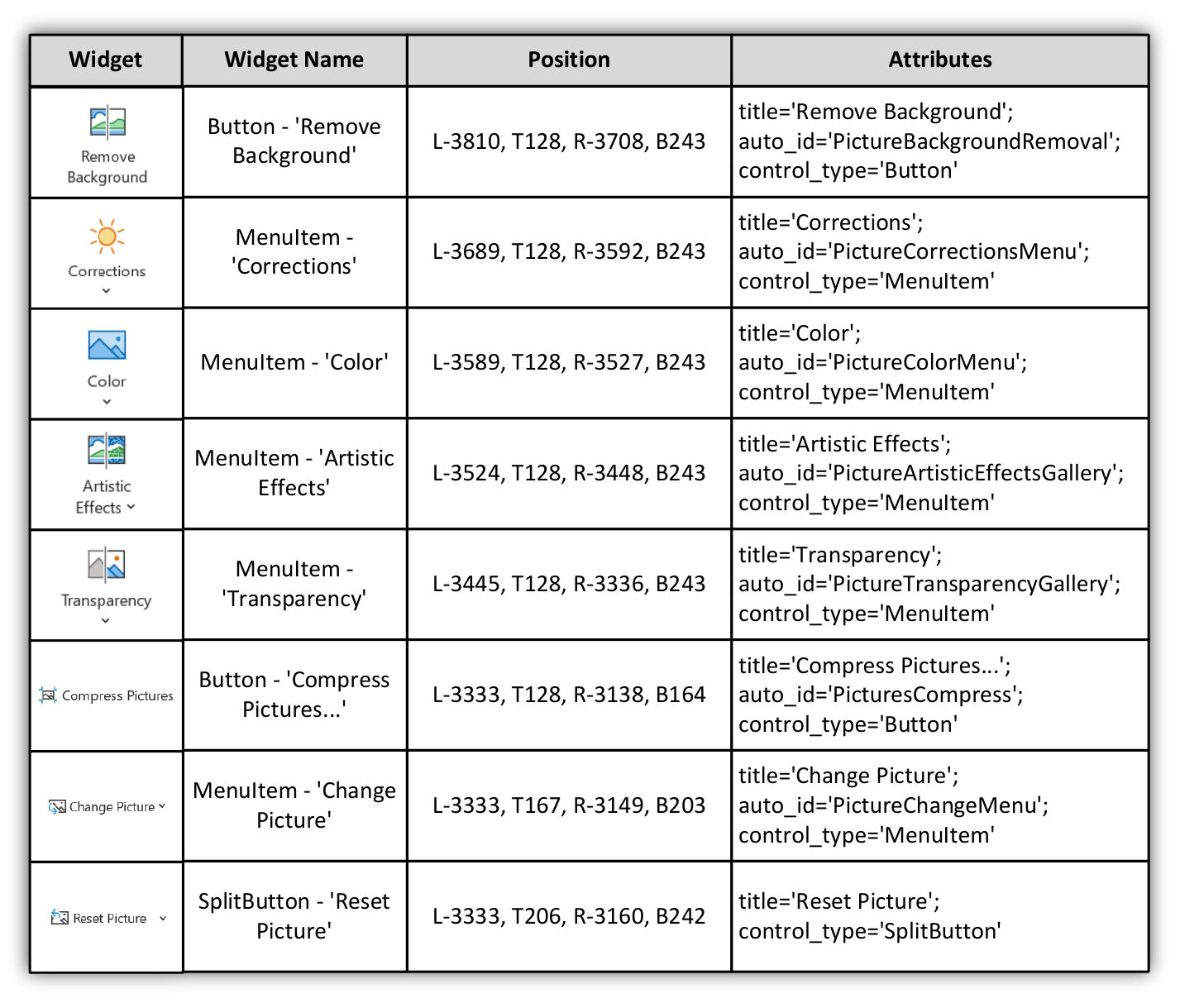}
    \vspace{-2em}
    \caption{Examples of UI element properties in the PowerPoint application for GUI Agent interaction.}
    \label{fig:widget_pro}
    % \vspace{-2em}
\end{figure}

\begin{figure*}[t]
    \centering
    \includegraphics[width=1\textwidth]{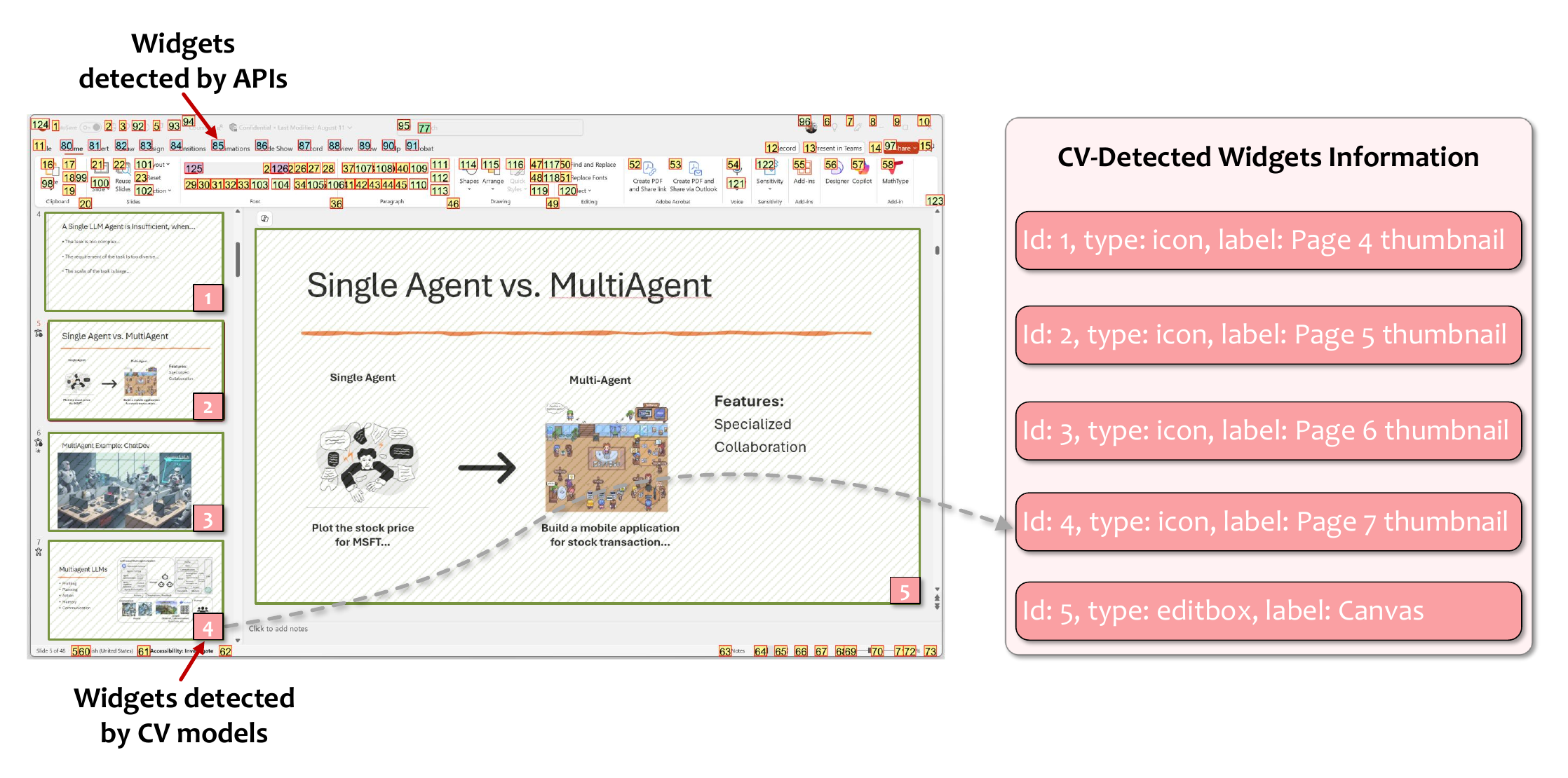}
    \vspace{-2em}
    \caption{An example illustrating the use of a CV approach to parse a PowerPoint GUI and detect non-standard widgets, inferring their types and labels.}
    \label{fig:cv}
    % \vspace{-2em}
\end{figure*}

Accurately perceiving the current state of the environment is essential for LLM-powered GUI agents, as it directly informs their decision-making and action-planning processes. This perception is enabled by gathering a combination of structured data, such as widget trees, and unstructured data, like screenshots, to capture a complete representation of the interface and its components. In Table~\ref{table:key-toolkits}, we outline key toolkits available for collecting GUI environment data across various platforms, and below we discuss their roles in detail:
\begin{enumerate}
    \item \textbf{GUI Screenshots:} Screenshots provide a visual snapshot of the application, capturing the entire state of the GUI at a given moment. They offer agents a reference for layout, design, and visual content, which is crucial when structural details about UI elements are either limited or unavailable. Visual elements like icons, images, and other graphical cues that may hold important context can be analyzed directly from screenshots. Many platforms have built-in tools to capture screenshots (\eg Windows Snipping Tool\footnote{\url{https://support.microsoft.com/en-us/windows/use-snipping-tool-to-capture\%2Dscreenshots\%2D00246869\%2D1843\%2D655f\%2Df220\%2D97299b865f6b}}, macOS Screenshot Utility\footnote{\url{https://support.apple.com/guide/mac-help/take-a-screenshot-mh26782/mac}}, and Android's MediaProjection API\footnote{\url{https://developer.android.com/reference/android/media/projection/MediaProjection}}), and screenshots can be enhanced with additional annotations, such as Set-of-Mark (SoM) highlights \cite{yang2023set} or bounding boxes \cite{wu2023widget} around key UI components, to streamline agent decisions. Figure~\ref{fig:screenshots} illustrates various screenshots of the VS Code GUI, including a clean version, as well as ones with SoM and bounding boxes that highlight actionable components, helping the agent focus on the most critical areas of the interface. 
    \item \textbf{Widget Trees:} Widget trees present a hierarchical view of interface elements, providing structured data about the layout and relationships between components \cite{gamma1995design}. We show an example of a GUI and its widget tree in Figure~\ref{fig:widget_tree}. By accessing the widget tree, agents can identify attributes such as element type, label, role, and relationships within the interface, all of which are essential for contextual understanding. Tools like Windows UI Automation and macOS's Accessibility API\footnote{\url{https://developer.apple.com/library/archive/documentation/Accessibility/Conceptual/AccessibilityMacOSX/}} provide structured views for desktop applications, while Android's Accessibility API and HTML DOM structures serve mobile and web platforms, respectively. This hierarchical data is indispensable for agents to map out logical interactions and make informed choices based on the UI structure. 
    \item \textbf{UI Element Properties:} Each UI element in the interface contains specific properties, such as control type, label text, position, and bounding box dimensions, that help agents target the appropriate components. These properties are instrumental for agents to make decisions about spatial relationships (\eg adjacent elements) and functional purposes (\eg distinguishing between buttons and text fields). For instance, web applications reveal properties like DOM attributes (id, class, name) and CSS styles that provide context and control information. These attributes assist agents in pinpointing precise elements for interaction, enhancing their ability to navigate and operate within diverse UI environments. Figure~\ref{fig:widget_pro} illustrates examples of selected UI element properties extracted by the Windows UI Automation API, which support GUI agents in decision-making.
    \item \textbf{Complementary CV Approaches:} When structured information is incomplete or unavailable, computer vision techniques can provide additional insights \cite{wang2024comprehensivesurveysmalllanguage}. For instance, OCR allows agents to extract text content directly from screenshots, facilitating the reading of labels, error messages, and instructions \cite{qian2022accelerating}. Furthermore, advanced object detection \cite{chen2020object} models like SAM (Segment Anything Model) \cite{kirillov2023segment}, DINO \cite{liu2023grounding} and OmniParser \cite{lu2024omniparserpurevisionbased} can identify and classify UI components in various layouts, supporting the agent in dynamic environments where UI elements may frequently change. These vision-based methods ensure robustness, enabling agents to function effectively even in settings where standard UI APIs are insufficient. We illustrate an example of this complementary information in Figure~\ref{fig:cv} and further detail these advanced computer vision approaches in Section~\ref{sec:agent_foundation:enhancement:cv}.
\end{enumerate}
Together, these elements create a comprehensive, multimodal representation of the GUI environment's current state, delivering both structured and visual data. By incorporating this information into prompt construction, agents are empowered to make well-informed, contextually aware decisions without missing critical environmental cues.

\subsubsection{Environment Feedback\label{sec:agent_foundation:env:feedback}}
\begin{figure*}[t]
    \centering
    \includegraphics[width=0.8\textwidth]{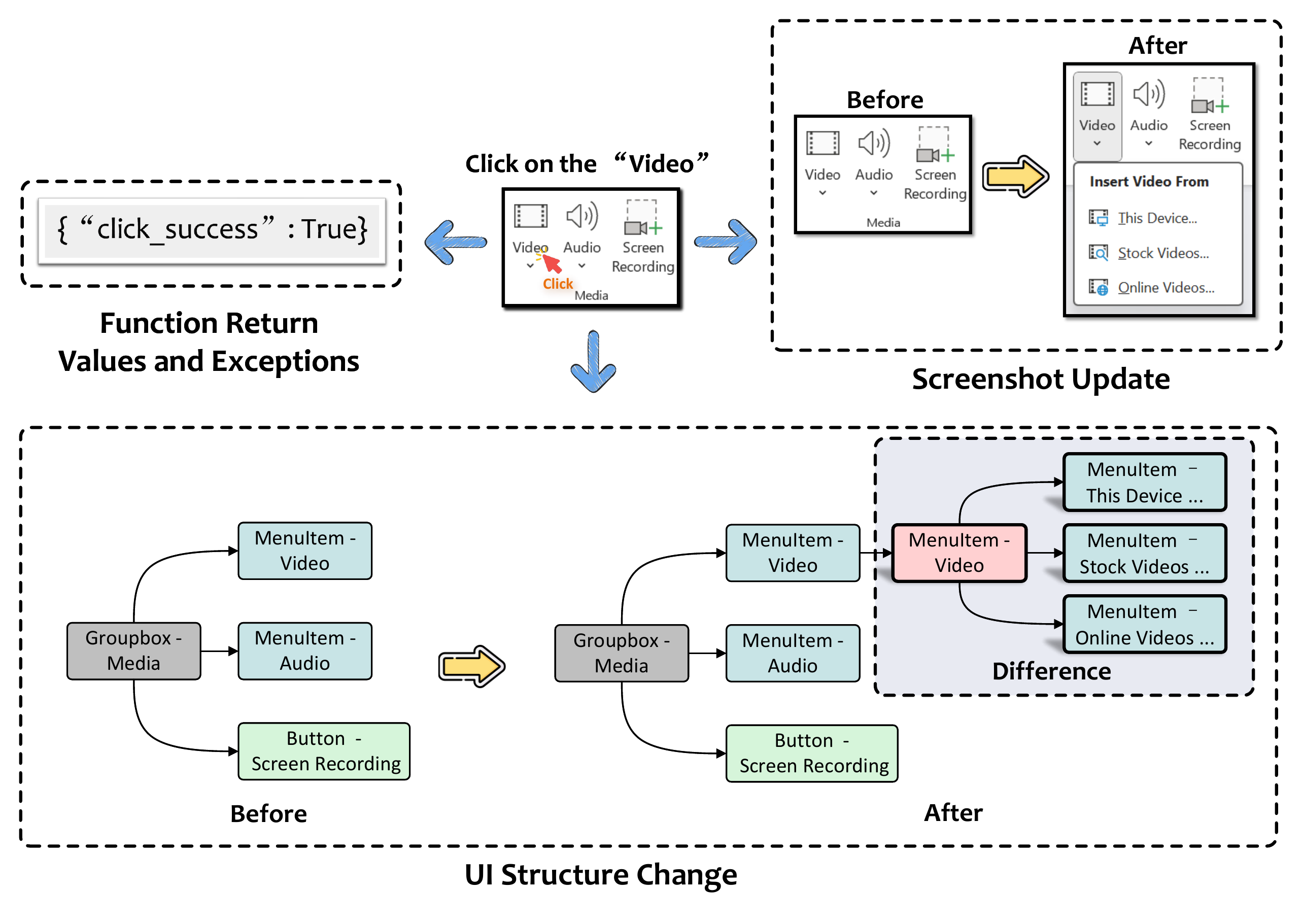}
    \vspace{-1em}
    \caption{Examples of various types of feedback obtained from a PowerPoint application environment.}
    \label{fig:feedback}
    % \vspace{-2em}
\end{figure*}
Effective feedback mechanisms are essential for GUI agents to assess the success of each action and make informed decisions for subsequent steps. Feedback can take several forms, depending on the platform and interaction type. Figure~\ref{fig:feedback} presents examples of various types of feedback obtained from the environment.
\begin{enumerate}
    \item \textbf{Screenshot Update:} By comparing before-and-after screenshots, agents can identify visual differences that signify state changes in the application. Screenshot analysis can reveal subtle variations in the interface, such as the appearance of a notification, visual cues, or confirmation messages, that may not be captured by structured data \cite{moran2018detecting}.
    \item \textbf{UI Structure Change:} After executing an action, agents can detect modifications in the widget tree structure, such as the appearance or disappearance of elements, updates to element properties, or hierarchical shifts \cite{ricos2023using}. These changes indicate successful interactions (\eg opening a dropdown or clicking a button) and help the agent determine the next steps based on the updated environment state.
    \item \textbf{Function Return Values and Exceptions:} Certain platforms offer direct feedback on action outcomes through function return values or system-generated exceptions \cite{du2024anytool}. For example, API responses or JavaScript return values can confirm action success on web platforms, while exceptions or error codes can signal failed interactions, guiding the agent to retry or select an alternative approach.
\end{enumerate}
These feedback provided by the environment is crucial for GUI agents to assess the outcomes of their previous actions. This real-time information enables agents to evaluate the effectiveness of their interventions and determine whether to adhere to their initial plans or pivot towards alternative strategies. Through this process of self-reflection, agents can adapt their decision-making, optimizing task execution and enhancing overall performance in dynamic and varied application environments.

\subsection{Prompt Engineering\label{sec:agent_foundation:prompt}}
\begin{figure*}[t]
    \centering
    \includegraphics[width=\textwidth]{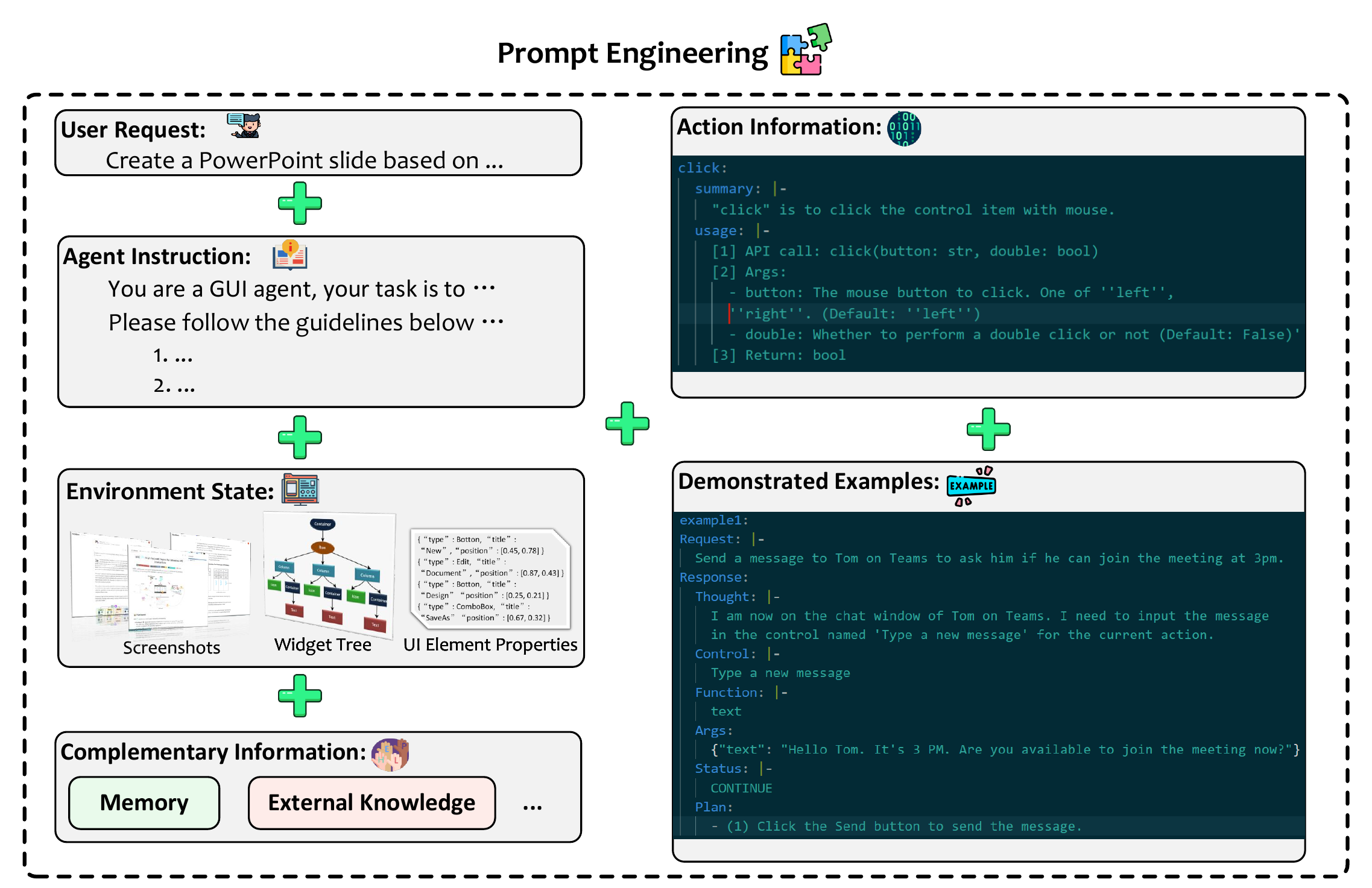}
    % \vspace{-2em}
    \caption{A basic example of prompt construction in a LLM-brained GUI agent.}
    \label{fig:prompt}
    % \vspace{-2em}
\end{figure*}
In the operation of LLM-powered GUI agents, effective prompt construction is a crucial step that encapsulates all necessary information for the agent to generate appropriate responses and execute tasks successfully \cite{wang2023review}. After gathering the relevant data from the environment, the agent formulates a comprehensive prompt that combines various components essential for inference by the LLM. Each component serves a specific purpose, and together they enable the agent to execute the user's request efficiently. Figure~\ref{fig:prompt} illustrates a basic example of prompt construction in an LLM-brained GUI agent. The key elements of the prompt are summarized as follows: 
\begin{enumerate}
    \item \textbf{User Request:} 
    This is the original task description provided by the user, outlining the objective and desired outcome. It serves as the foundation for the agent's actions and is critical for ensuring that the LLM understands the context and scope of the task.
    
    \item \textbf{Agent Instruction:} 
    This section provides guidance for the agent's operation, detailing its role, rules to follow, and specific objectives. Instructions clarify what inputs the agent will receive and outline the expected outputs from the LLM, establishing a framework for the inference process. The core agent instructions are usually embedded within the base system prompt of the LLM, with supplementary instructions dynamically injected or updated based on environmental feedback and contextual adaptation.
    
    \item \textbf{Environment States:} 
    The agent includes perceived GUI screenshots and UI information, as introduced in Section~\ref{sec:agent_foundation:env:state}. This multimodal data may consist of various versions of screenshots (\eg a clean version and a SoM annotated version) to ensure clarity and mitigate the risk of UI controls being obscured by annotations. This comprehensive representation of the environment is vital for accurate decision-making.
    
    \item \textbf{Action Documents:} 
    This component outlines the available actions the agent can take, detailing relevant documentation, function names, arguments, return values, and any other necessary parameters. Providing this information equips the LLM with the context needed to select and generate appropriate actions for the task at hand.
    
    \item \textbf{Demonstrated Examples:}  
    Including example input/output pairs is essential to activate the in-context learning \cite{dong2022survey} capability of the LLM. These examples help the model comprehend and generalize the task requirements, enhancing its performance in executing the GUI agent's responsibilities.
    
    \item \textbf{Complementary Information:} 
    Additional context that aids in planning and inference may also be included. This can consist of historical data retrieved from the agent's memory (as detailed in Section~\ref{sec:agent_foundation:memory}) and external knowledge sources, such as documents obtained through retrieval-augmented generation (RAG) methods \cite{lewis2020retrieval, gao2023retrieval}. This supplemental information can provide valuable insights that further refine the agent's decision-making processes.
\end{enumerate}

The construction of an effective prompt is foundational for the performance of LLM-powered GUI agents. By systematically incorporating aforementioned information, the agent ensures that the LLM is equipped with the necessary context and guidance to execute tasks accurately and efficiently. 

\subsection{Model Inference\label{sec:agent_foundation:inference}}
\begin{figure}[t]
    \centering
    \includegraphics[width=\columnwidth]{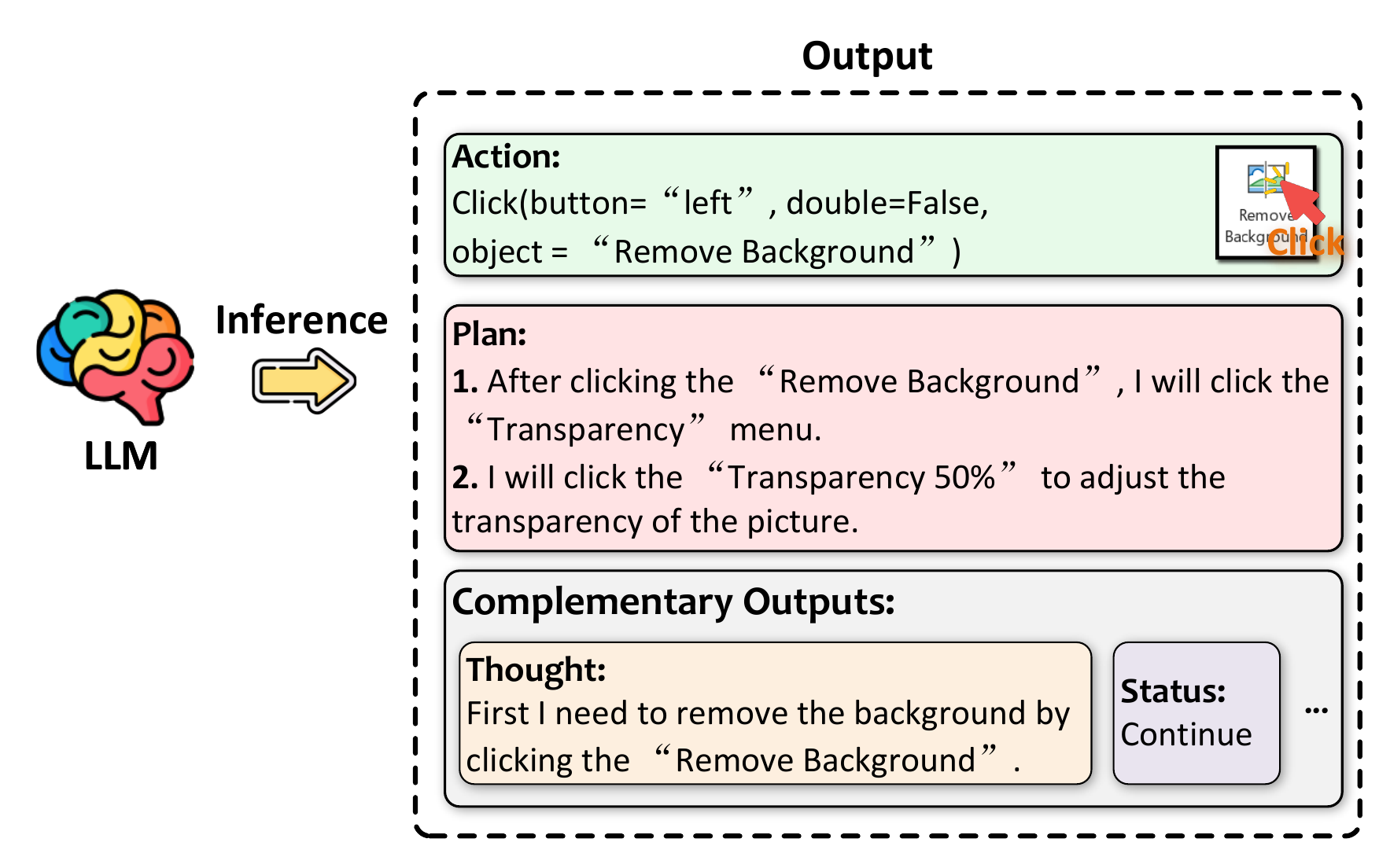}
    % \vspace{-2em}
    \caption{An example of the LLM's inference output in a GUI agent.}
    \label{fig:inference}
    % \vspace{-2em}
\end{figure}
The constructed prompt is submitted to the LLM for inference, where the LLM is tasked with generating both a plan and the specific actions required to execute the user's request. This inference process is critical as it dictates how effectively the GUI agent will perform in dynamic environments. It typically involves two main components: planning and action inference, as well as the generation of complementary outputs. Figure~\ref{fig:inference}  shows an example of the LLM's inference output.

\subsubsection{Planning\label{sec:agent_foundation:inference:plan}}
Successful execution of GUI tasks often necessitates a series of sequential actions, requiring the agent to engage in effective planning \cite{zhang2024dynamic, aghzal2025survey}. Analogous to human cognitive processes, thoughtful planning is essential to organize tasks, schedule actions, and ensure successful completion \cite{huang2024understanding, cho2024caap}. The LLM must initially conceptualize a long-term goal while simultaneously focusing on short-term actions to initiate progress toward that goal \cite{dagan2023dynamic}.

To effectively navigate the complexity of multi-step tasks, the agent should decompose the overarching task into manageable subtasks and establish a timeline for their execution \cite{khot2022decomposed}. Techniques such as CoT reasoning \cite{wei2022chain} can be employed, enabling the LLM to develop a structured plan that guides the execution of actions. This plan, which can be stored for reference during future inference steps, enhances the organization and focus of the agent's activities.

The granularity of planning may vary based on the nature of the task and the role of the agent \cite{huang2024understanding}. For complex tasks, a hierarchical approach that combines global planning (identifying broad subgoals) with local planning (defining detailed steps for those subgoals) can significantly improve the agent's ability to manage long-term objectives effectively \cite{chen2024can}.

\begin{table*}[t]
\caption{Overview of actions for GUI agents.}
\label{tab:actions}
\centering
\resizebox{\textwidth}{!}{ % Resize the entire figure to fit \textwidth
\begin{tabular}{l|l|p{1.5cm}|p{2.5cm}|l|l|p{2cm}}
\hline
\multicolumn{1}{c|}{\textbf{Action}} & \multicolumn{1}{c|}{\textbf{Category}} & \multicolumn{1}{c|}{\textbf{Original Executor}} & \multicolumn{1}{c|}{\textbf{Examples}}                                                           & \multicolumn{1}{c|}{\textbf{Platform}} & \multicolumn{1}{c|}{\textbf{Environment}} & \multicolumn{1}{c}{\textbf{Toolkit}} \\ \hline
Mouse actions                         & UI Operations                          & Mouse                                           & Click, scroll, hover, drag                                                                       & Computer                               & Windows                                   & UI Automation \ref{uia}, Pywinauto            \\ \hline
Mouse actions                         & UI Operations                          & Mouse                                           & Click, scroll, hover, drag                                                                       & Computer                               & macOS                                     & AppleScript \footnotemark[10], Automator \footnotemark[11]                \\ \hline
Mouse actions                         & UI Operations                          & Mouse                                           & Click, scroll, hover, drag                                                                       & Web                                    & Browser                                   &  \hyperlink{url:sele}{Selenium}, \hyperlink{url:pup}{Puppeteer}                  \\ \hline
Keyboard actions                      & UI Operations                          & Keyboard                                        & Typing, key presses, shortcuts                                                                   & Computer                               & Windows                                   & UI Automation \ref{uia}, Pywinauto            \\ \hline
Keyboard actions                      & UI Operations                          & Keyboard                                        & Typing, key presses, shortcuts                                                                   & Computer                               & macOS                                     & AppleScript \footnotemark[10], Automator \footnotemark[11]               \\ \hline
Keyboard actions                      & UI Operations                          & Keyboard                                        & Typing, key presses, shortcuts                                                                   & Web                                    & Browser                                   & \hyperlink{url:sele}{Selenium}, \hyperlink{url:pup}{Puppeteer}                   \\ \hline
Touch actions                         & UI Operations                          & Touchscreen                                     & Tap, swipe, pinch, zoom                                                                          & Mobile                                 & Android                                   &  \hyperlink{url:appium}{Appium}, \hyperlink{uia}{UIAutomator}                  \\ \hline
Touch actions                         & UI Operations                          & Touchscreen                                     & Tap, swipe, pinch, zoom                                                                          & Mobile                                 & iOS                                       & \hyperlink{url:appium}{Appium}, \hyperlink{url:xcui}{XCUITest}                      \\ \hline
Gesture actions                       & UI Operations                          & User hand                                       & Rotate, multi-finger gestures                                                                    & Mobile                                 & Android, iOS                              & \hyperlink{url:appium}{Appium} , GestureTools \footnotemark[12]                 \\ \hline
Voice commands                        & UI Operations                          & User voice                                      & Speech input, voice commands                                                                     & Mobile                                 & Android                                   & SpeechRecognizer  \footnotemark[13]                  \\ \hline
Voice commands                        & UI Operations                          & User voice                                      & Speech input, voice commands                                                                     & Mobile                                 & iOS                                       & SiriKit \footnotemark[14]                              \\ \hline
Clipboard operations                  & UI Operations                          & System clipboard                                & Copy, paste                                                                                      & Cross-platform                         & Cross-OS                                  & Pyperclip \footnotemark[15], Clipboard.js \footnotemark[16]              \\ \hline
Screen interactions                   & UI Operations                          & User                                            & Screen rotation, shake                                                                           & Mobile                                 & Android, iOS                              & Device sensors APIs \footnotemark[17]                  \\ \hline\hline
Shell Commands & Native API Calls & Command Line Interface & File manipulation, system operations, script execution & Computer & Unix/Linux, macOS & Bash, Terminal \\ \hline
Application APIs                      & Native API Calls                       & Application APIs                                & Send email, create document, fetch data                                                          & Computer                               & Windows                                   & Microsoft Office COM APIs \footnotemark[18]             \\ \hline
Application APIs                      & Native API Calls                       & Application APIs                                & Access calendar, send messages                                                                   & Mobile                                 & Android                                   & Android SDK APIs \footnotemark[19]                      \\ \hline
Application APIs                      & Native API Calls                       & Application APIs                                & Access calendar, send messages                                                                   & Mobile                                 & iOS                                       & iOS SDK APIs \footnotemark[20]                          \\ \hline
System APIs                           & Native API Calls                       & System APIs                                     & File operations, network requests                                                                & Computer                               & Windows                                   & Win32 API \footnotemark[21]                             \\ \hline
System APIs                           & Native API Calls                       & System APIs                                     & File operations, network requests                                                                & Computer                               & macOS                                     & Cocoa APIs \footnotemark[22]                            \\ \hline
Web APIs                              & Native API Calls                       & Web Services                                    & Fetch data, submit forms                                                                         & Web                                    & Browser                                   & Fetch API \footnotemark[23] , Axios \footnotemark[24]                      \\ \hline\hline
AI Models                             & AI Tools                               & AI Models                                       & Screen understanding,  summarization, image generation & Cross-platform                         & Cross-OS                                  & DALL·E \cite{ramesh2021zero} , OpenAI APIs \footnotemark[26]            \\ \hline
\end{tabular}
}
\end{table*}
\footnotetext[10]{\url{https://developer.apple.com/library/archive/documentation/AppleScript/Conceptual/AppleScriptLangGuide/introduction/ASLR_intro.html}\label{foot:as}}
\footnotetext[11]{{\url{https://www.macosxautomation.com/automator/}\label{foot:automator}}}
\footnotetext[12]{{\url{https://docs.blender.org/manual/en/latest/sculpt_paint/sculpting/introduction/gesture_tools.html}\label{foot:gesture}}}
\footnotetext[13]{{\url{https://developer.android.com/reference/android/speech/SpeechRecognizer}\label{foot:speech}}}
\footnotetext[14]{{\url{https://developer.apple.com/documentation/sirikit/}\label{foot:siri}}}
\footnotetext[15]{{\url{https://pypi.org/project/pyperclip/}\label{foot:Pyperclip}}}
\footnotetext[16]{{\url{https://clipboardjs.com/}\label{foot:Clipboard}}}
\footnotetext[17]{{\url{https://developer.android.com/develop/sensors-and-location/sensors/sensors_overview}\label{foot:sensor}}}
\footnotetext[18]{{\url{https://learn.microsoft.com/en-us/previous-versions/office/office-365-api/}\label{foot:msapi}}}
\footnotetext[19]{{\url{https://developer.android.com/reference}\label{foot:sdkapi}}}
\footnotetext[20]{{\url{https://developer.apple.com/ios/}\label{foot:iosapi}}}
\footnotetext[21]{{\url{https://learn.microsoft.com/en-us/windows/win32/api/}\label{foot:win32api}}}
\footnotetext[22]{{\url{https://developer.apple.com/library/archive/documentation/Cocoa/Conceptual/CocoaFundamentals/WhatIsCocoa/WhatIsCocoa.html}\label{foot:cocoaapi}}}
\footnotetext[23]{{\url{https://developer.mozilla.org/en-US/docs/Web/API/Fetch_API}\label{foot:fetchapi}}}
\footnotetext[24]{{\url{https://axios-http.com/docs/api_intro}\label{foot:axios}}}
% \footnotetext[25]{{\url{https://openai.com/index/gpt-4/}\label{foot:gpt}}}
% \footnotetext[25]{{\url{https://openai.com/index/dall-e-3/}\label{foot:dalle}}}
\footnotetext[25]{{\url{https://platform.openai.com/docs/overview}\label{foot:openai}}}
\addtocounter{footnote}{17}

\subsubsection{Action Inference\label{sec:agent_foundation:inference:action}}
Action inference is the core objective of the inference stage, as it translates the planning into executable tasks. The inferred actions are typically expressed as function call strings, encompassing the function name and relevant parameters. These strings can be readily converted into real-world interactions with the environment, such as clicks, keyboard inputs, mobile gestures, or API calls. A detailed discussion of these action types is presented in Section~\ref{sec:agent_foundation:action_exe}.

The input prompt must include a predefined set of actions available for the agent to select from. The agent can choose an action from this set or, if allowed, generate custom code or API calls to interact with the environment \cite{tan2024cradleempoweringfoundationagents}. This flexibility can enhance the agent's adaptability to unforeseen circumstances; however, it may introduce reliability concerns, as the generated code may be prone to errors.

\subsubsection{Complementary Outputs\label{sec:agent_foundation:inference:complementary}}
In addition to planning and action inference, the LLM can also generate complementary outputs that enhance the agent's capabilities. These outputs may include reasoning processes that clarify the agent's decision-making (\eg CoT reasoning), messages for user interaction, or communication with other agents or systems, or the status of the task (\eg continue or finished). The design of these functionalities can be tailored to meet specific needs, thereby enriching the overall performance of the GUI agent.

By effectively balancing planning and action inference while incorporating complementary outputs, agents can navigate complex tasks with a higher degree of organization and adaptability.

\subsection{Actions Execution\label{sec:agent_foundation:action_exe}}

Following the inference process, a crucial next step is for the GUI agent to execute the actions derived from the inferred commands within the GUI environment and subsequently gather feedback. Although the term ``GUI agent'' might suggest a focus solely on user interface actions, the action space can be greatly expanded by incorporating various toolboxes that enhance the agent's versatility. Broadly, the actions available to GUI agents fall into three main categories: \textit{(i)} UI operations \cite{mappingnaturallanguageinstructions}, \textit{(ii)} native API calls \cite{gu2016deep}, and \textit{(iii)} AI tools \cite{masterman2024landscape}. Each category offers unique advantages and challenges, enabling the agent to tackle a diverse range of tasks more effectively. We summarize the various actions commonly used in GUI agents, categorized into distinct types, in Table~\ref{tab:actions}, and provide detailed explanations of each category below.

\subsubsection{UI Operations\label{sec:agent_foundation:action_exe:ui}}
UI operations encompass the fundamental interactions that users typically perform with GUIs in software applications. These operations include various forms of input, such as mouse actions (clicks, drags, hovers), keyboard actions (key presses, combinations), touch actions (taps, swipes), and gestures (pinching, rotating). The specifics of these actions may differ across platforms and applications, necessitating a tailored approach for each environment.

While UI operations form the foundation of agent interactions with the GUI, they can be relatively slow due to the sequential nature of these tasks. Each operation must be executed step by step, which can lead to increased latency, especially for complex workflows that involve numerous interactions. Despite this drawback, UI operations are crucial for maintaining a broad compatibility across various applications, as they leverage standard user interface elements and interactions.

\subsubsection{Native API Calls\label{sec:agent_foundation:action_exe:api}}
In contrast to UI operations, some applications provide native APIs that allow GUI agents to perform actions more efficiently. These APIs offer direct access to specific functionalities within the application, enabling the agent to execute complex tasks with a single command \cite{lu2024turn}. For instance, calling the Outlook API allows an agent to send an email in one operation, whereas using UI operations would require a series of steps, such as navigating through menus and filling out forms \cite{song2024beyond}.

While native APIs can significantly enhance the speed and reliability of action execution, their availability is limited. Not all applications or platforms expose APIs for external use, and developing these interfaces can require substantial effort and expertise. Consequently, while native APIs present a powerful means for efficient task completion, they may not be as generalized across different applications as UI operations.

\subsubsection{AI Tools\label{sec:agent_foundation:action_exe:ai}}
The integration of AI tools into GUI agents represents a transformative advancement in their capabilities. These tools can assist with a wide range of tasks, including content summarization from screenshots or text, document enhancement, image or video generation (\eg calling ChatGPT \cite{wu2023brief}, DALL·E \cite{ramesh2021zero}), and even invoking other agents or Copilot tools for collaborative assistance. The rapid development of generative AI technologies enables GUI agents to tackle complex challenges that were previously beyond their capabilities.

By incorporating AI tools, agents can extend their functionality and enhance their performance in diverse contexts. For example, a GUI agent could use an AI summarization tool to quickly extract key information from a lengthy document or leverage an image generation tool to create custom visuals for user presentations. This integration not only streamlines workflows but also empowers agents to deliver high-quality outcomes in a fraction of the time traditionally required.

\subsubsection{Summary\label{sec:agent_foundation:action_exe:summary}}
An advanced GUI agent should adeptly leverage all three categories of actions: UI operations for broad compatibility, native APIs for efficient execution, and AI tools for enhanced capabilities. This multifaceted approach enables the agent to operate reliably across various applications while maximizing efficiency and effectiveness. By skillfully navigating these action types, GUI agents can fulfill user requests more proficiently, ultimately leading to a more seamless and productive user experience.

\subsection{Memory \label{sec:agent_foundation:memory}}

\begin{figure*}[t]
    \centering
    \includegraphics[width=\textwidth]{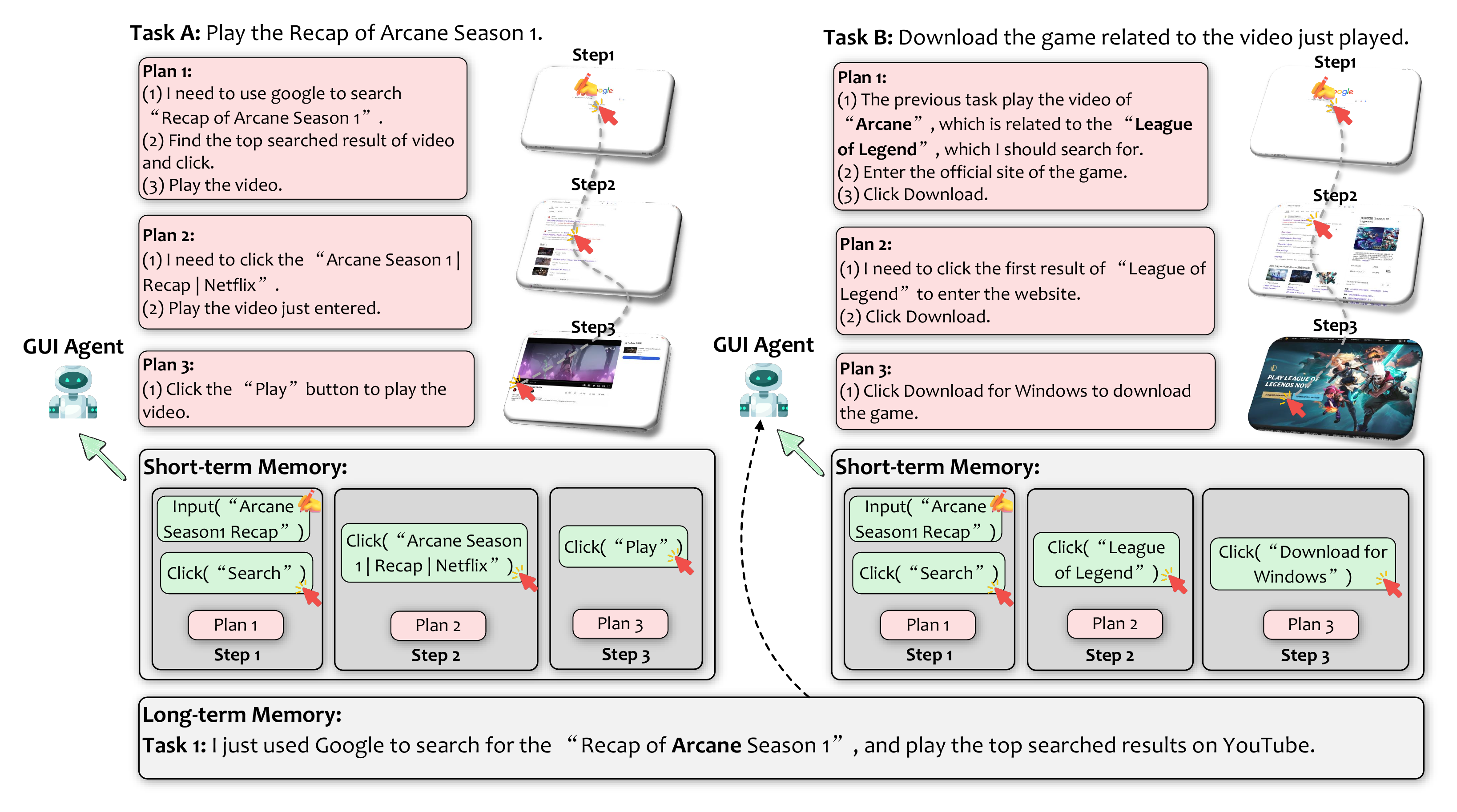}
    \vspace{-2em}
    \caption{Illustration of short-term memory and long-term memory in an LLM-brained GUI agent.}
    \label{fig:memory}
    % \vspace{-2em}
\end{figure*}

\begin{table*}[t]
\centering
\caption{Summary of memory in GUI agents.\label{tab:memory}}
\resizebox{\textwidth}{!}{ % Resize the entire figure to fit \textwidth
\begin{tabular}{l|l|l|l}
\hline
\multicolumn{1}{c|}{\textbf{Memory Element}} & \multicolumn{1}{c|}{\textbf{Memory Type}} & \multicolumn{1}{c|}{\textbf{Description}}                     & \multicolumn{1}{c}{\textbf{Storage Medium/Method}} \\ \hline
Action                                        & Short-term                                & Historical actions trajectory taken in the environment        & In-memory, Context window                           \\ \hline
Plan                                          & Short-term                                & Plan passed from previous step                                & In-memory, Context window                           \\ \hline
Execution Results                             & Short-term                                & Return values, error traces, and other environmental feedback & In-memory, Context window                           \\ \hline
Environment State                             & Short-term                                & Important environment state data, e.g., UI elements           & In-memory, Context window                           \\ \hline
Self-experience                               & Long-term                                 & Task completion trajectories from historical tasks            & Database, Disk                                      \\ \hline
Self-guidance                                 & Long-term                                 & Guidance and rules summarized from historical trajectories    & Database, Disk                                      \\ \hline
External Knowledge                            & Long-term                                 & Other external knowledge sources aiding task completion       & External Knowledge Base                             \\ \hline
Task Success Metrics                          & Long-term                                 & Metrics from task success or failure rates across sessions    & Database, Disk                                      \\ \hline
\end{tabular}
}
\end{table*}

For a GUI agent to achieve robust performance in complex, multi-step tasks, it must retain memory, enabling it to manage states in otherwise stateless environments. Memory allows the agent to track its prior actions, their outcomes, and the task's overall status, all of which are crucial for informed decision-making in subsequent steps \cite{lee2023explore}. By establishing continuity, memory transforms the agent from a reactive system into a proactive, stateful one, capable of self-adjustment based on accumulated knowledge. The agent's memory is generally divided into two main types: Short-Term Memory \cite{lu2023memochat} and Long-Term Memory \cite{wang2024augmenting}. We show an overview of different types of memory in GUI agents in Table~\ref{tab:memory}.

\subsubsection{Short-Term Memory\label{sec:agent_foundation:memory:short}}
Short-Term Memory (STM) provides the primary, ephemeral context used by the LLM during runtime \cite{tack2024online}. STM stores information pertinent to the current task, such as recent plans, actions, results, and environmental states, and continuously updates to reflect the task's ongoing status. This memory is particularly valuable in multi-step tasks, where each decision builds on the previous one, requiring the agent to maintain a clear understanding of the task's trajectory. As illustrated in Figure~\ref{fig:memory}, during the completion of independent tasks, the task trajectory, comprising actions and plans—is stored in the STM. This allows the agent to track task progress effectively and make more informed decisions.

However, STM is constrained by the LLM's context window, limiting the amount of information it can carry forward. To manage this limitation, agents can employ selective memory management strategies, such as selectively discarding or summarizing less relevant details to prioritize the most impactful information. Despite its limited size, STM is essential for ensuring coherent, contextually aware interactions and supporting the agent's capacity to execute complex workflows with immediate, relevant feedback.

\subsubsection{Long-Term Memory\label{sec:agent_foundation:memory:long}}
Long-Term Memory (LTM) serves as an external storage repository for contextual information that extends beyond the immediate runtime \cite{zhu2023ghost}. Unlike STM, which is transient, LTM retains historical task data, including previously completed tasks, successful action sequences, contextual tips, and learned insights. LTM can be stored on disk or in a database, enabling it to retain larger volumes of information than what is feasible within the LLM's immediate context window. 
In the example shown in Figure~\ref{fig:memory}, when the second task requests downloading a game related to the previous task, the agent retrieves relevant information from its LTM. This enables the agent to accurately identify the correct game, facilitating efficient task completion.

LTM contributes to the agent's self-improvement over time by preserving examples of successful task trajectories, operational guidelines, and common interaction patterns. When approaching a new task, the agent can leverage RAG techniques to retrieve relevant historical data, which enhances its ability to adapt strategies based on prior success. This is similar to the lifelong learning \cite{zheng2025lifelong}, which makes LTM instrumental in fostering an agent's capacity to ``learn'' from experience, enabling it to perform tasks with greater accuracy and efficiency as it accumulates insights across sessions. For instance, \cite{zheng2024synapsetrajectoryasexemplarpromptingmemory} provides an illustrative example of using past task trajectories stored in memory to guide and enhance future decision-making, a technique that is highly adaptable for GUI agents. It also enables better personalization by retaining information about previous tasks.

\subsection{Advanced Enhancements\label{sec:agent_foundation:enhancement}}

While most LLM-brained GUI agents incorporate fundamental components such as perception, planning, action execution, and memory, several advanced techniques have been developed to significantly improve the reasoning and overall capabilities of these agents. Here, we outline shared advancements widely adopted in research to guide the development of more specialized and capable LLM-brained GUI agents.

\begin{table*}[!t]
\centering
\caption{A summary of of GUI grounding models (Part I).}
\label{tab:gui-grounding1}
\resizebox{\textwidth}{!}{ % Resize the entire figure to fit \textwidth
\begin{tabular}{p{1.5cm}|p{1.5cm}|p{1.5cm}|p{1.5cm}|p{2.5cm}|p{2cm}|p{2cm}|p{3cm}|p{2cm}}
\toprule
\textbf{Model/ Benchmark} & \textbf{Platform} & \textbf{Foundation Model} & \textbf{Size} & \textbf{Dataset} & \textbf{Input} & \textbf{Output} & \textbf{Highlight} & \textbf{Link} \\ 
\midrule

OmniParser \cite{lu2024omniparserpurevisionbased} 
& Mobile, Desktop, and Web 
& BLIP-2 \cite{li2023blip}, YOLOv8 \cite{reis2023real} 
& / 
& 67{,}000 UI screenshots with bounding box annotations and 7{,}185 icon--description pairs generated using GPT-4 
& UI screenshots 
& IDs, bounding boxes, and descriptions of interactable elements 
& Introduces a purely vision-based screen parsing framework for general UI understanding without external information, significantly improving action prediction accuracy for LLM-driven agents 
& \url{https://github.com/microsoft/OmniParser} \\\hline

Iterative Narrowing \cite{nguyen2024improvedguigroundingiterative}
& Mobile, Web, and Desktop
& Qwen2-VL and OS-Atlas-Base
& /
& ScreenSpot \cite{cheng2024seeclickharnessingguigrounding}
& A GUI screenshot and a natural language query
& (x,y) coordinates representing the target location in the GUI
& Progressively crops regions of the GUI to refine predictions, enhancing precision for GUI grounding tasks
& \url{https://github.com/ant-8/GUI-Grounding-via-Iterative-Narrowing} \\\hline

Iris \cite{ge2024irisbreakingguicomplexity} & 
Mobile (iOS, Android), Desktop (Windows, macOS), and Web & 
Qwen-VL \cite{bai2023qwen} & 
9.6B & 
850K GUI-specific annotations and 150K vision–language instructions & 
High-resolution GUI screenshots with natural language instructions & 
Referring: Generates detailed descriptions of UI elements. 
Grounding: Locates UI elements on the screen. & 
Information-Sensitive Cropping for efficient handling 
of high-resolution GUI images, and Self-Refining Dual Learning 
to iteratively enhance GUI grounding and referring tasks 
without additional annotations & 
/ \\\hline

Attention-driven Grounding \cite{xu2024attentiondrivenguigroundingleveraging}
& Mobile, Web, and Desktop
& MiniCPM-Llama3-V 2.5
& 8.5B
& Mind2Web \cite{mind2webgeneralistagentweb}, ScreenSpot \cite{cheng2024seeclickharnessingguigrounding}, VisualWebBench \cite{liu2024visualwebbenchfarmultimodalllms}
& GUI screenshots and textual user queries
& Element localization via bounding boxes, text-to-image mapping for grounding, and actionable descriptions of GUI components
& Utilizes attention mechanisms in pre-trained MLLMs without fine-tuning
& \url{https://github.com/HeimingX/TAG} \\\hline

Aria-UI \cite{yang2024ariauivisualgroundinggui}
& Web, Desktop, and Mobile
& Aria \cite{li2024aria}
& 3.9B
& 3.9 million elements and 11.5 million samples
& GUI screenshots, user instructions, and action histories
& Pixel coordinates for GUI elements and corresponding actions
& A purely vision-based approach avoiding reliance on AXTree-like inputs
& \url{https://ariaui.github.io} \\\hline

UGround \cite{gou2024navigatingdigitalworldhumans} & 
Web, Desktop (Windows, MacOS, Linux), Mobile (Android, iOS) & 
LLaVA-NeXT-7B \cite{liu2024visual} & 
7B & 
Web-Hybrid and other existing datasets & 
GUI screenshots, user queries & 
Pixel coordinates of GUI elements & 
A universal GUI grounding model that relies solely on vision, eliminating the need for text-based representations & 
\url{https://osu-nlp-group.github.io/UGround/} \\\hline
        
GUI-Bee \cite{fan2025guibeealignguiaction} & Web & SeeClick \cite{cheng2024seeclickharnessingguigrounding}, QwenGUI \cite{chen2024guicoursegeneralvisionlanguage}, and UIX-7B \cite{liu2024harnessing} & 7B-13B & NovelScreenSpot & GUI screenshots, user queries, accessibility tree & GUI element grounding locations, actions and function calls, navigation steps, predicted GUI changes after interaction & Autonomously explores GUI environments, with Q-ICRL optimizing exploration efficiency and enhancing data diversity. & \url{https://gui-bee.github.io} \\\hline

\end{tabular}
}
\end{table*}

\begin{table*}[!t]
\centering
\caption{A summary of of GUI grounding models (Part II).}
\label{tab:gui-grounding2}
\resizebox{\textwidth}{!}{ % Resize the entire figure to fit \textwidth
\begin{tabular}{p{1.5cm}|p{1.5cm}|p{1.5cm}|p{1.5cm}|p{2.5cm}|p{2cm}|p{2cm}|p{3cm}|p{2cm}}
\toprule
\textbf{Model/ Benchmark} & \textbf{Platform} & \textbf{Foundation Model} & \textbf{Size} & \textbf{Dataset} & \textbf{Input} & \textbf{Output} & \textbf{Highlight} & \textbf{Link} \\ 
\midrule

RWKV-UI \cite{yang2025rwkv}  & Web &
1.6B & SIGLIP \cite{zhai2023sigmoid}, DINOv2 \cite{oquab2023dinov2}, SAM \cite{kirillov2023segment} & 
Websight \cite{laurenccon2024unlocking}, WebUI-7kbal \cite{wu2023webui}, Web2Code \cite{yun2024web2code} & 
High-resolution webpage images & 
Element grounding, Action prediction, CoT reasoning & 
Introduces a high-resolution three-encoder architecture with visual prompt engineering and CoT reasoning. & 
/ \\ \hline

TRISHUL \cite{singh2025trishulregionidentificationscreen} & 
Web, Desktop, and Mobile platforms & 
/ (Training-Free) & 
/ (Training-Free) & 
/ (Training-Free) & 
GUI Screenshots, user instructions/queries, hierarchical screen parsing outputs, OCR-extracted text descriptors & 
Action grounding, functionality descriptions of GUI elements, GUI referring, and SoMs & 
Utilizes hierarchical screen parsing and spatially enhanced element descriptions to enhance LVLMs without additional training. & 
/  \\ \hline

AutoGUI \cite{li2025autogui} & Web, Mobile & Qwen-VL-10B \cite{bai2023qwen}, SliME-8B \cite{zhang2024beyond} & 10B / 8B & AutoGUI-704k & GUI screenshots, User queries & Element functionalities, Element locations & Automatically labels UI elements based on interaction-induced changes, making it scalable and high-quality. & \url{https://autogui-project.github.io/} \\ \hline
Query Inference \cite{wu2025smoothing} & Mobile Android & Qwen2-VL-7B-Instruct \cite{wang2024qwen2vlenhancingvisionlanguagemodels} & 7B & UIBERT \cite{wu2024atlas} & GUI screenshots & Action-oriented queries, Coordinates & Improves reasoning without requiring large-scale training data. & \url{https://github.com/ZrW00/GUIPivot} \\\hline

WinClick \cite{hui2025winclickguigroundingmultimodal} & Windows OS & Phi3-Vision \cite{abdin2024phi} & 4.2B & WinSpot Benchmark & GUI screenshots, Natural language instructions & Element locations & The first GUI grounding model specifically tailored for Windows. & \url{https://github.com/zackhuiiiii/WinSpot} \\ \hline

FOCUS \cite{tang2025think} & Web, mobile applications, and desktop & Qwen2-VL-2B-Instruct \cite{wang2024qwen2vlenhancingvisionlanguagemodels} & 2B & GUICourse \cite{chen2024guicoursegeneralvisionlanguage}, Aguvis-stage1 \cite{xu2024aguvisunifiedpurevision}, Wave-UI \cite{zheng2024agentstudiotoolkitbuildinggeneral}, Desktop-UI \cite{lin2024training} & GUI screenshot + task instruction & Normalized coordinates (x, y) & A dual-system GUI grounding architecture inspired by human cognition, which dynamically switches between fast (intuitive) and slow (analytical) grounding modes based on task complexity & \url{https://github.com/sugarandgugu/Focus} \\ \hline

UI-E2I-Synth \cite{liu2025ui} & Web, Windows, and Android & InternVL2-4B and Qwen2-VL-7B & 4B and 7B & 1.6M screenshots, 9.9M instructions & GUI screenshot & Element coordinates & Introduces a three-stage synthetic data pipeline for GUI grounding with both explicit and implicit instruction synthesis & \url{https://colmon46.github.io/i2e-bench-leaderboard/} \\ \hline

RegionFocus \cite{luo2025visualtesttimescalinggui} & Web-based and Desktop interfaces & UI-TARS and Qwen2.5-VL & 72B & None (test-time only) & GUI screenshots with a point of interest & Coordinate-based actions & Introduces a visual test-time scaling framework that zooms into salient UI regions and integrates an image-as-map mechanism to track history and avoid repeated mistakes—boosting grounding accuracy without model retraining & \url{https://github.com/tiangeluo/RegionFocus} \\\hline

\end{tabular}
}
\end{table*}

\begin{table*}[!t]
\centering
\caption{A summary of of GUI grounding benchmarks.}
\label{tab:gui-grounding_data1}
\resizebox{\textwidth}{!}{ % Resize the entire figure to fit \textwidth
\begin{tabular}{p{1.5cm}|p{2cm}|p{3cm}|p{2cm}|p{2cm}|p{3cm}|p{2cm}}
\toprule
\textbf{Benchmark} & \textbf{Platform} & \textbf{Dataset} & \textbf{Input} & \textbf{Output} & \textbf{Highlight} & \textbf{Link} \\ 
\midrule

ScreenSpot \cite{cheng2024seeclickharnessingguigrounding}
& iOS, Android, macOS, and Windows
& Over 600 screenshots and 1{,}200 instructions
& GUI screenshots accompanied by user instructions
& Bounding boxes or coordinates of actionable GUI elements
& A realistic and diverse GUI grounding benchmark covering multiple platforms and a variety of elements
& \url{https://github.com/njucckevin/SeeClick} \\\hline

ScreenSpot-Pro \cite{li2024screenspot-pro}
& Windows, macOS, and Linux
& 1{,}581 instruction--screenshot pairs covering 23 applications across 5 industries and 3 operating systems
& High-resolution GUI screenshots paired with natural language instructions
& Bounding boxes for locating target UI elements
& Introduces a high-resolution benchmark for professional environments
& \url{https://github.com/likaixin2000/ScreenSpot-Pro-GUI-Grounding} \\\hline

PixelWeb \cite{yang2025pixelwebwebguidataset} & Web & 100,000 webpages & Rendered webpage screenshots and DOM information & BBox, mask, contour & The first GUI dataset to provide pixel-level annotations—including mask and contour—for web UIs, enabling high-precision GUI grounding and detection tasks & \url{https://huggingface.co/datasets/cyberalchemist/PixelWeb} \\\hline

\end{tabular}
}
\end{table*}

\subsubsection{Computer Vision-Based GUI Grounding\label{sec:agent_foundation:enhancement:cv}}
Although various tools (Section~\ref{table:key-toolkits}) enable GUI agents to access information like widget location, captions, and properties, certain non-standard GUIs or widgets may not adhere to these tools' protocols \cite{zhan2021research}, rendering their information inaccessible. Additionally, due to permission management, these tools are not always usable. Such incomplete information can present significant challenges for GUI agents, as the LLM may need to independently locate and interact with required widgets by estimating their coordinates to perform actions like clicking—a task that is inherently difficult without precise GUI data.

CV models offer a non-intrusive solution for GUI grounding directly from screenshots, enabling the detection, localization, segmentation, and even functional estimation of widgets \cite{li2020widget, white2019improving, wang2021screen2words, bai2021uibert}. This approach allows agents to interpret the visual structure and elements of the GUI without relying on system-level tools or internal metadata, which may be unavailable or incomplete. CV-based GUI parsing provides agents with valuable insights into interactive components, screen layout, and widget functionalities based solely on visual cues, enhancing their ability to recognize and act upon elements on the screen. Figure~\ref{fig:cv} provides an illustrative example of how a CV-based GUI parser works. While standard API-based detection captures predefined widgets, the CV model can identify additional elements, such as thumbnails and canvases, which may not have explicit API representations in the PowerPoint interface. This enhances widget recognition, allowing the agent to detect components beyond the scope of API detection. We show an overview of related GUI grounding models and benchmarks in Table~\ref{tab:gui-grounding1}, \ref{tab:gui-grounding2} and \ref{tab:gui-grounding_data1}.

A notable example is OmniParser~\cite{lu2024omniparserpurevisionbased}, which implements a multi-stage parsing technique involving a fine-tuned model for detecting interactable icons, an OCR module for extracting text, and an icon description model that generates localized semantic descriptions for each UI element. By integrating these components, OmniParser constructs a structured representation of the GUI, enhancing an agent's understanding of interactive regions and functional elements. This comprehensive parsing strategy has shown to significantly improve GPT-4V's screen comprehension and interaction accuracy. 

Such CV-based GUI grounding layers provide critical grounding information that significantly enhances an agent's ability to interact accurately and intuitively with diverse GUIs. This is particularly beneficial for handling custom or non-standard elements that deviate from typical accessibility protocols. Additionally, prompting methods like iterative narrowing have shown promise in improving the widget grounding capabilities of VLMs \cite{nguyen2024improvedguigroundingiterative}. Together, these approaches pave the way for more adaptable and resilient GUI agents, capable of operating effectively across a broader range of screen environments and application contexts.

Several works have introduced benchmarks to evaluate the GUI grounding capabilities of models and agents. For instance, ScreenSpot \cite{cheng2024seeclickharnessingguigrounding} serves as a pioneering benchmark designed to assess the GUI grounding performance of LLM-powered agents across diverse platforms, including iOS, Android, macOS, Windows, and web environments. It features a dataset with over 600 screenshots and 1,200 instructions, focusing on complex GUI components such as widgets and icons. This benchmark emphasizes the importance of GUI grounding in enhancing downstream tasks like web automation and mobile UI interaction. Building upon this, ScreenSpot-Pro \cite{li2024screenspot-pro} extends the scope to more professional, high-resolution environments. This evolved version includes 1,581 tasks with high-quality annotations, encompassing domains such as software development, creative tools, CAD, scientific applications, and office productivity. Key features of ScreenSpot-Pro include authentic high-resolution screenshots and meticulous annotations provided by domain experts.

These benchmarks provide critical evaluation criteria for assessing GUI grounding capabilities, thereby advancing the development of GUI agents for improved GUI understanding and interaction.

\subsubsection{Multi-Agent Framework\label{sec:agent_foundation:enhancement:multiagent}}
\begin{figure*}[t]
    \centering
    \includegraphics[width=1\textwidth]{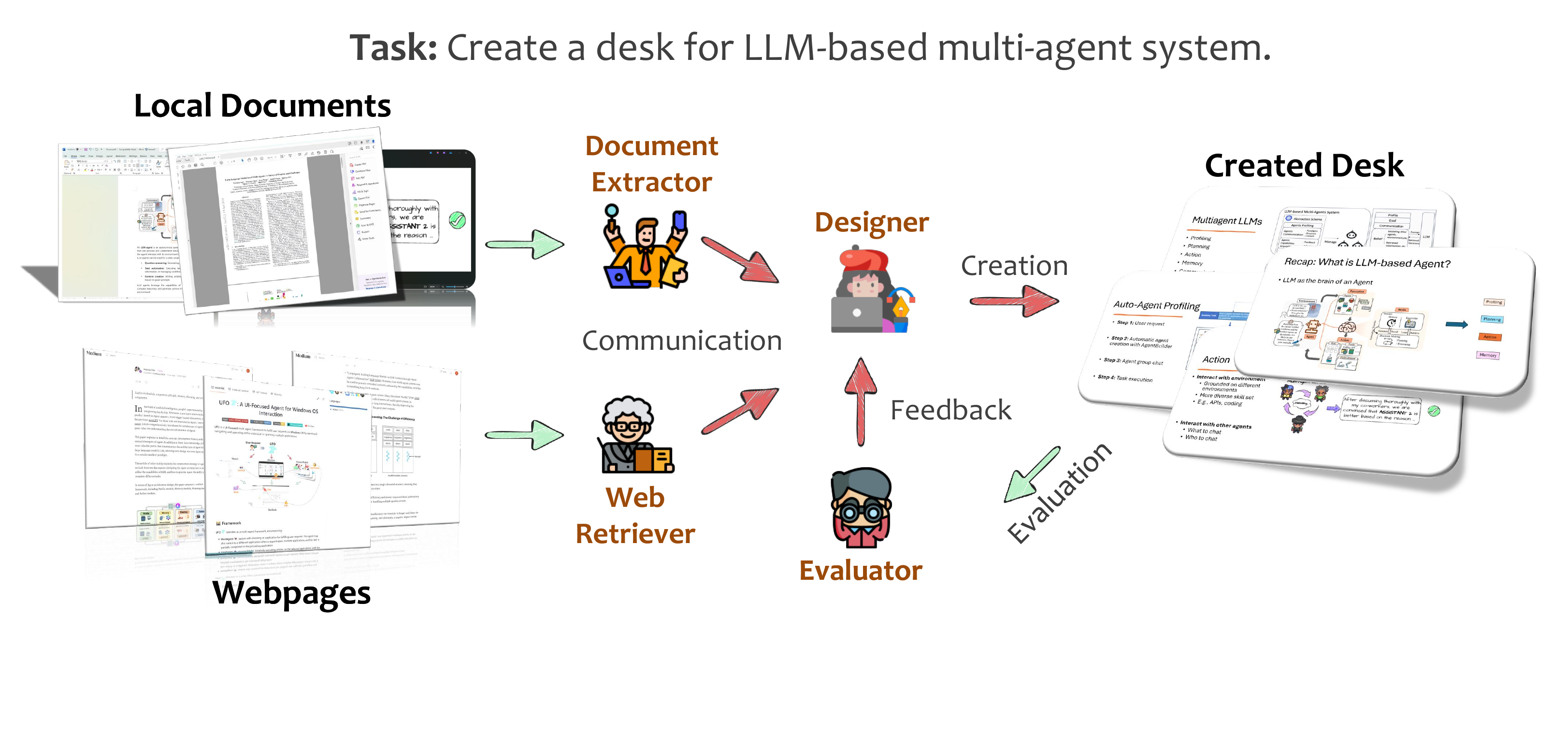}
    \vspace{-6em}
    \caption{An example of multi-agent system collaboration in creating a desk.}
    \label{fig:multiagent}
    % \vspace{-2em}
\end{figure*}
The adage ``two heads are better than one'' holds particular relevance for GUI automation tasks, where a single agent, though capable, can be significantly enhanced within a multi-agent framework \cite{li2023camel, chen2024internetagentsweavingweb}. Multi-agent systems leverage the collective intelligence, specialized skills, and complementary strengths of multiple agents to tackle complex tasks more effectively than any individual agent could alone. In the context of GUI agents, multi-agent systems offer advanced capabilities through two primary mechanisms: \textit{(i)} specialization and \textit{(ii)} inter-agent collaboration. Figure~\ref{fig:multiagent} illustrates an example of how an LLM-powered multi-agent collaborates to create a desk.
\begin{enumerate}
    \item \textbf{Specialization of Agents:} In a multi-agent framework, each agent is designed to specialize in a specific role or function, leveraging its unique capabilities to contribute to the overall task. As illustrated in the Figure~\ref{fig:multiagent}, specialization enables distinct agents to focus on different aspects of the task pipeline. For instance, the ``Document Extractor'' specializes in extracting relevant content from local documents, such as PDFs, while the ``Web Retriever'' focuses on gathering additional information from online sources. Similarly, the ``Designer'' transforms the retrieved information into visually appealing slides, and the "Evaluator" provides feedback to refine and improve the output. This functional separation ensures that each agent becomes highly adept at its designated task, leading to improved efficiency and quality of results \cite{song2024mmac}.
    \item \textbf{Collaborative Inter-Agent Dynamics:} The multi-agent system shown in the Figure~\ref{fig:multiagent} exemplifies how agents collaborate dynamically to handle complex tasks. The process begins with the ``Document Extractor'' and ``Web Retriever'', which work in parallel to collect information from local and online sources. The retrieved data is communicated to the ``Designer'', who synthesizes it into a cohesive set of slides. Once the slides are created, the ``Evaluator'' reviews the output, providing feedback for refinement. These agents share information, exchange context, and operate in a coordinated manner, reflecting a human-like teamwork dynamic. For example, as depicted, the agents' roles are tightly integrated—each output feeds into the next stage, creating a streamlined workflow that mirrors real-world collaborative environments \cite{zhang2024ufouifocusedagentwindows}.
\end{enumerate}
In such a system, agents can collectively engage in tasks requiring planning, discussion, and decision-making. Through these interactions, the system taps into each agent's domain expertise and latent potential for specialization, maximizing overall performance across diverse, multi-step processes.

\subsubsection{Self-Reflection\label{sec:agent_foundation:enhancement:self_reflection}}
\begin{figure*}[t]
    \centering
    \includegraphics[width=\textwidth]{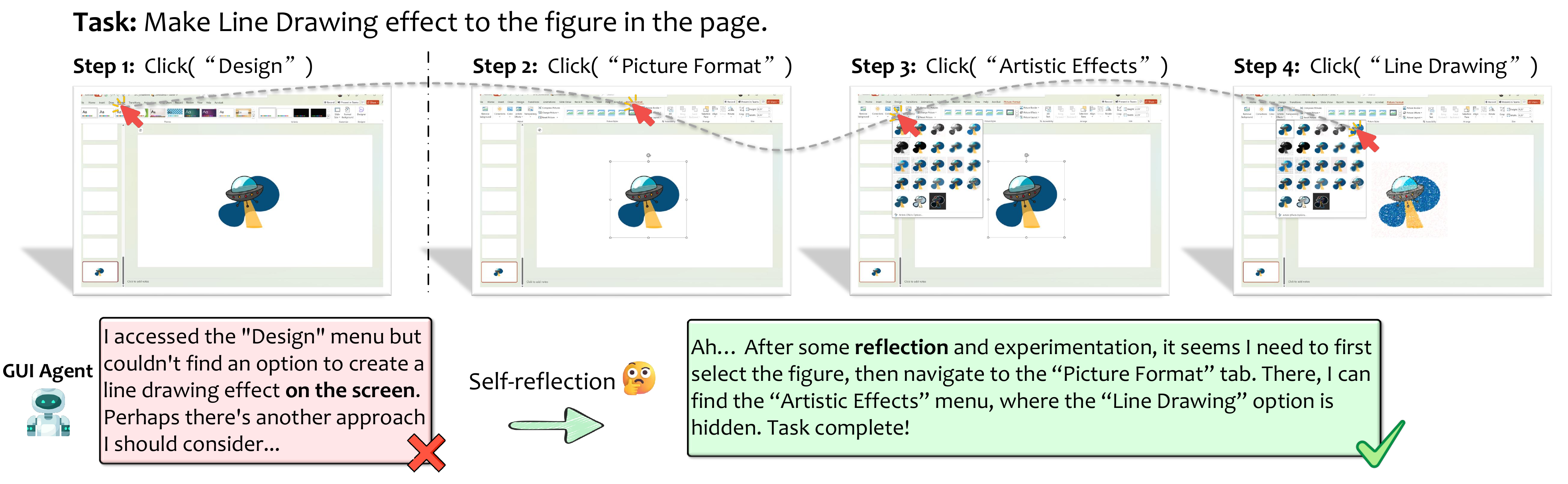}
    \vspace{-1em}
    \caption{An example of self-reflection in task completion of an LLM-powered GUI agent.}
    \label{fig:reflection}
    % \vspace{-2em}
\end{figure*}
``A fault confessed is half redressed''. In the context of GUI multi-agent systems, self-reflection refers to the agents' capacity to introspectively assess their reasoning, actions, and decisions throughout the task execution process \cite{renze2024self}. This capability allows agents to detect potential mistakes, adjust strategies, and refine actions, thereby improving the quality and robustness of their decisions, especially in complex or unfamiliar GUI environments. By periodically evaluating their own performance, self-reflective agents can adapt dynamically to produce more accurate and effective results \cite{pan2024autonomous}.

Self-reflection is particularly critical for GUI agents due to the variable nature of user interfaces and the potential for errors, even in human-operated systems. GUI agents frequently encounter situations that deviate from expectations, such as clicking the wrong button, encountering unexpected advertisements, navigating unfamiliar interfaces, receiving error messages from API calls, or even responding to user feedback on task outcomes. To ensure task success, a GUI agent must quickly reflect on its actions, assess these feedback signals, and adjust its plans to better align with the desired objectives. 

As illustrated in Figure~\ref{fig:reflection}, when the agent initially fails to locate the ``Line Drawing'' option in the Design menu, self-reflection enables it to reconsider and identify its correct location under Artistic Effects'' in the ``Picture Format'' menu, thereby successfully completing the task.

In practice, self-reflection techniques for GUI agents typically involve two main approaches: \textit{(i)} \textbf{ReAct} \cite{yao2022react} and \textit{(ii)} \textbf{Reflexion} \cite{shinn2024reflexion}.
\begin{enumerate}
    \item \textbf{ReAct (Reasoning and Acting):} ReAct integrates self-reflection into the agent's action chain by having the agent evaluate each action's outcome and reason about the next best step. In this framework, the agent doesn't simply follow a linear sequence of actions; instead, it adapts dynamically, continuously reassessing its strategy in response to feedback from each action. For example, if a GUI agent attempting to fill a form realizes it has clicked the wrong field, it can adjust by backtracking and selecting the correct element. Through ReAct, the agent achieves higher consistency and accuracy, as it learns to refine its behavior with each completed step.
    \item \textbf{Reflexion:} Reflexion emphasizes language-based feedback, where agents receive and process feedback from the environment as linguistic input, referred to as self-reflective feedback. This feedback is contextualized and used as input in subsequent interactions, helping the agent to learn rapidly from prior mistakes. For instance, if a GUI agent receives an error message from an application, Reflexion enables the agent to process this message, update its understanding of the interface, and avoid similar mistakes in future interactions. Reflexion's iterative feedback loop promotes continuous improvement and is particularly valuable for GUI agents navigating complex, multi-step tasks.
\end{enumerate}
Overall, self-reflection serves as an essential enhancement in GUI multi-agent systems, enabling agents to better navigate the variability and unpredictability of GUI environments. This introspective capability not only boosts individual agent performance but also promotes resilience, adaptability, and long-term learning in a collaborative setting.

\subsubsection{Self-Evolution\label{sec:agent_foundation:enhancement:self_evolution}}
\begin{figure*}[t]
    \centering
    \includegraphics[width=0.8\textwidth]{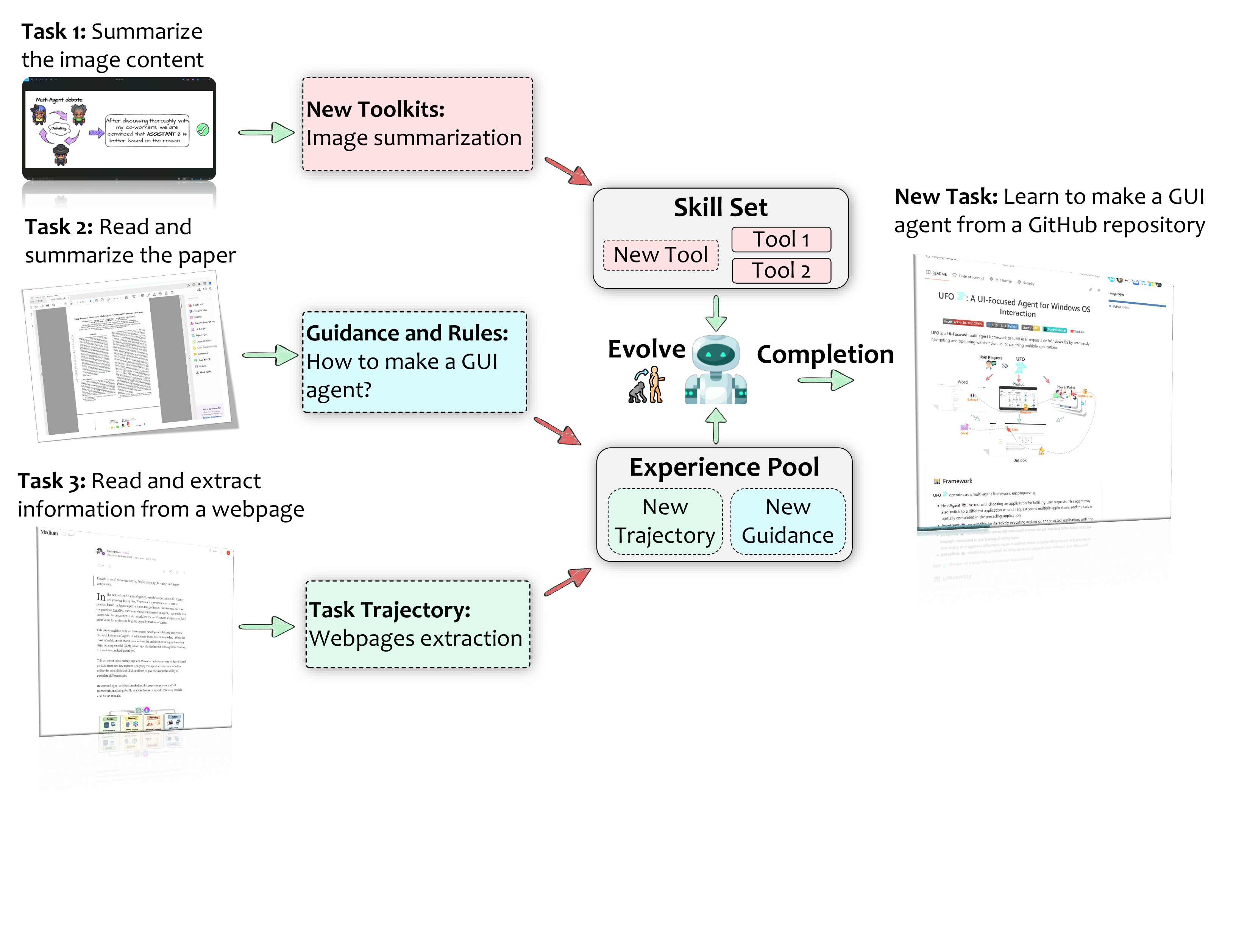}
    \vspace{-7.5em}
    \caption{An example self-evolution in a LLM-powered GUI agent with task completion.}
    \label{fig:evolve}
    % \vspace{-2em}
\end{figure*}
Self-evolution \cite{tao2024survey} is a crucial attribute that GUI agents should possess, enabling them to enhance their performance progressively through accumulated experience. In the context of GUI multi-agent systems, self-evolution allows not only individual agents to improve but also facilitates collective learning and adaptation by sharing knowledge and strategies among agents. During task execution, GUI agents generate detailed action trajectories accompanied by complementary information such as environment states, internal reasoning processes (the agent's thought processes), and evaluation results. This rich data serves as a valuable knowledge base from which GUI agents can learn and evolve. The knowledge extracted from this experience can be categorized into three main areas:
\begin{enumerate}
    \item \textbf{Task Trajectories}: The sequences of actions executed by agents, along with the corresponding environment states, are instrumental for learning \cite{zhao2024expel}. These successful trajectories can be leveraged in two significant ways. First, they can be used to fine-tune the core LLMs that underpin the agents. Fine-tuning with such domain-specific and task-relevant data enhances the model's ability to generalize and improves performance on similar tasks in the future. Second, these trajectories can be utilized as demonstration examples to activate the in-context learning capabilities of LLMs during prompt engineering. By including examples of successful task executions in the prompts, agents can better understand and replicate the desired behaviors without additional model training.

    For instance, suppose an agent successfully completes a complex task that involves automating data entry across multiple applications. The recorded action trajectory—comprising the steps taken, decisions made, and contextual cues—can be shared with other agents. These agents can then use this trajectory as a guide when faced with similar tasks, reducing the learning curve and improving efficiency.

    \item \textbf{Guidance and Rules:} From the accumulated experiences, agents can extract high-level rules or guidelines that encapsulate best practices, successful strategies, and lessons learned from past mistakes \cite{zhu2023large, zhang2024ruaglearnedruleaugmentedgenerationlarge}. Such guidance can be acquired by the LLM itself through trajectory summarization~\cite{zhu2023large}, or even via search-based algorithms, such as Monte Carlo Tree Search (MCTS)~\cite{zhang2024ruaglearnedruleaugmentedgenerationlarge}. This knowledge can be formalized into policies or heuristics that agents consult during decision-making processes, thereby enhancing their reasoning capabilities.

    For example, if agents repeatedly encounter errors when attempting to perform certain actions without proper prerequisites (\eg trying to save a file before specifying a file path), they can formulate a rule to check for these prerequisites before executing the action. This proactive approach reduces the likelihood of errors and improves task success rates.

    \item \textbf{New Toolkits:} Throughout their interactions, GUI agents may discover or develop more efficient methods, tools, or sequences of actions that streamline task execution \cite{tan2024cradleempoweringfoundationagents}. These may include optimized API calls, macros, or combinations of UI operations that accomplish tasks more effectively than previous approaches. LLMs can be leveraged to automatically analyze execution trajectories in order to summarize, discover, and generate high-level shortcuts or frequently used fast APIs, which can then be reused for future executions~\cite{jiang2025appagentx}. By incorporating these new tools into their repertoire, agents expand their capabilities and enhance overall efficiency.

    As an example, an agent might find that using a batch processing API can automate repetitive tasks more efficiently than performing individual UI operations in a loop. This new approach can be shared among agents within the multi-agent system, allowing all agents to benefit from the improved method and apply it to relevant tasks.
\end{enumerate}
Figure~\ref{fig:evolve} illustrates how a GUI agent evolves through task completion. During its operations, the agent adds new capabilities to its skill set, such as an image summarization toolkit, gains insights from reading a paper on creating GUI agents, and stores task trajectories like webpage extraction in its experience pool. When assigned a new task, such as ``Learn to make a GUI agent from a GitHub repository'', the agent draws on its acquired skills and past experiences to adapt and perform effectively.

This dynamic evolution highlights the agent's ability to continually learn, grow, and refine its capabilities. By leveraging past experiences, incorporating new knowledge, and expanding its toolset, GUI agents can adapt to diverse challenges, improve task execution, and significantly enhance the overall performance of the system, fostering a collaborative and ever-improving environment.

\subsubsection{Reinforcement Learning\label{sec:agent_foundation:enhancement:rl}}
\begin{figure*}[t]
    \centering
    \includegraphics[width=\textwidth]{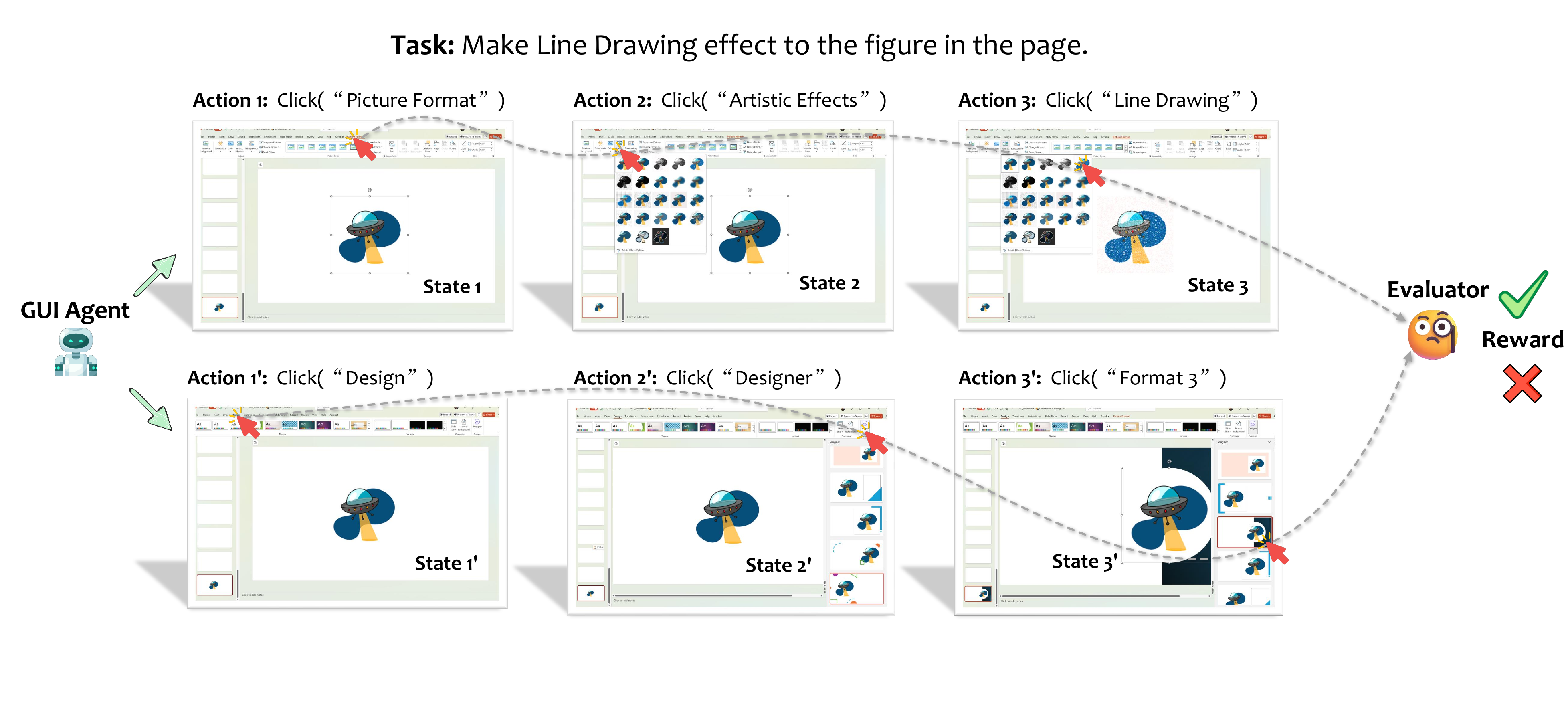}
    \vspace{-4em}
    \caption{An example of MDP modeling for task completion in a GUI agent.}
    \label{fig:mdp}
    % \vspace{-2em}
\end{figure*}
Reinforcement Learning (RL) \cite{kaelbling1996reinforcement} has witnessed significant advancements in aligning LLMs with desired behaviors \cite{wang2023aligning}, and has recently been employed in the development of LLM agents \cite{sun2024llm, zhai2024fine}. In the context of GUI multi-agent systems, RL offers substantial potential to enhance the performance, adaptability, and collaboration of GUI agents. GUI automation tasks naturally align with the structure of a Markov Decision Process (MDP) \cite{puterman1990markov}, making them particularly well-suited for solutions based on RL. In this context, the \textit{state} corresponds to the environment perception (such as GUI screenshots, UI element properties, and layout configurations), while \textit{actions} map directly to UI operations, including mouse clicks, keyboard inputs, and API calls. \textit{Rewards} can be explicitly defined based on various performance metrics, such as task completion, efficiency, and accuracy, allowing the agent to optimize its actions for maximal effectiveness. Figure~\ref{fig:mdp} illustrates an example of MDP modeling for task completion in a GUI agent, where state, action and reward are clearly defined.

By formulating GUI agent interactions as an MDP, we can leverage RL techniques to train agents that learn optimal policies for task execution through trial and error \cite{androidenv}. This approach enables agents to make decisions that maximize cumulative rewards over time, leading to more efficient and effective task completion. For example, an agent learning to automate form filling in a web application can use RL to discover the most efficient sequence of actions to input data and submit the form successfully, minimizing errors and redundant steps. This process helps align the agents more closely with desired behaviors in GUI automation tasks, especially in complex or ambiguous situations where predefined action sequences are insufficient.

As a representative approach, Bai \etal introduce DigiRL \cite{bai2024digirltraininginthewilddevicecontrol}, a two-phase RL framework for training GUI agents in dynamic environments. DigiRL begins with an offline RL phase that uses offline data to initialize the agent model, followed by online fine-tuning, where the model interacts directly with an environment to refine its strategies through live data within an Android learning environment using an LLM evaluator that provides reliable reward signals. This adaptive setting enables the agent to learn and respond effectively to the complexities of dynamic GUIs. Wang \etal propose DistRL \cite{wang2024distrl}, an RL fine-tuning pipeline specifically designed for on-device mobile control agents operating within Android. DistRL employs an asynchronous architecture, deploying RL fine-tuned agents across heterogeneous worker devices and environments for decentralized data collection. By leveraging off-policy RL techniques, DistRL enables centralized training with data gathered remotely from diverse environments, significantly enhancing the scalability and robustness of the model. These representative methods illustrate the potential of RL to improve GUI agents, demonstrating how both centralized and distributed RL frameworks can enable more responsive, adaptable, and effective GUI automation models in real-world applications.

\subsubsection{Summary \& Takeaways\label{sec:agent_foundation:enhancement:summary}}
In conclusion, the advanced techniques significantly enhance the capabilities of LLM-brained GUI agents, making them more versatile, efficient, and adaptive within multi-agent frameworks. Importantly, these techniques are not mutually exclusive—many can be integrated to create more powerful agents. For instance, incorporating self-reflection within a multi-agent framework allows agents to collaboratively improve task strategies and recover from errors. By leveraging these advancements, developers can design LLM-brained GUI agents that are not only adept at automating complex, multi-step tasks but also capable of continuously improving through self-evolution, adaptability to dynamic environments, and effective inter-agent collaboration. Future research is expected to yield even more sophisticated techniques, further extending the scope and robustness of GUI automation.

\subsection{From Foundations to Innovations: A Roadmap\label{sec:agent_foundation:roadmap}}

{\NoHyper
\begin{figure*}[t]
    \centering
    \resizebox{\textwidth}{!}{ % Resize the entire figure to fit \textwidth
    \begin{tikzpicture}[
        font=\sf\scriptsize,
        myarrow/.style={thick, -latex},
        Center/.style={circle, fill=white, text=root, align=center, font=\bf\scriptsize, inner sep=1pt},
    ]
    
    % Center Node
    \node[Center](ROOT) at (0,0) {Foundations\\to\\Innovations};
    
    % Node 1 Framework

    \arctextnode[N1][colorn1][\nodeHeight pt](\nodeRadius)(145+\offet)(75+\offet)[\levelOneFontSize]{Frameworks};
    \arctextnode[N1S1][colorn1][\subnodeHeight pt](\subnodeRadius)(145+\offet)(130+\offet)[\levelTwoFontSize]{Web};
        \arctextnode[N1S1c][colorn1][\subsubsubnodeHeight pt](\subsubnodeRadius)(145+\offet)(130+\offet)[\levelThreeFontSize]
        {
            \cite{zheng2024gpt4visiongeneralistwebagent, song2024beyond}\\
            \cite{chae2024webagentsworldmodels, gur2024realworldwebagentplanninglong, ma2024laserllmagentstatespace, he2024webvoyagerbuildingendtoendweb,
                  lai2024autowebglmbootstrapreinforcelarge, xie2023openagentsopenplatformlanguage, kil2024dualviewvisualcontextualizationweb,
                  abuelsaad2024agenteautonomouswebnavigation, koh2024tree, zhang2024webpilot, yang2024agentoccamsimplestrongbaseline,
                  murty2024nnetscape, shahbandeh2024naviqate, iong2024openwebagent, tang2024steward, putta2024agent, gu2024your,
                  verma2024adaptagent, kim2024auto, shen2024scribeagentspecializedwebagents, 
                  zhou2024proposeragentevaluatorpaeautonomousskilldiscovery, liu2024wepowebelementpreference,
                  huang2025r2d2, zhang2025symbiotic, pahuja2025explorerscalingexplorationdrivenweb, wornow2024automating,
                  zhang2025litewebagent, dammu2025towards, erdogan2025plan, zheng2025skillweaver, wang2025inducing, zhang2025enhancing}
        };
    \arctextnode[N1S2][colorn1][\subnodeHeight pt](\subnodeRadius)(130+\offet)(110+\offet)[\levelTwoFontSize]{Mobile};
        \arctextnode[N1S2c][colorn1][\subsubsubnodeHeight pt](\subsubnodeRadius)(130+\offet)(110+\offet)[\levelThreeFontSize]
        {
            \cite{zhang2023appagentmultimodalagentssmartphone, wang2024mobileagentautonomousmultimodalmobile}\\
            \cite{wen2024autodroid, zhang2024androidzoochainofactionthoughtgui}\\
            \cite{Song_2024, wen2024droidbotgptgptpowereduiautomation, ma2024cocoagentcomprehensivecognitivemllm,
                  zhang2024lookscreensmultimodalchainofaction, yan2023gpt4vwonderlandlargemultimodal,
                  li2024appagentv2advancedagent, wen2024autodroidv2boostingslmbasedgui,
                  wang2024mobileagentv2mobiledeviceoperation, zhang2024mobileexpertsdynamictoolenabledagent,
                  christianos2024lightweightneuralappcontrol, zhu2024moba, lee2024mobilegpt, 
                  wang2025mobileagenteselfevolvingmobileassistant, hoscilowicz2024clickagent, wu2025reachagent,
                  wang2025fedmobileagent, huangprompt2task, wang2025mobilev, MobileSteward, jiang2025appagentx, 
                  zhou2025chop, cheng2025kairos, dai2025advancing, liu2025learnactfewshotmobilegui, lai2025androidgen, wang2023enabling, kahlon2025agent, bishop2024latent}
        };
    \arctextnode[N1S3][colorn1][\subnodeHeight pt](\subnodeRadius)(110+\offet)(90+\offet)[\levelTwoFontSize]{Computer};
        \arctextnode[N1S3c][colorn1][\subsubsubnodeHeight pt](\subsubnodeRadius)(110+\offet)(90+\offet)[\levelThreeFontSize]
        {
            \cite{zhang2024ufouifocusedagentwindows, tan2024cradleempoweringfoundationagents}\\
            \cite{wu2024oscopilotgeneralistcomputeragents, li2024appagentv2advancedagent}\\
            \cite{agashe2024agentsopenagentic, wu2024guiactionnarratordid, li2023zero, he2024pcagentsleepai,
                  liu2025pc, aggarwal2025programming, zhao2025cola, lu2025steve, zhang2025ufo2, yin2025operation}
        };
    \arctextnode[N1S4][colorn1][\subnodeHeight pt](\subnodeRadius)(90+\offet)(75+\offet)[\levelTwoFontSize]{Cross-\\Platform};
        \arctextnode[N1S4c][colorn1][\subsubsubnodeHeight pt](\subsubnodeRadius)(90+\offet)(75+\offet)[\levelThreeFontSize]
        {
            \cite{liu2024autoglm, xu2024aguvisunifiedpurevision}\\
            \cite{pawlowski2024tinyclick, song2024mmac}\\
            \cite{su2025learnbyinteractdatacentricframeworkselfadaptive, he2025enhancing, wang2024oscar, jia2024agentstore,wang2024ponderpressadvancing, liu2025infiguiagentmultimodalgeneralistgui, agashe2025agent, hu2025guidingvlmagentsprocess, huang2025scaletrackscalingbacktrackingautomated}
        };

    % Node 2 Data
    \arctextnode[N2][colorn2][\nodeHeight pt](\nodeRadius)(75+\offet)(15+\offet)[\levelOneFontSize]{Data};
    \arctextnode[N2S1][colorn2][\subnodeHeight pt](\subnodeRadius)(75+\offet)(62+\offet)[\levelTwoFontSize]{Web};
        \arctextnode[N2S1c][colorn2][\subsubsubnodeHeight pt](\subsubnodeRadius)(75+\offet)(62+\offet)[\levelThreeFontSize]{
            \cite{mind2webgeneralistagentweb}\\
            \cite{Chen_Pitawela_Zhao_Zhou_Chen_Wu_2024,lù2024weblinxrealworldwebsitenavigation, pan2024webcanvasbenchmarkingwebagents, xu2024agenttrekagenttrajectorysynthesis, trabucco2025towards}
        };
    \arctextnode[N2S2][colorn2][\subnodeHeight pt](\subnodeRadius)(62+\offet)(47+\offet)[\levelTwoFontSize]{Mobile};
        \arctextnode[N2S2c][colorn2][\subsubsubnodeHeight pt](\subsubnodeRadius)(62+\offet)(47+\offet)[\levelThreeFontSize]{
            \cite{mappingnaturallanguageinstructions, zhang2024androidzoochainofactionthoughtgui}\\
            \cite{you2024ferretuigroundedmobileui, wu2024mobilevlmvisionlanguagemodelbetter,meng2024vgavisionguiassistant,rico,adatasetforinteractivevisionlanguagenavigation,metaguimultimodalconversationalagents,androidinwildlargescaledataset,lu2024guiodysseycomprehensivedataset,chai2024amexandroidmultiannotationexpo,chen2024octoplannerondevicelanguagemodel,wang2024eantlargescaledatasetefficient,xu2024androidlabtrainingsystematicbenchmarking,gao2024mobileviews,sun2025gui}
        };
    \arctextnode[N2S3][colorn2][\subnodeHeight pt](\subnodeRadius)(47+\offet)(30+\offet)[\levelTwoFontSize]{Computer};
        \arctextnode[N2S3c][colorn2][\subsubsubnodeHeight pt](\subsubnodeRadius)(47+\offet)(30+\offet)[\levelThreeFontSize]{
            \cite{niu2024screenagentvisionlanguagemodeldriven, wang2024lam, xu2025deskvision}\\
        };
    \arctextnode[N2S4][colorn2][\subnodeHeight pt](\subnodeRadius)(30+\offet)(15+\offet)[\levelTwoFontSize]{Cross-\\Platform};
        \arctextnode[N2S4c][colorn2][\subsubsubnodeHeight pt](\subsubnodeRadius)(30+\offet)(15+\offet)[\levelThreeFontSize]{
        \cite{lu2024omniparserpurevisionbased}\\
            \cite{chen2024guicoursegeneralvisionlanguage, gou2024navigatingdigitalworldhumans}\\
            \cite{sun2024genesis, chawla2024guide, chen2024guiworlddatasetguiorientedmultimodal, zhang2024xlam, shen2024falcon, liu2024visualagentbenchlargemultimodalmodels, baechler2024screenaivisionlanguagemodelui, chaimalas2025explorer, cheng2025navi}\\
        };

    % Node 3 Models

    \arctextnode[N3][colorn3][\nodeHeight pt](\nodeRadius)(15+\offet)(-85+\offet)[\levelOneFontSize]{Models};
        \arctextnode[N3S1][colorn3][\subnodeHeight pt](\subnodeRadius)(15+\offet)(-5+\offet)[\levelTwoFontSize]{Foundation\\Models};
                \arctextnode[N3S1c][colorn3][\subsubsubnodeHeight pt](\subsubnodeRadius)(15+\offet)(-5+\offet)[\levelThreeFontSize]{
                \cite{wang2024qwen2vlenhancingvisionlanguagemodels, abdin2024phi}\\
                \cite{cua2025, li2023blip, bai2023qwen}\\
                \cite{team2023gemini, hurst2024gpt, anthropic2024}\\                \cite{openai2023gpt4v,chen2024internvl,chen2024far,wang2024cogvlmvisualexpertpretrained,you2023ferret,liu2024visual,liu2024improved, huang2025spiritsight}
            };
        \arctextnode[N3S2][colorn3][\subnodeHeight pt](\subnodeRadius)(-5+\offet)(-25+\offet)[\levelTwoFontSize]{Web};
            \arctextnode[N3S2c][colorn3][\subsubsubnodeHeight pt](\subsubnodeRadius)(-5+\offet)(-25+\offet)[\levelThreeFontSize]{
                \cite{furuta2023multimodal, putta2024agent}\\
                \cite{fereidouni2024searchqueriestrainingsmaller, Thil_2024, he2024openwebvoyagerbuildingmultimodalweb,
                      qi2024webrltrainingllmweb, zhang2025breaking}
            };
        \arctextnode[N3S3][colorn3][\subnodeHeight pt](\subnodeRadius)(-25+\offet)(-45+\offet)[\levelTwoFontSize]{Mobile};
            \arctextnode[N3S3c][colorn3][\subsubsubnodeHeight pt](\subsubnodeRadius)(-25+\offet)(-45+\offet)[\levelThreeFontSize]{
                \cite{chen2024octoplannerondevicelanguagemodel, bai2024digirltraininginthewilddevicecontrol}\\
                \cite{you2024ferretuigroundedmobileui, wu2024mobilevlmvisionlanguagemodelbetter, meng2024vgavisionguiassistant}\\
                \cite{qian-etal-2024-visual, 
                      chen2024octopusondevicelanguagemodel, chen2024octopusv2ondevicelanguage, chen2024octopusv3technicalreport,
                      chen2024octopusv4graphlanguage, nong2024mobileflowmultimodalllmmobile, li2024uinav, wu2025vsc, papoudakis2025appvlm, bai2025digi, zheng2025vem, wang2025mp, lu2025ui,
                      zhang2025breaking, luo2025vimogenerativevisualgui}\\
                % \cite{wu2025vsc, papoudakis2025appvlm, bai2025digi, zheng2025vem, wang2025mp, lu2025ui,
                %       zhang2025breaking, luo2025vimogenerativevisualgui}
            };
        \arctextnode[N3S4][colorn3][\subnodeHeight pt](\subnodeRadius)(-45+\offet)(-65+\offet)[\levelTwoFontSize]{Computer};
            \arctextnode[N3S4c][colorn3][\subsubsubnodeHeight pt](\subsubnodeRadius)(-45+\offet)(-65+\offet)[\levelThreeFontSize]{
                \cite{niu2024screenagentvisionlanguagemodeldriven, yang2025octopus}\\
                \cite{wang2024lam, screenllm}
            };
        \arctextnode[N3S5][colorn3][\subnodeHeight pt](\subnodeRadius)(-65+\offet)(-85+\offet)[\levelTwoFontSize]{Cross-\\Platform};
            \arctextnode[N3S5c][colorn3][\subsubsubnodeHeight pt](\subsubnodeRadius)(-65+\offet)(-85+\offet)[\levelThreeFontSize]{
                \cite{hong2023cogagentvisuallanguagemodel}\\
                \cite{cheng2024seeclickharnessingguigrounding, shen2024falcon, zhang2023reinforceduiinstructiongrounding}\\
                \cite{lu2024omniparserpurevisionbased, baechler2024screenaivisionlanguagemodelui, zhang2024xlam}\\
                \cite{li2024ferret, lin2024training, wu2024atlas}\\
                \cite{qin2025uitarspioneeringautomatedgui, rahman2024v, yang2025magma,
                       huang2025spiritsight, xia2025gui, liu2025infiguir1advancingmultimodalgui}
            };

    % Node 4 Benchmark
    \arctextnode[N4][colorn4][\nodeHeight pt](\nodeRadius)(-85+\offet)(-170+\offet)[\levelOneFontSize]{Evaluation};

    % -------------------- Web --------------------
    \arctextnode[N4S1][colorn4][\subnodeHeight pt](\subnodeRadius)(-85+\offet)(-105+\offet)[\levelTwoFontSize]{Web};
        \arctextnode[N4S1c][colorn4][\subsubsubnodeHeight pt](\subsubnodeRadius)(-85+\offet)(-105+\offet)[\levelThreeFontSize]
        {
            \cite{liu2024visualwebbenchfarmultimodalllms}\\
            \cite{shahbandeh2024naviqate, lai2024autowebglmbootstrapreinforcelarge}\\
            \cite{xu2021groundingopendomaininstructionsautomate, wob, reinforcementlearningonwebinterfaces}\\
            \cite{mind2webgeneralistagentweb, Chen_Pitawela_Zhao_Zhou_Chen_Wu_2024, lù2024weblinxrealworldwebsitenavigation}\\
            \cite{webarenarealisticwebenvironment,koh2024visualwebarenaevaluatingmultimodalagents,deng2024multiturninstructionfollowingconversational,
                   zhang2024mminabenchmarkingmultihopmultimodal,pan2024webcanvasbenchmarkingwebagents,St-webagentbench,
                   furuta2024exposing,xu2024tur,xie2024osworldbenchmarkingmultimodalagents,
                   drouin2024workarena,jang2024videowebarena,ma2024cautionenvironmentmultimodalagents,wornowwonderbread,
                   zheng2024webolympus, webshopscalablerealworldweb, chezelles2024browsergym,
                   wu2025webwalkerbenchmarkingllmsweb, thomas2025webgames, tur2025safearenaevaluatingsafetyautonomous,
                   kara2025waber, xue2025illusion, zharmagambetov2025agentdam, lu2025agentrewardbench,
                   ye2025realwebassist, garg2025real, song2025bearcubs, evtimov2025wasp}
        };

    % -------------------- Mobile --------------------
    \arctextnode[N4S2][colorn4][\subnodeHeight pt](\subnodeRadius)(-105+\offet)(-130+\offet)[\levelTwoFontSize]{Mobile};
        \arctextnode[N4S2c][colorn4][\subsubsubnodeHeight pt](\subsubnodeRadius)(-105+\offet)(-130+\offet)[\levelThreeFontSize]
        {
            \cite{androidinwildlargescaledataset}\\
            \cite{xu2024androidlabtrainingsystematicbenchmarking, wen2024autodroid, mappingnaturallanguageinstructions}\\
            \cite{androidenv,wang2024mobileagentautonomousmultimodalmobile,mobileenvbuildingqualifiedevaluation,
                   lee2024benchmarkingmobiledevicecontrol,rawles2024androidworlddynamicbenchmarkingenvironment,
                   xing2024understanding,deng2024mobile,lee2024mobilesafetybench,
                   chen2024spa,zhang2024llamatouchfaithfulscalabletestbed,wang2024mobileagentbenchefficientuserfriendlybenchmark,
                   zhao2024gui, chai2025a3androidagentarena, ran2025beyond, chen2025aeia,sun2025autoeval,
                   wang2025fedmabench}
        };

    % -------------------- Computer --------------------
    \arctextnode[N4S3][colorn4][\subnodeHeight pt](\subnodeRadius)(-130+\offet)(-150+\offet)[\levelTwoFontSize]{Computer};
        \arctextnode[N4S3c][colorn4][\subsubsubnodeHeight pt](\subsubnodeRadius)(-130+\offet)(-150+\offet)[\levelThreeFontSize]
        {
            \cite{gao2024assistgui, bonatti2024windowsagentarenaevaluating, xie2024osworldbenchmarkingmultimodalagents}\\
            \cite{cao2024spider2vfarmultimodalagents,wang2024officebenchbenchmarkinglanguageagents,
                   zheng2024agentstudiotoolkitbuildinggeneral, zhao2025worldguidynamictestingcomprehensive,
                   nayak2025ui, wang2025computer}
        };

    % -------------------- Cross-Platform --------------------
    \arctextnode[N4S4][colorn4][\subnodeHeight pt](\subnodeRadius)(-150+\offet)(-170+\offet)[\levelTwoFontSize]{Cross-\\Platform};
        \arctextnode[N4S4c][colorn4][\subsubsubnodeHeight pt](\subsubnodeRadius)(-150+\offet)(-170+\offet)[\levelThreeFontSize]
        {
            \cite{wu2024guiactionnarratordid}\\
            \cite{liu2024visualagentbenchlargemultimodalmodels,kapoor2024omniactdatasetbenchmarkenabling,
                   lin2024videoguibenchmarkguiautomation}\\
            \cite{xu2024crabcrossenvironmentagentbenchmark,cheng2024seeclickharnessingguigrounding,
                   fan2024readpointedlayoutawaregui}
        };

    % Node 5
    \arctextnode[N5][colorn5][\nodeHeight pt](\nodeRadius)(-170+\offet)(-215+\offet)[\levelOneFontSize]{Applications};
        \arctextnode[N5S1][colorn5][\subnodeHeight pt](\subnodeRadius)(-170+\offet)(-192+\offet)[\levelTwoFontSize]{GUI\\ Testing};
            \arctextnode[N5S1c][colorn5][\subsubsubnodeHeight pt](\subsubnodeRadius)(-170+\offet)(-192+\offet)[\levelThreeFontSize]{
                \cite{zimmermann2023gui,liu2024make}\\        \cite{yoon2024intent,hu2024auitestagentautomaticrequirementsoriented,liu2024visiondrivenautomatedmobilegui,taeb2024axnav,cui2024large,liu2023fill,huang2024crashtranslator,feng2024prompting,ding2024improving, beyzaei2024automated, lu2025uxagent, ran2024guardian, li2024test, demissie2025vlm, yapaugci2025bugcraft, li2025reusedroid, chevrot2025autonomous, rosenbach2025automated, kongprophetagent, feng2025agent}
            };
        \arctextnode[N5S2][colorn5][\subnodeHeight pt](\subnodeRadius)(-192+\offet)(-215+\offet)[\levelTwoFontSize]{Virtual\\Assistants};
            \arctextnode[N5S2c][colorn5][\subsubsubnodeHeight pt](\subsubnodeRadius)(-192+\offet)(-215+\offet)[\levelThreeFontSize]{
                \cite{powerautomate}\\            \cite{gorniak2024vizability,ye2023proagent,guan2024intelligent,vu2024gptvoicetasker,pan2023autotask,gao2024assisteditor,huang2024promptrpa,gao2024easyask,openadapt2024,agentsea2024,openinterpreter2024,multion2024,magicos,srinivasan2025webnav}
            };
            % \arctextnode[N5S2S1][colorn5][8pt](3.5)(-165)(-190)[\levelTwoFontSize]{Research};
            % \arctextnode[N5S2S2][colorn5][8pt](3.5)(-190)(-205)[\levelTwoFontSize]{Open-Source\\Projects};
            % \arctextnode[N5S2S3][colorn5][8pt](3.5)(-205)(-230)[\levelTwoFontSize]{Production};

    % \arcarrow[colorn5](below N5S2S1)(above N5S2)[25];
    % \arcarrow[colorn5](below N5S2S2)(above N5S2)[5];
    % \arcarrow[colorn5](below N5S2S3)(above N5S2)[0];
    
    \end{tikzpicture}
    }
    \captionsetup{justification=centerlast} % Center the caption
    \caption{A Taxonomy of frameworks, data, models, evaluations, and applications: from foundations to innovations in LLM-brained GUI agents\label{fig:taxonomy}.}
\end{figure*}
\endNoHyper
}
% \endNoHyper

Building robust, adaptable, and effective LLM-powered GUI agents is a multifaceted process that requires careful integration of several core components. With a solid foundation in architecture, design, environment interaction, and memory, as outlined in Section~\ref{sec:agent_foundation}, we now shift our focus to the critical elements required for deploying these agents in practical scenarios. This exploration begins with an in-depth review of state-of-the-art LLM-brained GUI agent frameworks in Section~\ref{sec:framework}, highlighting their advancements and unique contributions to the field. Building on this, we delve into the methodologies for optimizing LLMs for GUI agents, starting with data collection and processing strategies in Section~\ref{sec:data}, and progressing to model optimization techniques in Section~\ref{sec:model}. To ensure robust development and validation, we then examine evaluation methodologies and benchmarks in Section~\ref{sec:evaluation}, which are essential for assessing agent performance and reliability. Finally, we explore a diverse range of practical applications in Section~\ref{sec:applications}, demonstrating the transformative impact of these agents across various domains. 

Together, these sections provide a comprehensive roadmap for advancing LLM-brained GUI agents from foundational concepts to real-world implementation and innovation. This roadmap, spanning from foundational components to real-world deployment, encapsulates the essential pipeline required to bring an LLM-powered GUI agent concept from ideation to implementation. 

To provide a comprehensive view, we first introduce a taxonomy in Figure~\ref{fig:taxonomy}, which categorizes recent work on LLM-brained GUI agents across frameworks, data, models, evaluation, and applications. This taxonomy serves as a blueprint for navigating the extensive research and development efforts within each field, while acknowledging overlaps among categories where certain models, frameworks, or datasets contribute to multiple aspects of GUI agent functionality. 
\section{LLM-Brained GUI Agent Framework\label{sec:framework}}

The integration of LLMs has unlocked new possibilities for constructing GUI agents, enabling them to interpret user requests, analyze GUI components, and autonomously perform actions across diverse environments. By equipping these models with essential components and functionalities, as outlined in Section~\ref{sec:agent_foundation}, researchers have created sophisticated frameworks tailored to various platforms and applications. These frameworks represent a rapidly evolving area of research, with each introducing innovative techniques and specialized capabilities that push the boundaries of what GUI agents can achieve.

We offer a detailed discussion of each framework, examining their foundational design principles, technical advancements, and the specific challenges they address in the realm of GUI automation. By delving into these aspects, we aim to provide deeper insights into how these agents are shaping the future of human-computer interaction and task automation, and the critical role they play in advancing this transformative field.

\begin{figure*}[t]
    \centering
    \includegraphics[width=\textwidth]{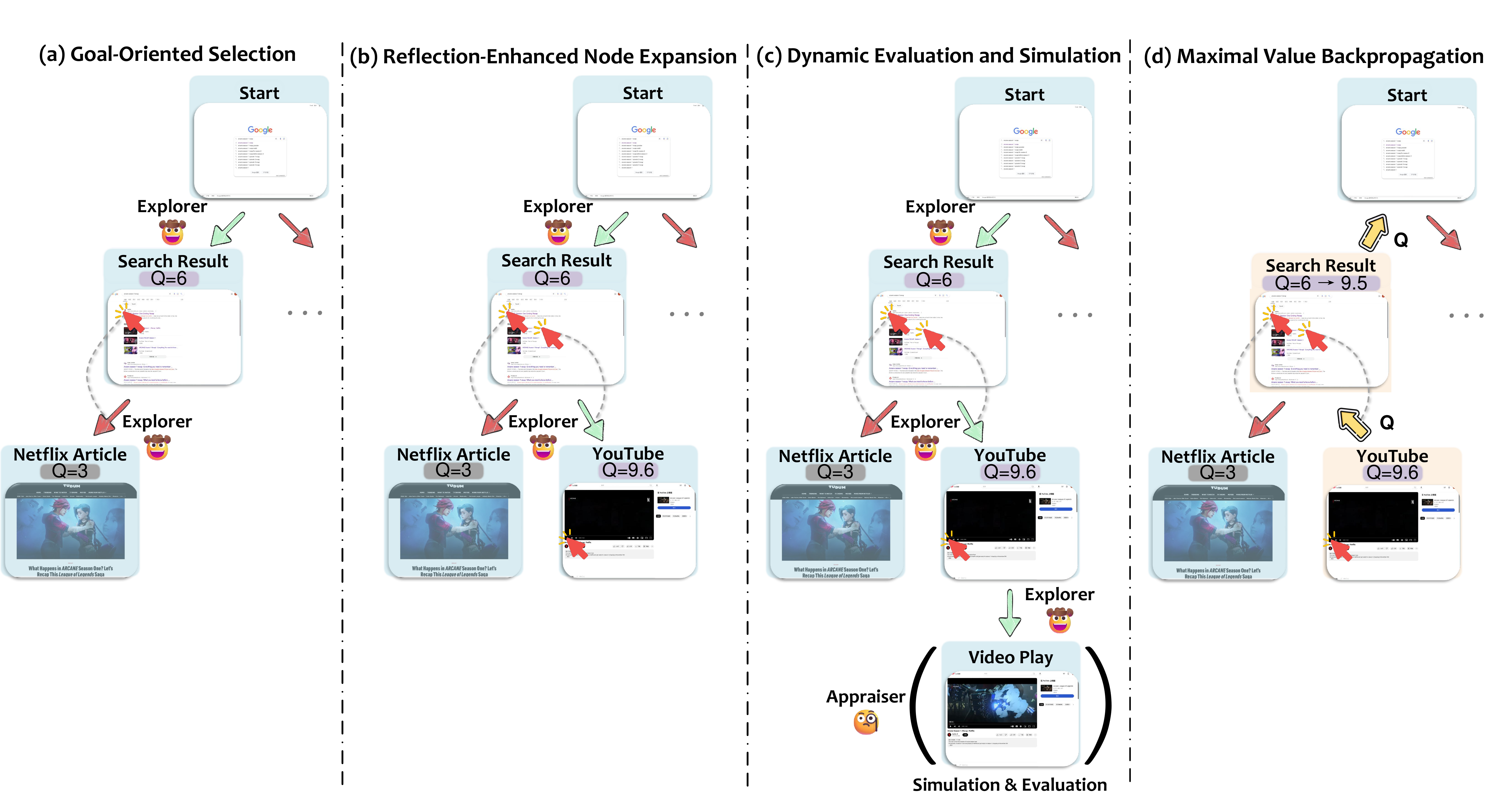}
    \vspace{-2em}
    \caption{An illustration of the local optimization stage in WebPilot \cite{zhang2024webpilot} using MCTS. Figure adapted from the original paper.}
    \label{fig:webpilot}
    % \vspace{-2em}
\end{figure*}

\begin{figure*}[t]
    \centering
    \includegraphics[width=\textwidth]{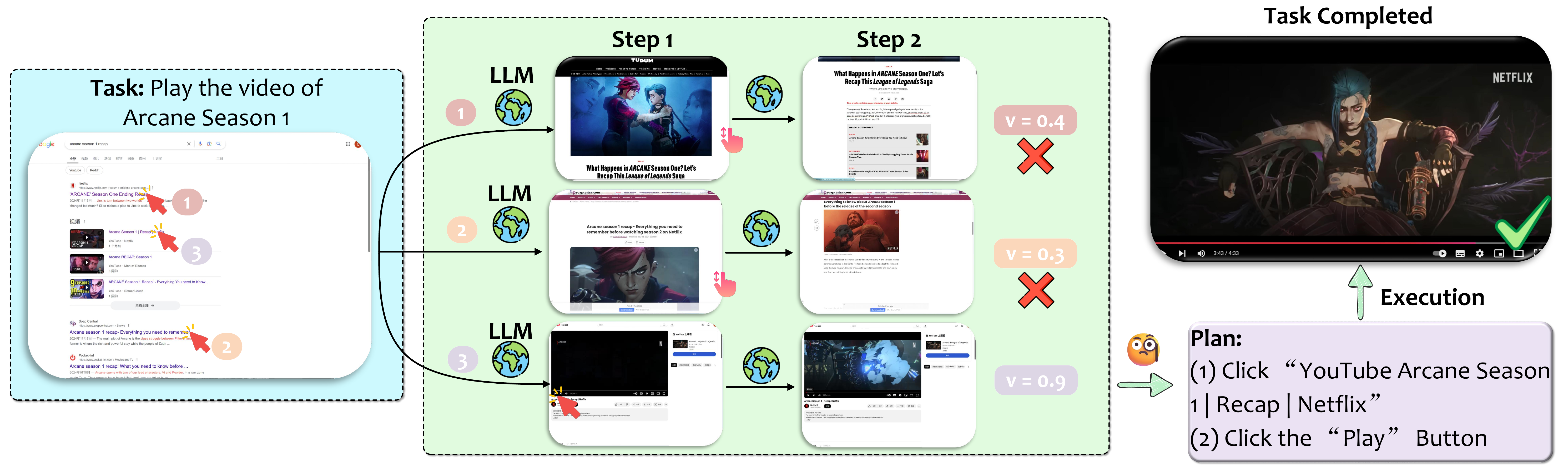}
    \vspace{-1em}
    \caption{An example illustrating how WebDreamer \cite{gu2024your} uses an LLM to simulate the outcome of each action. Figure adapted from the original paper.}
    \label{fig:world_model}
    % \vspace{-2em}
\end{figure*}

\begin{table*}[h!]
\centering
\caption{Overview of LLM-brained GUI agent frameworks on web platforms (Part I).\label{tab:web_framework1}}
\resizebox{\textwidth}{!}{ % Resize the entire figure to fit \textwidth
\begin{tabular}{p{1.5cm}|p{1.5cm}|p{2cm}|p{2.3cm}|p{2cm}|p{2.5cm}|p{3.5cm}|p{2cm}}
\hline
\textbf{Agent}                                              & \textbf{Platform} & \textbf{Perception}                                       & \textbf{Action}                                                                                                          & \textbf{Model}                                                                                          & \textbf{Architecture}                                                                             & \textbf{Highlight}                                                                                                                                                             & \textbf{Link}                                 \\ \hline
WMA \cite{chae2024webagentsworldmodels}                     & Web               & Accessibility tree from DOM                               & UI operations, \eg clock, type, and hover                                                                              & Llama-3.1-8B-Instruct \cite{dubey2024llama} for predicting observations and GPT-4 for policy modeling   & Single-agent with simulation-based observation                                                    & Uses a world model to predict state changes before committing actions, improving task success rates and minimizing unnecessary interactions with the environment               & \url{https://github.com/kyle8581/WMA-Agents}    \\ \hline
WebAgent \cite{gur2024realworldwebagentplanninglong}        & Web               & HTML structure                                            & UI interactions                                                                                                          & HTML-T5 for task planning and summarization and Flan-U-PaLM \cite{chung2024scaling} for code generation & Two-stage architecture for planning and program synthesis                                         & Leverages specialized LLMs to achieve HTML-based task planning and programmatic action execution                                                                               & /                                             \\ \hline
LASER \cite{ma2024laserllmagentstatespace}                  & Web               & GUI structure of the web environment, with defined states & Defined per state, such as searching, selecting items, navigating pages, and finalizing a purchase                       & GPT-4                                                                                                   & Single-agent                                                                                      & Uses a state-space exploration approach, allowing it to handle novel situations with flexible backtracking                                                                     & \url{https://github.com/Mayer123/LASER}                                             \\ \hline
WebVoyager \cite{he2024webvoyagerbuildingendtoendweb}       & Web               & Screenshots with numerical labels on interactive elements & Standard UI operations                                                                                                   & GPT-4V                                                                                                  & Single-agent                                                                                      & Integrates visual and textual cues within real-world, rendered web pages, enhancing its ability to navigate complex web structures                                             & \url{https://github.com/MinorJerry/WebVoyager}  \\ \hline
AutoWeb-GLM \cite{lai2024autowebglmbootstrapreinforcelarge} & Web               & Simplified HTML and OCR for text recognition              & UI operations such as clicking, typing, scrolling, and selecting, and advanced APIs like jumping to specific URLs        & ChatGLM3-6B \cite{glm2024chatglm}                                                                       & Single-agent                                                                                      & Its HTML simplification method for efficient webpage comprehension and its bilingual benchmark                                                                                 & \url{https://github.com/THUDM/AutoWebGLM}       \\ \hline
OpenAgents \cite{xie2023openagentsopenplatformlanguage}     & Web               & DOM elements                                              & Standard UI operations, browser-based actions controlled, API calls for tool execution, and structured data manipulation & GPT-4 and Claude \cite{anthropic2024}                                                                                       & Multi-agent architecture, with distinct agents (Data Agent, Plugins Agent, and Web Agent)         & Democratizes access to language agents by providing an open-source, multi-agent framework optimized for real-world tasks                                                       & \url{https://github.com/xlang-ai/OpenAgents}    \\ \hline
SeeAct \cite{zheng2024gpt4visiongeneralistwebagent}         & Web               & Screenshot images and HTML structure                      & Standard UI operations                                                                                                   & GPT-4V                                                                                                  & Single-agent                                                                                      & Its use of GPT-4V's multimodal capabilities to integrate both visual and HTML information, allowing for more accurate task performance on dynamic web content                  & \url{https://github.com/OSU-NLP-Group/SeeAct}   \\ \hline
DUAL-VCR \cite{kil2024dualviewvisualcontextualizationweb}   & Web               & HTML elements and screenshots                             & Standard UI operations                                                                                                   & Flan-T5-base \cite{chung2024scaling}                                                                    & Two-stage single-agent architecture                                                               & Dual-view contextualization                                                                                                                                                    & /                                             \\ \hline
Agent-E \cite{abuelsaad2024agenteautonomouswebnavigation}   & Web               & DOM structure and change observation                      & Standard UI operations                                                                                                   & GPT-4 Turbo                                                                                             & Hierarchical multi-agent architecture, composed of a planner agent and a browser navigation agent & Hierarchical architecture and adaptive DOM perception                                                                                                                          & \url{https://github.com/EmergenceAI/Agent-E}    \\ \hline
Search-Agent \cite{koh2024tree}                             & Web               & Screenshot and text descriptions                          & Standard UI operations                                                                                                   & GPT-4                                                                                                   & Single-agent with search                                                                          & Novel inference-time search algorithm that enhances the agent’s ability to perform multi-step planning and decision-making                                                     & \url{https://jykoh.com/search-agents}           \\ \hline
R2D2 \cite{huang2025r2d2} & Web & DOM & Standard UI operations & GPT-4o & Single-agent & Dynamically constructs an internal web environment representation for more robust decision-making. The integration of a replay buffer and error analysis reduces navigation errors and improves task completion rates. & \url{https://github.com/AmenRa/retriv} \\ \hline
\end{tabular}
}
\end{table*}

\begin{table*}[h!]
\centering
\caption{Overview of LLM-brained GUI agent frameworks on web platforms (Part II).\label{tab:web_framework2}}
\resizebox{\textwidth}{!}{ % Resize the entire figure to fit \textwidth
\begin{tabular}{p{1.5cm}|p{1.5cm}|p{2cm}|p{2.3cm}|p{2cm}|p{2.5cm}|p{3.5cm}|p{2cm}}
\hline
\textbf{Agent}                                                      & \textbf{Platform}                & \textbf{Perception}                                                                                & \textbf{Action}                                                            & \textbf{Model}                                                                                  & \textbf{Architecture}                                                              & \textbf{Highlight}                                                                                                                                        & \textbf{Link}                                    \\ \hline
ScribeAgent \cite{shen2024scribeagentspecializedwebagents} & Web & HTML-DOM & Standard UI operations & Single-agent architecture & Specialized fine-tuning approach using production-scale workflow data to outperform general-purpose LLMs like GPT-4 in web navigation tasks & \url{https://github.com/colonylabs/ScribeAgent} \\ \hline
PAE \cite{zhou2024proposeragentevaluatorpaeautonomousskilldiscovery} & Web & Screenshots & Standard UI Operations & Claude 3 Sonnet \cite{anthropic2024}, Qwen2VL-7B \cite{wang2024qwen2vlenhancingvisionlanguagemodels}, and LLaVa-1.6 \cite{liu2024visual} & A multi-agent architecture involving a task proposer to suggest tasks, an agent policy to perform tasks, and an autonomous evaluator to assess success and provide feedback using RL. & Autonomous skill discovery in real-world environments using task proposers and reward-based evaluation & \url{https://yanqval.github.io/PAE/}  \\ \hline
WebPilot \cite{zhang2024webpilot}                           & Web               & Accessibility trees (actrees) and dynamic observations    & Standard UI operations                                                                                                   & GPT-4                                                                                                   & Multi-agent architecture, with Global Optimization and Local Optimization                         & Dual optimization strategy (Global and Local) with Monte Carlo Tree Search (MCTS) \cite{browne2012survey}, allowing dynamic adaptation to complex, real-world web environments & \url{https://yaoz720.github.io/WebPilot/}       \\ \hline
Hybrid Agent \cite{song2024beyond}                          & Web               & Accessibility trees and screenshots                       & Standard UI operations, API calls, and generating code                                                                   & GPT-4                                                                                                   & Multi-agent system, combining both API and browsing capabilities                                  & Hybrid Agent seamlessly integrates web browsing and API calls                                                                                                                  & \url{https://github.com/yueqis/API-Based-Agent} \\ \hline
AgentOccam \cite{yang2024agentoccamsimplestrongbaseline}    & Web               & HTML                                                      & Standard UI operations                                                                                                   & GPT-4                                                                                                   & Single-agent                                                                                      & Simple design that optimizes the observation and action spaces                                                                                                                 & /                                             \\ \hline
NNetnav \cite{murty2024nnetscape}                           & Web               & DOM                                                       & Standard UI operations                                                                                                   & GPT-4                                                                                                   & Single-agent                                                                                      & Trains web agents using synthetic demonstrations, eliminating the need for expensive human input                                                                               & \url{https://github.com/MurtyShikhar/Nnetnav}   \\ \hline
NaviQAte \cite{shahbandeh2024naviqate}                      & Web               & Screenshots                                               & Standard UI operations                                                                                                   & GPT-4                                                                                                   & Single-agent system                                                                               & Frames web navigation as a question-and-answer task                                                                                                                            & /                                             \\ \hline
OpenWeb-Agent \cite{iong2024openwebagent}                   & Web               & HTML and screenshots                                      & UI operations, Web APIs, and self-generated code                                                                         & GPT-4 and AutoWebGLM \cite{lai2024autowebglmbootstrapreinforcelarge}                                    & Modular single-agent                                                                              & Modular design that allows developers to seamlessly integrate various models to automate web tasks                                                                             & \url{https://github.com/THUDM/OpenWebAgent/}    \\ \hline
Steward \cite{tang2024steward}                              & Web               & HTML and screenshots                                      & Standard UI operations                                                                                                   & GPT-4                                                                                                   & Single-agent                                                                                      & Ability to automate web interactions using natural language instructions                                                                                                       & /   
\\ \hline 
WebDreamer \cite{gu2024your}        & Web                             & Screenshots combined with SoM, and HTML                                                    & Standard UI operations and navigation actions & GPT-4o                  & Model-based single-agent architecture & Pioneers the use of LLMs as world models for planning in complex web   environments                                   & \url{https://github.com/OSU-NLP-Group/WebDreamer}     \\ \hline
Agent Q \cite{putta2024agent} & Web & DOM for textual input, screenshots for visual feedback & UI interactions, querying the user for help & LLaMA-3 70B \cite{dubey2024llama} for policy learning and execution, GPT-V for visual feedback & Single-agent with MCTS and RL & Combination of MCTS-guided search and self-critique mechanisms enables iterative improvement in reasoning and task execution & \url{https://github.com/sentient-engineering/agent-q} \\\hline
\end{tabular}
}
\end{table*}

\begin{table*}[h!]
\centering
\caption{Overview of LLM-brained GUI agent frameworks on web platforms (Part III).\label{tab:web_framework3}}
\resizebox{\textwidth}{!}{ % Resize the entire figure to fit \textwidth
\begin{tabular}{p{1.5cm}|p{1.5cm}|p{2cm}|p{2.3cm}|p{2cm}|p{2.5cm}|p{3.5cm}|p{2cm}}
\hline
\textbf{Agent}                                                      & \textbf{Platform}                & \textbf{Perception}                                                                                & \textbf{Action}                                                            & \textbf{Model}                                                                                  & \textbf{Architecture}                                                              & \textbf{Highlight}                                                                                                                                        & \textbf{Link}                                    \\ \hline
Auto-Intent \cite{kim2024auto}      & Web               & HTML structure                             & Standard UI Operations                       & GPT-3.5, GPT-4, Llama-3 \cite{dubey2024llama} for action inference; Mistral-7B \cite{jiang2023mistral} and Flan-T5XL \cite{chung2024scaling} for intent prediction & Single-agent with self-exploration                 & Introduces a unique self-exploration strategy to generate semantically diverse intent hints                     & /                    \\ \hline
AdaptAgent \cite{verma2024adaptagent} & Web               & GUI screenshots with HTML/DOM structures   & Standard UI Operations and Playwright scripts & GPT-4o and CogAgent \cite{hong2023cogagentvisuallanguagemodel}                                                                     & Single-agent                                        & Adapts to unseen tasks with just 1–2 multimodal human demonstrations                                             & /                    \\\hline 
WEPO \cite{liu2024wepowebelementpreference} & Web & HTML and DOM & Standard UI Operations & Llama3-8B \cite{dubey2024llama}, Mistral-7B \cite{jiang2023mistral}, and Gemma-2B \cite{team2024gemma} & Single-agent architecture. & Incorporates a distance-based sampling mechanism tailored to the DOM tree structure, enhancing preference learning by distinguishing between salient and non-salient web elements with DPO \cite{rafailov2024direct}. & / \\ \hline
AgentSymbiotic \cite{zhang2025symbiotic} & Web & Accessible tree structure of web elements & Standard UI operations & Large LLMs: GPT-4o, Claude-3.5. Small LLMs: LLaMA-3 \cite{dubey2024llama}, DeepSeek-R1 \cite{guo2025deepseek} & Multi-agent iterative architecture & Introduces an iterative, symbiotic learning process between large and small LLMs for web automation. Enhances both data synthesis and task performance through speculative data synthesis, multi-task learning, and privacy-preserving hybrid modes. & /  \\ \hline
LiteWebAgent \cite{zhang2025litewebagent} & Web & DOM, Screenshots & Standard UI operations, Playwright script & Any LLM and MLLM & Single-agent & First open-source, production-ready web agent integrating tree search for multi-step task execution. & \url{https://github.com/PathOnAI/LiteWebAgent} \\ \hline
ECLAIR \cite{wornow2024automating} & Web & Screenshots & Standard UI operations & GPT-4V, GPT-4o, CogAgent \cite{cogagent2024} & Single-agent architecture & Eliminates the high setup costs, brittle execution, and burdensome maintenance associated with traditional RPA by learning from video and text documentation. & \url{https://github.com/HazyResearch/eclair-agents} \\ \hline
Dammu \etal \cite{dammu2025towards} & Web & DOM elements, Webpage accessibility attributes & Standard UI operations & Not specified & Single-agent architecture & User-aligned task execution, where the agent adapts to individual user preferences in an ethical manner. & / \\ \hline
Plan-and-Act \cite{erdogan2025plan} & Web & HTML & Standard UI operations & LLaMA-3.3-70B-Instruct \cite{dubey2024llama} & Two-stage modular architecture: PLANNER + EXECUTOR & Decouples planning from execution in LLM-based GUI agents and introduces a scalable synthetic data generation pipeline to fine-tune each component & / \\ \hline
SkillWeaver \cite{zheng2025skillweaver} & Web & GUI screenshots and Accessibility Tree & Standard UI operations and high-level skill APIs & GPT-4o & Single-agent & Introduces a self-improvement framework for web agents that autonomously discover, synthesize, and refine reusable skill APIs through exploration & \url{https://github.com/OSU-NLP-Group/SkillWeaver} \\ \hline
ASI \cite{wang2025inducing} & Web & Webpage Accessibility Tree & Standard GUI actions & Claude-3.5-Sonnet & Single-agent & Introduces programmatic skills that are verified through execution to ensure quality and are used as callable actions to improve efficiency & \url{https://github.com/zorazrw/agent-skill-induction} \\ \hline
Rollback Agent \cite{zhang2025enhancing} & Web & Accessibility trees & Standard GUI actions & Multi-agent architecture & Multi-module, ReAct-inspired agent architecture & Introduces a modular rollback mechanism that enables multi-step rollback to avoid dead-end states & / \\ \hline

\end{tabular}
}
\end{table*}

\subsection{Web GUI Agents\label{sec:framework:web}}

\begin{figure*}[t]
    \centering
    \includegraphics[width=\textwidth]{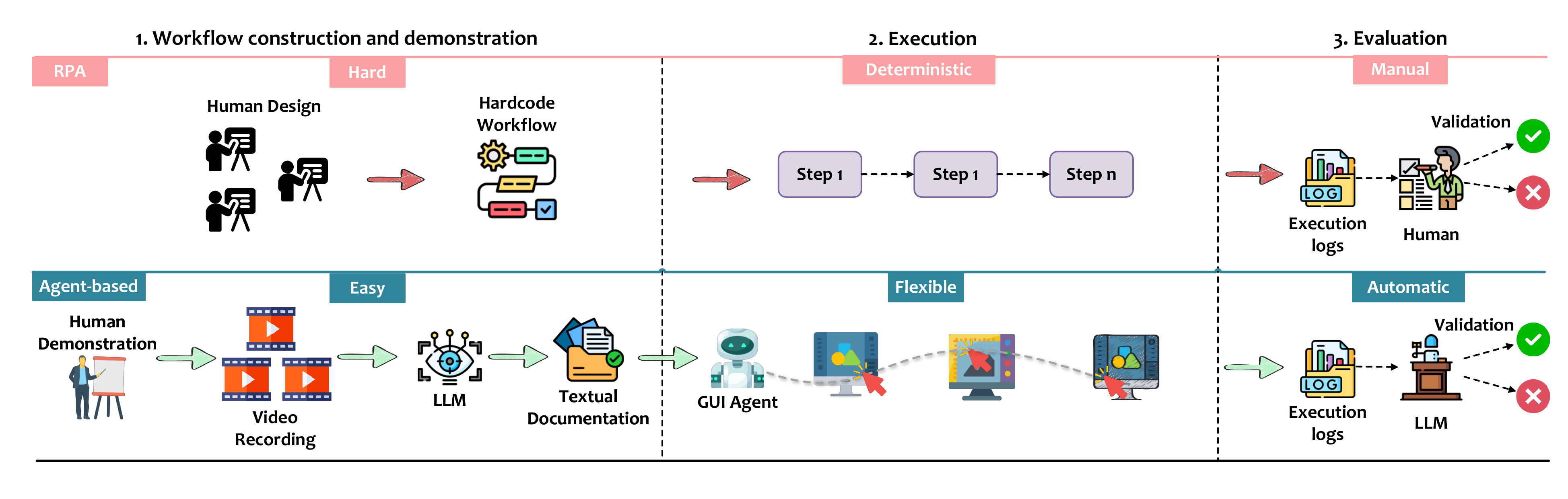}
    \vspace{-2em}
    \caption{Comparison of RPA and agent based automation. Figure adapted from \cite{wornow2024automating}.}
    \label{fig:rpa_agent}
\end{figure*}
Advancements in web GUI agents have led to significant strides in automating complex tasks within diverse and dynamic web environments. Recent frameworks have introduced innovative approaches that leverage multimodal inputs, predictive modeling, and task-specific optimizations to enhance performance, adaptability, and efficiency. In this subsection, we first summarize key web GUI agent frameworks in Tables~\ref{tab:web_framework1}, \ref{tab:web_framework2} and \ref{tab:web_framework3}, then delve into representative frameworks, highlighting their unique contributions and how they collectively push the boundaries of web-based GUI automation.

One prominent trend is the integration of multimodal capabilities to improve interaction with dynamic web content. For instance, \textbf{SeeAct}~\cite{zheng2024gpt4visiongeneralistwebagent} harnesses GPT-4V's multimodal capacities to ground actions on live websites effectively. By leveraging both visual data and HTML structure, SeeAct integrates grounding techniques using image annotations, HTML attributes, and textual choices, optimizing interactions with real-time web content. This approach allows SeeAct to achieve a task success rate of 51.1\% on real-time web tasks, highlighting the importance of dynamic evaluation in developing robust web agents.

Building upon the advantages of multimodal inputs, \textbf{WebVoyager}~\cite{he2024webvoyagerbuildingendtoendweb} advances autonomous web navigation by supporting end-to-end task completion across real-world web environments. Utilizing GPT-4V for both visual (screenshots) and textual (HTML elements) inputs, WebVoyager effectively interacts with dynamic web interfaces, including those with dynamically rendered content and intricate interactive elements. This multimodal capability allows WebVoyager to manage complex interfaces with a success rate notably surpassing traditional text-only methods, setting a new benchmark in web-based task automation.

In addition to multimodal integration, some frameworks focus on parsing intricate web structures and generating executable code to navigate complex websites. \textbf{WebAgent}~\cite{gur2024realworldwebagentplanninglong} employs a two-tiered model approach by combining HTML-T5 for parsing long, complex HTML documents with Flan-U-PaLM~\cite{chung2024scaling} for program synthesis. This modular design enables WebAgent to translate user instructions into executable Python code, autonomously handling complex, real-world websites through task-specific sub-instructions. WebAgent demonstrates a 50\% improvement in success rates on real websites compared to traditional single-agent models, showcasing the advantages of integrating HTML-specific parsing with code generation for diverse and dynamic web environments.

To enhance decision-making in web navigation, several frameworks introduce state-space exploration and search algorithms. \textbf{LASER}~\cite{ma2024laserllmagentstatespace} models web navigation as state-space exploration, allowing flexible backtracking and efficient decision-making without requiring extensive in-context examples. By associating actions with specific states and leveraging GPT-4's function-calling feature for state-based action selection, LASER minimizes errors and improves task success, particularly in e-commerce navigation tasks such as WebShop and Amazon. This state-based approach provides a scalable and efficient solution, advancing the efficiency of LLM agents in GUI navigation.

Similarly, \textbf{Search-Agent}~\cite{koh2024tree} innovatively introduces a best-first search algorithm to enhance multi-step reasoning in interactive web environments. By exploring multiple action paths, this approach improves decision-making, achieving up to a 39\% increase in success rates across benchmarks like WebArena~\cite{webarenarealisticwebenvironment}. Search-Agent's compatibility with existing multimodal LLMs demonstrates the effectiveness of search-based algorithms for complex, interactive web tasks.

Expanding on search-based strategies, \textbf{WebPilot}\cite{zhang2024webpilot} employs a dual optimization strategy combining global and local Monte Carlo Tree Search (MCTS) \cite{browne2012survey} to improve adaptability in complex and dynamic environments. As illustrated in Figure~\ref{fig:webpilot}, WebPilot decomposes overarching tasks into manageable sub-tasks, with each undergoing localized optimization. This approach allows WebPilot to continuously adjust its strategies in response to real-time observations, mimicking human-like decision-making and flexibility. Extensive testing on benchmarks like WebArena~\cite{webarenarealisticwebenvironment} and MiniWoB++~\cite{reinforcementlearningonwebinterfaces} demonstrates WebPilot's state-of-the-art performance, showcasing exceptional adaptability compared to existing methods.

Furthering the concept of predictive modeling, the \textbf{WMA}~\cite{chae2024webagentsworldmodels} introduces a world model to simulate and predict the outcomes of UI interactions. By focusing on transition-based observations, WMA allows agents to simulate action results before committing, reducing unnecessary actions and increasing task efficiency. This predictive capability is particularly effective in long-horizon tasks that require high accuracy, with WMA demonstrating strong performance on benchmarks such as WebArena~\cite{webarenarealisticwebenvironment} and Mind2Web~\cite{mind2webgeneralistagentweb}.

Along similar lines, \textbf{WebDreamer}\cite{gu2024your} introduces an innovative use of LLMs for model-based planning in web navigation, as depicted in Figure~\ref{fig:world_model}. WebDreamer simulates and evaluates potential actions and their multi-step outcomes using LLMs before execution~\cite{yao2024tree}, akin to a ``dreamer'' that envisions various scenarios. By preemptively assessing the potential value of different plans, WebDreamer selects and executes the plan with the highest expected value. This approach addresses critical challenges in web automation, such as safety concerns and the need for robust decision-making in complex and dynamic environments, demonstrating superiority over reactive agents in benchmarks like VisualWebArena~\cite{koh2024visualwebarenaevaluatingmultimodalagents} and Mind2Web-live~\cite{pan2024webcanvasbenchmarkingwebagents}.

Beyond predictive modeling, integrating API interactions into web navigation offers enhanced flexibility and efficiency. The \textbf{Hybrid Agent}~\cite{song2024beyond} combines web browsing and API interactions, dynamically switching between methods based on task requirements. By utilizing API calls for structured data interaction, the Hybrid Agent reduces the time and complexity involved in traditional web navigation, achieving higher accuracy and efficiency in task performance. This hybrid architecture underscores the benefits of integrating both structured API data and human-like browsing capabilities in AI agent systems.

Addressing the challenges of complex web structures and cross-domain interactions, \textbf{AutoWebGLM}~\cite{lai2024autowebglmbootstrapreinforcelarge} offers an efficient solution by simplifying HTML to focus on key webpage components, thereby improving task accuracy. Using reinforcement learning and rejection sampling for fine-tuning, AutoWebGLM excels in complex navigation tasks on both English and Chinese sites. Its bilingual dataset and structured action-perception modules make it practical for cross-domain web interactions, emphasizing the importance of efficient handling in diverse web tasks.

\textbf{ECLAIR \cite{wornow2024automating}} represents a pioneering application that replaces traditional RPA with a foundation model-powered GUI agent for enterprise automation. Unlike conventional RPA, which relies on manually programmed rules and rigid scripts, ECLAIR dynamically learns workflows from video demonstrations and textual SOPs (Standard Operating Procedures), significantly reducing setup time and improving adaptability. It operates on enterprise web applications, leveraging GPT-4V and CogAgent \cite{cogagent2024} to perceive GUI elements, plan actions, and execute workflows, and validate automatically. By eliminating the high maintenance costs and execution brittleness of RPA, ECLAIR introduces a more flexible and scalable approach to GUI automation. We show a comparison of such agent-based vs. RPA automation in Figure~\ref{fig:rpa_agent}. This work establishes an important foundation for LLM-powered GUI automation, demonstrating how multimodal foundation models can bridge the gap between process mining, RPA, and fully autonomous enterprise workflows.

In summary, recent frameworks for web GUI agents have made substantial progress by integrating multimodal inputs, predictive models, and advanced task-specific optimizations. These innovations enable robust solutions for real-world tasks, enhancing the capabilities of web-based GUI agents and marking significant steps forward in developing intelligent, adaptive web automation.

\begin{table*}[h!]
\centering
\caption{Overview of LLM-brained GUI agent frameworks on mobile platforms (Part I).\label{tab:mobile_framework1}}
\resizebox{\textwidth}{!}{ % Resize the entire figure to fit \textwidth
\begin{tabular}{p{1.5cm}|p{1.5cm}|p{2cm}|p{2.3cm}|p{2cm}|p{2.5cm}|p{3.5cm}|p{2cm}}
\hline
\textbf{Agent}                                                      & \textbf{Platform}                & \textbf{Perception}                                                                                & \textbf{Action}                                                            & \textbf{Model}                                                                                  & \textbf{Architecture}                                                              & \textbf{Highlight}                                                                                                                                        & \textbf{Link}                                    \\\hline
Wang \etal \cite{wang2023enabling} & Android Mobile & Android view hierarchy structure & (1) Screen Question Generation, (2) Screen Summarization, (3) Screen Question Answering, and (4) Mapping Instruction to UI Action & PaLM \cite{anil2023palm2technicalreport} & Single-agent & The first paper to study Screen Question Generation and Screen QA using LLMs & \url{https://github.com/google-research/google-research/tree/master/llm4mobile}
\\ \hline
VisionTasker \cite{Song_2024}                                       & Android mobile devices           & UI screenshots with widget detection and text extraction                                           & UI operations such as tapping, swiping, and entering text                  & ERNIE Bot \cite{ernie}                                                                          & Single-agent with vision-based UI understanding and sequential task planning       & Vision-based UI understanding approach, which allows it to interpret UI semantics directly from screenshots without view hierarchy dependencies           & \url{https://github.com/AkimotoAyako/VisionTasker} \\ \hline
DroidBot-GPT \cite{wen2024droidbotgptgptpowereduiautomation}        & Android mobile devices           & Translates the GUI state information of Android applications into natural language prompts         & UI operations, including actions like click, scroll, check, and edit       & GPT                                                                                             & Single-agent                                                                       & Automates Android applications without modifications to either the app or the model                                                                       & \url{https://github.com/MobileLLM/DroidBot-GPT}    \\ \hline
CoCo-Agent \cite{ma2024cocoagentcomprehensivecognitivemllm}         & Android mobile devices           & GUI screenshots, OCR layouts, and historical actions                                               & GUI actions, such as clicking, scrolling, and typing                       & CLIP \cite{radford2021learning} for vision encoding and LLaMA-2-chat-7B for language processing & Single-agent                                                                       & Its dual approach of Comprehensive Environment Perception and Conditional Action Prediction                                                               & \url{https://github.com/xbmxb/CoCo-Agent}          \\ \hline
Auto-GUI \cite{zhang2024lookscreensmultimodalchainofaction}         & Android mobile devices           & GUI screenshots                                                                                    & GUI operations                                                             & BLIP-2 vision encoder \cite{li2023blip} with a FLAN-Alpaca \cite{wei2021finetuned}              & Single-agent with chain-of-action                                                  & Its direct interaction with GUI elements. Its chain-of-action mechanism enables it to leverage both past and planned actions                              & \url{https://github.com/cooelf/Auto-GUI}           \\ \hline                                     
MobileGPT \cite{lee2024mobilegpt} & Android mobile devices & Simplified HTML representation & Standard UI operations and navigation actions & GPT-4-turbo for screen understanding and reasoning, GPT-3.5-turbo for slot-filling sub-task parameters & Single-agent architecture augmented by a hierarchical memory structure & Introduces a human-like app memory that allows for task decomposition into modular sub-tasks & \url{https://mobile-gpt.github.io} \\ \hline
MM-Navigator \cite{yan2023gpt4vwonderlandlargemultimodal}           & Mobile iOS and Android           & Smartphone screenshots with associated set-of-mark tags                                            & Clickable UI operations                                                    & GPT-4V                                                                                          & Single-agent                                                                       & Using set-of-mark prompting with GPT-4V for precise GUI navigation on smartphones                                                                         & \url{https://github.com/zzxslp/MM-Navigator}       \\ \hline
AppAgent \cite{zhang2023appagentmultimodalagentssmartphone}         & Android mobile devices           & Real-time screenshots and XML files detailing the interactive elements                             & User-like actions, like Tap, Long press, Swipe, Text input, Back and Exit  & GPT-4V                                                                                          & Single-agent                                                                       & Its ability to perform tasks on any smartphone app using a human-like interaction method                                                                  & \url{https://appagent-official.github.io/}         \\ \hline
AppAgent-V2 \cite{li2024appagentv2advancedagent}                    & Android mobile devices           & GUI screenshots with annotated elements, OCR for detecting text and icons, Structured XML metadata & Standard UI Operations: Tap, text input, long press, swipe, back, and stop & GPT-4                                                                                           & Multi-phase architecture with Exploration Phase and Deployment Phase               & Enhances adaptability and precision in mobile environments by combining structured data parsing with visual features                                      & /                                                \\ \hline
FedMobileAgent \cite{wang2025fedmobileagent} &  Android mobile devices &  GUI Screenshots &  Standard UI operations &  Qwen2-VL-Instruct-7B \cite{wang2024qwen2vlenhancingvisionlanguagemodels} & Multi-agent federated learning & Introduces privacy-preserving federated learning for mobile automation, enabling large-scale training without centralized human annotation.  & / \\ \hline
\end{tabular}
}
\end{table*}

\begin{table*}[h!]
\centering
\caption{Overview of LLM-brained GUI agent frameworks on mobile platforms (Part II).\label{tab:mobile_framework2}}
\resizebox{\textwidth}{!}{ % Resize the entire figure to fit \textwidth
\begin{tabular}{p{1.5cm}|p{1.5cm}|p{2cm}|p{2.3cm}|p{2cm}|p{2.5cm}|p{3.5cm}|p{2cm}}
\hline
\textbf{Agent}                                            & \textbf{Platform}           & \textbf{Perception}                                                                                    & \textbf{Action}                                                        & \textbf{Model}                                        & \textbf{Architecture}                                                                                                                                                & \textbf{Highlight}                                                                                                                                          & \textbf{Link}                                                                                      \\ \hline
Prompt2Task \cite{huangprompt2task} & Android mobile devices & GUI structure and layout hierarchy, full-page textual descriptions, OCR-based text extraction & Standard UI operations & GPT-4 & Multi-agent architecture & Enables UI automation through free-form textual prompts, eliminating the need for users to script automation tasks. & \url{https://github.com/PromptRPA/Prompt2TaskDataset} \\ \hline
ClickAgent \cite{hoscilowicz2024clickagent} & Android Mobile Devices & Screenshots & Standard UI operations & InternVL-2.0 \cite{chen2024internvl}, TinyClick \cite{pawlowski2024tinyclick}, SeeClick \cite{cheng2024seeclickharnessingguigrounding} & Single-agent & Combines MLLM reasoning with a dedicated UI location model to enhance UI interaction accuracy & \url{https://github.com/Samsung/ClickAgent} \\ \hline

AutoDroid \cite{wen2024autodroid}                                   & Android mobile devices           & Simplified HTML-style representation                                                               & Standard UI operations                                                     & GPT-3.5, GPT-4, and Vicuna-7B \cite{vicuna2023}                                                 & Single-agent architecture                                                          & Its use of app-specific knowledge and a multi-granularity query optimization module to reduce the computational cost                                      & \url{https://autodroid-sys.github.io/}             \\ \hline
AutoDroid-V2 \cite{wen2024autodroidv2boostingslmbasedgui} & Android Mobile Devices & Structured GUI Representations & Multi-step scripts of standard UI operations and API calls & Llama-3.1-8B \cite{dubey2024llama} & Script-based architecture. & Converts GUI task automation into a script generation problem, enhancing efficiency and task success rates. & /  \\ \hline
CoAT \cite{zhang2024androidzoochainofactionthoughtgui}              & Android mobile devices           & Screenshot-based context and semantic information                                                  & Standard UI operations                                                     & GPT-4V                                                                                          & Single-agent architecture                                                          & The integration of a chain-of-action-thought process, which explicitly maps each action to screen descriptions, reasoning steps, and anticipated outcomes & \url{https://github.com/ZhangL-HKU/CoAT}           \\ \hline
Mobile-Agent \cite{wang2024mobileagentautonomousmultimodalmobile}   & Mobile Android                   & Screenshots with icon detection                                                                    & Standard UI operations                                                     & GPT-4V with Grounding DINO \cite{liu2023grounding} and CLIP \cite{radford2021learning} for icon detection                                          & Single-agent                                                                       & Vision-centric approach that eliminates dependency on system-specific data                                                                                & \url{https://github.com/X-PLUG/MobileAgent}        \\ \hline
Mobile-Agent-v2 \cite{wang2024mobileagentv2mobiledeviceoperation}   & Mobile Android OS and Harmony OS & Screenshots with text, icon recognition, and description                                           & Standard UI operations on mobile phones                                    & GPT-4V with Grounding DINO \cite{liu2023grounding} and Qwen-VL-Int4 \cite{bai2023qwentechnicalreport}                   & Multi-agent architecture with Planning Agent, Decision Agent, and Reflection Agent & Multi-agent architecture enhances task navigation for long-sequence operations                                                                            & \url{https://github.com/X-PLUG/MobileAgent}        \\ \hline
Mobile-Experts \cite{zhang2024mobileexpertsdynamictoolenabledagent} & Mobile Android                   & Interface memory and procedural memory                                                             & Standard UI operations and code-combined tool formulation                  & VLMs                                                                                            & Multi-agent framework with double-layer planning                                   & Code-combined tool formulation method and double-layer planning mechanism for collaborative task execution                                                & /                                                \\ \hline
LiMAC \cite{christianos2024lightweightneuralappcontrol}             & Mobile Android                   & Screenshots and corresponding widget trees                                                             & Standard UI operations                                                     & Lightweight transformer and fine-tuned VLMs                                                     & Single-agent                                                                       & Balances computational efficiency and natural language understanding                                                                                      & /                                                \\ \hline
MobA \cite{zhu2024moba}                                             & Mobile Android                   & GUI structures, screenshots with annotation                                                        & Standard UI operations and API function calls                              & GPT-4                                                                                           & Two-level agent: a Global Agent and a Local Agent                                  & Two-level agent system that separates task planning and execution into two specialized agents                                                             & \url{https://github.com/OpenDFM/MobA}         
\\ \hline
Mobile-Agent-E \cite{wang2025mobileagenteselfevolvingmobileassistant} & Mobile Android & GUI screenshots, OCR for detecting text and icons & Standard UI operations and APIs & GPT-4o, Claude-3.5-Sonnet, Gemini-1.5-Pro & Hierarchical Multi-Agent System & Hierarchical multi-agent framework that separates planning from execution for improved long-term reasoning and self-evolution, enabling the system to learn reusable tips and shortcuts & \url{https://x-plug.github.io/MobileAgent}\\ \hline
\end{tabular}
}
\end{table*}

\begin{table*}[h!]
\centering
\caption{Overview of LLM-brained GUI agent frameworks on mobile platforms (Part III).\label{tab:mobile_framework3}}
\resizebox{\textwidth}{!}{ % Resize the entire figure to fit \textwidth
\begin{tabular}{p{1.5cm}|p{1.5cm}|p{2cm}|p{2.3cm}|p{2cm}|p{2.5cm}|p{3.5cm}|p{2cm}}
\hline
\textbf{Agent}                                            & \textbf{Platform}           & \textbf{Perception}                                                                                    & \textbf{Action}                                                        & \textbf{Model}                                        & \textbf{Architecture}                                                                                                                                                & \textbf{Highlight}                                                                                                                                          & \textbf{Link}   \\ \hline
ReachAgent \cite{wu2025reachagent} &  Android mobile devices &  GUI Screenshots, XML document & Standard UI operations & MobileVLM \cite{wu2024mobilevlmvisionlanguagemodelbetter} & Single-agent, two-stage training & Divides tasks into subtasks: ``Page Reaching'' (navigating to the correct screen) and ``Page Operation'' (performing actions on the screen), using RL with preference-based training to improve long-term task success. & / \\ \hline
Mobile-Agent-V \cite{wang2025mobilev} & Mobile Android & Video guidance, XML hierarchy & Standard UI operations & GPT-4o & Multi-agent system & Introduces video-guided learning, allowing the agent to acquire operational knowledge efficiently. & \url{https://github.com/X-PLUG/MobileAgent} \\ \hline
MobileSteward \cite{MobileSteward} & Mobile Android & XML layouts, Screenshots & Standard UI interactions, Code execution & GPT-4V, GPT-4o & App-oriented multi-agent framework & Introduces an app-oriented multi-agent framework with self-evolution, overcoming the complexity of cross-app interactions by dynamically recruiting specialized agents. & \url{https://github.com/XiaoMi/MobileSteward} \\ \hline
AppAgentX \cite{jiang2025appagentx} & Mobile Android & Screenshots & Standard UI operations & GPT-4o & Single-agent architecture & Introduces an evolutionary mechanism that enables dynamic learning from past interactions and replaces inefficient low-level operations with high-level actions. & \url{https://appagentx.github.io/} \\ \hline
CHOP \cite{zhou2025chop} & Mobile Android & Screenshots & Standard UI operations & GPT-4o & Multi-agent architecture & Introduces a basis subtask framework, where subtasks are predefined based on human task decomposition patterns, ensuring better executability and efficiency. & \url{https://github.com/Yuqi-Zhou/CHOP} \\ \hline
OS-Kairos \cite{cheng2025kairos} & Mobile Android & GUI screenshots & Standard UI operations & OS-Atlas-Pro-7B and GPT-4o & Single-agent with critic-in-the-loop design & Introduces an adaptive interaction framework where each GUI action is paired with a confidence score, dynamically deciding between autonomous execution and human intervention & \url{https://github.com/Wuzheng02/OS-Kairos}\\ \hline
V-Droid \cite{dai2025advancing} & Mobile Android & Android Accessibility Tree & Standard UI operations & LLaMA-3.1-8B-Instruct \cite{dubey2024llama} & Verifier-Driven Single-Agent Architecture & Introduces a novel verifier-driven architecture where the LLM does not generate actions directly but instead scores and selects from a finite set of extracted actions, improving task success rates and significantly reducing latency & /  \\ \hline
LearnAct \cite{liu2025learnactfewshotmobilegui} & Mobile Android & GUI screenshots, UI trees, and demonstration trajectories & Standard GUI actions & Gemini-1.5-Pro, UI-TARS-7B-SFT, Qwen2-VL-7B & Multi-agent & Introduces a structured, demonstration-based learning pipeline for mobile GUI agents. It addresses long-tail generalization via few-shot demonstrations, achieving substantial performance gains on complex real-world mobile tasks & \url{https://lgy0404.github.io/LearnAct} \\ \hline
AndroidGen \cite{lai2025androidgen} & Mobile Android & XML UI structure & Standard GUI actions & GLM-4-9B \cite{glm2024chatglm} / LLaMA-3-70B \cite{dubey2024llama} & Multi-module single-agent & Innovatively addresses data scarcity for Android agents through a self-improving architecture, a zero human-annotation training pipeline, and effective generalization from easy to hard tasks & \url{https://github.com/THUDM/AndroidGen} \\ \hline
Agent-Initiated Interaction \cite{kahlon2025agent} & Android Mobile & Accessibility tree and screenshots & Standard GUI operations & Gemini 1.5 & Single-agent architecture & Pioneers agent-initiated interaction in mobile UI automation & \url{https://github.com/google-research/google-research/tree/master/android_interaction} \\ \hline
Latent State Estimation \cite{bishop2024latent} & Android Mobile & Accessibility tree & Standard GUI operations & PaLM 2 & Two-module design with Reasoner and Grounder & First to formalize the estimation of latent UI states using LLMs to support UI automation & / \\ \hline

\end{tabular}
}
\end{table*}

\subsection{Mobile GUI Agents\label{sec:framework:mobile}}
The evolution of mobile GUI agents has been marked by significant advancements, leveraging multimodal models, complex architectures, and adaptive planning to address the unique challenges of mobile environments. These agents have progressed from basic interaction capabilities to sophisticated systems capable of dynamic, context-aware operations across diverse mobile applications. We first provide an overview of mobile GUI agent frameworks in Tables~\ref{tab:mobile_framework1} \ref{tab:mobile_framework2} and \ref{tab:mobile_framework3}.

\textbf{Wang \etal} \cite{wang2023enabling} pioneer the use of LLMs to enable conversational interaction with mobile UIs, establishing one of the earliest foundations for mobile GUI agents. Their approach involves directly prompting foundation models such as PaLM using structured representations of Android view hierarchies, which are transformed into HTML-like text to better align with the LLM's training distribution. The authors define and evaluate four core tasks, including Screen Summarization, Screen QA, Screen Question Generation, and Instruction-to-UI Mapping—demonstrating that strong performance can be achieved with as few as two prompt examples per task. Emphasizing practicality and accessibility, the work enables rapid prototyping without model fine-tuning, and stands out as a seminal effort in prompt-based evaluation of LLM-powered GUI agents for mobile applications.

Early efforts focused on enabling human-like GUI interactions without requiring backend system access. One such pioneering framework is \textbf{AppAgent}~\cite{zhang2023appagentmultimodalagentssmartphone}, which utilizes GPT-4V's multimodal capabilities to comprehend and respond to both visual and textual information. By performing actions like tapping and swiping using real-time screenshots and structured XML data, AppAgent can interact directly with the GUI across a variety of applications, from social media to complex image editing. Its unique approach of learning through autonomous exploration and observing human demonstrations allows for rapid adaptability to new apps, highlighting the effectiveness of multimodal capabilities in mobile agents.

Building upon this foundation, \textbf{AppAgent-V2}~\cite{li2024appagentv2advancedagent} advances the framework by enhancing visual recognition and incorporating structured data parsing. This enables precise, context-aware interactions and the ability to perform complex, multi-step operations across different applications. AppAgent-V2 also introduces safety checks to handle sensitive data and supports cross-app tasks by tracking and adapting to real-time interactions. This progression underscores the importance of advanced visual recognition and structured data processing in improving task precision and safety in real-time mobile environments.

Parallel to these developments, vision-centric approaches emerged to further enhance mobile task automation without relying on app-specific data. For instance, \textbf{Mobile-Agent}~\cite{wang2024mobileagentautonomousmultimodalmobile} leverages OCR, CLIP \cite{radford2021learning}, and Grounding DINO \cite{liu2023grounding} for visual perception. By using screenshots and visual tools, Mobile-Agent performs operations ranging from app navigation to complex multitasking, following instructions iteratively and adjusting for errors through a self-reflective mechanism. This vision-based method positions Mobile-Agent as a versatile and adaptable assistant for mobile tasks.

To address challenges in long-sequence navigation and complex, multi-app scenarios, \textbf{Mobile-Agent-v2}~\cite{wang2024mobileagentv2mobiledeviceoperation} introduces a multi-agent architecture that separates planning, decision-making, and reflection. By distributing responsibilities among three agents, this framework optimizes task progress tracking, retains memory of task-relevant information, and performs corrective actions when errors occur. Integrated with advanced visual perception tools like Grounding DINO \cite{liu2023grounding} and Qwen-VL-Int4~\cite{bai2023qwen}, Mobile-Agent-v2 showcases significant improvements in task completion rates on both Android and Harmony OS, highlighting the potential of multi-agent systems for handling complex mobile tasks.

In addition to vision-centric methods, some frameworks focus on translating GUI states into language to enable LLM-based action planning. \textbf{VisionTasker}~\cite{Song_2024} combines vision-based UI interpretation with sequential LLM task planning by processing mobile UI screenshots into structured natural language. Supported by YOLO-v8~\cite{reis2023real} and PaddleOCR\footnote{\url{https://github.com/PaddlePaddle/PaddleOCR}} for widget detection, VisionTasker allows the agent to automate complex tasks across unfamiliar apps, demonstrating higher accuracy than human operators on certain tasks. This two-stage design illustrates a versatile and adaptable framework, setting a strong precedent in mobile automation.

Similarly, \textbf{DroidBot-GPT}~\cite{wen2024droidbotgptgptpowereduiautomation} showcases an innovative approach by converting GUI states into natural language prompts, enabling LLMs to autonomously decide on action sequences. By interpreting the GUI structure and translating it into language that GPT models can understand, DroidBot-GPT generalizes across various apps without requiring app-specific modifications. This adaptability underscores the transformative role of LLMs in handling complex, multi-step tasks with minimal custom data.

To enhance action prediction and context awareness, advanced frameworks integrate perception and action systems within a multimodal LLM. \textbf{CoCo-Agent}~\cite{ma2024cocoagentcomprehensivecognitivemllm} exemplifies this by processing GUI elements like icons and layouts through its Comprehensive Event Perception and Comprehensive Action Planning  modules. By decomposing actions into manageable steps and leveraging high-quality data from benchmarks like Android in the Wild (AITW)~\cite{androidinwildlargescaledataset} and META-GUI~\cite{metaguimultimodalconversationalagents}, CoCo-Agent demonstrates its ability to automate mobile tasks reliably across varied smartphone applications.

Further advancing this integration, \textbf{CoAT}~\cite{zhang2024androidzoochainofactionthoughtgui} introduces a chain-of-action-thought process to enhance action prediction and context awareness. Utilizing sophisticated models such as GPT-4V and set-of-mark tagging, CoAT addresses the limitations of traditional coordinate-based action recognition. By leveraging the Android-In-The-Zoo (AITZ) dataset it builds, CoAT provides deep context awareness and improves both action prediction accuracy and task completion rates, highlighting its potential for accessibility and user convenience on Android platforms.

Addressing the need for efficient handling of multi-step tasks with lower computational costs, \textbf{AutoDroid}~\cite{wen2024autodroid} combines LLM-based comprehension with app-specific knowledge. Using an HTML-style GUI representation and a memory-based approach, AutoDroid reduces dependency on extensive LLM queries. Its hybrid architecture of cloud and on-device models enhances responsiveness and accessibility, making AutoDroid a practical solution for diverse mobile tasks. AutoDroid-V2 \cite{wen2024autodroidv2boostingslmbasedgui} enhances its predecessor AutoDroid, by utilizing on-device language models to generate and execute multi-step scripts for user task automation. By transforming dynamic and complex GUI elements of mobile apps into structured app documents, it achieves efficient and accurate automation without depending on cloud-based resources. The script-based approach reduces computational overhead by minimizing query frequency, thereby improving task efficiency and addressing the limitations of stepwise agents. This advancement enables privacy-preserving and scalable task automation on mobile platforms.

\textbf{MobileGPT \cite{lee2024mobilegpt}} automates tasks on Android devices using a human-like app memory system that emulates the cognitive process of task decomposition—Explore, Select, Derive, and Recall. This approach results in highly efficient and accurate task automation. Its hierarchical memory structure supports modular, reusable, and adaptable tasks and sub-tasks across diverse contexts. MobileGPT demonstrates superior performance over state-of-the-art systems in task success rates, cost efficiency, and adaptability, highlighting its potential for advancing mobile task automation.

In a more advanced distributed setting, \textbf{FedMobileAgent \cite{wang2025fedmobileagent}} employs a federated learning framework to train mobile automation agents using self-sourced data from users’ phone interactions. It addresses the high cost and privacy concerns associated with traditional human-annotated datasets by introducing Auto-Annotation, which leverages vision-language models (VLMs) to infer user intentions from screenshots and actions. The system enables decentralized training through federated learning while preserving user privacy, and its adaptive aggregation method enhances model performance under non-IID data conditions. Experimental results on several mobile benchmarks demonstrate that FedMobileAgent achieves performance comparable to human-annotated models at a fraction of the cost.

In summary, mobile GUI agents have evolved significantly, progressing from single-agent systems to complex, multi-agent frameworks capable of dynamic, context-aware operations. These innovations demonstrate that sophisticated architectures, multimodal processing, and advanced planning strategies are essential in handling the diverse challenges of mobile environments, marking significant advancements in mobile automation capabilities.

\begin{table*}[h!]
\centering
\caption{Overview of LLM-brained GUI agent frameworks on computer platforms (Part I)..\label{tab:computer_framework1}}
\resizebox{\textwidth}{!}{ % Resize the entire figure to fit \textwidth
\begin{tabular}{p{1.5cm}|p{1.5cm}|p{2cm}|p{2.3cm}|p{2cm}|p{2.5cm}|p{3.5cm}|p{2cm}}
\hline
\textbf{Agent}                          & \textbf{Platform}                                              & \textbf{Perception}                     & \textbf{Action}                                                            & \textbf{Model}                                                                                         & \textbf{Architecture}                                                                           & \textbf{Highlight}                                                                                                                                      & \textbf{Link}                                \\ \hline
UFO \cite{zhang2024ufouifocusedagentwindows}              & Windows computer            & Screenshots with annotated controls, and widget properties                                             & Standard UI operations with additional customized operations           & GPT-Vision                                            & Dual-agent architecture, consisting of a HostAgent (for application selection and global planning) and an AppAgent (for specific task execution within applications) & Its dual-agent system that seamlessly navigates and interacts with multiple applications to fulfill complex user requests in natural language on Windows OS & \url{https://github.com/microsoft/UFO}                                                               \\ \hline
UFO\textsuperscript{2} \cite{zhang2025ufo2} & Windows desktops & GUI screenshots and textual control properties list & Unified GUI–API action layer & GPT-4o (and GPT-4V, o1, Gemini-Flash); Vision grounding via OmniParser-v2 & Centralized HostAgent with application-specialized AppAgents & Transforms a conventional CUA into an OS-native, pluggable AgentOS with deep Windows integration, hybrid GUI–API actions, vision + UIA perception, speculative multi-action planning, retrieval-augmented knowledge, and a non-intrusive PiP virtual desktop & \url{https://github.com/microsoft/UFO/} \\ \hline
ScreenAgent \cite{niu2024screenagentvisionlanguagemodeldriven}      & Linux and Windows desktop        & Screenshots                                                                                        & Standard UI operations                                                     & ScreenAgent model                                                                               & Single-agent                                                                       & Integrated planning-acting-reflecting pipeline that simulates a continuous thought process                                                                & \url{https://github.com/niuzaisheng/ScreenAgent}   \\ \hline
OS-Copilot \cite{wu2024oscopilotgeneralistcomputeragents} & Linux and MacOS computer    & Unified interface that includes mouse and keyboard control, API calls, and Bash or Python interpreters & Standard UI operations, Bash and Python commands, as well as API calls & GPT-4                                                 & Multi-component architecture involving a planner, configurator, actor, and critic modules                                                                            & Self-directed learning capability, allowing it to adapt to new applications by autonomously generating and refining tools                                   & \url{https://os-copilot.github.io/}                                                                  \\ \hline
Cradle \cite{tan2024cradleempoweringfoundationagents}     & Windows computer            & Complete screen videos with Grounding DINO \cite{liu2023grounding}  and SAM \cite{kirillov2023segment} for object detection and localization               & Keyboard and mouse actions                                             & GPT-4                                                 & Modular single-agent architecture                                                                                                                                    & Its generalizability across various digital environments, allowing it to operate without relying on internal APIs                                           & \url{https://baai-agents.github.io/Cradle/}                                                          \\ \hline
Agent S \cite{agashe2024agentsopenagentic}                          & Ubuntu and Windows computer      & Screenshots and accessibility tree                                                                 & Standard UI operations and system-level controls                           & GPT-4 and Claude-3.5 Sonnet \cite{anthropic2024}                                                                      & Multi-agent architecture comprising a Manager and Worker structure                 & Experience-augmented hierarchical planning                                                                                                                & \url{https://github.com/simular-ai/Agent-S}        \\ \hline
GUI Narrator \cite{wu2024guiactionnarratordid}            & Windows computer            & High-resolution screenshots                                                                            & Standard UI operations                                                 & GPT-4 and QwenVL-7B \cite{bai2023qwentechnicalreport} & Two-stage architecture, detecting the cursor location and selecting keyframes, then generating action captions                                                       & Uses the cursor as a focal point to improve understanding of high-resolution GUI actions                                                                    & \url{https://showlab.github.io/GUI-Narrator}                                                         \\ \hline
PC Agent \cite{he2024pcagentsleepai} & Windows Computer & Screenshots and event-based tracking & Standard UI Operations & Qwen2-VL-72B-Instruct \cite{wang2024qwen2vlenhancingvisionlanguagemodels} and Molmo \cite{deitke2024molmopixmoopenweights} & A planning agent for decision-making combined with a grounding agent for executing actions. & Human cognition transfer framework, which transforms raw interaction data into cognitive trajectories to enable complex computer tasks. & \url{https://gair-nlp.github.io/PC-Agent/} \\ \hline
\end{tabular}
}
\end{table*}

\begin{table*}[h!]
\centering
\caption{Overview of LLM-brained GUI agent frameworks on computer platforms (Part II).\label{tab:computer_framework2}}
\resizebox{\textwidth}{!}{ % Resize the entire figure to fit \textwidth
\begin{tabular}{p{1.5cm}|p{1.5cm}|p{2cm}|p{2.3cm}|p{2cm}|p{2.5cm}|p{3.5cm}|p{2cm}}
\hline
\textbf{Agent}                          & \textbf{Platform}                                              & \textbf{Perception}                     & \textbf{Action}                                                            & \textbf{Model}                                                                                         & \textbf{Architecture}                                                                           & \textbf{Highlight}                                                                                                                                      & \textbf{Link}                                \\ \hline
Zero-shot Agent \cite{li2023zero}                         & Computer                    & HTML code and DOM                                                                                      & Standard UI operations                                                 & PaLM-2 \cite{anil2023palm2technicalreport}            & Single-agent                                                                                                                                                         & Zero-shot capability in performing computer control tasks                                                                                                   & \url{https://github.com/google-research/google-research/tree/master/zero_shot_structured_reflection} 
\\ \hline
PC-Agent \cite{liu2025pc} & Windows computers & UI tree, Screenshots & Standard UI operations & GPT-4o & Hierarchical Multi-Agent & PC-Agent's hierarchical multi-agent design enables efficient decomposition of complex PC tasks. Its Active Perception Module enhances fine-grained GUI understanding by combining accessibility structures, OCR, and intention grounding. & \url{https://github.com/X-PLUG/MobileAgent/tree/main/PC-Agent} \\ \hline
PwP \cite{aggarwal2025programming} & VSCode-based IDE in Computers & Screenshots, File system access, Terminal outputs & Standard UI interactions, File operations, Bash commands, Tools in VSCode & GPT-4o, Claude-3.5 Sonnet, Gemini-1.5 & Single-agent architecture & Shifts software engineering agents from API-based tool interactions to direct GUI-based computer use, allowing agents to interact with an IDE as a human developer would. & \url{https://programmingwithpixels.com} \\ \hline
COLA \cite{zhao2025cola} & Windows computers & GUI structure, properties and screenshots & Standard UI operations and system APIs & GPT-4o & Hierarchical Multi-Agent & A dynamic task scheduling mechanism with a plug-and-play agent pool, enabling adaptive handling of GUI tasks & \url{https://github.com/Alokia/COLA-demo} \\ \hline
STEVE \cite{lu2025steve} & Windows Desktop & GUI screenshots and A11y Tree & Standard UI operations & Qwen2-VL \cite{wang2024qwen2vlenhancingvisionlanguagemodels} and GPT-4o & Single-agent & Introduces a scalable step verification pipeline using GPT-4o to generate binary labels for agent actions, and applies KTO optimization to incorporate both positive and negative actions into agent learning & \url{https://github.com/FanbinLu/STEVE} \\ \hline
TaskMind \cite{yin2025operation} & Windows Computer & Standard GUI actions & GPT-3.5 / GPT-4 & Single-agent architecture & Introduces a novel task graph representation with cognitive dependencies, enabling LLMs to better generalize demonstrated GUI tasks & \url{https://github.com/Evennaire/TaskMind} \\ \hline

\end{tabular}
}
\end{table*}

\subsection{Computer GUI Agents\label{sec:framework:computer}}

\begin{figure}[t]
    \centering
    \includegraphics[width=\columnwidth]{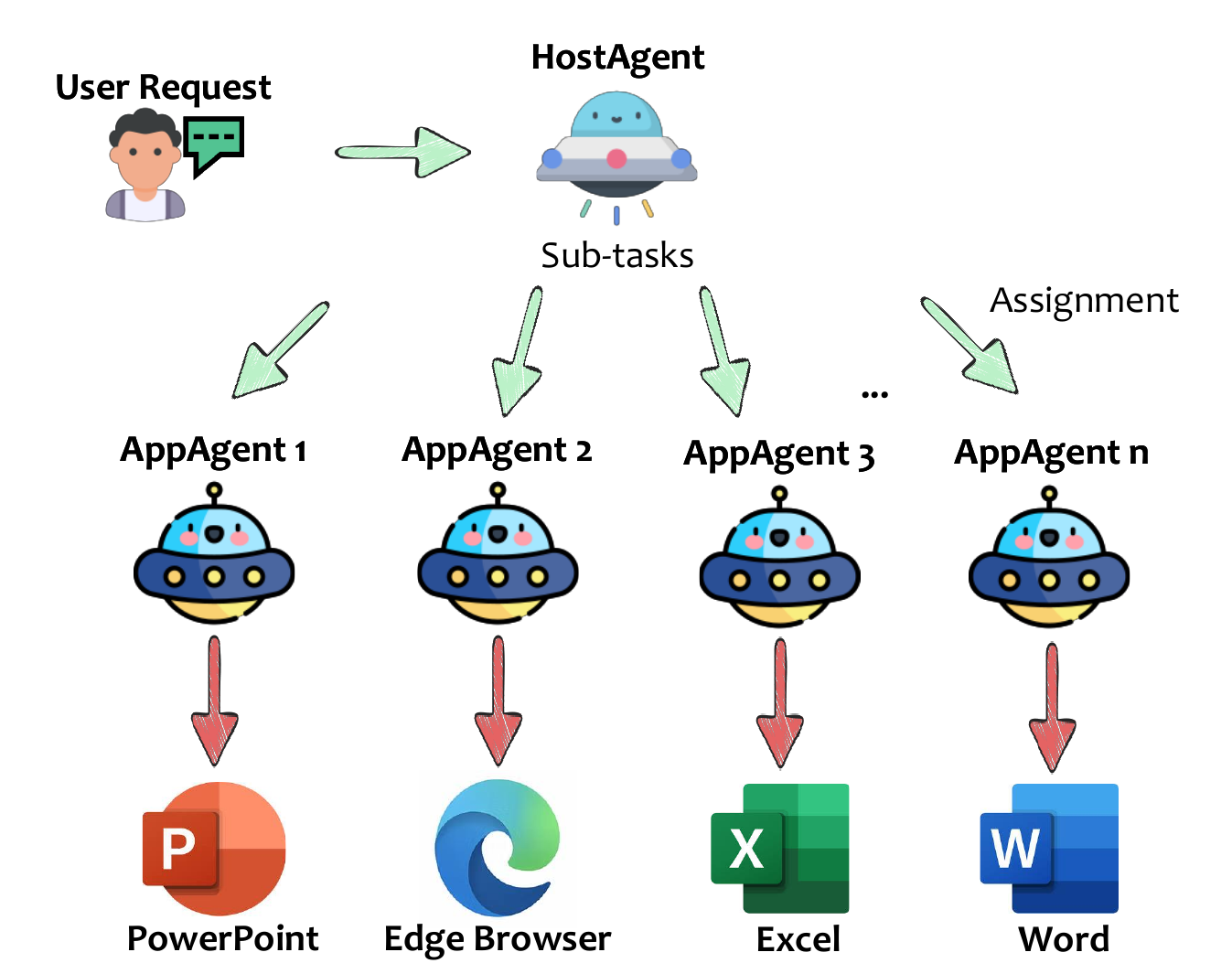}
    \vspace{-1.5em}
    \caption{The multi-agent architecture employed in UFO \cite{zhang2024ufouifocusedagentwindows}. Figure adapted from the original paper.}
    \label{fig:ufo}
\end{figure}

Computer GUI agents have evolved to offer complex automation capabilities across diverse operating systems, addressing challenges such as cross-application interaction, task generalization, and high-level task planning. They have led to the development of sophisticated frameworks capable of handling complex tasks across desktop environments. These agents have evolved from simple automation tools to intelligent systems that leverage multimodal inputs, advanced architectures, and adaptive learning to perform multi-application tasks with high efficiency and adaptability.  We provide an overview of computer GUI agent frameworks in Table~\ref{tab:computer_framework1} and \ref{tab:computer_framework2}.

One significant development in this area is the introduction of multi-agent architectures that enhance task management and execution. For instance, the UI-Focused Agent, \textbf{UFO} \cite{zhang2024ufouifocusedagentwindows} represents a pioneering framework specifically designed for the Windows operating system. UFO redefines UI-focused automation through its advanced dual-agent architecture, leveraging GPT-Vision to interpret GUI elements and execute actions autonomously across multiple applications. The framework comprises a HostAgent, responsible for global planning, task decomposition, and application selection, and an AppAgent, tasked with executing assigned subtasks within individual applications, as illustrated in Figure~\ref{fig:ufo}. This centralized structure enables UFO to manage complex, multi-application workflows such as aggregating information and generating reports. Similar architectural approach has also been adopted by other GUI agent frameworks \cite{agentsea2024, zhu2024moba, zhang2024mobileexpertsdynamictoolenabledagent}. By incorporating safeguards and customizable actions, UFO ensures efficiency and security when handling intricate commands, positioning itself as a cutting-edge assistant for Windows OS. Its architecture, exemplifies dynamic adaptability and robust task-solving capabilities across diverse applications, demonstrating the potential of multi-agent systems in desktop automation.

\begin{figure}[t]
    \centering
    \includegraphics[width=\columnwidth]{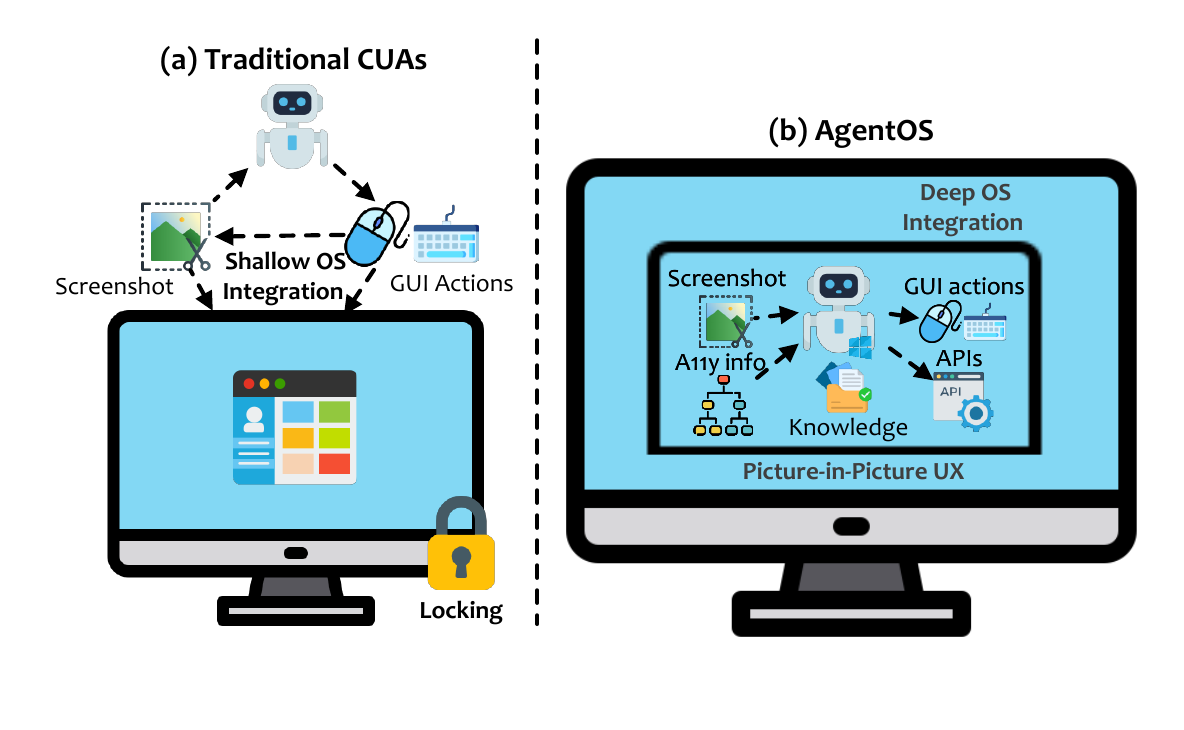}
    \vspace{-1.5em}
    \caption{The comparison of traditional CUAs and the Desktop AgentOS UFO\textsuperscript{2}. Figure adapted from the original paper.}
    \label{fig:agentos}
\end{figure}

\textbf{UFO\textsuperscript{2}} \cite{zhang2025ufo2}, the successor to UFO, elevates GUI automation from a vision-only prototype to a deeply integrated, Windows-native \emph{AgentOS} (Figure~\ref{fig:agentos}). It coordinates tasks through a centralized \emph{HostAgent}, which delegates subtasks to application-specialized \emph{AppAgents}. A hybrid perception pipeline that fuses Windows UI Automation (UIA) metadata with OmniParser-v2 visual grounding delivers robust control identification even for custom widgets. Via a unified GUI–API action layer, AppAgents preferentially invoke high-level application APIs and fall back to pixel-level clicks only when necessary, cutting both latency and brittleness. A picture-in-picture virtual desktop cleanly isolates agent execution from the user’s main session, enabling non-intrusive multitasking. Runtime performance is further boosted by retrieval-augmented help documents and execution logs, coupled with speculative multi-action planning that executes several steps per single LLM invocation. Tested on 20+ real Windows applications, \textbf{UFO\textsuperscript{2}} exceeds Operator \cite{openai2025operator} and other CUAs by more than 10 percentage points in success rate while halving LLM calls. Because the framework is model-agnostic, swapping GPT-4o for a stronger LLM such as \emph{o1} yields additional gains without code changes.

Building upon the theme of adaptability and generalist capabilities, \textbf{Cradle}~\cite{tan2024cradleempoweringfoundationagents} pushes the boundaries of general computer control by utilizing VLMs for interacting with various software, ranging from games to professional applications, without the need for API access. Cradle employs GPT-4o to interpret screen inputs and perform low-level actions, making it versatile across different types of software environments. Its six-module structure, covering functions such as information gathering and self-reflection, enables the agent to execute tasks, reason about actions, and utilize past interactions to inform future decisions. Cradle's capacity to function in dynamic environments, including complex software, marks it as a significant step toward creating generalist agents with broad applicability across desktop environments.

Extending the capabilities of computer GUI agents to multiple operating systems, \textbf{OS-Copilot}~\cite{wu2024oscopilotgeneralistcomputeragents} introduces a general-purpose framework designed to operate across Linux and macOS systems. Its notable feature, \textbf{FRIDAY}, showcases the potential of self-directed learning by adapting to various applications and performing tasks without explicit training for each app. Unlike application-specific agents, FRIDAY integrates APIs, keyboard and mouse controls, and command-line operations, creating a flexible platform that can autonomously generate and refine tools as it interacts with new applications. OS-Copilot's ability to generalize across unseen applications, validated by its performance on the GAIA benchmark, provides a foundational model for OS-level agents capable of evolving in complex environments. This demonstrates promising directions for creating adaptable digital assistants that can handle diverse desktop environments and complex task requirements.

In the emerging field of LLM-powered GUI agents for desktop environments, Programming with Pixels (PwP) \cite{aggarwal2025programming} introduces a compelling alternative to traditional tool-based software engineering agents. Rather than relying on predefined API calls, PwP enables agents to interact directly with an IDE using visual perception, keyboard inputs, and mouse clicks, mimicking the way human developers operate within an IDE. This approach allows for generalization beyond predefined APIs, providing a highly expressive environment where agents can execute a wide range of software engineering tasks, including debugging, UI generation, and code editing. Evaluations conducted on PwP-Bench demonstrate that computer-use agents, despite lacking direct access to structured APIs, can match or even surpass traditional tool-based approaches in certain scenarios.

In summary, computer GUI agents have evolved significantly, progressing from single-task automation tools to advanced multi-agent systems capable of performing complex, multi-application tasks and learning from interactions. Frameworks like UFO, Cradle, and OS-Copilot illustrate the potential of adaptable, generalist agents in desktop automation, paving the way for the evolution of more intelligent and versatile AgentOS frameworks.

\begin{table*}[h!]
\centering
\caption{Overview of LLM-brained cross-platform GUI agent frameworks.\label{tab:cross_framework1}}
\resizebox{\textwidth}{!}{ % Resize the entire figure to fit \textwidth
\begin{tabular}{p{1.5cm}|p{1.5cm}|p{2cm}|p{2.3cm}|p{2cm}|p{2.5cm}|p{3.5cm}|p{2cm}}
\hline
\textbf{Agent}                          & \textbf{Platform}                                              & \textbf{Perception}                     & \textbf{Action}                                                            & \textbf{Model}                                                                                         & \textbf{Architecture}                                                                           & \textbf{Highlight}                                                                                                                                      & \textbf{Link}                                \\ \hline
AutoGLM \cite{liu2024autoglm}           & Web and Mobile Android                                         & Screenshots with SoM annotation and OCR & Standard UI operations, Native API interactions, and AI-driven actions     & ChatGLM \cite{glm2024chatglm}                                                                                                & Single-agent architecture                                                                       & Self-evolving online curriculum RL framework, which enables continuous improvement by interacting with real-world environments & \url{https://xiao9905.github.io/AutoGLM/}      \\ \hline
TinyClick \cite{pawlowski2024tinyclick} & Web, Mobile, and Windows platforms                             & GUI screenshots                         & Standard UI operations, Native API interactions, and AI-driven actions     & Florence-2-Base VLM \cite{xiao2023florence2advancingunifiedrepresentation}                             & Single-agent, with single-turn tasks                                                            & Compact size (0.27B parameters) with high performance                                                                                                   & \url{https://huggingface.co/Samsung/TinyClick} \\ \hline
OSCAR \cite{wang2024oscar}              & Desktop and Mobile                                             & Screenshots                             & Standard UI operations                                                     & GPT-4                                                                                                  & Single-agent architecture                                                                       & Ability to adapt to real-time feedback and dynamically adjust its actions                                                                               & /                                            \\ \hline
AgentStore \cite{jia2024agentstore} & Desktop and mobile environments & GUI structures and properties, accessibility   trees, screenshots and terminal output \etc & Standard UI operations, API calls             & GPT-4o and InternVL2-8B \cite{chen2024internvl} & Multi-agent architecture              & Dynamically integrate a wide variety of   heterogeneous agents, enabling both specialized and generalist capabilities & \url{https://chengyou-jia.github.io/AgentStore-Home/} \\ \hline
MMAC-Copilot \cite{song2024mmac}        & Windows OS Desktop, mobile applications, and game environments & Screenshots                             & Standard UI operations, Native APIs, and Collaborative multi-agent actions & GPT-4V, SeeClick \cite{cheng2024seeclickharnessingguigrounding} and Genimi Vision for different agents & Multi-agent architecture with Planner, Programmer, Viewer, Mentor, Video Analyst, and Librarian & Collaborative multi-agent architecture where agents specialize in specific tasks                                                                        & /                                            \\ \hline
AGUVIS~\cite{xu2024aguvisunifiedpurevision} & Web, desktop, and mobile & Image-based observations & Standard UI operations & Fine-tuned Qwen2-VL \cite{wang2024qwen2vlenhancingvisionlanguagemodels} & Single-agent architecture & Pure vision-based approach for GUI interaction, bypassing textual UI representations and enabling robust cross-platform generalization & \url{https://aguvis-project.github.io} \\ \hline
Ponder \& Press~\cite{wang2024ponderpressadvancing} & Web, Android, iOS Mobile, Windows, and macOS & Purely visual inputs & Standard UI operations & GPT-4o and Claude 3.5 Sonnet for high-level task decomposition, a fine-tuned Qwen2-VL-Instruct \cite{wang2024qwen2vlenhancingvisionlanguagemodels} for GUI element grounding & Divide-and-conquer architecture & Purely vision-based GUI agent that does not require non-visual inputs & \url{https://invinciblewyq.github.io/ponder-press-page/} \\ \hline
InfiGUIAgent \cite{liu2025infiguiagentmultimodalgeneralistgui} & Mobile, Web, Desktop & Raw screenshots & Standard UI operations. & Qwen2-VL-2B \cite{wang2024qwen2vlenhancingvisionlanguagemodels} & Single-agent architecture enhanced by hierarchical reasoning. & Introduces native reasoning skills, such as hierarchical and expectation-reflection reasoning, enabling advanced and adaptive task handling. & \url{https://github.com/Reallm-Labs/InfiGUIAgent} \\  \hline
Learn-by-Interact \cite{su2025learnbyinteractdatacentricframeworkselfadaptive} & Web, code development, and desktops & GUI screenshots with SoM and accessibility tree & Standard UI interactions and code execution & Claude-3.5-Sonnet, Gemini-1.5-Pro \cite{team2023gemini}, CodeGemma-7B, CodeStral-22B & Multi-agent & Introduces a fully autonomous data synthesis process, eliminating the need for human-labeled agentic data & / \\  \hline
CollabUIAgents \cite{he2025enhancing} & Mobile Android, Web & Screenshots, UI trees & Standard UI operations & Qwen2-7B \cite{wang2024qwen2vlenhancingvisionlanguagemodels}, GPT-4 & Multi-agent system & A multi-agent reinforcement learning framework that introduces a Credit Re-Assignment (CR) strategy, using LLMs instead of environment-specific rewards to enhance performance and generalization. & \url{https://github.com/THUNLP-MT/CollabUIAgents} \\ \hline
Agent S2 \cite{agashe2025agent} & Ubuntu, Windows, Android & GUI screenshot & Standard UI operations and system APIs & Claude-3.7-Sonnet, Claude-3.5-Sonnet, GPT-4o (for Manager and Worker roles), UI-TARS-72B-DPO, Tesseract OCR, and UNO (for grounding experts) & Compositional multi-agent architecture with a Manager for planning, a Worker for execution, and a Mixture of Grounding experts & Features a Mixture of Grounding technique and Proactive Hierarchical Planning, enabling more accurate grounding and adaptive replanning in long-horizon tasks & \url{https://github.com/simular-ai/Agent-S} \\ \hline
GuidNav \cite{hu2025guidingvlmagentsprocess} & Android and Web & GUI screenshots & Standard UI operations and system APIs & GPT-4o, Gemini 2.0 Flash, Qwen-VL-Plus & Single-agent & Introduces a novel process reward model that provides fine-grained, step-level feedback to enhance GUI task accuracy and success & / \\ \hline
ScaleTrack \cite{huang2025scaletrackscalingbacktrackingautomated} & Web, Android Mobile, and Desktop Computers & GUI screenshots & Standard GUI operations & Qwen2-VL-7B & Single-agent & First GUI agent framework to introduce backtracking—learning not only the next action but also historical action sequences & / \\ \hline

\end{tabular}
}
\end{table*}

\subsection{Cross-Platform GUI Agents\label{sec:framework:cross}}
Cross-platform GUI agents have emerged as versatile solutions capable of interacting with various environments, from desktop and mobile platforms to more complex systems. These frameworks prioritize adaptability and efficiency, leveraging both lightweight models and multi-agent architectures to enhance cross-platform operability. In this subsection, we first  We overview cross-platform GUI agent frameworks in Table~\ref{tab:cross_framework1}, then explore key frameworks that exemplify the advancements in cross-platform GUI automation.

A significant stride in this domain is represented by \textbf{AutoGLM}~\cite{liu2024autoglm}, which bridges the gap between web browsing and Android control by integrating large multimodal models for seamless GUI interactions across platforms. AutoGLM introduces an Intermediate Interface Design that separates planning and grounding tasks, improving dynamic decision-making and adaptability. By employing a self-evolving online curriculum with reinforcement learning, the agent learns incrementally from real-world feedback and can recover from errors. This adaptability and robustness make AutoGLM ideal for real-world deployment in diverse user applications, setting a new standard in cross-platform automation and offering promising directions for future research in foundation agents.

While some frameworks focus on integrating advanced models for cross-platform interactions, others emphasize efficiency and accessibility. \textbf{TinyClick}~\cite{pawlowski2024tinyclick} addresses the need for lightweight solutions by focusing on single-turn interactions within GUIs. Utilizing the Florence-2-Base Vision-Language Model, TinyClick executes tasks based on user commands and screenshots with only 0.27 billion parameters. Despite its compact size, it achieves high accuracy—73\% on Screenspot~\cite{cheng2024seeclickharnessingguigrounding} and 58.3\% on OmniAct~\cite{kapoor2024omniactdatasetbenchmarkenabling} —outperforming larger multimodal models like GPT-4V while maintaining efficiency. Its multi-task training and MLLM-based data augmentation enable precise UI element localization, making it suitable for low-resource environments and addressing latency and resource constraints in UI grounding and action execution.

In addition to lightweight models, multi-agent architectures play a crucial role in enhancing cross-platform GUI interactions. \textbf{OSCAR}~\cite{wang2024oscar} exemplifies this approach by introducing a generalist GUI agent capable of autonomously navigating and controlling both desktop and mobile applications. By utilizing a state machine architecture, OSCAR dynamically handles errors and adjusts its actions based on real-time feedback, making it suitable for automating complex workflows guided by natural language. The integration of standardized OS controls, such as keyboard and mouse inputs, allows OSCAR to interact with applications in a generalized manner, improving productivity across diverse GUI environments. Its open-source design promotes broad adoption and seamless integration, offering a versatile tool for cross-platform task automation and productivity enhancement.

Expanding on the concept of multi-agent systems, \textbf{AgentStore}~\cite{jia2024agentstore} introduces a flexible and scalable framework for integrating heterogeneous agents to automate tasks across operating systems. The key feature of AgentStore is the MetaAgent, which uses the innovative AgentToken strategy to dynamically manage a growing number of specialized agents. By enabling dynamic agent enrollment, the framework fosters adaptability and scalability, allowing both specialized and generalist capabilities to coexist. This multi-agent architecture supports diverse platforms, including desktop and mobile environments, leveraging multimodal perceptions such as GUI structures and system states. AgentStore's contributions highlight the importance of combining specialization with generalist capabilities to overcome the limitations of previous systems.

Further advancing cross-platform GUI interaction, \textbf{MMAC-Copilot}~\cite{song2024mmac} employs a multi-agent, multimodal approach to handle tasks across 3D gaming, office, and mobile applications without relying on APIs. By utilizing specialized agents like Planner, Viewer, and Programmer, MMAC-Copilot collaborates to adapt to the complexities of visually rich environments. Using GPT-4V for visual recognition and OCR for text analysis, it achieves high task completion rates in visually complex environments. The framework's integration with VIBench, a benchmark for non-API applications, underscores its real-world relevance and adaptability. MMAC-Copilot's robust foundation for dynamic interaction across platforms extends applications to industries like gaming, healthcare, and productivity.

\textbf{AGUVIS}~\cite{xu2024aguvisunifiedpurevision} leverages a pure vision approach to automate GUI interactions, overcoming limitations of text-based systems like HTML or accessibility trees. Its platform-agnostic design supports web, desktop, and mobile applications while reducing inference costs. AGUVIS employs a two-stage training process: the first focuses on GUI grounding, and the second integrates planning and reasoning within a unified model. This approach delivers state-of-the-art performance in both offline and online scenarios, streamlining decision-making and execution. 

\textbf{Agent S2 \cite{agashe2025agent}} builds upon its predecessor, \textbf{Agent S \cite{agashe2024agentsopenagentic}}, by introducing a hierarchical and compositional framework for GUI agents that integrates generalist models with specialized grounding modules. Departing from monolithic architectures, it employs a Mixture of Grounding (MoG) strategy to delegate fine-grained grounding tasks to expert modules, and adopts Proactive Hierarchical Planning (PHP) to dynamically revise action plans based on evolving observations. Relying solely on GUI screenshots, Agent S2 generalizes effectively across Ubuntu, Windows, and Android platforms. It demonstrates strong scalability and consistently outperforms larger monolithic models by strategically distributing cognitive responsibilities. The design of Agent S2 underscores the advantages of modular architectures for handling long-horizon, high-fidelity GUI interactions.

In summary, cross-platform GUI agents exemplify the future of versatile automation, offering solutions ranging from lightweight models like TinyClick to sophisticated multi-agent systems such as MMAC-Copilot. Each framework brings unique innovations, contributing to a diverse ecosystem of GUI automation tools that enhance interaction capabilities across varying platforms, and marking significant advancements in cross-platform GUI automation.

\subsection{Takeaways\label{sec:framework:takeaways}}

The landscape of GUI agent frameworks has seen notable advancements, particularly in terms of multi-agent architectures, multimodal inputs, and enhanced action sets. These developments are laying the groundwork for more versatile and powerful agents capable of handling complex, dynamic environments. Key takeaways from recent advancements include:

\begin{enumerate}
    \item \textbf{Multi-Agent Synergy:} Multi-agent systems, such as those in UFO \cite{zhang2024ufouifocusedagentwindows} and MMAC-Copilot \cite{song2024mmac}, represent a significant trend in GUI agent development. By assigning specialized roles to different agents within a framework, multi-agent systems can enhance task efficiency, adaptability, and overall performance. As agents take on more complex tasks across diverse platforms, the coordinated use of multiple agents is proving to be a powerful approach, enabling agents to handle intricate workflows with greater precision and speed.

    \item \textbf{Multimodal Input Benefits:} While some agents still rely solely on text-based inputs (\eg DOM structures or HTML), incorporating visual inputs, such as screenshots, has shown clear performance advantages. Agents like WebVoyager \cite{he2024webvoyagerbuildingendtoendweb} and SeeAct \cite{zheng2024gpt4visiongeneralistwebagent} highlight how visual data, combined with textual inputs, provides a richer representation of the environment state, helping agents make better-informed decisions. This integration of multimodal inputs is essential for accurate interpretation in visually complex or dynamic environments where text alone may not capture all necessary context.

    \item \textbf{Expanding Action Sets Beyond UI Operations:} Recent agents have expanded their action sets beyond standard UI operations to include API calls and AI-driven actions, as seen in Hybrid Agent \cite{song2024beyond} and AutoWebGLM \cite{lai2024autowebglmbootstrapreinforcelarge}. Incorporating diverse actions allows agents to achieve higher levels of interaction and task completion, particularly in environments where data can be directly retrieved or manipulated through API calls. This flexibility enhances agent capabilities, making them more efficient and adaptable across a wider range of applications.

    \item \textbf{Emerging Techniques for Improved Decision-Making:} Novel approaches such as world models in WMA \cite{chae2024webagentsworldmodels}  and search-based strategies in Search-Agent \cite{koh2024tree} represent promising directions for more advanced decision-making. World models allow agents to simulate action outcomes, reducing unnecessary interactions and improving efficiency, especially in long-horizon tasks. Similarly, search-based algorithms like best-first and MCTS help agents explore action pathways more effectively, enhancing their adaptability in complex, real-time environments.

    \item \textbf{Toward Cross-Platform Generalization:} Cross-platform frameworks, such as AutoGLM \cite{liu2024autoglm} and OSCAR \cite{wang2024oscar}, underscore the value of generalizability in GUI agent design. These agents are pioneering efforts to create solutions that work seamlessly across mobile, desktop, and web platforms, moving closer to the goal of a one-stop GUI agent that can operate across multiple ecosystems. Cross-platform flexibility will be crucial for agents that aim to assist users consistently across their digital interactions.

    \item \textbf{Pure Vision-Based Agent:} To enable universal GUI control, pure vision-based frameworks have emerged as a prominent solution. These agents rely solely on screenshots for decision-making, eliminating the need for access to metadata such as widget trees or element properties. Notable work like AGUVIS~\cite{xu2024aguvisunifiedpurevision} exemplifies this approach. While pure vision-based methods offer greater generalizability and bypass system API limitations, they require strong ``grounding'' capabilities to accurately locate and interact with UI elements—an ability often lacking in many foundational models. Fine-tuning models specifically for visual grounding and GUI understanding, or integrating GUI parsing techniques like OmniParser~\cite{lu2024omniparserpurevisionbased}, can address this challenge and enhance the agent's ability to perform precise interactions.

\end{enumerate}

The field of GUI agents is moving towards multi-agent architectures, multimodal capabilities, diverse action sets, and novel decision-making strategies. These innovations mark significant steps toward creating intelligent, adaptable agents capable of high performance across varied and dynamic environments. The future of GUI agents lies in the continued refinement of these trends, driving agents towards broader applicability and more sophisticated, human-like interactions across platforms.

\section{Data for Optimizing LLM-Brained GUI Agents\label{sec:data}}
In the previous section, we explored general frameworks for LLM-brained GUI agents, most of which rely on foundational LLMs such as GPT-4V and GPT-4o. However, to elevate these agents' performance and efficiency, optimizing their ``brain'', the underlying model is crucial. Achieving this often involves fine-tuning foundational models using large-scale, diverse, and high-quality contextual GUI datasets \cite{li2024effects}, which are specifically curated to enable these models to excel in GUI-specific tasks. Collecting such datasets, particularly those rich in GUI screenshots, metadata, and interactions, necessitates an elaborate process of data acquisition, filtering, and preprocessing, each requiring substantial effort and resources \cite{chen2024data}.

As GUI agents continue to gain traction, researchers have focused on assembling datasets that represent a broad spectrum of platforms and capture the diverse intricacies of GUI environments. These datasets are pivotal in training models that can generalize effectively, thanks to their coverage of varied interfaces, workflows, and user interactions. To ensure comprehensive representation, innovative methodologies have been employed to collect and structure these data assets. In the sections that follow, we detail an end-to-end pipeline for data collection and processing tailored to training GUI-specific LLMs. We also examine significant datasets from various platforms, providing insights into their unique features, the methodologies used in their creation, and their potential applications in advancing the field of LLM-brained GUI agents.

\subsection{Data Collection\label{sec:data:collection}}

Data is pivotal in training a purpose-built GUI agent, yet gathering it requires substantial time and effort due to the task's complexity and the varied environments involved.

\subsubsection{Data Composition and Sources\label{sec:data:collection:source}}

The essential data components for GUI agent training closely align with the agent's perception and inference requirements discussed in Sections~\ref{sec:agent_foundation:env:state} and~\ref{sec:agent_foundation:inference}. At a high level, this data comprises:
\begin{enumerate} 
    \item \textbf{User Instructions:} These provide the task's overarching goal, purpose, and specific details, typically in natural language, offering a clear target for the agent to accomplish, \eg ``change the font size of all text to 12''.
    \item \textbf{Environment Perception:} This typically includes GUI screenshots, often with various visual augmentations, as well as optional supplementary data like widget trees and UI element properties to enrich the context. This should encompass both the static assessment of environment states (Section~\ref{sec:agent_foundation:env:state}) and the dynamic environment feedback that captures post-action changes (Section~\ref{sec:agent_foundation:env:feedback}), thereby providing sufficient contextual information.
    \item \textbf{Task Trajectory:} This contains the critical action sequence required to accomplish the task, along with supplementary information, such as the agent's plan. A trajectory usually involves multiple steps and actions to navigate through the task. 
\end{enumerate}
While user instructions and environmental perception serve as the model's input, the expected model output is the task trajectory. This trajectory's action sequence is then grounded within the environment to complete the task.

For \textbf{user instructions}, it is crucial to ensure that they are realistic and reflective of actual user scenarios. Instructions can be sourced in several ways: \textit{(i)} directly from human designers, who can provide insights based on real-world applications; \textit{(ii)} extracted from existing, relevant datasets if suitable data is available; \textit{(iii)} sourcing from public materials, such as websites, application help documentation, and other publicly available resources; and \textit{(iv)} generated by LLMs, which can simulate a broad range of user requests across different contexts. Additionally, LLMs can be employed for data augmentation \cite{ding2024data}, increasing both the quality and diversity of instructions derived from the original data.

For gathering \textbf{environment perception} data, various toolkits—such as those discussed in Section~\ref{sec:agent_foundation:env:state}—can be used to capture the required GUI data. This can be done within an environment emulator (\eg Android Studio Emulator\footnote{\url{https://developer.android.com/studio}}, Selenium WebDriver\footnote{\url{https://www.selenium.dev/}}, Windows Sandbox\footnote{\url{https://learn.microsoft.com/en-us/windows/security/application-security/application-isolation/windows-sandbox/windows-sandbox-overview}}) or by directly interfacing with a real environment to capture the state of GUI elements, including screenshots, widget trees, and other metadata essential for the agent's operation.

Collecting \textbf{task trajectories}, which represent the agent's action sequence to complete a task, is often the most challenging aspect. Task trajectories need to be accurate, executable, and well-validated. Collection methods include \textit{(i)} using programmatically generated scripts, which define action sequences for predefined tasks, providing a highly controlled data source; \textit{(ii)} employing human annotators, who complete tasks in a crowdsourced manner with each step recorded, allowing for rich, authentic action data; and \textit{(iii)} leveraging model or agent bootstrapping \cite{tan2024large}, where an existing LLM or GUI agent attempts to complete the task and logs its actions, though this method may require additional validation due to potential inaccuracies. All these methods demand considerable effort, reflecting the complexities of gathering reliable, task-accurate data for training GUI agents.

\subsubsection{Collection Pipeline\label{sec:data:collection:pipeline}}
\begin{figure*}[t]
    \centering
    \includegraphics[width=\textwidth]{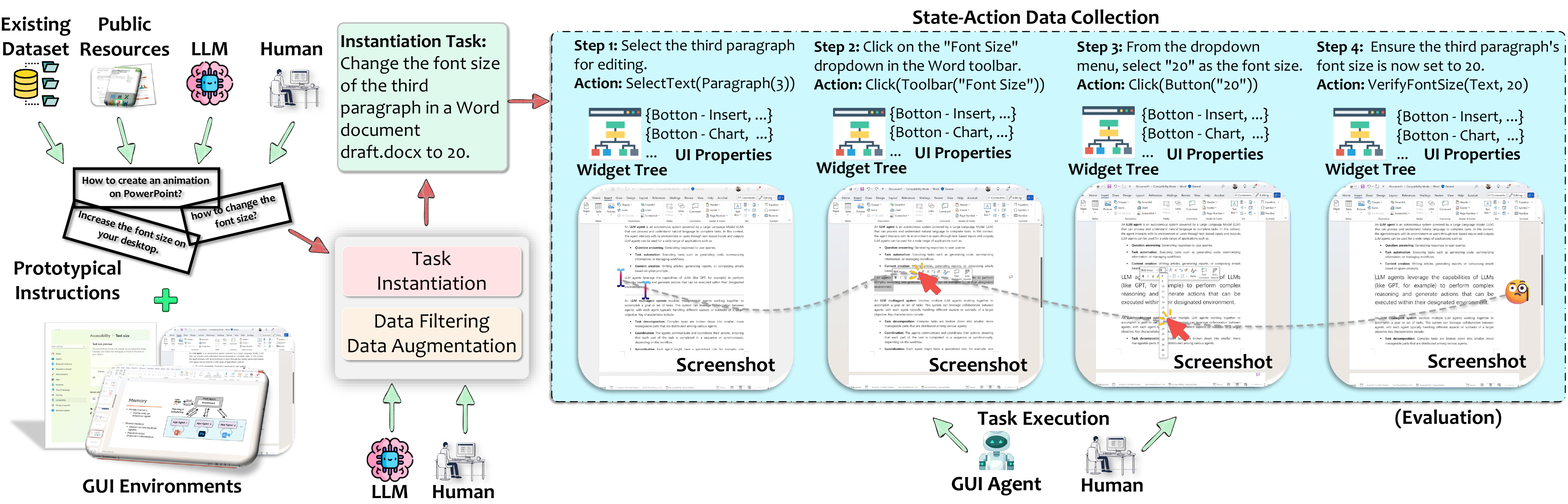}
    % \vspace{-2em}
    \caption{A complete pipeline for data collection for training a GUI agent model.}
    \label{fig:data_collection}
    % \vspace{-2em}
\end{figure*}

Figure~\ref{fig:data_collection} presents a complete pipeline for data collection aimed at training a GUI agent model. The process begins with gathering initial user instructions, which may come from various aforementioned sources. These instructions are typically prototypical, not yet tailored or grounded to a specific environment \cite{liu2024visualagentbenchlargemultimodalmodels}. For instance, an instruction like ``how to change the font size?'' from a general website lacks specificity and doesn't align with the concrete requests a user might make within a particular application. To address this, an instantiation step is required \cite{liu2024visualagentbenchlargemultimodalmodels}, where instructions are contextualized within a specific environment, making them more actionable. For example, the instruction might be refined to ``Change the font size of the third paragraph in a Word document of \texttt{draft.docx} to 20.'', giving it a clear, environment-specific goal. This instantiation process can be conducted either manually by humans or programmatically with an LLM.

Following instantiation, instructions may undergo a filtering step to remove low-quality data, ensuring only relevant and actionable instructions remain. Additionally, data augmentation techniques can be applied to expand and diversify the dataset, improving robustness. Both of these processes can involve human validation or leverage LLMs for efficiency.

Once instruction refinement is complete, task trajectories and environment perceptions are collected simultaneously. As actions are performed within the environment, each step is logged, providing a record of the environment's state and the specific actions taken. After a full task trajectory is recorded, an evaluation phase is necessary to identify and remove any failed or inaccurate sequences, preserving the quality of the dataset. By iterating this pipeline, a high-quality dataset of GUI agent data can be compiled, which is crucial for training optimized models. 

In the following sections, we review existing GUI agent datasets across various platforms, offering insights into current practices and potential areas for improvement.

\begin{table*}[t]
\centering
\caption{Overview of datasets for optimizing LLMs tailored for web GUI agents.\label{tab:web_data1}}
\resizebox{\textwidth}{!}{ % Resize the entire figure to fit \textwidth
\begin{tabular}{p{1.5cm}|p{1cm}|p{2cm}|p{2.3cm}|p{2cm}|p{1.5cm}|p{2.5cm}|p{2cm}}
\hline
\textbf{Dataset} & \textbf{Platform} & \textbf{Source} & \textbf{Content} & \textbf{Scale} & \textbf{Collection Method} & \textbf{Highlight} & \textbf{Link} \\\hline

Mind2Web \cite{mind2webgeneralistagentweb} & Web & Crowdsourced & Task descriptions, action sequences, webpage snapshots & 2,350 tasks from 137 websites & Human demonstrations & Develops generalist web agents with diverse user interactions on real-world websites & \url{https://osu-nlp-group.github.io/Mind2Web/} \\\hline

Mind2Web-Live~\cite{pan2024webcanvasbenchmarkingwebagents} & Web & Sampled and re-annotated from the Mind2Web~\cite{mind2webgeneralistagentweb} & Textual task descriptions, intermediate evaluation states, action sequences, and metadata, GUI screenshots & 542 tasks, with 4,550 detailed annotation steps. & Annotated by human experts. & Emphasis on dynamic evaluation using ``key nodes'', which represent critical intermediate states in web tasks. & \url{https://huggingface.co/datasets/iMeanAI/Mind2Web-Live} \\ \hline

WebVLN \cite{Chen_Pitawela_Zhao_Zhou_Chen_Wu_2024} & Web & Human-designed, LLM-generated & Text instructions, plans, GUI screenshots, HTML content & 8,990 navigation paths, 14,825 QA pairs & WebVLN simulator, LLM-generated QA pairs & Vision-and-language navigation for human-like web browsing & \url{https://github.com/WebVLN/WebVLN} \\\hline

WebLINX \cite{lù2024weblinxrealworldwebsitenavigation} &	Web &	From human experts&	Conversational interactions, action sequences, DOM and screenshots	&2,337 demonstrations with over 100,000 interactions &	Annotated by human experts	&The first large-scale dataset designed to evaluate agents in real-world conversational web navigation	& \url{https://mcgill-nlp.github.io/weblinx/}\\\hline
AgentTrek \cite{xu2024agenttrekagenttrajectorysynthesis} & Web & Web tutorials & Task metadata, step-by-step instructions, action sequences, visual observations, reproducible native traces & 4,902 trajectories & VLM agent guided by tutorials, with Playwright capturing the traces & Synthesizes high-quality trajectory data by leveraging web tutorials & / \\ \hline

MultiUI \cite{liu2024harnessing} & Web & Combination of human-designed instructions and automated extraction from web structures & Textual task descriptions, plans, action sequences, GUI screenshots, accessibility trees, bounding box annotations & 7.3 million instruction samples from 1 million websites & LLMs and Playwright & Supports a broad range of UI-related tasks, including GUI understanding, action prediction, and element grounding. & \url{https://neulab.github.io/MultiUI/} \\ \hline

Explorer \cite{pahuja2025explorerscalingexplorationdrivenweb} & Web & Popular URLs with systematic web exploration by LLMs & Textual task descriptions, Action sequences, GUI screenshots, Accessibility trees, HTML content & 94K successful web trajectories, 49K unique URLs, 720K screenshots & Generated by a multi-agent LLM pipeline & Largest-scale web trajectory dataset to date; dynamically explores web pages to create contextually relevant tasks & / \\ \hline
InSTA \cite{trabucco2025towards} & Web & Automatically generated by LLMs across 1M websites from Common Crawl & Web navigation tasks in natural language, task plans and action sequences, HTML-based observations converted to markdown, and evaluations from LLM-based judges & 150,000 tasks across 150,000 websites & Generated by LLMs using the Playwright API and filtered by LLM-based judges & Presents a fully automated three-stage data generation pipeline—task generation, action execution, and evaluation—using only language models without any human annotations & \url{https://data-for-agents.github.io} \\ \hline

\end{tabular}
}
\end{table*}

\subsection{Web Agent Data\label{sec:data:web}}
Web-based GUI agents demand datasets that capture the intricate complexity and diversity of real-world web interactions. These datasets often encompass varied website structures, including DOM trees and HTML content, as well as multi-step task annotations that reflect realistic user navigation and interaction patterns. Developing agents that can generalize across different websites and perform complex tasks requires comprehensive datasets that provide rich contextual information. We provide an overview of web-based GUI agents dataset in Table~\ref{tab:web_data1}.

Building upon this need, several significant datasets have been developed to advance web-based GUI agents. Unlike traditional datasets focusing on narrow, predefined tasks, \textbf{Mind2Web}~\cite{mind2webgeneralistagentweb} represents a significant step forward by emphasizing open-ended task descriptions, pushing agents to interpret high-level goals independently. It offers over 2,350 human-annotated tasks across 137 diverse websites, capturing complex interaction patterns and sequences typical in web navigation. This setup aids in evaluating agents' generalization across unseen domains and serves as a benchmark for language grounding in web-based GUIs, enhancing adaptability for real-world applications.

Similarly, \textbf{WebVLN}~\cite{Chen_Pitawela_Zhao_Zhou_Chen_Wu_2024} expands on web GUI tasks by combining navigation with question-answering. It provides agents with text-based queries that guide them to locate relevant web pages and extract information. By leveraging both HTML and visual content from websites, WebVLN aligns with real-world challenges of web browsing. This dataset is particularly valuable for researchers aiming to develop agents capable of complex, human-like interactions in GUI-driven web spaces.

Moreover, \textbf{WebLINX}~\cite{lù2024weblinxrealworldwebsitenavigation} focuses on conversational GUI agents, particularly emphasizing real-world web navigation through multi-turn dialogue. Featuring over 2,300 expert demonstrations across 155 real-world websites, WebLINX creates a rich environment with DOM trees and screenshots for training and evaluating agents capable of dynamic, user-guided navigation tasks. This dataset promotes agent generalization across new sites and tasks, with comprehensive action and dialogue data that provide insights into enhancing agent responsiveness in realistic web-based scenarios.

\textbf{MultiUI \cite{liu2024harnessing}} is a large-scale dataset designed to enhance GUI agents' text-rich visual understanding. It comprises 7.3 million multimodal instruction samples collected from 1 million websites, covering key web UI tasks such as element grounding, action prediction, and interaction modeling. Unlike traditional datasets that rely on raw HTML, MultiUI utilizes structured accessibility trees to generate high-quality multimodal instructions. Models trained on MultiUI demonstrate substantial performance improvements, achieving a 48\% gain on VisualWebBench \cite{liu2024visualwebbenchfarmultimodalllms} and a 19.1\% increase in element accuracy on Mind2Web \cite{mind2webgeneralistagentweb}.

\textbf{InSTA \cite{trabucco2025towards}} is an Internet-scale dataset for training GUI-based web agents, generated entirely through an automated LLM pipeline without human annotations. It covers 150k diverse websites sourced from Common Crawl and includes rich web navigation tasks, trajectories in Playwright API calls, and evaluations using LLM-based judges. The dataset highlights strong generalization capabilities and data efficiency, significantly outperforming human-collected datasets like Mind2Web \cite{mind2webgeneralistagentweb} and WebLINX \cite{lù2024weblinxrealworldwebsitenavigation} in zero-shot and low-resource settings. InSTA represents a key advancement in scalable data curation for LLM-powered GUI agents, offering unprecedented coverage across real-world web interfaces.

Collectively, these datasets represent essential resources that enable advancements in web agent capabilities, supporting the development of adaptable and intelligent agents for diverse web applications.

\begin{table*}[t]
\centering
\caption{Overview of datasets for optimizing LLMs tailored for mobile GUI agents (Part I).\label{tab:mobile_data1}}
\resizebox{\textwidth}{!}{ % Resize the entire figure to fit \textwidth
\begin{tabular}{p{1.5cm}|p{1cm}|p{2cm}|p{2.3cm}|p{2cm}|p{1.5cm}|p{2.5cm}|p{2cm}}
\hline
\textbf{Dataset} & \textbf{Platform} & \textbf{Source} & \textbf{Content} & \textbf{Scale} & \textbf{Collection Method} & \textbf{Highlight} & \textbf{Link} \\\hline
VGA~\cite{meng2024vgavisionguiassistant} & Android Mobile & Rico~\cite{rico} & GUI screenshots, task descriptions, action sequences, bounds, layout, and functions of GUI elements & 63.8k instances, 22.3k instruction-following data pairs, 41.4k conversation data pairs & Generated by GPT-4 models & Prioritizes visual content to reduce inaccuracies & \url{https://github.com/Linziyang1999/Vision\%2DGUI\%2Dassistant} \\\hline

Rico~\cite{rico} & Android Mobile & Gathered from real Android apps on Google Play Store & Textual data, screenshots, action sequences, UI structure, annotated UI representations & 72,219 unique UI screens, 10,811 user interaction traces & Crowdsourcing, automated exploration & Comprehensive dataset for mobile UI design, interaction modeling, layout generation & \url{http://www.interactionmining.org/} \\\hline

PixelHelp \cite{mappingnaturallanguageinstructions} & Android Mobile & Human, web “How-to”, Rico UI corpus synthetic & Natural language instructions, action sequences, GUI screenshots, structured UI data & 187 multi-step instructions, 295,476 synthetic single-step commands & Human annotation and synthetic generation & Pioneering method for grounding natural language instructions to executable mobile UI actions & \url{https://github.com/google-research/google-research/tree/master/seq2act} \\\hline

MoTIF~\cite{adatasetforinteractivevisionlanguagenavigation} & Android Mobile & Human-written & Natural language instructions, action sequences, GUI screenshots, structured UI data & 6,100 tasks across 125 Android apps & Human annotation & Task feasibility prediction for interactive GUI in mobile apps & \url{https://github.com/aburns4/MoTIF} \\\hline

META-GUI~\cite{metaguimultimodalconversationalagents} & Android Mobile & SMCalFlow~\cite{andreas2020task} & Dialogues, action sequences, screenshots, Android view hierarchies & 1,125 dialogues and 4,684 turns & Human annotation & Task-oriented dialogue system for mobile GUI without relying on back-end APIs & \url{https://x-lance.github.io/META-GU} \\\hline

AITW~\cite{androidinwildlargescaledataset} & Android Mobile & Human-generated instructions, LLM-generated prompts & Natural language instructions, UI screenshots, observation-action pairs & 715,142 episodes and 30,378 unique instructions & Human raters using Android emulators & Large-scale dataset for device control research with extensive app and UI diversity & \url{https://github.com/google-research/google-research/tree/master/android_in_the_wild} \\\hline

GUI-Xplore \cite{sun2025gui} & Mobile Android & Combination of automated exploration and manual design & Exploration videos, textual tasks, QA pairs, view hierarchies, GUI screenshots, action sequences, and GUI transition graphs & 312 apps, 115 hours of video, 32,569 QA pairs, 41,293 actions, about 200 pages per app & Automated and human exploration & Introduces an exploration-based pretraining paradigm that provides rich app-specific priors through video data & \url{https://github.com/921112343/GUI-Xplore} \\\hline

\end{tabular}
}
\end{table*}

\begin{table*}[h!]
\centering
\caption{Overview of datasets for optimizing LLMs tailored for mobile GUI agents (Part II).\label{tab:mobile_data2}}
\resizebox{\textwidth}{!}{ % Resize the entire figure to fit \textwidth
\begin{tabular}{p{1.5cm}|p{1cm}|p{2cm}|p{2.3cm}|p{2cm}|p{1.5cm}|p{2.5cm}|p{2cm}}
\hline
\textbf{Dataset} & \textbf{Platform} & \textbf{Source} & \textbf{Content} & \textbf{Scale} & \textbf{Collection Method} & \textbf{Highlight} & \textbf{Link} \\\hline

GUI Odyssey~\cite{lu2024guiodysseycomprehensivedataset} & Android Mobile & Human designers, GPT-4 & Textual tasks, plans, action sequences, GUI screenshots & 7,735 episodes across 201 apps & Human demonstrations & Focuses on cross-app navigation tasks on mobile devices & \url{https://github.com/OpenGVLab/GUI-Odyssey} \\\hline

Amex~\cite{chai2024amexandroidmultiannotationexpo} & Android Mobile & Human-designed, ChatGPT-generated & Text tasks, action sequences, high-res screenshots with multi-level annotations & 104,000 screenshots, 1.6 million interactive elements, 2,946 instructions & Human annotations, autonomous scripts & Multi-level, large-scale annotations supporting complex mobile GUI tasks & \url{https://yuxiangchai.github.io/AMEX/} \\\hline

Ferret-UI~\cite{you2024ferretuigroundedmobileui} & iOS, Android Mobile & Spotlight dataset, GPT-4 & Text tasks, action plans, GUI element annotations, bounding boxes & 40,000 elementary tasks, 10,000 advanced tasks & GPT-generated & Benchmark for UI-centric tasks with adjustable screen aspect ratios & \url{https://github.com/apple/ml-ferret} \\\hline

AITZ~\cite{zhang2024androidzoochainofactionthoughtgui} & Android Mobile & AITW~\cite{androidinwildlargescaledataset} & Screen-action pairs, action descriptions & 18,643 screen-action pairs across 70+ apps, 2,504 episodes & GPT-4V, icon detection models & Structured "Chain-of-Action-Thought" enhancing GUI navigation & \url{https://github.com/IMNearth/CoAT} \\\hline

Octo-planner~\cite{chen2024octoplannerondevicelanguagemodel} & Android Mobile & GPT-4 generated & Text tasks, decomposed plans, action sequences & 1,000 data points & GPT-4 generated & Optimized for task planning with GUI actions & \url{https://huggingface.co/NexaAIDev/octopus-planning} \\\hline

E-ANT~\cite{wang2024eantlargescaledatasetefficient} & Android tiny-apps & Human behaviors & Task descriptions, screenshots, action sequences, page element data & 40,000+ traces, 10,000 action intents & Human annotation & First large-scale Chinese dataset for GUI navigation with real human interactions & / \\\hline

Mobile3M \cite{wu2024mobilevlmvisionlanguagemodelbetter} & Android Mobile & Real-world interactions, simulations & UI screenshots, XML documents, action sequences & 3,098,786 UI pages, 20,138,332 actions & Simulation algorithm & Large-scale Chinese mobile GUI dataset with unique navigation graph & \url{https://github.com/Meituan-AutoML/MobileVLM} \\\hline

AndroidLab \cite{xu2024androidlabtrainingsystematicbenchmarking} & Android Mobile & Human design, LLM self-exploration, academic datasets & Text instructions, action sequences, XML data, screenshots & 10.5k traces, 94.3k steps & Human annotation, LLM self-exploration & XML-based interaction data with unified action space & \url{https://github.com/THUDM/Android-Lab} \\\hline

MobileViews \cite{gao2024mobileviews} & Android Mobile & LLM-enhanced app traversal tool & Screenshot-view hierarchy pairs & 600,000 screenshots, VH pairs from 20,000+ apps & LLM-enhanced crawler & Largest open-source mobile screen dataset & \url{https://huggingface.co/datasets/mllmTeam/MobileViews} \\\hline
FedMABench \cite{wang2025fedmabench} & Android Mobile & AndroidControl \cite{li2024effects}, AITW \cite{androidinwildlargescaledataset} & Textual task descriptions, action sequences, and GUI screenshots & 6 dataset series with over 30 subsets & Inferred from existing Android datasets & The first dataset designed to benchmark federated mobile GUI agents & \url{https://github.com/wwh0411/FedMABench} \\\hline

\end{tabular}
}
\end{table*}
\subsection{Mobile Agent Data\label{sec:data:mobile}}
Mobile platforms are critical for GUI agents due to the diverse range of apps and unique user interactions they involve. To develop agents that can effectively navigate and interact with mobile interfaces, datasets must offer a mix of single and multi-step tasks, focusing on natural language instructions, UI layouts, and user interactions.  We first overview mobile GUI agents dataset in Tables~\ref{tab:mobile_data1} and \ref{tab:mobile_data2}.

An early and foundational contribution in this domain is the \textbf{Rico} dataset~\cite{rico}, which provides over 72,000 unique UI screens and 10,811 user interaction traces from more than 9,700 Android apps. Rico has been instrumental for tasks such as UI layout similarity, interaction modeling, and perceptual modeling, laying the groundwork for mobile interface understanding and GUI agent development.

Building upon the need for grounding natural language instructions to mobile UI actions, \textbf{PIXELHELP}~\cite{mappingnaturallanguageinstructions} introduces a dataset specifically designed for this purpose. It includes multi-step instructions, screenshots, and structured UI element data, enabling detailed analysis of how verbal instructions can be converted into mobile actions. This dataset has significant applications in accessibility and task automation, supporting agents that autonomously execute tasks based on verbal cues.

Further expanding the scope, the \textbf{Android in the Wild (AITW)} dataset~\cite{androidinwildlargescaledataset} offers one of the most extensive collections of natural device interactions. Covering a broad spectrum of Android applications and diverse UI states, AITW captures multi-step tasks emulating real-world device usage. Collected through interactions with Android emulators, it includes both screenshots and action sequences, making it ideal for developing GUI agents that navigate app interfaces without relying on app-specific APIs. Due to its scale and diversity, AITW has become a widely used standard in the field.

In addition, \textbf{META-GUI}~\cite{metaguimultimodalconversationalagents} provides a unique dataset for mobile task-oriented dialogue systems by enabling direct interaction with mobile GUIs, bypassing the need for API-based controls. This approach allows agents to interact across various mobile applications using multi-turn dialogues and GUI traces, broadening their capabilities in real-world applications without custom API dependencies. The dataset's support for complex interactions and multi-turn dialogue scenarios makes it valuable for building robust conversational agents.

Recently, \textbf{MobileViews}~\cite{gao2024mobileviews} emerged as the largest mobile screen dataset to date, offering over 600,000 screenshot--view hierarchy pairs from 20,000 Android apps. Collected with an LLM-enhanced app traversal tool, it provides a high-fidelity resource for mobile GUI agents in tasks such as screen summarization, tappability prediction, and UI component identification. Its scale and comprehensive coverage of screen states make MobileViews a key resource for advancing mobile GUI agent capabilities.

Collectively, mobile platforms currently boast the richest set of datasets due to their versatile tools, emulator support, and diverse use cases, reflecting the demand for high-quality, adaptive GUI agents in mobile applications.

\begin{table*}[t]
\centering
\caption{Overview of datasets for optimizing LLMs tailored for computer GUI agents.\label{tab:computer_data1}}
\resizebox{\textwidth}{!}{ % Resize the entire figure to fit \textwidth
\begin{tabular}{p{1.5cm}|p{1cm}|p{2cm}|p{2.3cm}|p{2cm}|p{1.5cm}|p{2.5cm}|p{2cm}}
\hline
\textbf{Dataset} & \textbf{Platform} & \textbf{Source} & \textbf{Content} & \textbf{Scale} & \textbf{Collection Method} & \textbf{Highlight} & \textbf{Link} \\\hline
ScreenAgent \cite{niu2024screenagentvisionlanguagemodeldriven} & Linux, Windows OS & Human-designed & GUI screenshots, action sequences & 273 task sessions, 3,005 training screenshots, 898 test screenshots & Human annotation & VLM-based agent across multiple desktop environments & \url{https://github.com/niuzaisheng/ScreenAgent} \\\hline

LAM~\cite{wang2024lam} & Windows OS & Application documentation, WikiHow articles, Bing search queries & Task descriptions in natural language, step-by-step plans, action sequences, GUI screenshots & 76,672 task-plan pairs, 2,192 task-action trajectories & Instantiated using GPT-4, with actions tested and validated in the Windows environment using UFO~\cite{zhang2024ufouifocusedagentwindows} & Provides a structured pipeline for collecting, validating, and augmenting data, enabling high-quality training for action-oriented AI models. & \url{https://github.com/microsoft/UFO/tree/main/dataflow} \\ \hline
DeskVision \cite{xu2025deskvision} & Windows, macOS, and Linux desktops & Internet & GUI screenshots with annotated bounding boxes for UI elements and detailed region captions & 54,855 screenshots with 303,622 UI element annotations & UI elements detected using OmniParser and PaddleOCR & The first large-scale, open-source dataset focusing on real-world desktop GUI scenarios across operating systems & / \\ \hline
\end{tabular}
}
\end{table*}
\subsection{Computer Agent Data\label{sec:data:computer}}

In contrast to mobile and web platforms, the desktop domain for GUI agents has relatively fewer dedicated datasets, despite its critical importance for applications like productivity tools and enterprise software. However, notable efforts have been made to support the development and evaluation of agents designed for complex, multi-step desktop tasks. We show related dataset for computer GUI agents in Table~\ref{tab:computer_data1}.

A significant contribution in this area is \textbf{ScreenAgent}~\cite{niu2024screenagentvisionlanguagemodeldriven}, a dedicated dataset and model designed to facilitate GUI control in Linux and Windows desktop environments. ScreenAgent provides a comprehensive pipeline that enables agents to perform multi-step task execution autonomously, encompassing planning, action, and reflection phases. By leveraging annotated screenshots and detailed action sequences, it allows for high precision in UI element positioning and task completion, surpassing previous models in accuracy. This dataset is invaluable for researchers aiming to advance GUI agent capabilities in the desktop domain, enhancing agents' decision-making accuracy and user interface interactions.

The \textbf{LAM}  \cite{wang2024lam} is specifically designed to train and evaluate Large Action Models (LAMs) for GUI environments, bridging natural language task understanding and action execution. It comprises two core components: Task-Plan data, detailing user tasks with step-by-step plans, and Task-Action data, translating these plans into executable GUI actions. Sourced from application documentation, WikiHow articles, and Bing search queries, the dataset is enriched and structured using GPT-4. Targeting the Windows OS, with a focus on automating tasks in Microsoft Word, it includes 76,672 task-plan pairs and 2,688 task-action trajectories, making it one of the largest collections for GUI-based action learning. Data quality is ensured through a robust validation pipeline that combines LLM-based instantiation, GUI interaction testing, and manual review. Each entry is complemented with GUI screenshots and metadata, enabling models to learn both high-level task planning and low-level execution. The dataset’s modular design supports fine-tuning for specific GUI tasks and serves as a replicable framework for building datasets in other environments, marking a significant contribution to advancing GUI-based automation.

Although the desktop domain has fewer datasets compared to mobile and web, efforts like ScreenAgent and LAMs highlight the growing interest and potential for developing sophisticated GUI agents for computer systems.

\begin{table*}[h!]
\centering
\caption{Overview of datasets for optimizing LLMs tailored for cross-platform GUI agents (Part I).\label{tab:cross_data1}}
\resizebox{\textwidth}{!}{ % Resize the entire figure to fit \textwidth
\begin{tabular}{p{1.5cm}|p{1cm}|p{2cm}|p{2.3cm}|p{2cm}|p{1.5cm}|p{2.5cm}|p{2cm}}
\hline
\textbf{Dataset} & \textbf{Platform} & \textbf{Source} & \textbf{Content} & \textbf{Scale} & \textbf{Collection Method} & \textbf{Highlight} & \textbf{Link} \\\hline
Visual-AgentBench \cite{liu2024visualagentbenchlargemultimodalmodels} & Android Mobile, Web & VAB-Mobile: Android Virtual Device, VAB-WebArena-Lite: WebArena~\cite{koh2024visualwebarenaevaluatingmultimodalagents} & Task instructions, action sequences, screen observations & VAB-Mobile: 1,213 trajectories, 10,175 steps; VAB-WebArena-Lite: 1,186 trajectories, 9,522 steps & Program-based solvers, agent bootstrapping, human demonstrations & Systematic evaluation of VLM as a visual foundation agent across multiple scenarios & \url{https://github.com/THUDM/VisualAgentBench} \\\hline

GUICourse \cite{chen2024guicoursegeneralvisionlanguage} & Android Mobile, Web & Web scraping, simulation, manual design & GUI screenshots, action sequences, OCR tasks, QA pairs & 10 million website page-annotation pairs, 67,000 action instructions & LLM-based auto-annotation, crowd-sourcing & Dataset suite for enhancing VLM GUI navigation on web and mobile platforms & \url{https://github.com/yiye3/GUICourse} \\\hline

GUI-World~\cite{chen2024guiworlddatasetguiorientedmultimodal} & OS, Mobile, Web, XR & Student workers, YouTube instructional videos & GUI videos, human-annotated keyframes, captions, QA data, action sequences & 12,000 videos, 83,176 frames & Human annotation & Designed for dynamic, sequential GUI tasks with video data & \url{https://gui-world.github.io/} \\\hline

ScreenAI \cite{baechler2024screenaivisionlanguagemodelui} & Android, iOS, Desktop/Web & Crawling apps and webpages, synthetic QA & Screen annotation, screen QA, navigation, summarization & Annotation: hundreds of millions; QA: tens of millions; Navigation: millions & Model, human annotation & Comprehensive pretraining and fine-tuning for GUI tasks across platforms & \url{https://github.com/google\%2Dresearch\%2Ddatasets/screen_annotation} \\\hline

OmniParser \cite{lu2024omniparserpurevisionbased} & Web, Desktop, Mobile & Popular webpages & UI screenshots, bounding boxes, icon descriptions, OCR-derived text & 67,000+ screenshots, 7,000 icon-description pairs & Finetuned detection model, OCR, human descriptions & Vision-based parsing of UI screenshots into structured elements & \url{https://github.com/microsoft/OmniParser} \\\hline
\end{tabular}
}
\end{table*}

\begin{table*}[h!]
\centering
\caption{Overview of datasets for optimizing LLMs tailored for cross-platform GUI agents (Part II).\label{tab:cross_data2}}
\resizebox{\textwidth}{!}{ % Resize the entire figure to fit \textwidth
\begin{tabular}{p{1.5cm}|p{1cm}|p{2cm}|p{2.3cm}|p{2cm}|p{1.5cm}|p{2.5cm}|p{2cm}}
\hline
\textbf{Dataset} & \textbf{Platform} & \textbf{Source} & \textbf{Content} & \textbf{Scale} & \textbf{Collection Method} & \textbf{Highlight} & \textbf{Link} \\\hline
Web-Hybrid \cite{gou2024navigatingdigitalworldhumans} & Web, Android Mobile & Web-synthetic data & Screenshots, text-based referring expressions, coordinates on GUIs & 10 million GUI elements, 1.3 million screenshots & Rule-based synthesis, LLMs for referring expressions & Largest dataset for GUI visual grounding & \url{https://osu-nlp-group.github.io/UGround/} \\\hline

GUIDE \cite{chawla2024guide} & Computer and Web & Direct submissions from businesses and survey responses & Task descriptions, GUI screenshots, action sequences, CoT reasoning, spatial grounding & N/A & Collected through NEXTAG, an automated annotation tool & Integrates images, action sequences, task descriptions, and spatial grounding into a unified dataset & \url{https://github.com/superagi/GUIDE} \\ \hline

xLAM~\cite{zhang2024xlam} & Web and tools used & Synthesized data, and existing dataset & Textual tasks, action sequences, function-calling data & 60,000 data points & Collected using AI models with human verification steps & Provides a \emph{unified format} across diverse environments, enhancing generalizability and error detection for GUI agents & \url{https://github.com/SalesforceAIResearch/xLAM} \\\hline
Insight-UI~\cite{shen2024falcon} & iOS, Android, Windows, Linux, Web & Common Crawl corpus & Textual tasks, plans, action sequences, GUI screenshots & 434,000 episodes, 1,456,000 images & Automatic simulations performed by a browser API & Instruction-free paradigm and entirely auto-generated & / \\ \hline
OS-Genesis \cite{sun2024genesis} & Web and Android & Reverse task synthesis, where the GUI environment is explored interactively without predefined tasks or human annotations. & High-level instructions, low-level instructions, action sequences, and environment states. & 1,000 synthesized trajectories. & Model-based interaction-driven approach with GPT-4o. & Reverses the conventional task-driven data collection process by enabling exploration-first trajectory synthesis. & \url{https://qiushisun.github.io/OS-Genesis-Home/}  \\ \hline
Navi-plus \cite{cheng2025navi} & Web and Android &  AndroidControl \cite{li2024effects} and Mind2Web \cite{mind2webgeneralistagentweb} & Task descriptions, GUI action trajectories, low-level step instructions, screenshots, and follow-up ASK/SAY interaction pairs & / & LLM-automated with human validation & Introduces a Self-Correction GUI Navigation task featuring the novel ASK action for recovering missing information & /
\\ \hline
Explorer \cite{chaimalas2025explorer} & Web and Android & Automated traversal of real websites and Android apps & UI screenshots, bounding boxes of interactable elements, screen similarity labels, and user actions & KhanAcademy (Web): 2,841 interactables, 378 screen similarity samples; Spotify (Android): 1,207 interactables, 451 screen similarity samples & Automated tools, HTML parsing, Accessibility Tree & Platform-independent, supports auto-labeling, and enables trace recording and voice-controlled GUI navigation & \url{https://github.com/varnelis/Explorer}\\ \hline

\end{tabular}
}
\end{table*}

\subsection{Cross-Platform Agent Data\label{sec:data:cross}}

Cross-platform datasets play a pivotal role in developing versatile GUI agents that can operate seamlessly across mobile, computer, and web environments. Such datasets support generalizability and adaptability, enabling agents to handle varied interfaces and tasks in real-world applications. We provide an overview of related dataset for cross-platform GUI agents in Table~\ref{tab:cross_data1} and \ref{tab:cross_data2}.

One significant contribution is \textbf{ScreenAI}~\cite{baechler2024screenaivisionlanguagemodelui}, which extends the scope of data collection to include both mobile and desktop interfaces. Covering tasks such as screen annotation, question-answering, and navigation, ScreenAI offers hundreds of millions of annotated samples. Its comprehensive scale and mixed-platform coverage make it a robust foundation for GUI agents that need to manage complex layouts and interactions across diverse interfaces. By emphasizing element recognition and screen summarization, ScreenAI advances the development of multi-platform GUI agents capable of handling varied visual structures.

Building upon the need for evaluating visual foundation models across environments, \textbf{VisualAgentBench}~\cite{liu2024visualagentbenchlargemultimodalmodels} is a groundbreaking cross-platform benchmark designed to assess GUI agents in both mobile and web settings. It emphasizes interaction-focused tasks, using environments like Android Virtual Device and WebArena-Lite~\cite{webarenarealisticwebenvironment} to evaluate and improve agent responses to GUI layouts and user interface actions. The dataset's innovative collection method, which combines program-based solvers and large multimodal model bootstrapping, facilitates robust training trajectories that enhance adaptability and error recovery in GUI agent tasks.

Furthermore, \textbf{GUI-World}~\cite{chen2024guiworlddatasetguiorientedmultimodal} spans multiple platforms, including desktop, mobile, and XR environments, with over 12,000 annotated videos. Designed to address the challenges of dynamic and sequential GUI tasks, GUI-World allows researchers to benchmark GUI agent capabilities across diverse interfaces. By providing detailed action sequences and QA pairs, it sets a high standard for evaluating agents in complex, real-world scenarios.

Additionally, \textbf{xLAM}\cite{zhang2024xlam} contributes significantly to actionable agent development by providing a unified dataset format designed to support multi-turn interactions, reasoning, and function-calling tasks. Sourced from datasets like WebShop\cite{webshopscalablerealworldweb}, ToolBench~\cite{guo2024stabletoolbench}, and AgentBoard \cite{ma2024agentboard}, xLAM standardizes data formats across diverse environments, addressing the common issue of inconsistent data structures that hinder agent training and cross-environment compatibility. By offering a consistent structure, xLAM enhances the adaptability and error detection capabilities of GUI agents, allowing for more seamless integration and performance across different applications.

OS-Genesis \cite{sun2024genesis} adopts a reverse task synthesis approach for the Android and web platforms. It leverages GPT-4o to interactively explore the environment and generate instructions in a reverse manner. This process constructs high-quality, diverse GUI trajectories without relying on human annotations or predefined tasks. By eliminating these dependencies, OS-Genesis achieves scalable and efficient training for GUI agents while significantly enhancing the diversity and quality of the generated data.

Collectively, these cross-platform datasets contribute to building multi-platform GUI agents, paving the way for agents that can seamlessly navigate and perform tasks across different interfaces, fostering more generalized and adaptable systems.

\subsection{Takeaways\label{sec:data:takeaways}}
Data collection and curation for LLM-powered GUI agents is an intensive process, often requiring substantial human involvement, particularly for generating accurate action sequences and annotations. While early datasets were limited in scale and task diversity, recent advancements have led to large-scale, multi-platform datasets that support more complex and realistic GUI interactions. Key insights from these developments include:

\begin{enumerate}
    \item \textbf{Scale and Diversity:} High-quality, large-scale data is essential for training robust GUI agents capable of handling diverse UI states and tasks. Datasets like MobileViews~\cite{gao2024mobileviews} and ScreenAI~\cite{baechler2024screenaivisionlanguagemodelui} illustrate the importance of vast and varied data to accommodate the dynamic nature of mobile and desktop applications, enhancing the agent's resilience across different environments.

    \item \textbf{Cross-Platform Flexibility:} Cross-platform datasets such as VisualAgentBench~\cite{liu2024visualagentbenchlargemultimodalmodels} and GUI-World~\cite{chen2024guiworlddatasetguiorientedmultimodal} underscore the value of generalizability, enabling agents to perform consistently across mobile, web, and desktop environments. This cross-platform adaptability is a crucial step towards creating one-stop solutions where a single GUI agent can operate seamlessly across multiple platforms.
   
    \item \textbf{Automated Data Collection:} AI-driven data collection tools, as exemplified by OmniParser~\cite{lu2024omniparserpurevisionbased} and MobileViews~\cite{gao2024mobileviews}, showcase the potential to significantly reduce manual efforts and accelerate scalable dataset creation. By automating the annotation process, these tools pave the way for more efficient data pipelines, moving towards a future where AI supports AI by expediting data gathering and labeling for complex GUI interactions.
   
    \item \textbf{Unified Data Formats and Protocols:} xLAM's unified data format is an essential innovation that improves compatibility across diverse platforms~\cite{zhang2024xlam}, addressing a significant bottleneck in cross-platform GUI agent development. Establishing standardized protocols or action spaces for data collection, particularly given the varied data formats, action spaces, and environment representations across platforms, will be vital in furthering agent generalization and consistency.

\end{enumerate}

In summary, the evolving landscape of datasets for LLM-powered GUI agents spans multiple platforms, with each dataset addressing unique challenges and requirements specific to its environment. These foundational resources are key to enabling agents to understand complex UIs, perform nuanced interactions, and improve generalization across diverse applications. The push towards cross-platform adaptability, automated data collection, and standardized data formats will continue to shape the future of GUI agents.
\section{Models for Optimizing LLM-Brained GUI Agents\label{sec:model}}
LLMs act as the ``brain'' of GUI agents, empowering them to interpret user intents, comprehend GUI screens, and execute actions that directly impact their environments. While several existing foundation models are robust enough to serve as this core, they can be further fine-tuned and optimized to evolve into Large Action Models (LAMs)—specialized models tailored to improve the performance and efficiency of GUI agents. These LAMs bridge the gap between general-purpose capabilities and the specific demands of GUI-based interactions.

In this section, we first introduce the foundation models that currently form the backbone of GUI agents, highlighting their strengths and limitations. We then delve into the concept of LAMs, discussing how these models are fine-tuned with GUI-specific datasets to enhance their adaptability, accuracy, and action-orientation in GUI environments. Through this exploration, we illustrate the progression from general-purpose LLMs to purpose-built LAMs, laying the foundation for advanced, intelligent GUI agents.

% \usepackage[normalem]{ulem}
\begin{table*}[!h]
\centering
\caption{Overview of foundation models for LLM-brained GUI agents.}
\label{tab:foundation_model}
\resizebox{\textwidth}{!}{ % Resize the entire figure to fit \textwidth
\begin{tabular}{p{1.5cm}|p{1.2cm}|p{1cm}|p{3cm}|p{4.cm}|p{4cm}|p{0.8cm}|p{2cm}}
\hline
\multicolumn{1}{c|}{\textbf{Model}}                                   & \multicolumn{1}{c|}{\textbf{Modality}} & \multicolumn{1}{c|}{\textbf{Model Size}} & \multicolumn{1}{c|}{\textbf{Architecture}}                                                                                                 & \multicolumn{1}{c|}{\textbf{Training Methods}}                                                                                                                                                                                                                                                                      & \multicolumn{1}{c|}{\textbf{Highlights}}                                                                                                                                                        & \multicolumn{1}{p{0.8cm}|}{\textbf{Open-Source}} & \multicolumn{1}{c}{\textbf{Link}}                                \\ \hline
GPT-4o \cite{hurst2024gpt}                                            & Text, audio, image, and video          & -                                        & Multimodal autoregressive architecture                                                                                                     & Pre-trained on a mix of public data, further trained for alignment with human preferences and safety considerations                                                                                                                                                                                                 & Unified multimodal architecture that seamlessly processes and generates outputs across text, audio, image, and video, offering faster and more cost-effective operation than its predecessors   & No                                        & /                                                                \\ \hline
GPT-4V \cite{openai2023gpt4v}                                         & Text and image                         & -                                        & -                                                                                                                                          & Pre-trained on a large dataset of text and image data, followed by fine-tuning with reinforcement learning from human feedback (RLHF)                                                                                                                                                                               & Notable for its multimodal capabilities, allowing it to analyze and understand images alongside text                                                                                            & No                                        & /                                                                \\ \hline
Gemini \cite{team2023gemini}                                          & Text, image, audio, and video          & Nano versions: 1.8B/3.25B                & Enhanced Transformer decoders                                                                                                              & Large-scale pre-training on multimodal data, followed by supervised fine-tuning, reward modeling, and RLHF                                                                                                                                                                                                          & Achieves state-of-the-art performance across multimodal tasks, including a groundbreaking 90\% on the MMLU benchmark, and demonstrates capacity for on-device deployment with small model sizes & No                                        & /                                                                \\ \hline
Claude 3.5 Sonnet (Computer Use) \cite{anthropic2024, hu2024dawnguiagentpreliminary} & Text and image                         & -                                        & ReAct-based reasoning                                                                                                                      & -                                                                                                                                                                                                                                                                                                                   & Pioneering role in GUI automation as the first public beta model to utilize a vision-only paradigm for desktop task automation                                                                  & No                                        & /                                                                \\\hline
Operator \cite{cua2025, openai2025operator} & Text and Image & - & Built on GPT-4o & Supervised learning and reinforcement learning & Trained to use a computer like a human, achieving remarkable performance on benchmarks & No & /
\\\hline\hline
Qwen-VL \cite{bai2023qwen}                                            & Text and image                         & 9.6B                                     & A Vision Transformer (ViT) \cite{dosovitskiy2021imageworth16x16words} as the visual encoder, with a large language model based on the Qwen-7B architecture                            & Two stages of pre-training and a final stage of instruction fine-tuning                                                                                                                                                                                                                                             & Achieves state-of-the-art performance on vision-language benchmarks and supports fine-grained visual understanding                                                                              & Yes                                       & \url{httpss://github.com/QwenLM/Qwen-VL}                            \\ \hline
Qwen2-VL \cite{wang2024qwen2vlenhancingvisionlanguagemodels}          & Text, image, and video                 & 2B/7B/72B                                & ViT \cite{dosovitskiy2021imageworth16x16words} as the vision encoder, paired with the Qwen2 series of language models & The ViT is trained with image-text pairs; all parameters are unfrozen for broader multimodal learning with various datasets; fine-tuning the LLM on instruction datasets                                                                                                                                            & Introduces Naive Dynamic Resolution for variable resolution image processing and Multimodal Rotary Position Embedding for enhanced multimodal integration                              & Yes                                       & \url{httpss://github.com/QwenLM/Qwen2-VL}                           \\ \hline
InternVL-2 \cite{chen2024internvl,chen2024far}                        & Text, image, video, and medical data   & 1B/2B/4B/ 8B/26B/40B                      & ViT as the vision encoder and a LLM as the language component                                  & Progressive alignment strategy, starting with coarse data and moving to fine data                                                                                                                                                                                                                                   & Demonstrates powerful capabilities in handling complex multimodal tasks with various model sizes                                                                                                & Yes                                       & \url{httpss://internvl.github.io/blog/2024-07-02-InternVL-2.0/}     \\ \hline
CogVLM \cite{wang2024cogvlmvisualexpertpretrained}                    & Text and image                         & 17B                                      & A ViT encoder, a two-layer MLP adapter, a pretrained large language model, and a visual expert module                                      & Stage 1 focuses on image captioning; Stage 2 combines image captioning and referring expression comprehension tasks                                                                                                                                                                                                 & Achieves deep integration of visual and language features while preserving the full capabilities of large language models                                                                       & Yes                                       & \url{httpss://github.com/THUDM/CogVLM}                              \\ \hline
Ferret \cite{you2023ferret}                                           & Text and image                         & 7B/13B                                   & Decoder-only architecture based on the Vicuna model, combined with a visual encoder                                                        & A combination of supervised training and additional instruction tuning                                                                                                                                                                                                                                              & Ability to handle free-form region inputs via its hybrid region representation, enabling versatile spatial understanding and grounding                                                          & Yes                                       & \url{httpss://github.com/apple/ml-ferret}                           \\ \hline
LLaVA \cite{liu2024visual}                                            & Text and image                         & 7B/13B                                   & A vision encoder (CLIP ViT-L/14), a language decoder (Vicuna)                                                                              & Pre-training using filtered image-text pairs, fine-tuning with a multimodal instruction-following dataset                                                                                                                                                                                                           & Its lightweight architecture enables quick experimentation, demonstrating capabilities close to GPT-4 in multimodal reasoning                                                                   & Yes                                       & \url{httpss://llava-vl.github.io}                                   \\ \hline
LLaVA-1.5 \cite{liu2024improved}                                      & Text and image                         & 7B/13B                                   & A vision encoder (CLIP-ViT) and an encoder-decoder LLM architecture (\eg Vicuna or LLaMA)                                                & Pre-training on vision-language alignment with image-text pairs; visual instruction tuning with specific task-oriented data                                                                                                                                                                                         & Notable for its data efficiency and scaling to high-resolution image inputs                                                                                                                     & Yes                                       & \url{httpss://llava-vl.github.io}                                   \\ \hline
BLIP-2 \cite{li2023blip}                                              & Text and image                         & 3.4B/12.1B                               & A frozen image encoder, a lightweight Querying Transformer to bridge the modality gap, and a frozen large language model                   & Vision-language representation learning: trains the Q-Former with a frozen image encoder; Vision-to-language generative learning: connects the Q-Former to a frozen LLM to enable image-to-text generation                                                                                                          & Achieves state-of-the-art performance on various vision-language tasks with a compute-efficient strategy by leveraging frozen pre-trained models                                                & Yes                                       & \url{httpss://github.com/salesforce/LAVIS/tree/main/projects/blip2} \\ \hline
Phi-3.5-Vision \cite{abdin2024phi}                                    & Text and image                         & 4.2B                                     & Image encoder: CLIP ViT-L/14 to process visual inputs, and transformer decoder based on the Phi-3.5-mini model for textual outputs         & Pre-training on a combination of interleaved image-text datasets, synthetic OCR data, chart/table comprehension data, and text-only data; supervised fine-tuning using large-scale multimodal and text datasets; Direct Preference Optimization (DPO) to improve alignment, safety, and multimodal task performance & Excels in reasoning over visual and textual inputs, demonstrating competitive performance on single-image and multi-image tasks while being compact                                             & Yes                                       & \url{httpss://github.com/microsoft/Phi-3CookBook/tree/main}         \\ \hline
\end{tabular}
}
\end{table*}

\subsection{Foundation Models\label{sec:model:foundation}}
Foundation models serve as the core of LLM-powered GUI agents, providing the essential capabilities for understanding and interacting with graphical user interfaces. Recent advancements in both close-source and open-source MLLMs have significantly enhanced the potential of GUI agents, offering improvements in efficiency, scalability, and multimodal reasoning. This subsection explores these foundation models, highlighting their innovations, contributions, and suitability for GUI agent applications. For a quick reference, Table~\ref{tab:foundation_model} presents an overview of the key models and their characteristics.

\subsubsection{Close-Source Models\label{sec:model:foundation:close}}
While proprietary models are not openly available for customization, they offer powerful capabilities that can be directly utilized as the ``brain'' of GUI agents.

Among these, GPT-4V~\cite{openai2023gpt4v} and GPT-4o~\cite{hurst2024gpt} are most commonly used in existing GUI agent frameworks due to their strong abilities, as discussed in Section~\ref{sec:framework}. GPT-4V represents a significant advancement in multimodal AI, combining text and image analysis to expand the functionality of traditional LLMs. Its ability to understand and generate responses based on both textual and visual inputs makes it well-suited for GUI agent tasks that require deep multimodal reasoning. Although its deployment is limited due to safety and ethical considerations, GPT-4V underscores the potential of foundation models to revolutionize GUI agent development with enhanced efficiency and flexibility.

Similarly, GPT-4o offers a unified multimodal autoregressive architecture capable of processing text, audio, images, and video. This model excels in generating diverse outputs efficiently, achieving faster response times at lower costs compared to its predecessors. Its rigorous safety and alignment practices make it reliable for sensitive tasks, positioning it as a robust tool for intelligent GUI agents that require comprehensive multimodal comprehension.

The \textbf{Gemini} model family~\cite{team2023gemini} advances multimodal AI modeling by offering versions tailored for high-complexity tasks, scalable performance, and on-device efficiency. Notably, the Nano models demonstrate significant capability in reasoning and coding tasks despite their small size, making them suitable for resource-constrained devices. Gemini's versatility and efficiency make it a compelling choice for powering GUI agents that require both performance and adaptability.

Emphasizing industry investment in GUI automation, \textbf{Claude 3.5 Sonnet (Computer Use)} introduces a pioneering approach by utilizing a vision-only paradigm for desktop task automation \cite{anthropic2024, hu2024dawnguiagentpreliminary}. It leverages real-time screenshots to observe the GUI state and generate actions, eliminating the need for metadata or underlying GUI structure. This model effectively automates GUI tasks by interpreting the screen, moving the cursor, clicking buttons, and typing text. Its unique architecture integrates a ReAct-based~\cite{yao2022react} reasoning paradigm with selective observation, reducing computational overhead by observing the environment only when necessary. Additionally, Claude 3.5 maintains a history of GUI screenshots, enhancing task adaptability and enabling dynamic interaction with software environments in a human-like manner. Despite challenges in handling dynamic interfaces and error recovery, this model represents a significant step forward in creating general-purpose GUI agents. Its development highlights substantial industry investment in this area, indicating a growing focus on leveraging LLMs for advanced GUI automation.

The Operator model \cite{cua2025, openai2025operator}, developed by OpenAI, represents a new frontier in Computer-Using Agents (CUA), akin to Claude 3.5 Sonnet (Computer Use). Designed to interact with GUI environments through LLM-powered reasoning and vision capabilities, Operator builds upon GPT-4o, integrating reinforcement learning to navigate and execute tasks across digital interfaces such as browsers, forms, and applications. By perceiving screenshots, interpreting UI elements, and performing actions via a virtual cursor and keyboard, Operator enables the automation of complex GUI-based workflows, including online purchases, email management, and document editing. Notably, Operator excels in understanding and manipulating digital environments, establishing itself as a powerful tool for human-computer interaction automation. Its exceptional performance on various benchmarks underscores its leading capabilities in GUI-based task automation.

\subsubsection{Open-Source Models\label{sec:model:foundation:open}}
Open-source models provide flexibility for customization and optimization, allowing developers to tailor GUI agents with contextual data and deploy them on devices with limited resources.

The \textbf{Qwen-VL} series~\cite{bai2023qwen} is notable for its fine-grained visual understanding and multimodal capabilities. With a Vision Transformer-based visual encoder and the Qwen-7B language model \cite{bai2023qwentechnicalreport}, it achieves state-of-the-art results on vision-language benchmarks while supporting multilingual interactions. Its efficiency and open-source availability, along with quantized versions for resource efficiency, make it suitable for developing GUI agents that require precise visual comprehension.

Building upon this, \textbf{Qwen2-VL}~\cite{wang2024qwen2vlenhancingvisionlanguagemodels} introduces innovations like Naive Dynamic Resolution and Multimodal Rotary Position Embedding, enabling efficient processing of diverse modalities including extended-length videos. The scalable versions of Qwen2-VL balance computational efficiency and performance, making them adaptable for both on-device applications and complex multimodal tasks in GUI environments.

\textbf{InternVL-2}~\cite{chen2024internvl,chen2024far} combines a Vision Transformer with a Large Language Model to handle text, images, video, and medical data inputs. Its progressive alignment strategy and availability in various sizes allow for flexibility in deployment. By achieving state-of-the-art performance in complex multimodal tasks, InternVL-2 demonstrates powerful capabilities that are valuable for GUI agents requiring comprehensive multimodal understanding.

Advancing efficient integration of visual and linguistic information, \textbf{CogVLM}~\cite{wang2024cogvlmvisualexpertpretrained} excels in cross-modal tasks with a relatively small number of trainable parameters. Its ability to deeply integrate visual and language features while preserving the full capabilities of large language models makes it a cornerstone for GUI agent development, especially in applications where resource efficiency is critical.

Enhancing spatial understanding and grounding, \textbf{Ferret}~\cite{you2023ferret} offers an innovative approach tailored for GUI agents. By unifying referring and grounding tasks within a single framework and employing a hybrid region representation, it provides precise interaction with graphical interfaces. Its robustness against object hallucinations and efficient architecture make it ideal for on-device deployment in real-time GUI applications.

The \textbf{LLaVA} model~\cite{liu2024visual} integrates a visual encoder with a language decoder, facilitating efficient alignment between modalities. Its lightweight projection layer and modular design enable quick experimentation and adaptation, making it suitable for GUI agents that require fast development cycles and strong multimodal reasoning abilities. Building on this, \textbf{LLaVA-1.5}~\cite{liu2024improved} introduces a novel MLP-based cross-modal connector and scales to high-resolution image inputs, achieving impressive performance with minimal training data. Its data efficiency and open-source availability pave the way for widespread use in GUI applications requiring detailed visual reasoning.

\textbf{BLIP-2}~\cite{li2023blip} employs a compute-efficient strategy by leveraging frozen pre-trained models and introducing a lightweight Querying Transformer. This design allows for state-of-the-art performance on vision-language tasks with fewer trainable parameters. BLIP-2's modularity and efficiency make it suitable for resource-constrained environments, highlighting its potential for on-device GUI agents.

Finally, \textbf{Phi-3.5-Vision}~\cite{abdin2024phi} achieves competitive performance in multimodal reasoning within a compact model size. Its innovative training methodology and efficient integration of image and text understanding make it a robust candidate for GUI agents that require multimodal reasoning and on-device inference without the computational overhead of larger models.

In summary, both close-source and open-source foundation models have significantly advanced the capabilities of LLM-powered GUI agents. While proprietary models offer powerful out-of-the-box performance, open-source models provide flexibility for customization and optimization, enabling tailored solutions for diverse GUI agent applications. The innovations in multimodal reasoning, efficiency, and scalability across these models highlight the evolving landscape of foundation models, paving the way for more intelligent and accessible GUI agents.

\begin{figure*}[t]
    \centering
    \includegraphics[width=\textwidth]{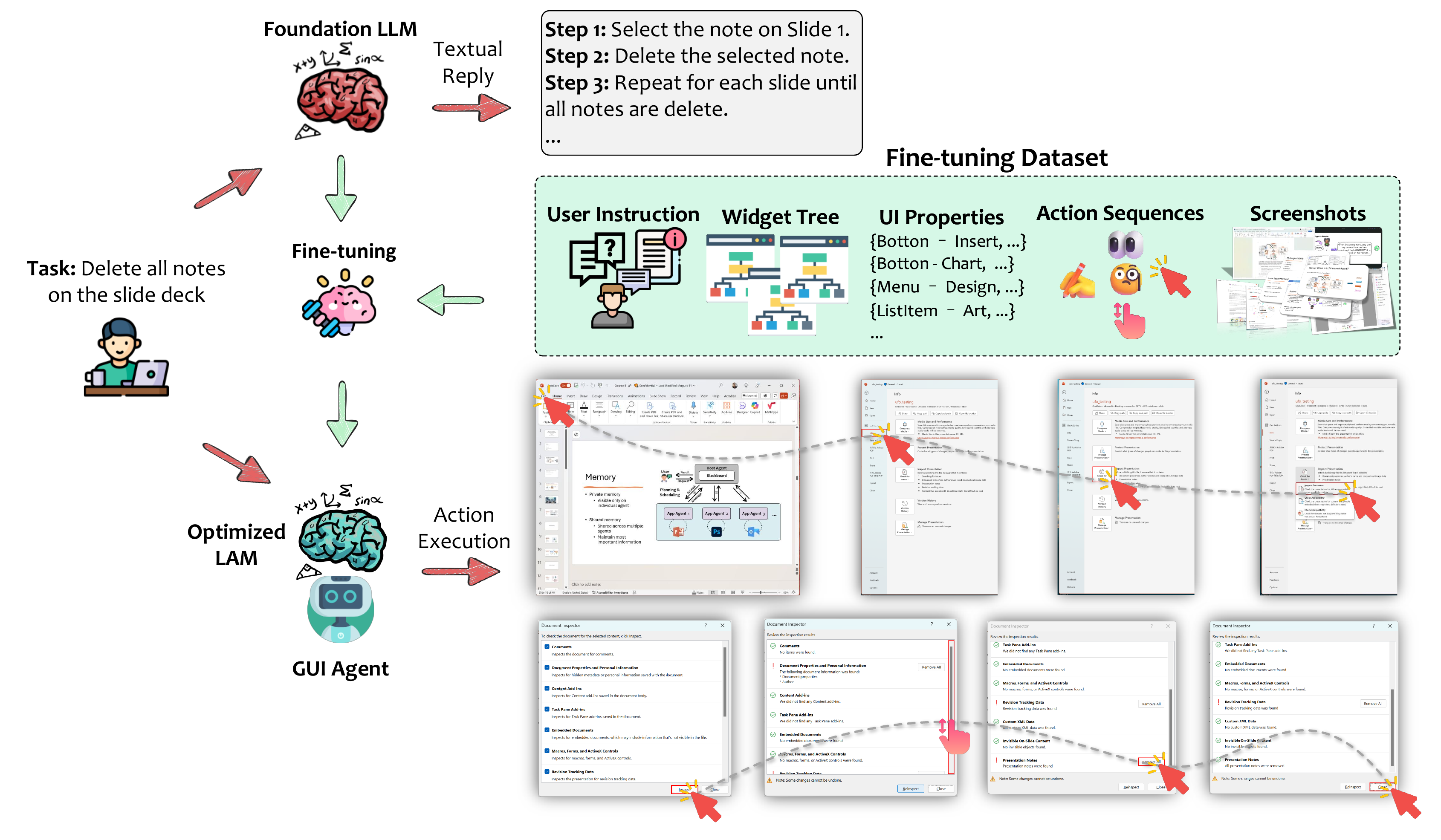}
    % \vspace{-2em}
    \caption{The evolution from foundation LLMs to GUI agent-optimized LAM with fine-tuning.}
    \label{fig:lam}
    % \vspace{-2em}
\end{figure*}

\subsection{Large Action Models\label{sec:model:lam}}
While general-purpose foundation LLMs excel in capabilities like multimodal understanding, task planning, and tool utilization, they often lack the specialized optimizations required for GUI-oriented tasks. To address this, researchers have introduced \emph{Large Action Models} (LAMs)—foundation LLMs fine-tuned with contextual, GUI-specific datasets (as outlined in Section~\ref{sec:data}) to enhance their action-driven capabilities. These models represent a significant step forward in refining the ``brain'' of GUI agents for superior performance.

In the realm of GUI agents, LAMs provide several transformative advantages:
\begin{enumerate}
    \item \textbf{Enhanced Action Orientation:} By specializing in action-oriented tasks, LAMs enable accurate interpretation of user intentions and generation of precise action sequences. This fine-tuning ensures that LAMs can seamlessly align their outputs with GUI operations, delivering actionable steps tailored to user requests.
    
    \item \textbf{Specialized Planning for Long, Complex Tasks:} LAMs excel in devising and executing intricate, multi-step workflows. Whether the tasks span multiple applications or involve interdependent operations, LAMs leverage their training on extensive action sequence datasets to create coherent, long-term plans. This makes them ideal for productivity-focused tasks requiring sophisticated planning across various tools.
    
    \item \textbf{Improved GUI Comprehension and Visual Grounding:} Training on datasets that incorporate GUI screenshots allows LAMs to advance their abilities in detecting, localizing, and interpreting UI components such as buttons, menus, and forms. By utilizing visual cues instead of relying solely on structured UI metadata, LAMs become highly adaptable, performing effectively across diverse software environments.
    
    \item \textbf{Efficiency through Model Size Reduction:} Many LAMs are built on smaller foundational models—typically around 7 billion parameters—that are optimized for GUI-specific tasks. This compact, purpose-driven design reduces computational overhead, enabling efficient operation even in resource-constrained environments, such as on-device inference.
\end{enumerate}
As illustrated in Figure~\ref{fig:lam}, the process of developing a purpose-built LAM for GUI agents begins with a robust, general-purpose foundation model, ideally with VLM capabilities. Fine-tuning these models on comprehensive, specialized GUI datasets—including user instructions, widget trees, UI properties, action sequences, and annotated screenshots—transforms them into optimized LAMs, effectively equipping them to serve as the ``brain'' of GUI agents.

This optimization bridges the gap between planning and execution. A general-purpose LLM might provide only textual plans or abstract instructions in response to user queries, which may lack precision. In contrast, a LAM-empowered GUI agent moves beyond planning to actively and intelligently execute tasks on GUIs. By interacting directly with application interfaces, these agents perform tasks with remarkable precision and adaptability. This paradigm shift marks the evolution of GUI agents from passive task planners to active, intelligent executors.
% For a broader understanding of LAMs and their applications, readers are encouraged to refer to \cite{LAM}.

% In the following sections, we present an analysis of LAMs tailored for GUI agents across different platforms, summarized in Tables~\ref{tab:model1}, \ref{tab:model2}, and \ref{tab:model3}, followed by an in-depth discussion in the subsequent subsections.

\begin{table*}[t]
\centering
\caption{An overview of GUI-optimized models on web platforms.}
\label{tab:web_model1}
\resizebox{\textwidth}{!}{ % Resize the entire figure to fit \textwidth
\begin{tabular}{p{1cm}|p{1cm}|p{1.5cm}|p{1cm}|p{2.5cm}|p{2cm}|p{2.5cm}|p{4cm}|p{2cm}}
\hline
\multicolumn{1}{c|}{\textbf{Model}}                            & \multicolumn{1}{c|}{\textbf{Platform}} & \multicolumn{1}{p{1.5cm}|}{\textbf{Foundation Model}}                                                              & \multicolumn{1}{c|}{\textbf{Size}} & \multicolumn{1}{c|}{\textbf{Input}}                  & \multicolumn{1}{c|}{\textbf{Output}}                        & \multicolumn{1}{c|}{\textbf{Dataset}}                                                                  & \multicolumn{1}{c|}{\textbf{Highlights}}                                                                                                                                               & \multicolumn{1}{c}{\textbf{Link}}                                                                                             \\ \hline
Agent Q \cite{putta2024agent}                                   & Web                                    & LLaMA-3 70B \cite{dubey2024llama}                                                                                                 & 70B                                      & HTML DOM representations                             & Plans, thoughts, actions, and action explanations           & WebShop benchmark and OpenTable dataset                                                                & Combines Monte Carlo Tree Search (MCTS) with self-critique mechanisms, leveraging reinforcement learning to achieve exceptional performance                                            & \url{https://github.com/sentient\%2Dengineering/agent-q}                                                                            \\ \hline
GLAINTEL \cite{fereidouni2024searchqueriestrainingsmaller}      & Web                                    & Flan-T5 \cite{chung2024scaling}                                                                                                     & 780M                                     & User instructions and observations of webpage state  & GUI actions                                                 & 1.18M real-world products, 12,087 crowd-sourced natural language intents, 1,010 human demonstrations   & Efficient use of smaller LLMs, and integration of RL and human demonstrations for robust performance                                                                                   & /                                                                                                                              \\ \hline
WebN-T5 \cite{Thil_2024}                                        & Web                                    & T5 \cite{raffel2020exploring}                                                                               & -                                        & HTML and DOM with screenshots                        & Hierarchical navigation plans and GUI interactions          & MiniWoB++, 13,000 human-made demonstrations                                                            & Combines supervised learning and reinforcement learning to address limitations of previous models in memorization and generalization                                                   & /                                                                                                                              \\ \hline
OpenWeb-Voyager \cite{he2024openwebvoyagerbuildingmultimodalweb} & Web                                    & Idefics2-8b-instruct \cite{laurenccon2024matters}                                                           & 8B                                       & GUI screenshots, accessibility trees                 & Actions on GUI, planning and thought, answers to queries    & Mind2Web and WebVoyager datasets, and generated queries for real-world web navigation                  & Combining imitation learning with a feedback loop for continuous improvement                                                                                                           & \url{https://github.com/MinorJerry/OpenWebVoyager}                                                                               \\ \hline
WebRL \cite{qi2024webrltrainingllmweb}                          & Web                                    & Llama-3.1 \cite{dubey2024llama} and GLM-4 \cite{du2021glm}                                                  & 8B/9B/ 70B                                & Task instructions, action history, HTML content      & Actions, element identifiers, explanations or notes         & WebArena-Lite                                                                                          & Introduces a self-evolving online curriculum reinforcement learning framework, which dynamically generates tasks based on past failures and adapts to the agent's skill level          & \url{https://github.com/THUDM/WebRL}                                                                                             \\ \hline
WebGUM \cite{furuta2023multimodal}                              & Web                                    & Flan-T5 \cite{chung2024scaling} and Vision Transformer (ViT) \cite{dosovitskiy2021imageworth16x16words}     & 3B                                       & HTML, screenshots, interaction history, instructions & Web navigation actions and free-form text                   & MiniWoB++ and WebShop benchmarks                                                                       & Integrates temporal and local multimodal perception, combining HTML and visual tokens, and uses an instruction-finetuned language model for enhanced reasoning and task generalization & \url{https://console.cloud.google.com/storage/browser/gresearch/webllm} \\ \hline
\end{tabular}
}
\end{table*}

\subsection{LAMs for Web GUI Agents\label{sec:model:web}}

In the domain of web-based GUI agents, researchers have developed specialized LAMs that enhance interaction and navigation within web environments. These models are tailored to understand the complexities of web GUIs, including dynamic content and diverse interaction patterns. We present an analysis of LAMs tailored for web GUI agents in Table~\ref{tab:web_model1}.

Building upon the need for multimodal understanding, \textbf{WebGUM}~\cite{furuta2023multimodal} integrates HTML understanding with visual perception through temporal and local tokens. It leverages Flan-T5 \cite{chung2024scaling}  for instruction fine-tuning and ViT \cite{dosovitskiy2021imageworth16x16words} for visual inputs, enabling it to process both textual and visual information efficiently. This multimodal grounding allows WebGUM to generalize tasks effectively, significantly outperforming prior models on benchmarks like MiniWoB++~\cite{reinforcementlearningonwebinterfaces} and WebShop~\cite{webshopscalablerealworldweb}. With its data-efficient design and capacity for multi-step reasoning, WebGUM underscores the importance of combining multimodal inputs in enhancing GUI agent performance.

Addressing the challenge of multi-step reasoning and planning in GUI environments, researchers have introduced frameworks that incorporate advanced search and learning mechanisms. For instance, \textbf{Agent Q}~\cite{putta2024agent} employs MCTS combined with self-critique mechanisms and Direct Preference Optimization (DPO)~\cite{rafailov2024direct} to improve success rates in complex tasks such as product search and reservation booking. By fine-tuning the LLaMA-3 70B model~\cite{dubey2024llama} to process HTML DOM representations and generate structured action plans, thoughts, and environment-specific commands, this framework showcases the power of integrating reasoning, search, and iterative fine-tuning for autonomous agent development.

Leveraging smaller models for efficient web interaction, \textbf{GLAINTEL}~\cite{fereidouni2024searchqueriestrainingsmaller} demonstrates that high performance can be achieved without large computational resources. Utilizing the Flan-T5~\cite{chung2024scaling} model with 780M parameters, it focuses on dynamic web environments like simulated e-commerce platforms. The model incorporates RL to optimize actions such as query formulation and navigation, effectively integrating human demonstrations and unsupervised learning. Achieving results comparable to GPT-4-based methods at a fraction of the computational cost, GLAINTEL underscores the potential of reinforcement learning in enhancing web-based GUI agents for task-specific optimization.

To enable continuous improvement and generalization across varied web domains, \textbf{OpenWebVoyager}~\cite{he2024openwebvoyagerbuildingmultimodalweb} combines imitation learning with an iterative exploration-feedback-optimization cycle. Leveraging large multimodal models like Idefics2-8B~\cite{laurenccon2024matters}, it performs autonomous web navigation tasks. By training on diverse datasets and fine-tuning using trajectories validated by GPT-4 feedback, the agent addresses real-world complexities without relying on synthetic environments. This approach significantly advances GUI agent frameworks by demonstrating the capability to generalize across varied web domains and tasks.

Moreover, tackling challenges such as sparse training data and policy distribution drift, \textbf{WebRL}~\cite{qi2024webrltrainingllmweb} introduces a self-evolving curriculum and robust reward mechanisms for training LLMs as proficient web agents. By dynamically generating tasks based on the agent's performance, WebRL fine-tunes models like Llama-3.1~\cite{dubey2024llama} and GLM-4~\cite{glm2024chatglm}, achieving significant success rates in web-based tasks within the WebArena environment. This framework outperforms both proprietary APIs and other open-source models, highlighting the effectiveness of adaptive task generation and sustained learning improvements in developing advanced GUI agents.

These advancements in LAMs for web GUI agents illustrate the importance of integrating multimodal inputs, efficient model designs, and innovative training frameworks to enhance agent capabilities in complex web environments.

\begin{table*}[!h]
\centering
\caption{An overview of GUI-optimized models on mobile platforms (Part I).}
\label{tab:mobile_model1}
\resizebox{\textwidth}{!}{ % Resize the entire figure to fit \textwidth
\begin{tabular}{p{1.cm}|p{1cm}|p{1.5cm}|p{1cm}|p{2.5cm}|p{2cm}|p{2.5cm}|p{4cm}|p{2cm}}
\hline
\multicolumn{1}{c|}{\textbf{Model}}                            & \multicolumn{1}{c|}{\textbf{Platform}} & \multicolumn{1}{p{1.5cm}|}{\textbf{Foundation Model}}                                                              & \multicolumn{1}{c|}{\textbf{Size}} & \multicolumn{1}{c|}{\textbf{Input}}                  & \multicolumn{1}{c|}{\textbf{Output}}                        & \multicolumn{1}{c|}{\textbf{Dataset}}                                                                  & \multicolumn{1}{c|}{\textbf{Highlights}}                                                                                                                                               & \multicolumn{1}{c}{\textbf{Link}}                                                                                             \\ \hline
Mobile-VLM \cite{wu2024mobilevlmvisionlanguagemodelbetter}       & Mobile Android                         & Qwen-VL-Chat \cite{bai2023qwen}                                                                             & 9.8B                                     & Screenshots and structured XML documents             & Action predictions, navigation steps, and element locations & Mobile3M, includes 3 million UI pages, 20+ million actions, and XML data structured as directed graphs & Mobile-specific pretraining tasks that enhance intra- and inter-UI understanding, with a uniquely large and graph-structured Chinese UI dataset (Mobile3M)                             & \url{https://github.com/XiaoMi/mobilevlm}                                                                                        \\ \hline
Octo-planner \cite{chen2024octoplannerondevicelanguagemodel}    & Mobile devices                         & Phi-3 Mini \cite{abdin2024phi}                                                                              & 3.8B                                     & User queries and available function descriptions     & Execution steps                                             & 1,000 data samples generated using GPT-4                                                               & Optimized for resource-constrained devices to ensure low latency, privacy, and offline functionality                                                                                   & \url{https://huggingface.co/NexaAIDev/octopus-planning}                                                                          \\ \hline
DigiRL \cite{bai2024digirltraininginthewilddevicecontrol}       & Mobile Android                         & AutoUI \cite{zhang2024lookscreensmultimodalchainofaction}                                                   & 1.3B                                     & Screenshots                                          & GUI actions                                                 & AiTW                                                                                                   & Offline-to-online reinforcement learning, bridging gaps in static and dynamic environments                                                                                             & \url{https://github.com/DigiRL-agent/digirl}                                                                                     \\ \hline
LVG \cite{qian-etal-2024-visual}                                & Mobile Android                         & Swin Transformer \cite{liu2021swin} and BERT \cite{devlin2018bert} & -                                        & UI screenshots and free-form language expressions    & Bounding box coordinates                                    & UIBert dataset and synthetic dataset                                                                   & Unifies detection and grounding tasks through layout-guided contrastive learning                                                                                                       & /                                                                                                                              \\ \hline
Ferret-UI \cite{you2024ferretuigroundedmobileui}               & Android and iPhone platforms                      & Ferret \cite{you2023ferret}                                                                                                & 7B/13B                                                                     & Raw screen pixels, sub-images divided for finer resolution, bounding boxes and regional annotations & Widget bounding boxes, text from OCR tasks, descriptions of UI elements or overall screen functionality, UI interaction actions & Generated from RICO (for Android) and AMP (for iPhone)                                                                                                    & Multi-platform support with high-resolution adaptive image encoding                                                                                                                                                              & \url{https://github.com/apple/ml-ferret/tree/main/ferretui}     \\ \hline
Octopus \cite{chen2024octopusondevicelanguagemodel}            & Mobile devices                                    & CodeLlama-7B \cite{rozière2024codellamaopenfoundation}, Google Gemma 2B \cite{team2024gemma}                                                                                             & 7B, 2B                                                                     & API documentation examples                                                                          & Function names with arguments for API calls                                                                                     & RapidAPI Hub                                                                                                                                              & Use of conditional masking to enforce correct output formatting                                                                                                                                                                  & /                                                             \\ \hline
Octopus v2 \cite{chen2024octopusv2ondevicelanguage}            & Edge devices                                      & Gemma-2B \cite{team2024gemma}                                                                                                                   & 2B                                                                         & User queries and descriptions of available functions                                                & Function calls with precise parameters                                                                                          & 20 Android APIs, with up to 1,000 data points generated for training                                                                                      & Functional tokenization strategy, which assigns unique tokens to function calls, significantly reducing the context length required for accurate prediction                                                                      & /                                                             \\ \hline
Octopus v3 \cite{chen2024octopusv3technicalreport}             & Edge devices                                      & CLIP-based model  and a causal language model                                         & Less than 1 billion parameters                                             & Queries and commands, images and functional tokens                                                  & Functional tokens for actions                                                                                                   & Leveraged from Octopus v2 \cite{chen2024octopusv2ondevicelanguage}                                                                                        & Introduction of functional tokens for multimodal applications enables the representation of any function as a token, enhancing the model's flexibility                                                                           & /                                                             \\ \hline
Octopus v4 \cite{chen2024octopusv4graphlanguage}               & Serverless cloud-based platforms and edge devices & 17 models                                                                                                                  & Varies                                                                     & User queries                                                                                        & Domain-specific answers, actions                                                                                                & Synthetic datasets similar to Octopus v2                                                                                                                  & Graph-based framework integrating multiple specialized models for optimized performance                                                                                                                                          & \url{https://github.com/NexaAI/octopus-v4}                      \\ \hline
\end{tabular}
}
\end{table*}

\begin{table*}[!h]
\centering
\caption{An overview of GUI-optimized models on mobile platforms (Part II).}
\label{tab:mobile_model2}
\resizebox{\textwidth}{!}{ % Resize the entire figure to fit \textwidth
\begin{tabular}{p{1.cm}|p{1cm}|p{1.5cm}|p{1cm}|p{2.5cm}|p{2cm}|p{2.5cm}|p{4cm}|p{2cm}}
\hline
\multicolumn{1}{c|}{\textbf{Model}}                            & \multicolumn{1}{c|}{\textbf{Platform}} & \multicolumn{1}{p{1.5cm}|}{\textbf{Foundation Model}}                                                              & \multicolumn{1}{c|}{\textbf{Size}} & \multicolumn{1}{c|}{\textbf{Input}}                  & \multicolumn{1}{c|}{\textbf{Output}}                        & \multicolumn{1}{c|}{\textbf{Dataset}}                                                                  & \multicolumn{1}{c|}{\textbf{Highlights}}                                                                                                                                               & \multicolumn{1}{c}{\textbf{Link}}                                                                                             \\ \hline
VGA \cite{meng2024vgavisionguiassistant}                       & Mobile Android                                    & LLaVA-v1.6-mistral-7B \cite{liu2024visual}                                                                                                     & 7B                                                                         & GUI screenshots with positional, visual, and hierarchical data                                      & Actions and function calls, descriptions of GUI components, navigation and task planning                                        & 63.8k-image dataset constructed from the RICO                                                                                                             & Minimizes hallucinations in GUI comprehension by employing an image-centric fine-tuning approach, ensuring balanced attention between text and visual content                                                                    & \url{https://github.com/Linziyang1999/VGA\%2Dvisual\%2DGUI\%2Dassistant} \\ \hline
MobileFlow \cite{nong2024mobileflowmultimodalllmmobile}        & Mobile phones                                     & Qwen-VL-Chat \cite{bai2023qwen}                                                                                                               & 21B                                                                        & GUI screenshots with OCR textual information and bounding boxes                                     & GUI actions and question answering                                                                                              & 70k manually labeled business-specific data spanning 10 business sectors, and datasets like RefCOCO, ScreenQA, Flickr30K                                  & Hybrid visual encoder capable of variable-resolution input and Mixture of Experts (MoE) \cite{cai2024survey} for enhanced performance and efficiency                                                                             & /                                                             \\ \hline
UINav \cite{li2024uinav}                                       & Mobile Android                                    & SmallBERT \cite{turc2019well}                                                                                              & Agent model: 320k, Referee model: 430k, SmallBERT model: 17.6MB & UI elements, utterance, screen representation                                                       & Predicted actions and element to act upon                                                                                       & 43 tasks across 128 Android apps and websites, collecting 3,661 demonstrations                                                                            & Introduces a macro action framework and an error-driven demonstration collection process, significantly reducing training effort while enabling robust task performance with small, efficient models suitable for mobile devices & /                                                             \\ \hline
AppVLM \cite{papoudakis2025appvlm} & Android mobile devices & Paligemma-3B-896 \cite{beyer2024paligemma} & 3B & Annotated screenshots with bounding boxes and UI labels & GUI actions & AndroidControl \cite{li2024effects}, AndroidWorld \cite{rawles2024androidworlddynamicbenchmarkingenvironment} & A lightweight model that achieves near-GPT-4o performance in Android control tasks while being 10× faster and more resource-efficient. & / 
 \\ \hline
VSC-RL \cite{wu2025vsc} & Mobile Android & AutoUI \cite{zhang2024lookscreensmultimodalchainofaction}, Gemini-1.5-Pro & / & Screenshots & GUI actions & AitW & Addresses sparse-reward, long-horizon tasks for RL by autonomously breaking a complicated goal into subgoals  & \url{https://ai-agents-2030.github.io/VSC-RL}\\ \hline
Digi-Q \cite{bai2025digi} & Mobile Android & LLaVA-1.5 \cite{liu2024improved} & 7B & GUI screenshots & GUI actions, Q-values & AitW  \cite{androidinwildlargescaledataset} & Introduces a VLM-based Q-function for GUI agent training, enabling reinforcement learning without online interactions. & \url{https://github.com/DigiRL-agent/digiq} \\ \hline
VEM \cite{zheng2025vem} & Mobile Android & Qwen2VL \cite{wang2024qwen2vlenhancingvisionlanguagemodels} & 7B & GUI screenshots & GUI actions, Q-values & AitW  \cite{androidinwildlargescaledataset} & Unlike traditional RL methods that require environment interactions, VEM enables training purely on offline data with a Value Environment Model. & \url{https://github.com/microsoft/GUI-Agent-RL} \\ \hline
MP-GUI \cite{wang2025mp} & Mobile Android & InternViT-300M and InternLM2.5-7B-chat \cite{cai2024internlm2} & 8B & GUI screenshots & Natural language output, element grounding, captioning, and semantic navigation & 680K mixed-modality dataset & Introduces a tri-perceiver architecture that models textual, graphical, and spatial modalities to enhance GUI reasoning & \url{https://github.com/BigTaige/MP-GUI} \\ \hline
UI-R1 \cite{lu2025ui} & Mobile Android & Qwen2.5-VL-3B & 3B & GUI screenshots & Reasoning text and GUI actions & ScreenSpo and AndroidControl & Introduces a rule-based reinforcement learning approach using GRPO to enhance reasoning and action prediction in GUI tasks with only 136 examples & \url{https://github.com/lll6gg/UI-R1} \\\hline
ViMo \cite{luo2025vimogenerativevisualgui} & Mobile Android  & Pre-trained Stable Diffusion model \cite{rombach2022high} & / & Current GUI image, user action (in natural language), GUI text representation & GUI text representation of the next state and reconstructed full GUI image (visual prediction of the next screen) & Android Control and AITW & First GUI world model that predicts future visual GUI states & \url{https://ai-agents-2030.github.io/ViMo/} \\ \hline
\end{tabular}
}
\end{table*}

\subsection{LAMs for Mobile GUI Agents\label{sec:model:mobile}}

\begin{figure}[t]
    \centering
    \includegraphics[width=0.9\columnwidth]{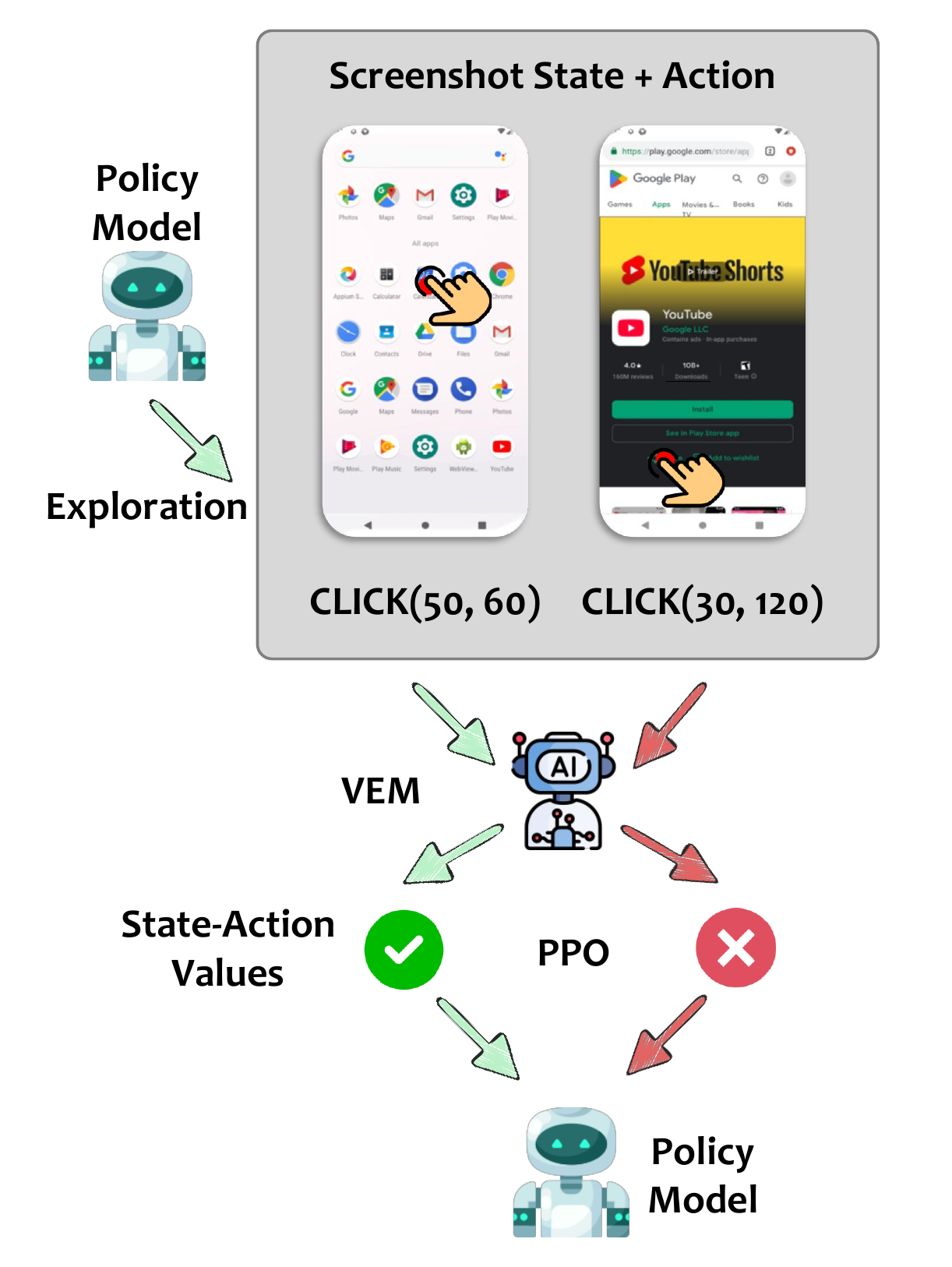}
    \vspace{-1em}
    \caption{The PPO training process of VEM \cite{zheng2025vem}. Figure adapted from the original paper.}
    \label{fig:vem}
\end{figure}

Mobile platforms present unique challenges for GUI agents, including diverse screen sizes, touch interactions, and resource constraints. Researchers have developed specialized LAMs to address these challenges, enhancing interaction and navigation within mobile environments.  We present an overview of LAMs tailored for mobile GUI agents in Table~\ref{tab:mobile_model1} and \ref{tab:mobile_model2}.

Focusing on detailed UI understanding, \textbf{MobileVLM}~\cite{wu2024mobilevlmvisionlanguagemodelbetter} introduces an advanced vision-language model designed specifically for mobile UI manipulation tasks. Built on Qwen-VL-Chat~\cite{bai2023qwen}, it incorporates mobile-specific pretraining tasks for intra- and inter-UI comprehension. By leveraging the Mobile3M dataset—a comprehensive corpus of 3 million UI pages and interaction traces organized into directed graphs—the model excels in action prediction and navigation tasks. MobileVLM's novel two-stage pretraining framework significantly enhances its adaptability to mobile UIs, outperforming existing VLMs in benchmarks like ScreenQA \cite{hsiao2024screenqalargescalequestionanswerpairs} and Auto-UI~\cite{zhang2024lookscreensmultimodalchainofaction}. This work highlights the effectiveness of tailored pretraining in improving mobile GUI agent performance.

Addressing the need for robust interaction in dynamic environments, \textbf{DigiRL}~\cite{bai2024digirltraininginthewilddevicecontrol} presents a reinforcement learning-based framework tailored for training GUI agents in Android environments. By leveraging offline-to-online RL, DigiRL adapts to real-world stochasticity, making it suitable for diverse, multi-step tasks. Unlike prior models reliant on imitation learning, DigiRL autonomously learns from interaction data, refining itself to recover from errors and adapt to new scenarios. The use of a pre-trained Vision-Language Model with 1.3 billion parameters enables efficient processing of GUI screenshots and navigation commands. Its performance on the AITW dataset demonstrates a significant improvement over baseline methods, positioning DigiRL as a benchmark in the development of intelligent agents optimized for complex GUI interactions.

Both \textbf{Digi-Q} \cite{bai2025digi} and \textbf{VEM} \cite{zheng2025vem} investigate the use of offline RL to enhance the performance of GUI agents without requiring direct interaction with the environment. Digi-Q employs temporal-difference learning to train a Q-function offline and derives policies through a Best-of-N selection strategy based on the predicted Q-values. Similarly, VEM introduces an environment-free RL framework tailored for training LLM-powered GUI agents using PPO. It directly estimates state-action values from offline data by fine-tuning with annotated value data from GPT-4o, thereby enabling policy training without real-time execution in a GUI environment. At inference time, only the policy model is utilized. Figure~\ref{fig:vem} illustrates the overall architecture of VEM. The study further demonstrates that offline RL with structured credit assignment can achieve performance comparable to interactive RL models. Overall, VEM offers a scalable and layout-agnostic approach for training GUI agents while minimizing interaction costs. Both works underscore the potential of offline RL for GUI agent training.

To enhance GUI comprehension and reduce reliance on textual data, \textbf{VGA}~\cite{meng2024vgavisionguiassistant} employs fine-tuned vision-language models that prioritize image-based cues such as shapes, colors, and positions. Utilizing the RICO \cite{rico} dataset for training, VGA is tailored for Android GUIs and employs a two-stage fine-tuning process to align responses with both visual data and human intent. The model excels in understanding GUI layouts, predicting design intents, and facilitating precise user interactions. By outperforming existing models like GPT-4V in GUI comprehension benchmarks, VGA sets a new standard for accuracy and efficiency in mobile GUI agents.

In the context of lightweight and efficient models, \textbf{UINav}~\cite{li2024uinav} demonstrates a practical system for training neural agents to automate UI tasks on mobile devices. It balances accuracy, generalizability, and computational efficiency through macro actions and an error-driven demonstration collection process. UINav uses a compact encoder-decoder architecture and SmallBERT \cite{turc2019well} for text and screen element encoding, making it suitable for on-device inference. A key innovation is its ability to generalize across diverse tasks and apps with minimal demonstrations, addressing key challenges in UI automation with a versatile framework.

\textbf{UI-R1} \cite{lu2025ui} introduces a RL–based training paradigm aimed at enhancing GUI action prediction for multimodal large language models (MLLMs). The resulting model, UI-R1-3B, fine-tunes Qwen2.5-VL-3B using a novel rule-based reward function that jointly evaluates action type correctness and click coordinate accuracy, while also enabling o1-style \cite{jaech2024openai} chain-of-thought (CoT) reasoning through structured \texttt{<think>} tags. UI-R1 relies on only 136 high-quality samples selected via a three-stage filtering strategy. Despite this limited supervision, UI-R1-3B achieves significant improvements on both in-domain and out-of-domain benchmarks. By leveraging Group Relative Policy Optimization (GRPO) \cite{shao2024deepseekmath}, the framework aligns policy optimization with the goals of GUI grounding and task execution. UI-R1 establishes a scalable and data-efficient approach for training GUI agents via RL and paves the way for lightweight yet effective agent design. Its methodology has also been successfully extended to cross-platform agents \cite{xia2025gui, liu2025infiguir1advancingmultimodalgui}, demonstrating strong generalization capabilities.

In addition to action models, \textbf{ViMo} \cite{luo2025vimogenerativevisualgui} introduces a novel generative visual world model for GUI agents, aimed at improving App agent decision-making by predicting the next GUI state as an image rather than a textual description. A key innovation of ViMo is the Symbolic Text Representation (STR), which replaces GUI text regions with structured placeholders to facilitate accurate and legible text synthesis. This decoupled design allows the system to handle GUI graphics generation using a fine-tuned diffusion model, and text generation through an LLM, thereby achieving high visual fidelity and semantic precision. ViMo significantly boosts both GUI prediction quality and downstream agent performance, with a reported 29.14\% relative improvement in GUI generation metrics and enhanced planning accuracy for long-horizon tasks. As a forward simulator, ViMo represents a crucial advancement toward reliable world models for mobile GUI agents, supporting more effective decision evaluation and trajectory planning in visual environments.

\begin{table*}[t]
\centering
\caption{An overview of GUI-optimized models on computer platforms.}
\label{tab:computer_model1}
\resizebox{\textwidth}{!}{ % Resize the entire figure to fit \textwidth
\begin{tabular}{p{1cm}|p{1cm}|p{1.5cm}|p{1cm}|p{2.5cm}|p{2cm}|p{2.5cm}|p{4cm}|p{2cm}}
\hline
\multicolumn{1}{c|}{\textbf{Model}}                            & \multicolumn{1}{c|}{\textbf{Platform}} & \multicolumn{1}{p{1.5cm}|}{\textbf{Foundation Model}}                                                              & \multicolumn{1}{c|}{\textbf{Size}} & \multicolumn{1}{c|}{\textbf{Input}}                  & \multicolumn{1}{c|}{\textbf{Output}}                        & \multicolumn{1}{c|}{\textbf{Dataset}}                                                                  & \multicolumn{1}{c|}{\textbf{Highlights}}                                                                                                                                               & \multicolumn{1}{c}{\textbf{Link}}                                                                                             \\ \hline
Screen-Agent \cite{niu2024screenagentvisionlanguagemodeldriven} & Linux and Windows desktop                         & CogAgent \cite{hong2023cogagentvisuallanguagemodel}                                                                                                                  & 18B                                                                        & GUI screenshots                                                                                     & Mouse and keyboard actions                                                                                                      & 273 task sessions                                                                                                                                         & Comprehensive pipeline of planning, acting, and reflecting to handle real computer screen operations autonomously                                                                                                                & \url{https://github.com/niuzaisheng/ScreenAgent}                \\ \hline
Octopus \cite{yang2025octopus}                                 & Desktop                                           & MPT-7B \cite{mosaicml2023mpt7b}  and CLIP ViT-L/14 \cite{radford2021learning}  & 7B                                                                         & Visual images, scene graphs containing objects and relations, environment messages                  & Executable action code and plans                                                                                                & OctoGibson: 476 tasks with structured initial and goal states; OctoMC: 40 tasks across biomes; OctoGTA: 25 crafted tasks spanning different game settings & Incorporates reinforcement learning with environmental feedback                                                                                                                                                                  & \url{https://choiszt.github.io/Octopus/}          \\ \hline
LAM~\cite{wang2024lam} & Windows OS & Mistral-7B \cite{jiang2023mistral} & 7B & Task requests in natural language, application environmental data & Plans, actions & 76,672 task-plan pairs, 2,192 task-action trajectories & The LAM model bridges the gap between planning and action execution in GUI environments. It introduces a multi-phase training pipeline combining task planning, imitation learning, self-boosting exploration, and reward-based optimization for robust action-oriented performance. & \url{https://github.com/microsoft/UFO/tree/main/dataflow} \\ \hline
ScreenLLM \cite{screenllm} & Desktop & LLaVA & 7B, 13B & GUI screenshots & Predicted GUI actions & High-resolution YouTube tutorials & Introduces a novel stateful screen schema to compactly represent GUI interactions over time, enabling fine-grained understanding and accurate action prediction & / \\ \hline
\end{tabular}
}
\end{table*}

\subsection{LAMs for Computer GUI Agents\label{sec:model:computer}}

For desktop and laptop environments, GUI agents must handle complex applications, multitasking, and varied interaction modalities. Specialized LAMs for computer GUI agents enhance capabilities in these settings, enabling more sophisticated task execution.  We overview of LAMs for computer GUI agents across in Table~\ref{tab:computer_model1}.

Integrating planning, acting, and reflecting phases, \textbf{ScreenAgent}~\cite{niu2024screenagentvisionlanguagemodeldriven} is designed for autonomous interaction with computer screens. Based on CogAgent \cite{hong2023cogagentvisuallanguagemodel}, it is fine-tuned using the ScreenAgent Dataset, providing comprehensive GUI interaction data across diverse tasks. With inputs as screenshots and outputs formatted in JSON for mouse and keyboard actions, ScreenAgent achieves precise UI element localization and handles continuous multi-step tasks. Its capability to process real-time GUI interactions using a foundation model sets a new benchmark for LLM-powered GUI agents, making it an ideal reference for future research in building more generalized intelligent agents.

Bridging high-level planning with real-world manipulation, \textbf{Octopus}~\cite{yang2025octopus} represents a pioneering step in embodied vision-language programming. Leveraging the MPT-7B~\cite{mosaicml2023mpt7b} and CLIP ViT-L/14~\cite{radford2021learning}, Octopus integrates egocentric and bird’s-eye views for visual comprehension, generating executable action code. Trained using the OctoVerse suite, its datasets encompass richly annotated environments like OmniGibson, Minecraft, and GTA-V, covering routine and reasoning-intensive tasks. Notably, Octopus innovates through Reinforcement Learning with Environmental Feedback, ensuring adaptive planning and execution. Its vision-dependent functionality offers seamless task generalization in unseen scenarios, underscoring its capability as a unified model for embodied agents operating in complex GUI environments.

Wang \etal \cite{wang2024lam} present a comprehensive overview of \textbf{LAMs}, a new paradigm in AI designed to perform tangible actions in GUI environments, using UFO~\cite{zhang2024ufouifocusedagentwindows} at Windows OS as a case study platform. Built on the Mistral-7B~\cite{jiang2023mistral} foundation, LAMs advance beyond traditional LLMs by integrating task planning with actionable outputs. Leveraging structured inputs from tools like the UI Automation (UIA) API, LAMs generate executable steps for dynamic planning and adaptive responses. A multi-phase training strategy—encompassing task-plan pretraining, imitation learning, self-boosting exploration, and reinforcement learning—ensures robustness and accuracy. Evaluations on real-world GUI tasks highlight LAMs' superior task success rates compared to standard models. This innovation establishes a foundation for intelligent GUI agents capable of transforming user requests into real-world actions, driving significant progress in productivity and automation.

These developments in computer GUI agents highlight the integration of advanced visual comprehension, planning, and action execution, paving the way for more sophisticated and capable desktop agents.

\begin{table*}[!h]
\centering
\caption{An overview of GUI-optimized models on cross-platform agents (Part I).}
\label{tab:cross_model1}
\resizebox{\textwidth}{!}{ % Resize the entire figure to fit \textwidth
\begin{tabular}{p{1cm}|p{1cm}|p{1.5cm}|p{1cm}|p{2cm}|p{2cm}|p{3cm}|p{4cm}|p{2cm}}
\hline
\multicolumn{1}{c|}{\textbf{Model}}                            & \multicolumn{1}{c|}{\textbf{Platform}} & \multicolumn{1}{p{1.5cm}|}{\textbf{Foundation Model}}                                                              & \multicolumn{1}{c|}{\textbf{Size}} & \multicolumn{1}{c|}{\textbf{Input}}                  & \multicolumn{1}{c|}{\textbf{Output}}                        & \multicolumn{1}{c|}{\textbf{Dataset}}                                                                  & \multicolumn{1}{c|}{\textbf{Highlights}}                                                                                                                                               & \multicolumn{1}{c}{\textbf{Link}}                                                                                             \\ \hline
RUIG \cite{zhang2023reinforceduiinstructiongrounding}     & Mobile and desktop                                 & Swin Transformer \cite{liu2021swin}  and BART \cite{lewis2019bart}  & 4 decoder layers                                                                             & UI screenshots and text instructions                                                                                                                                 & Bounding box predictions in linguistic form                                                                                                                  & MoTIF dataset and RicoSCA dataset for mobile UI data and Common Crawl for desktop UI data                                            & Innovatively uses policy gradients to improve the spatial decoding in the pixel-to-sequence paradigm                                                                                                     & /                                                \\ \hline
CogAgent \cite{hong2023cogagentvisuallanguagemodel}       & PC, web, and Android platforms                     & CogVLM-17B \cite{wang2024cogvlmvisualexpertpretrained}                                                    & 18B                                                                                          & GUI screenshots combined with OCR-derived text                                                                                                                       & Task plans, action sequences, and textual descriptions                                                                                                       & CCS400K, text recognition datasets: 80M synthetic text images, visual grounding datasets and GUI dataset Mind2Web and AiTW           & High-resolution cross-module to balance computational efficiency and high-resolution input processing                                                                                                    & \url{https://github.com/THUDM/CogVLM}              \\ \hline
SeeClick \cite{cheng2024seeclickharnessingguigrounding}   & iOS, Android, macOS, Windows, and web & Qwen-VL \cite{bai2023qwen}                                                                                                  & 9.6B                                                                                         & GUI screenshots and textual instructions                                                                                                                             & GUI actions and element locations for interaction                                                                                                            & 300k webpages with text and icons, RICO, and data from LLaVA                                                                         & Ability to perform GUI tasks purely from screenshots and its novel GUI grounding pre-training approach                                                                                                   & \url{https://github.com/njucckevin/SeeClick}       \\ \hline
ScreenAI \cite{baechler2024screenaivisionlanguagemodelui} & Mobile, desktop, and tablet UIs                    & PaLI-3 \cite{chen2023pali}                                                                                & 5B                                                                                           & GUI screenshots with OCR text, image captions, and other visual elements                                                                                             & Text-based answers for questions, screen annotations with bounding box coordinates and labels, navigation instructions, summaries of screen content          & 262M mobile web screenshots and 54M mobile app screenshots                                                                           & Unified representation of UIs and infographics, combining visual and textual elements                                                                                                                    & \url{https://github.com/kyegomez/ScreenAI}         \\ \hline
V-Zen \cite{rahman2024v} & Computers and Web & Vicuna-7B \cite{vicuna2023}, DINO \cite{liu2023grounding}, EVA-2-CLIP \cite{sun2023eva} & 7B & Text, GUI Images & Action Prediction, GUI Bounding Box & GUIDE \cite{chawla2024guide} & Dual-resolution visual encoding for precise GUI grounding and task execution & \url{https://github.com/abdur75648/V-Zen} \\ \hline
Ferret-UI 2 \cite{li2024ferret}                           & iPhone, Android, iPad, Web, AppleTV                & Vicuna-13B \cite{vicuna2023}, Gemma-2B \cite{team2024gemma}, Llama3-8B \cite{dubey2024llama}              & Vicuna-13B \cite{vicuna2023}, Gemma-2B \cite{team2024gemma}, Llama3-8B \cite{dubey2024llama} & UI screenshots, annotated bounding boxes and labels for UI widgets, OCR-detected text and bounding boxes for text elements, source HTML hierarchy trees for web data & Descriptions of UI elements, widget classification, OCR, tapability, and text/widget location, interaction instructions and multi-round interaction-based QA & Core-set, GroundUI-18k, GUIDE, Spotlight                                                                                             & Multi-platform support with high-resolution adaptive image encoding                                                                                                                                      & /                                                \\ \hline
% OmniParser \cite{lu2024omniparserpurevisionbased}         & Mobile, desktop, and webpage                       & BLIP-2 \cite{li2023blip}, YOLOv8 \cite{reis2023real}                                                      & -                                                                                            & UI screenshots with bounding boxes and numeric IDs, and structured representations including text from OCR and descriptions of functionality                         & IDs, bounding boxes, and descriptions for interactable elements                                                                                              & 67,000 UI screenshots labeled with bounding boxes and 7,185 icon-description pairs using GPT-4                                       & Introduces a vision-only screen parsing framework, enabling general UI understanding without reliance on external information, significantly improving action prediction accuracy for LLM-powered agents & \url{https://github.com/microsoft/OmniParser}      \\ \hline
ShowUI \cite{lin2024training}                             & Websites, desktops, and mobile phones              & Phi-3.5-Vision \cite{abdin2024phi}                                                                                   & 4.2B                                                                                         & GUI screenshots with OCR for text-based UI elements and visual grounding for icons and widgets                                                                       & GUI actions, navigation, element location                                                                                                                    & ScreenSpot, RICO, GUIEnv, GUIAct, AiTW, AiTZ, GUI-World                                                                              & Interleaved Vision-Language-Action approach, allowing seamless navigation, grounding, and understanding of GUI environments                                                                        & \url{https://github.com/showlab/ShowUI}            \\ \hline
OS-ATLAS \cite{wu2024atlas}                               & Windows, macOS, Linux, Android, and the web        & InternVL-2 \cite{chen2024internvl} and Qwen2-VL \cite{bai2023qwen}                                        & 4B/7B                                                                                        & GUI screenshots                                                                                                                                                      & GUI actions                                                                                                                                                  & AndroidControl, SeeClick, and others annotated with GPT-4, over 13 million GUI elements and 2.3 million screenshots                  & The first foundation action model designed for generalist GUI agents, supporting cross-platform GUI tasks, and introducing a unified action space                                                        & \url{https://osatlas.github.io/}                   \\ \hline
xLAM \cite{zhang2024xlam}                                 & Diverse environments                               & Mistral-7B \cite{jiang2023mistral} and DeepSeek-Coder-7B \cite{guo2024deepseek}                           & Range from 1B to 8×22B                                                                       & Unified function-calling data formats                                                                                                                                & Function calls, thought processes                                                                                                                            & Synthetic and augmented data, including over 60,000 high-quality samples generated using APIGen from 3,673 APIs across 21 categories & Excels in function-calling tasks by leveraging unified and scalable data pipelines                                                                                                                       & \url{https://github.com/SalesforceAIResearch/xLAM}
\\ \hline
SpiritSight \cite{huang2025spiritsight} & Web, Android, Windows Desktop & InternVL \cite{chen2024internvl} & 2B, 8B, and 26B & GUI screenshots & GUI actions & AitW \cite{androidinwildlargescaledataset}, CommonCrawl websites, and custom annotations & Introduces a Universal Block Parsing (UBP) method to resolve positional ambiguity in high-resolution visual inputs. & \url{https://hzhiyuan.github.io/SpiritSight-Agent}
\\ \hline
\end{tabular}
}
\end{table*}

\begin{table*}[!h]
\centering
\caption{An overview of GUI-optimized models on cross-platform agents (Part II).}
\label{tab:cross_model2}
\resizebox{\textwidth}{!}{ % Resize the entire figure to fit \textwidth
\begin{tabular}{p{1cm}|p{1cm}|p{1.5cm}|p{1cm}|p{2cm}|p{2cm}|p{3cm}|p{4cm}|p{2cm}}
\hline
\multicolumn{1}{c|}{\textbf{Model}}                            & \multicolumn{1}{c|}{\textbf{Platform}} & \multicolumn{1}{p{1.5cm}|}{\textbf{Foundation Model}}                                                              & \multicolumn{1}{c|}{\textbf{Size}} & \multicolumn{1}{c|}{\textbf{Input}}                  & \multicolumn{1}{c|}{\textbf{Output}}                        & \multicolumn{1}{c|}{\textbf{Dataset}}                                                                  & \multicolumn{1}{c|}{\textbf{Highlights}}                                                                                                                                               & \multicolumn{1}{c}{\textbf{Link}}                                                                                             \\ \hline
Falcon-UI~\cite{shen2024falcon} & iOS, Android, Windows, Linux, Web & Qwen2-VL-7B & 7B & Screenshots of GUI with node information and OCR annotations for visible elements & GUI actions and coordinates or bounding boxes for interaction elements & Insight-UI dataset, further fine-tuned on datasets such as AITW, AITZ, Android Control, and Mind2Web & Decouples GUI context comprehension from instruction-following tasks, leveraging an instruction-free pretraining approach. & / \\ \hline
UI-TARS \cite{qin2025uitarspioneeringautomatedgui} & Web, Desktop (Windows, macOS), and Mobile (Android) & Qwen-2-VL 7B and 72B \cite{wang2024qwen2vlenhancingvisionlanguagemodels} & 7B / 72B & GUI screenshots & GUI actions & GUI screenshots and metadata collected from websites, apps, and operating systems; action trace datasets from various GUI agent benchmarks; 6M GUI tutorials for reasoning enhancement; multiple open-source datasets & Pure vision-based perception with standardized GUI actions across platforms (Web, Mobile, Desktop). & \url{https://github.com/bytedance/UI-TARS} \\ \hline
Magma \cite{yang2025magma} & Web, Mobile, Desktop, Robotics & LLaMA-3-8B  \cite{dubey2024llama}, ConvNeXt-Xxlarge \cite{liu2022convnet} & 8.6B & GUI screenshots, textual task descriptions & GUI actions, robotic manipulation & UI, robotics data, human instructional videos & Jointly trains on heterogeneous datasets, enabling generalization across digital and physical tasks & \url{https://microsoft.github.io/Magma/} \\ \hline
GUI-R1   \cite{xia2025gui} & Windows, Linux, MacOS, Android, and Web & QwenVL2.5 \cite{bai2025qwen2}    & 3B and 7B & GUI screenshots & Reasoning text and GUI actions & Mixture of 3K high-quality samples & first framework   to apply rule-based reinforcement learning (RFT) to high-level GUI tasks   across platforms.                    & \url{https://github.com/ritzz-ai/GUI-R1.git} \\ \hline
InfiGUI-R1 \cite{liu2025infiguir1advancingmultimodalgui} & Web, Desktop, and Android & Qwen2.5-VL-3B-Instruct & 3B & GUI screenshots, Accessibility Tree & Reasoning text and GUI actions & Diverse dataset mixture & Two-stage training framework Actor2Reasoner: (1) Reasoning Injection via Spatial Reasoning Distillation, and (2) Deliberation Enhancement via Reinforcement Learning with Sub-goal Guidance and Error Recovery Scenario Construction & \url{https://github.com/Reallm-Labs/InfiGUI-R1} \\ \hline

Task Generalization \cite{zhang2025breaking} & Web and Android (Mobile) & Qwen2-VL-7B-Instruct \cite{wang2024qwen2vlenhancingvisionlanguagemodels} & 7B & GUI screenshots & Thoughts and grounded coordinate-based actions & 11 domain datasets with 56K GUI trajectory samples & Introduces mid-training on diverse non-GUI reasoning tasks (particularly math and code) to substantially enhance GUI agent planning capabilities & \url{https://github.com/hkust-nlp/GUIMid} \\ \hline

\end{tabular}
}
\end{table*}

\subsection{Cross-Platform Large Action Models\label{sec:model:cross}}

To achieve versatility across various platforms, cross-platform LAMs have been developed, enabling GUI agents to operate seamlessly in multiple environments such as mobile devices, desktops, and web interfaces.  We provide an analysis of  LAMs tailored for cross-platform GUI agents in Table~\ref{tab:cross_model1} and \ref{tab:cross_model2}.

\textbf{CogAgent}~\cite{hong2023cogagentvisuallanguagemodel} stands out as an advanced visual language model specializing in GUI understanding and navigation across PC, web, and Android platforms. Built on CogVLM~\cite{wang2024cogvlmvisualexpertpretrained}, it incorporates a novel high-resolution cross-module to process GUI screenshots efficiently, enabling detailed comprehension of GUI elements and their spatial relationships. Excelling in tasks requiring OCR and GUI grounding, CogAgent achieves state-of-the-art performance on benchmarks like Mind2Web~\cite{mind2webgeneralistagentweb} and AITW \cite{androidinwildlargescaledataset}. Its ability to generate accurate action plans and interface with GUIs positions it as a pivotal step in developing intelligent agents optimized for GUI environments. CogAgent has further evolved into its beta version, GLM-PC~\cite{cogagent2024}, offering enhanced control capabilities.

Focusing on universal GUI understanding, \textbf{Ferret-UI 2}~\cite{li2024ferret} from Apple is a state-of-the-art multimodal large language model designed to master UI comprehension across diverse platforms, including iPhones, Android devices, iPads, web, and AppleTV. By employing dynamic high-resolution image encoding, adaptive gridding, and high-quality multimodal training data generated through GPT-4, it outperforms its predecessor and other competing models in UI referring, grounding, and interaction tasks. Ferret-UI 2's advanced datasets and innovative training techniques ensure high accuracy in spatial understanding and user-centered interactions, setting a new benchmark for cross-platform UI adaptability and performance.

Advancing GUI automation, \textbf{ShowUI}~\cite{lin2024training} introduces a pioneering Vision-Language-Action  model that integrates high-resolution visual inputs with textual understanding to perform grounding, navigation, and task planning. Optimized for web, desktop, and mobile environments, ShowUI leverages the Phi-3.5-vision-instruct backbone and comprehensive datasets to achieve robust results across benchmarks like ScreenSpot \cite{cheng2024seeclickharnessingguigrounding} and GUI-Odyssey \cite{lu2024guiodysseycomprehensivedataset}. Its ability to process multi-frame and dynamic visual inputs alongside JSON-structured output actions highlights its versatility. With innovations in interleaved image-text processing and function-calling capabilities, ShowUI sets a new standard for LLM-powered GUI agents.

Addressing the need for a unified action space, \textbf{OS-ATLAS}~\cite{wu2024atlas} introduces a foundational action model specifically designed for GUI agents across platforms like Windows, macOS, Linux, Android, and the web. By leveraging a massive multi-platform dataset and implementing a unified action space, OS-ATLAS achieves state-of-the-art performance in GUI grounding and out-of-distribution generalization tasks. Its scalable configurations adapt to varying computational needs while maintaining versatility in handling natural language instructions and GUI elements. As a powerful open-source alternative to commercial solutions, OS-ATLAS marks a significant step toward democratizing access to advanced GUI agents.

Magma \cite{yang2025magma} is a foundation model for multimodal AI agents that integrates LLMs with vision and action understanding to complete UI navigation and robotic manipulation tasks. Unlike previous models optimized for either UI automation or robotics, Magma jointly trains on a heterogeneous dataset (about 39M samples) spanning UI screenshots, web navigation, robot trajectories, and instructional videos. It employs SoM and Trace-of-Mark techniques, which enhance action grounding and prediction by labeling actionable elements in GUI environments and tracking motion traces in robotic tasks.

UI-TARS \cite{qin2025uitarspioneeringautomatedgui} is an advanced, vision-based Large Action Model (LAM) optimized for multi-platform GUI agents. Unlike traditional approaches, it relies solely on GUI screenshots for perception, eliminating the need for structured representations. By incorporating a unified action space, UI-TARS enables seamless execution across Web, Windows, macOS, and Android environments. Built on Qwen-2-VL, it is trained on 6 million GUI tutorials, large-scale screenshot datasets, and multiple open-source benchmarks. A key innovation of UI-TARS is its System-2 reasoning capability, which allows it to generate explicit reasoning steps before executing actions, enhancing decision-making in dynamic environments. Additionally, it employs an iterative self-improvement framework, refining its performance through reflection-based learning. Experimental results demonstrate that UI-TARS outperforms existing models, including GPT-4o and Claude, in task execution benchmarks.

These cross-platform LAMs demonstrate the potential of unified models that can adapt to diverse environments, enhancing the scalability and applicability of GUI agents in various contexts.

\subsection{Takeaways\label{sec:model:takeaways}}

The exploration of LAMs for GUI agents has revealed several key insights that are shaping the future of intelligent interaction with graphical user interfaces:

\begin{enumerate}
    \item \textbf{Smaller Models for On-Device Inference:} Many of the optimized LAMs are built from smaller foundational models, often ranging from 1 billion to 7 billion parameters. This reduction in model size enhances computational efficiency, making it feasible to deploy these models on resource-constrained devices such as mobile phones and edge devices. The ability to perform on-device inference without relying on cloud services addresses privacy concerns and reduces latency, leading to a more responsive user experience.

    \item \textbf{Enhanced GUI Comprehension Reduces Reliance on Structured Data:} Models like VGA \cite{meng2024vgavisionguiassistant} and OmniParser \cite{lu2024omniparserpurevisionbased} emphasize the importance of visual grounding and image-centric fine-tuning to reduce dependency on structured UI metadata. By improving GUI comprehension directly from visual inputs, agents become more adaptable to different software environments, including those where structured data may be inaccessible or inconsistent.

    \item \textbf{Reinforcement Learning Bridges Static and Dynamic Environments:} The application of reinforcement learning in models like DigiRL \cite{bai2024digirltraininginthewilddevicecontrol} demonstrates the effectiveness of bridging static training data with dynamic real-world environments. This approach allows agents to learn from interactions, recover from errors, and adapt to changes, enhancing their robustness and reliability in practical applications.

    \item \textbf{Unified Function-Calling Enhances Interoperability:} Efforts to standardize data formats and function-calling mechanisms, as seen in models like xLAM \cite{zhang2024xlam}, facilitate multi-turn interactions and reasoning across different platforms. This unification addresses compatibility issues and enhances the agent's ability to perform complex tasks involving multiple APIs and services.

    \item \textbf{Inference-Time Computing and Reasoning Models:} Recent work highlights the importance of inference-time computing, where models plan, reason, and decompose tasks on the fly without architectural changes. Techniques such as extended context windows and chain-of-thought prompting (\eg ``o1-style'' reasoning) enable more robust, long-horizon decision-making. UI-R1 \cite{lu2025ui}, GUI-R1 \cite{xia2025gui} and InfiGUI-R1 \cite{liu2025infiguir1advancingmultimodalgui} are pioneering efforts in this direction. There is also growing interest in rule-based rewards and cost functions that guide inference-time behavior, integrating explicit heuristics to improve the stability, interpretability, and generalization of GUI agents.
\end{enumerate}
The advancements in LAMs for GUI agents highlight a trend toward specialized, efficient, and adaptable models capable of performing complex tasks across various platforms. By focusing on specialization, multimodal integration, and innovative training methodologies, researchers are overcoming the limitations of general-purpose LLMs. These insights pave the way for more intelligent, responsive, and user-friendly GUI agents that can transform interactions with software applications.

\section{Evaluation for LLM-Brained GUI Agents\label{sec:evaluation}}
\begin{figure*}[t]
    \centering
    \includegraphics[width=0.8\textwidth]{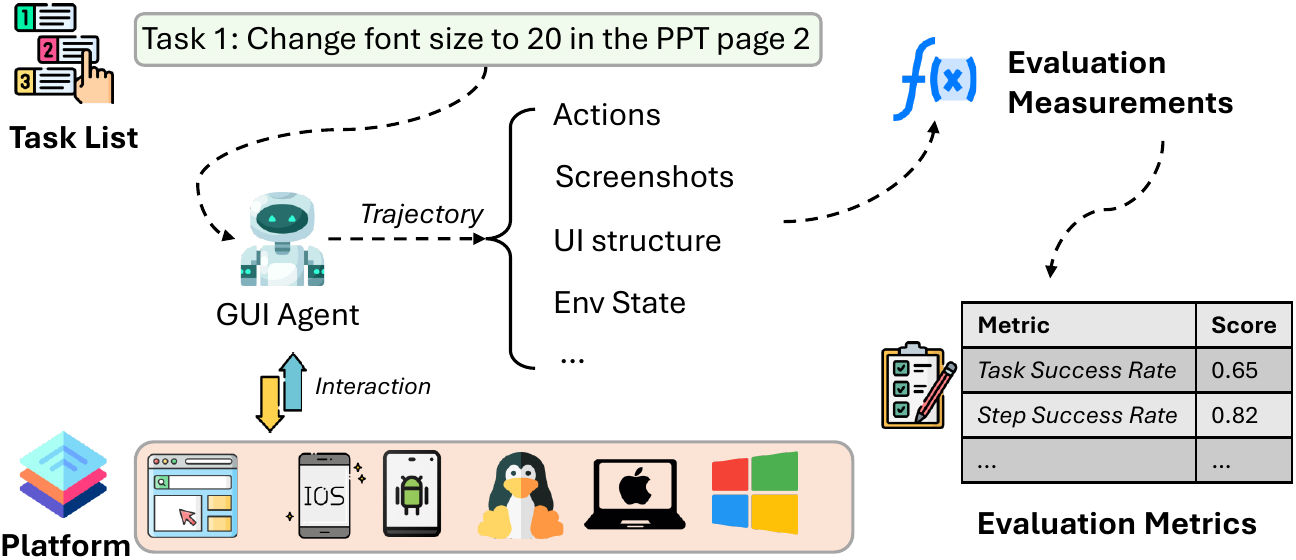}
    % \vspace{-5.5em}
    \caption{An illustrative example of evaluation of task completion by a GUI agent.}
    \label{fig:benchmark}
    % \vspace{-2em}
\end{figure*}

In the domain of GUI agents, evaluation is crucial for enhancing both functionality and user experience~\cite{li2024survey, huang2024survey2} and should be conducted across multiple aspects. By systematically assessing these agents' effectiveness across various tasks, evaluation   t only gauges their performance in different dimensions but also provides a framework for their continuous improvement~\cite{liu2023agentbench}. Furthermore, it encourages in  vation by identifying areas for potential development, ensuring that GUI agents evolve in tandem with advancements in LLMs and align with user expectations.

As illustrated in Figure~\ref{fig:benchmark}, when a GUI agent completes a task, it produces an action sequence, captures screenshots, extracts UI structures, and logs the resulting environment states. These outputs serve as the foundation for evaluating the agent's performance through various metrics and measurements across diverse platforms. In the subsequent sections, we delve into these evaluation methodologies, discussing the metrics and measurements used to assess GUI agents comprehensively. We also provide an overview of existing benchmarks tailored for GUI agents across different platforms, highlighting their key features and the challenges they address.

\subsection{Evaluation Metrics\label{sec:evaluation:metric}}
Evaluating GUI agents requires robust and multidimensional metrics to assess their performance across various dimensions, including accuracy, efficiency, and compliance (\eg safety). In a typical benchmarking setup, the GUI agent is provided with a natural language instruction as input and is expected to auto  mously execute actions until the task is completed. During this process, various assets can be collected, such as the sequence of actions taken by the agent, step-wise observations (\eg DOM or HTML structures), screenshots, runtime logs, final states, and execution time. These assets enable evaluators to determine whether the task has been completed successfully and to analyze the agent's performance. In this section, we summarize the key evaluation metrics commonly used for benchmarking GUI agents.   te that different research works may use different names for these metrics, but with similar calculations. We align their names in this section.

\begin{enumerate}
    \item \textbf{Step Success Rate:} Completing a task may require multiple steps. This metric measures the ratio of the number of steps that are successful over the total steps within a task. A high step success rate indicates precise and accurate execution of granular steps, which is essential for the reliable performance of tasks involving multiple steps~\cite{mind2webgeneralistagentweb, pan2024webcanvasbenchmarkingwebagents, androidinwildlargescaledataset}.
    
    \item \textbf{Turn Success Rate:} A \textit{turn} indicates a single interaction between the user and the agent. A turn may consist of multiple steps, and completing a task may consist of multiple turns. This metric measures the ratio of turns that successfully address the request in that interaction over all turns. It focuses on the agent's ability to understand and fulfill user expectations during interactive or dialog-based tasks, ensuring the agent's responsiveness and reliability across iterative interactions, particularly in tasks requiring dynamic user-agent communication~\cite{lù2024weblinxrealworldwebsitenavigation, deng2024multi}.
    
    \item \textbf{Task Success Rate:} Task success rate measures the successful task completion over all tasks set in the benchmark. It evaluates whether the final task completion state is achieved while ig  ring the intermediate steps. This metric provides an overall measure of end-to-end task completion, reflecting the agent's ability to handle complex workflows holistically~\cite{webshopscalablerealworldweb, mobileenvbuildingqualifiedevaluation, xie2024osworldbenchmarkingmultimodalagents}.
    
    \item \textbf{Efficiency Score:} Efficiency score evaluates how effectively the agent completes tasks while considering resource consumption, execution time, or total steps the agent might take. This metric can be broken down into the following sub-metrics:
    \begin{itemize}
        \item \textbf{Time Cost:} Measures the time taken to complete tasks.
        \item \textbf{Resource Cost:} Measures the memory/CPU/GPU usage to complete tasks.
        \item \textbf{LLM Cost:} Evaluates the computational or monetary cost of LLM calls used during task execution.
        \item \textbf{Step Cost:} Measures the total steps required to complete tasks.
    \end{itemize}
    Depending on the specific metrics used, the efficiency score can be interpreted differently in different papers~\cite{chen2024spa, deng2024mobile}.
    
    \item \textbf{Completion under Policy:} This metric measures the rate at which tasks are completed successfully while adhering to policy constraints. It ensures that the agent complies with user-defined or organizational rules, such as security, ethical, safety, privacy, or business guidelines, during task execution. This metric is particularly relevant for applications where compliance is as critical as task success~\cite{St-webagentbench}.
    
    \item \textbf{Risk Ratio:} Similar to the previous metric, the risk ratio evaluates the potential risk associated with the agent's actions during task execution. It identifies vulnerabilities, errors, or security concerns that could arise during task handling. A lower risk ratio indicates higher trustworthiness and reliability, while a higher ratio may suggest areas needing improvement to minimize risks and enhance robustness~\cite{St-webagentbench}.
\end{enumerate}
The implementation of metrics in each GUI agent benchmark might vary depending on the platform and the task formulation. In all tables in this section, we mapped the original metrics used in the benchmarks, which may possess different names, to the categories that we defined above.

\subsection{Evaluation Measurements\label{sec:evaluation:measurement}}
To effectively evaluate GUI agents, various measurement techniques are employed to assess their accuracy and alignment with expected outputs. These measurements validate different aspects of agent performance, ranging from textual and visual correctness to interaction accuracy and system state awareness, using code, models, and even agents as evaluators~\cite{Zhuge2024AgentasaJudgeEA}. Below, we summarize key measurement approaches used in benchmarking GUI agents. Based on these measurements, the evaluation metrics defined beforehand can be calculated accordingly.

\begin{enumerate}
\item \textbf{Text Match:} This measurement evaluates whether the text-based outputs of the agent match the expected results. For example, whether a target product name is reached when the agent is browsing an e-commerce website. It can involve different levels of strictness, including:
\begin{itemize}
    \item \textbf{Exact Match:} Ensures the output perfectly matches the expected result.
    \item \textbf{Partial or Fuzzy Match:} Allows for approximate matches, which are useful for handling mi  r variations such as typos or sy  nyms.
    \item \textbf{Semantic Similarity:} Measures deeper alignment in semantic meaning using techniques like cosine similarity of text embeddings or other semantic similarity measures.
\end{itemize}
Text Match is widely applied in tasks involving textual selections, data entry, or natural language responses.

\item \textbf{Image Match:} Image Match focuses on validating whether the agent acts or stops on the expected page (\eg webpage, app UI), or selects the right image. It involves comparing screenshots, selected graphical elements, or visual outcomes against ground truth images using image similarity metrics or visual question answering (VQA) methods. This measurement is particularly crucial for tasks requiring precise visual identification.

\item \textbf{Element Match:} This measurement checks whether specific widget elements (\eg those in HTML, DOM, or application UI hierarchies) interacted with by the agent align with the expected elements. These may include:
\begin{itemize}
    \item \textbf{HTML Tags and Attributes:} Ensuring the agent identifies and interacts with the correct structural elements.
    \item \textbf{URLs and Links:} Validating navigation-related elements.
    \item \textbf{DOM Hierarchies:} Confirming alignment with expected DOM structures in dynamic or complex web interfaces.
    \item \textbf{UI Controls and Widgets:} Verifying interactions with platform-specific controls such as buttons, sliders, checkboxes, dropdown menus, or other GUI components in desktop and mobile applications.
    \item \textbf{Accessibility Identifiers:} Utilizing accessibility identifiers or resource IDs in mobile platforms like Android and iOS to ensure correct element selection.
    \item \textbf{View Hierarchies:} Assessing alignment with expected view hierarchies in mobile applications, similar to DOM hierarchies in web applications.
    \item \textbf{System Controls and APIs:} Ensuring correct interaction with operating system controls or APIs, such as file dialogs, system menus, or   tifications in desktop environments.
\end{itemize}
Element Match ensures robust interaction with user interface components across different platforms during task execution.

\item \textbf{Action Match:} This measurement assesses the accuracy of the agent's actions, such as clicks, scrolls, or keystrokes, by comparing them against an expected sequence. It involves:
\begin{itemize}
    \item \textbf{Action Accuracy:} Validates that each action (including action type and its arguments) is performed correctly (\eg clicking the correct button, typing the right input).
    \item \textbf{Action Sequence Alignment:} Ensures actions occur in the correct order to meet task requirements.
    \item \textbf{Location Prediction:} Checks that spatial actions, such as mouse clicks or touch gestures, target the intended regions of the interface.
\end{itemize}
Action Match is vital for evaluating step-wise correctness in task completion.

\item \textbf{State Information:} State Information captures runtime data related to the system's environment during task execution. It provides insights into contextual factors that may influence the agent's behavior, such as:
\begin{itemize}
    \item \textbf{Application State:} Information about the state of the application being interacted with (\eg open files, active windows, saved files in given locations).
    \item \textbf{System Logs:} Detailed logs recording the agent's decisions and interactions.
    \item \textbf{Environment Variables:} Contextual data about the operating system or runtime environment.
\end{itemize}
This measurement is valuable for debugging, performance analysis, and ensuring reliability under diverse conditions.
\end{enumerate}

Each of these measurement techniques contributes to a comprehensive evaluation framework, ensuring that the agent   t only completes tasks but does so with precision, efficiency, and adaptability. Together, they help build trust in the agent's ability to perform reliably in real-world scenarios while maintaining compliance with policy constraints.

\subsection{Evaluation Platforms\label{sec:evaluation:platform}}
Evaluating GUI agents requires diverse platforms to capture the varying environments in which these agents operate. The platforms span web, mobile, and desktop environments, each with unique characteristics, challenges, and tools for evaluation. This section summarizes the key aspects of these platforms and their role in benchmarking GUI agents.

\begin{enumerate}
\item \textbf{Web:} Web platforms are among the most common environments for GUI agents, reflecting their prevalence in everyday tasks such as browsing, form filling, and data scraping. Key characteristics of web platforms for evaluation include:
    \begin{itemize}
    \item \textbf{Dynamic Content:} Web applications often involve dynamic elements generated through JavaScript, AJAX, or similar tech  logies, requiring agents to handle asynchro  us updates effectively.
    \item \textbf{Diverse Frameworks:} The variety of web tech  logies (\eg HTML, CSS, JavaScript frameworks) demands robust agents capable of interacting with a range of interface designs and structures.
    \item \textbf{Tools and Libraries:} Evaluation often uses tools such as Selenium, Puppeteer, or Playwright to emulate browser interactions, collect runtime information, and compare outcomes against expected results.
    \item \textbf{Accessibility Compliance:} Metrics like WCAG (Web Content Accessibility Guidelines) adherence can also be evaluated to ensure inclusivity.
    \end{itemize}

\item \textbf{Mobile:} Mobile platforms, particularly Android and iOS, pose unique challenges for GUI agents due to their constrained interfaces and touch-based interactions. Evaluating agents on mobile platforms involves:
    \begin{itemize}
    \item \textbf{Screen Size Constraints:} Agents must adapt to limited screen real estate, ensuring interactions remain accurate and efficient.
    \item \textbf{Touch Gestures:} Evaluating the agent's ability to simulate gestures such as taps, swipes, and pinches is essential.
    \item \textbf{Platform Diversity:} Android devices vary significantly in terms of screen sizes, resolutions, and system versions, while iOS offers more standardized conditions.
    \item \textbf{Evaluation Tools:} Tools like Appium and Espresso (for Android) or XCTest (for iOS) and emulators are commonly used for testing and evaluation.
    \end{itemize}

\item \textbf{Desktop:} Desktop platforms provide a richer and more complex environment for GUI agents, spanning multiple operating systems such as Windows, macOS, and Linux. Evaluations on desktop platforms often emphasize:
    \begin{itemize}
    \item \textbf{Application Diversity:} Agents must handle a wide range of desktop applications, including productivity tools, web browsers, and custom enterprise software.
    \item \textbf{Interaction Complexity:} Desktop interfaces often include advanced features such as keyboard shortcuts, drag-and-drop, and context menus, which agents must handle correctly.
    \item \textbf{Cross-Platform Compatibility:} Evaluations may involve ensuring agents can operate across multiple operating systems and versions.
    \item \textbf{Automation Frameworks:} Tools such as Windows UI Automation, macOS Accessibility APIs, and Linux's AT-SPI are used to automate and monitor agent interactions.
    \item \textbf{Resource Usage:} Memory and CPU usage are significant metrics, particularly for long-running tasks or resource-intensive applications.
    \end{itemize}
\end{enumerate}

Each platform presents distinct challenges and opportunities for evaluating GUI agents. Web platforms emphasize scalability and dynamic interactions, mobile platforms focus on touch interfaces and performance, and desktop platforms require handling complex workflows and cross-application tasks. Some benchmarks are cross-platform, requiring agents to be robust, adaptable, and capable of generalizing across different environments.

All the metrics, measurements, and platforms discussed are essential for a comprehensive evaluation of GUI agents across multiple aspects. Most existing benchmarks rely on them for evaluation. In what follows, detail these benchmarks for GUI agents selectively.

\begin{table*}[!h]
\centering
\caption{Overview of web GUI agent benchmarks (Part I).\label{tab:web_benchmark1}}
\resizebox{\textwidth}{!}{%
\begin{tabular}{p{2cm}|p{1.5cm}|p{1cm}|p{1cm}|p{3.5cm}|p{1.5cm}|p{2cm}|p{2cm}|p{2cm}}
\hline
\multicolumn{1}{c|}{\textbf{Benchmark}}     & \multicolumn{1}{c|}{\textbf{Platform}}  & \multicolumn{1}{c|}{\textbf{Year}} & \multicolumn{1}{c|}{\textbf{Live}} & \multicolumn{1}{c|}{\textbf{Highlight}}      & \multicolumn{1}{c|}{\textbf{Data Size}}    & \multicolumn{1}{c|}{\textbf{Metric}}  & \multicolumn{1}{c|}{\textbf{Measurement}}  & \multicolumn{1}{c}{\textbf{Link}}   \\ \hline
MiniWoB++~\cite{wob,reinforcementlearningonwebinterfaces} & Web & 2017 & Yes & Evaluates agents on basic web interactions like clicking, typing, and form navigation. & 100 web interaction tasks & Task Success Rate & Element Match & \url{https://github.com/Farama\%2DFoundation/miniwob\%2Dplusplus} \\ \hline
RUSS~\cite{xu2021groundingopendomaininstructionsautomate} & Web & 2021 & No & Uses ThingTalk for mapping natural language to web actions, enabling precise web-based task execution in real HTML environments. & 741 instructions & Task Success Rate & Text Match, Element Match & \url{https://github.com/xnancy/russ} \\ \hline
WebShop~\cite{webshopscalablerealworldweb} & Web & 2022 & Yes & Simulates e-commerce navigation with real-world products, challenging agents with instruction comprehension, multi-page navigation, and strategic exploration. & 12,087 instructions & Task Success Rate, Step Success Rate" & Text Match & \url{https://webshop-pnlp.github.io/} \\ \hline
Mind2Web~\cite{mind2webgeneralistagentweb} & Web & 2023 & No & Tests adaptability on real-world, dynamic websites across domains. & 2,000 tasks & Step Success Rate, Task Success Rate  &  Element Match, Action Match & \url{https://github.com/OSU-NLP-Group/Mind2Web} \\ \hline
Mind2Web-Live~\cite{pan2024webcanvasbenchmarkingwebagents} & Web & 2024 & Yes & Provides intermediate action tracking for realistic task assessment, along with an updated Mind2Web-Live dataset and tools for an  tation. & 542 tasks & Step Success Rate, Task Success Rate, Efficiency Score & Element Match, Text Match, trajectory length  & \url{https://huggingface.co/datasets/iMeanAI/Mind2Web-Live} \\ \hline
Mind2Web-Live-Abstracted \cite{shahbandeh2024naviqate} & Web & 2024 & Yes & Abstract the descriptions by omitting task-specific details and user input information in Mind2Web-Live, which are more streamlined and less time-consuming to compose. & 104 samples & Task Success Rate, Efficiency Score & Text Match, Image Match, Element Match, Path Length & \url{https://anonymous.4open.science/r/naviqate} \\ \hline
WebArena~\cite{webarenarealisticwebenvironment} & Web & 2023 & Yes & Simulates realistic, multi-tab browsing on Docker-hosted websites, focusing on complex, long-horizon tasks that mirror real online interactions. & 812 long-horizon tasks & Step Success Rate & Text Match & \url{https://webarena.dev/} \\ \hline
VisualWebArena \cite{koh2024visualwebarenaevaluatingmultimodalagents} & Web & 2024 & Yes & Assesses multimodal agents on visually grounded tasks, requiring both visual and textual interaction capabilities in web environments. & 910 tasks & Step Success Rate & Text Match, Image Match & \url{https://jykoh.com/vwa} \\ \hline
MT-Mind2Web~\cite{deng2024multiturninstructionfollowingconversational} & Web & 2024 & No & Introduces conversational web navigation with multi-turn interactions, supported by a specialized multi-turn web dataset. & 720 sessions/3525 instructions & Step Success Rate, Turn Success Rate & Element Match, Action Match & \url{https://github.com/magicgh/self-map} \\ \hline
\end{tabular}}
\end{table*}

\begin{table*}[!h]
\centering
\caption{Overview of web GUI agent benchmarks (Part II).\label{tab:web_benchmark2}}
\resizebox{\textwidth}{!}{%
\begin{tabular}{p{2cm}|p{1.5cm}|p{1cm}|p{1cm}|p{3.5cm}|p{1.5cm}|p{2cm}|p{2cm}|p{2cm}}
\hline
\multicolumn{1}{c|}{\textbf{Benchmark}}     & \multicolumn{1}{c|}{\textbf{Platform}}  & \multicolumn{1}{c|}{\textbf{Year}} & \multicolumn{1}{c|}{\textbf{Live}} & \multicolumn{1}{c|}{\textbf{Highlight}}      & \multicolumn{1}{c|}{\textbf{Data Size}}    & \multicolumn{1}{c|}{\textbf{Metric}}  & \multicolumn{1}{c|}{\textbf{Measurement}}  & \multicolumn{1}{c}{\textbf{Link}}   \\ \hline
MMInA~\cite{zhang2024mminabenchmarkingmultihopmultimodal} & Web & 2024 & Yes & Tests multihop, multimodal tasks on real-world websites, requiring agents to handle cross-page information extraction and reasoning for complex tasks. & 1,050 tasks & Step Success Rate, Task Success Rate & Text Match, Element Match & \url{https://mmina.cliangyu.com/} \\ \hline
AutoWebBench \cite{lai2024autowebglmbootstrapreinforcelarge} & Web & 2024 & No & Bilingual web browsing benchmark with 10,000 browsing traces, supporting evaluation across language-specific environments. & 10,000 traces & Step Success Rate, Efficiency Score &  Element Match, Action Match, Time & \url{https://github.com/THUDM/AutoWebGLM}\\ \hline
WorkArena \cite{drouin2024workarena}                                           & Web                                & 2024 & Yes & Focuses on real-world enterprise software interactions, targeting tasks frequently performed by knowledge workers                                                                          & 19,912 unique task instances                                            & Task Success Rate, Efficiency Score, Completion under Policy, Turn Success Rate & Element Match, Text Match, Execution-based Validation          & \url{https://github.com/ServiceNow/WorkArena}     \\ \hline
VideoWebArena \cite{jang2024videowebarena}                                     & Web                                & 2024 & Yes & Focuses on long-context multimodal agents using video tutorials for task completion                                                                                                        & 74 videos amounting to approximately 4 hours, with 2,021 tasks in total & Task Success Rate, Intermediate Intent Success Rate, Efficiency Scores          & Element Match, State Information, Exact and Fuzzy Text Matches & \url{https://github.com/ljang0/videowebarena}     \\ \hline
EnvDistraction \cite{ma2024cautionenvironmentmultimodalagents}                 & Web                                & 2024 & No & Evaluates the ``faithfulness'' of multimodal GUI agents by assessing their susceptibility to environmental distractions, such as pop-ups, fake search results, or misleading recommendations & 1,198 tasks                                                             & Task Success Rate                                                               & Text Match, Element Match, State Information                   & \url{https://github.com/xbmxb/EnvDistraction}     \\ \hline
WebVLN-v1 \cite{Chen_Pitawela_Zhao_Zhou_Chen_Wu_2024} & Web & 2024 & No & Combines navigation and question-answering on shopping sites, integrating visual and textual content for unified web interaction evaluation. & 8,990 paths and 14,825 QA pairs & Task Success Rate, Efficiency Score & Element Match, Path Length, Trajectory Length & \url{https://github.com/WebVLN/WebVLN} \\ \hline
WEBLINX \cite{lù2024weblinxrealworldwebsitenavigation} & Web & 2024 & No & Focuses on conversational navigation, requiring agents to follow multi-turn user instructions in realistic, dialogue-based web tasks. & 100k interactions & Turn Success Rate & Element Match,  Text Match, Action Match & \url{https://mcgill-nlp.github.io/weblinx/} \\ \hline
ST-WebAgentBench \cite{St-webagentbench} & Web  & 2024 & Yes & Evaluates policy-driven safety in web agents, using the Completion under Policy metric to ensure compliance in enterprise-like environments. & 235 tasks & Task Success Rate, Completion under Policy (CuP), Risk Ratio & Element Match, Action Match, Text Match & \url{https://sites.google.com/view/st-webagentbench/home} \\ \hline
CompWoB \cite{furuta2024exposing} & Web & 2023 & No & Tests agents on sequential, compositional tasks that require state management across multiple steps, simulating real-world automation scenarios. & 50 compositional tasks & Task Success Rate & Element Match & \url{https://github.com/google-research/google-research/tree/master/compositional\_rl/compwob} \\ \hline
\end{tabular}}
\end{table*}

\begin{table*}[!h]
\centering
\caption{Overview of web GUI agent benchmarks (Part III).\label{tab:web_benchmark3}}
\resizebox{\textwidth}{!}{%
\begin{tabular}{p{2cm}|p{1.5cm}|p{1cm}|p{1cm}|p{3.5cm}|p{1.5cm}|p{2cm}|p{2cm}|p{2cm}}
\hline
\multicolumn{1}{c|}{\textbf{Benchmark}}     & \multicolumn{1}{c|}{\textbf{Platform}}  & \multicolumn{1}{c|}{\textbf{Year}} & \multicolumn{1}{c|}{\textbf{Live}} & \multicolumn{1}{c|}{\textbf{Highlight}}      & \multicolumn{1}{c|}{\textbf{Data Size}}    & \multicolumn{1}{c|}{\textbf{Metric}}  & \multicolumn{1}{c|}{\textbf{Measurement}}  & \multicolumn{1}{c}{\textbf{Link}}   \\ \hline
TURKING BENCH \cite{xu2024tur} & Web & 2024 & Yes & Uses natural HTML tasks from crowdsourcing to assess interaction skills with real-world web layouts and elements. & 32.2K instances & Task Success Rate & Text Match, Element Match, Image Match & \url{https://turkingbench.github.io} \\ \hline
VisualWebBench \cite{liu2024visualwebbenchfarmultimodalllms} & Web & 2024 & No & Provides a fine-grained assessment of multimodal large language models (MLLMs) on web-specific tasks & 1,534 instances from 139 real websites across 87 sub-domains & Task Success Rate, Turn Success Rate, Efficiency Metrics & Text Match, Image Match, Element Match, Action Match & \url{https://visualwebbench.github.io/} \\ \hline
WONDERBREAD \cite{wornowwonderbread} & Web & 2024 & No & Focuses on business process management (BPM) tasks like documentation, knowledge transfer, and process improvement & 2,928 human demonstrations across 598 distinct workflows & Task Success Rate, Step Success Rate, Efficiency Score, Completion under Policy & Text Match, Action Match, State Information & \url{https://github.com/HazyResearch/wonderbread} \\ \hline
WebOlympus \cite{zheng2024webolympus} & Web & 2024 & Yes & An open platform for web agents that simplifies running demos, evaluations, and data collection for web agents on live websites & 50 tasks & Task Success Rate, Step Success Rate & Action Match & / \\ \hline
BrowserGym \cite{chezelles2024browsergym} & Web & 2024 & Yes & Provides a unified, extensible, and open-source environment for evaluating web agents with standardized APIs and observations. & Benchmarks include MiniWoB(++) with 125 tasks, WebArena with 812 tasks, and WorkArena with up to 341 tasks per level. & Task Success Rate, Step Success Rate, Turn Success Rate, Efficiency Metrics. & Text-based matching and element match. & \url{https://github.com/ServiceNow/BrowserGym} \\ \hline
WebWalkerQA \cite{wu2025webwalkerbenchmarkingllmsweb} & Web & 2025 & Yes & Benchmarks the capacity of LLMs to handle deep, structured, and realistic web-based navigation and reasoning tasks. & 680 high-quality QA pairs. & Task Success Rate, Efficiency Score. & Text Match, Action Match. & \url{https://github.com/Alibaba-NLP/WebWalker}  \\ \hline
WebGames \cite{thomas2025webgames} & Web & 2025 & Yes & A comprehensive benchmark designed to evaluate the capabilities of general-purpose web-browsing AI agents through 50+ interactive challenges. It uniquely provides a hermetic testing environment with verifiable ground-truth solutions. & 50+ challenges & Task Success Rate & Action Match & \url{https://github.com/convergence-ai/webgames} \\ \hline
\end{tabular}}
\end{table*}

\begin{table*}[!h]
\centering
\caption{Overview of web GUI agent benchmarks (Part IV).\label{tab:web_benchmark4}}
\resizebox{\textwidth}{!}{%
\begin{tabular}{p{2cm}|p{1.5cm}|p{1cm}|p{1cm}|p{3.5cm}|p{1.5cm}|p{2cm}|p{2cm}|p{2cm}}
\hline
\multicolumn{1}{c|}{\textbf{Benchmark}}     & \multicolumn{1}{c|}{\textbf{Platform}}  & \multicolumn{1}{c|}{\textbf{Year}} & \multicolumn{1}{c|}{\textbf{Live}} & \multicolumn{1}{c|}{\textbf{Highlight}}      & \multicolumn{1}{c|}{\textbf{Data Size}}    & \multicolumn{1}{c|}{\textbf{Metric}}  & \multicolumn{1}{c|}{\textbf{Measurement}}  & \multicolumn{1}{c}{\textbf{Link}}   \\ \hline
SafeArena \cite{tur2025safearenaevaluatingsafetyautonomous} & Web & 2025 & Yes & The first benchmark specifically designed to evaluate the deliberate misuse of web agents by testing their ability to complete both safe and harmful tasks. & 500 tasks & Task Success Rate, Completion under Policy, Risk Ratio & Text Match, State Information & \url{https://safearena.github.io} \\ \hline
WABER \cite{kara2025waber} & Web & 2025 & Yes & Introduces two new evaluation metrics—Efficiency and Reliability—that go beyond standard success rate measurements & 655 tasks & Task Success Rate, Efficiency Score & Action Match, State Information  & \url{https://github.com/SumanKNath/WABER} \\ \hline
Online-Mind2Web \cite{xue2025illusion} & Web & 2025 & Yes & A real-world online evaluation benchmark designed to reflect actual user interactions with live web interfaces & 300 tasks from 136 websites & Task Success Rate, Efficiency Score & Image Match, Action Match, State Information, LLM-as-a-Judge Evaluation & \url{https://github.com/OSU-NLP-Group/Online-Mind2Web} \\ \hline
AgentDAM \cite{zharmagambetov2025agentdam} & Web & 2025 & Yes & The first benchmark to evaluate privacy leakage risks in multimodal, realistic web environments using agentic models & 246 human-annotated test cases & Task Success Rate, Risk Ratio & Action Match, Text Match & \url{https://github.com/facebookresearch/ai-agent-privacy} \\ \hline
AgentRewardBench \cite{lu2025agentrewardbench} & Web & 2025 & No & The first benchmark to rigorously evaluate LLM-based judges against human expert annotations across multiple web agent tasks & 1,302 trajectories, 351 tasks & Task Success Rate, Completion under Policy & Image Match, Element/State Match & \url{https://agent-reward-bench.github.io}\\ \hline
RealWebAssist \cite{ye2025realwebassist} & Web & 2025 & No & The first benchmark to evaluate long-horizon web assistance using real-world users’ sequential instructions expressed in natural and often ambiguous language & 1,885 instructions & Task Success Rate, Step Success Rate, Efficiency Score & Action Match & \url{https://scai.cs.jhu.edu/projects/RealWebAssist/}\\ \hline
REAL \cite{garg2025real} & Web & 2025 & Yes & Fully deterministic, high-fidelity replicas of real-world websites (e.g., Airbnb, Amazon, Gmail), enabling safe, reproducible, and configurable testing for multi-turn GUI-based agents & 112 tasks across 11 deterministic websites & Task Success Rate & Text Match, Action Match, State Information Match & \url{https://github.com/agi-inc/agisdk} \\ \hline
BEARCUBS \cite{song2025bearcubs} & Web & 2025 & Yes & Emphasizes interaction with live web pages and includes multimodal tasks (e.g., video, audio, 3D) that cannot be solved by text-only methods, addressing limitations of prior benchmarks relying on static or simulated environments & 111 questions & Task Success Rate, Efficiency Score & Text Match, Action Match & \url{https://bear-cubs.github.io} \\ \hline
WASP \cite{evtimov2025wasp} & Web & 2025 & Yes & The first end-to-end benchmark for evaluating web agents' security under realistic prompt injection attacks, simulating attacker capabilities in live sandboxed web environments & 84 test cases & Task Success Rate, Completion under Policy, Risk Ratio & Action Match, State Information & / \\ \hline

\end{tabular}}
\end{table*}

\subsection{Web Agent Benchmarks\label{sec:evaluation:web}}

Evaluating GUI agents in web environments necessitates benchmarks that capture the complexities and nuances of web-based tasks. Over the years, several benchmarks have been developed, each contributing unique perspectives and challenges to advance the field. We first provide an overview of these benchmarks in Tables~\ref{tab:web_benchmark1}, \ref{tab:web_benchmark2}, \ref{tab:web_benchmark3} and \ref{tab:web_benchmark4}.

One of the pioneering efforts in this domain is \textbf{MiniWoB++}~\cite{wob, reinforcementlearningonwebinterfaces}, focusing on assessing reinforcement learning agents on web-based GUI tasks. It introduces realistic interaction scenarios, including clicking, typing, and navigating web elements, and leverages workflow-guided exploration (WGE) to improve efficiency in environments with sparse rewards. Agents are evaluated based on success rates, determined by their ability to achieve final goal states, highlighting adaptability and robustness across various complexities.

Building upon the need for more realistic environments, \textbf{Mind2Web}\cite{mind2webgeneralistagentweb} represents a significant advancement by enabling agents to handle real-world HTML environments rather than simplified simulations. Established after the advent of LLMs\cite{yan2023gpt}, it offers a large dataset of over 2,000 tasks spanning multiple domains, presenting challenges from basic actions to complex multi-page workflows. The benchmark emphasizes end-to-end task performance through metrics like Element Accuracy and Task Success Rate, encouraging rigorous evaluation of agents.

Extending Mind2Web's capabilities, \textbf{MT-Mind2Web}~\cite{deng2024multiturninstructionfollowingconversational} introduces conversational web navigation, requiring sophisticated interactions that span multiple turns with both users and the environment. This advanced benchmark includes 720 web navigation conversation sessions with 3,525 instruction and action sequence pairs, averaging five user-agent interactions per session, thereby testing agents' conversational abilities and adaptability.

To further enhance realism, \textbf{WebArena}~\cite{webarenarealisticwebenvironment} sets a new standard with its realistic web environment that mimics genuine human interactions. Featuring 812 tasks across multiple domains, it requires agents to perform complex, long-horizon interactions over multi-tab web interfaces. By focusing on functional correctness rather than surface-level matches, WebArena promotes thorough assessment of agents' practical abilities.

Recognizing the importance of multimodal capabilities, \textbf{VisualWebArena}, an extension of WebArena~\cite{webarenarealisticwebenvironment}, was designed to assess agents on realistic visually grounded web tasks. Comprising 910 diverse tasks in domains like Classifieds, Shopping, and Reddit, it adds new visual functions for measuring open-ended tasks such as visual question answering and fuzzy image matching, thereby challenging agents in multimodal understanding.

Similarly, \textbf{VideoWebArena}~\cite{jang2024videowebarena} focuses on evaluating agents' abilities to comprehend and interact with video content on the web. It presents 74 videos across 2,021 tasks, challenging agents in video-based information retrieval, contextual reasoning, and skill application. This benchmark highlights critical deficiencies in current models, emphasizing the need for advancements in agentic reasoning and video comprehension.

Complementing this, \textbf{VisualWebBench}~\cite{liu2024visualwebbenchfarmultimodalllms} offers a multimodal benchmark that assesses understanding, OCR, grounding, and reasoning across website, element, and action levels. Spanning 1.5K samples from real-world websites, it identifies challenges such as poor grounding and subpar OCR with low-resolution inputs, providing a crucial evaluation perspective distinct from general multimodal benchmarks.

Beyond the challenges of multimodality, understanding agents' resilience to environmental distractions is crucial. \textbf{EnvDistraction}~\cite{ma2024cautionenvironmentmultimodalagents} introduces a benchmark that evaluates the faithfulness of multimodal GUI agents under   n-malicious distractions, such as pop-ups and recommendations. The study demonstrates that even advanced agents are prone to such distractions, revealing vulnerabilities that necessitate robust multimodal perception for reliable automation.

Focusing on safety and trustworthiness, \textbf{ST-WebAgentBench}~\cite{St-webagentbench} takes a unique approach by emphasizing the management of unsafe behaviors in enterprise settings. It features a human-in-the-loop system and a policy-driven hierarchy, introducing the Completion under Policy (CuP) metric to evaluate agents' compliance with organizational, user, and task-specific policies. This benchmark operates in web environments using BrowserGym \cite{chezelles2024browsergym} and includes 235 tasks with policies addressing various safety dimensions, providing a comprehensive framework for evaluating agents in enterprise scenarios.

Addressing the automation of enterprise software tasks, \textbf{WorkArena}~\cite{drouin2024workarena} offers a benchmark emphasizing tasks commonly performed within the Service  w platform. With 19,912 unique instances across 33 tasks, it highlights the significant performance gap between current state-of-the-art agents and human capabilities in enterprise UI automation, setting a trajectory for future in  vation.

BrowserGym \cite{chezelles2024browsergym} builds ecosystem designed for web agent research. It unifies various benchmarks like  MiniWoB(++) \cite{reinforcementlearningonwebinterfaces}, WebArena \cite{webarenarealisticwebenvironment}, and WorkArena \cite{drouin2024workarena} under a single framework, addressing the issue of fragmentation in web agent evaluation. By leveraging standardized observation and action spaces, it enables consistent and reproducible experiments. BrowserGym's extensible architecture make it a vital tool for developing and testing GUI-driven agents powered by LLMs, significantly accelerating in  vation in web automation research.

In the realm of interacting with live websites, \textbf{WebOlympus}~\cite{zheng2024webolympus} introduces an open platform that enables web agents to interact with live websites through a Chrome extension-based interface. Supporting diverse tasks and integrating a safety monitor to prevent harmful actions, it promotes safer automation of web-based tasks and provides a critical tool for evaluating agent performance in realistic scenarios.

Collectively, these benchmarks have significantly contributed to advancing the evaluation of web-based GUI agents, each addressing different aspects such as realism, multimodality, safety, and enterprise applicability. Their developments reflect the evolving challenges and requirements in creating sophisticated agents capable of complex web interactions.

\begin{table*}[!h]
\centering
\caption{Overview of mobile GUI agent benchmarks (Part I).\label{tab:mobile_benchmark1}}
\resizebox{\textwidth}{!}{%
\begin{tabular}{p{2cm}|p{1.5cm}|p{1cm}|p{1cm}|p{3.5cm}|p{1.5cm}|p{2cm}|p{2cm}|p{2cm}}
\hline
\multicolumn{1}{c|}{\textbf{Benchmark}} & 
\multicolumn{1}{c|}{\textbf{Platform}}  & 
\multicolumn{1}{c|}{\textbf{Year}} & 
\multicolumn{1}{c|}{\textbf{Live}} & 
\multicolumn{1}{c|}{\textbf{Highlight}}  & 
\multicolumn{1}{c|}{\textbf{Data Size}} & 
\multicolumn{1}{c|}{\textbf{Metric}}  & 
\multicolumn{1}{c|}{\textbf{Measurement}} & 
\multicolumn{1}{c}{\textbf{Link}} \\ \hline

AndroidEnv \cite{androidenv} & Android & 2021 & Yes & Provides an open-source platform based on the Android ecosystem with over 100 tasks across approximately 30 apps, focusing on reinforcement learning for various Android interactions. & 100+ tasks & NA & NA & \url{https://github.com/google-deepmind/android\_env} \\ \hline

PIXELHELP \cite{mappingnaturallanguageinstructions} & Android & 2020 & No & Includes a corpus of natural language instructions paired with UI actions across four task categories, aiding in grounding language to UI interactions. & 187 multi-step instructions & Step Success Rate & Element Match, Action Match & \url{https://github.com/google-research/google-research/tree/master/seq2act} \\ \hline

Mobile-Env \cite{mobileenvbuildingqualifiedevaluation} & Android & 2024 & Yes & Comprehensive toolkit for Android GUI benchmarks to enable controlled evaluations of real-world app interactions. & 224 tasks & Task Success Rate, Step Success Rate & Text Match, Element Match, Image Match, State Information & \url{https://github.com/X-LANCE/Mobile-Env} \\ \hline

B-MOCA \cite{lee2024benchmarkingmobiledevicecontrol} & Android & 2024 & Yes & Benchmarks mobile device control agents on realistic tasks, incorporating UI layout and language randomization to evaluate generalization capabilities. & 131 tasks & Task Success Rate & Element Match, State Information & \url{https://b-moca.github.io/} \\ \hline

AndroidWorld \cite{rawles2024androidworlddynamicbenchmarkingenvironment} & Android & 2024 & Yes & Offers a dynamic Android environment, allowing for diverse natural language instruction testing. & 116 tasks & Task Success Rate & State Information & \url{https://github.com/google-research/android\_world} \\ \hline

Mobile-Eval \cite{wang2024mobileagentautonomousmultimodalmobile} & Android & 2024 & Yes & Benchmark based on mainstream Android apps, and designed to test common mobile interactions. & 30 instructions & Task Success Rate, Step Success Rate, Efficiency Score & Text Match, Path Length & \url{https://github.com/X-PLUG/MobileAgent} \\ \hline

DroidTask \cite{wen2024autodroid} & Android & 2024 & Yes & Android Task Automation benchmark supports exploration and task recording in real apps with corresponding GUI action traces. & 158 tasks & Step Success Rate, Task Success Rate & Element Match, Action Match & \url{https://github.com/MobileLLM/AutoDroid} \\ \hline

AITW \cite{androidinwildlargescaledataset} & Android & 2023 & No & A large-scale dataset, which is partly inspired by PIXELHELP, covering diverse Android interactions. & 715,142 episodes & Task Success Rate, Step Success Rate & Action Match & \url{https://github.com/google-research/google-research/tree/master/android\_in\_the\_wild} \\ \hline

AndroidArena \cite{xing2024understanding} & Android & 2024 & Yes & Focuses on daily cross-app and constrained tasks within the Android ecosystem, providing single-app and multi-app interaction scenarios. & 221 tasks & Task Success Rate, Step Success Rate, Efficiency Score & Action Match, Path Length & \url{https://github.com/AndroidArenaAgent/AndroidArena} \\ \hline

ANDROIDLAB \cite{xu2024androidlabtrainingsystematicbenchmarking} & Android & 2024 & Yes & Provides a structured evaluation framework with 138 tasks across nine apps, supporting both text-only and multimodal agent evaluations on Android. & 138 tasks & Task Success Rate, Step Success Rate, Efficiency Score & Element Match, Image Match & \url{https://github.com/THUDM/Android-Lab} \\ \hline

GTArena \cite{zhao2024gui} & Mobile applications & 2024 & No & Introduces a Transition Tuple for GUI defects, enabling large-scale defect dataset creation and reproducible, end-to-end automated testing. & 10,000+ GUI display and GUI interactions & Task Success Rate, Step Success Rate & Text Match, Element Match, Action Match & \url{https://github.com/ZJU-ACES-ISE/ChatUITest} \\ \hline
\end{tabular}}
\end{table*}

\begin{table*}[!h]
\centering
\caption{Overview of mobile GUI agent benchmarks (Part II).\label{tab:mobile_benchmark2}}
\resizebox{\textwidth}{!}{%
\begin{tabular}{p{2cm}|p{1.5cm}|p{1cm}|p{1cm}|p{3.5cm}|p{1.5cm}|p{2cm}|p{2cm}|p{2cm}}
\hline
\multicolumn{1}{c|}{\textbf{Benchmark}}     & \multicolumn{1}{c|}{\textbf{Platform}}  & \multicolumn{1}{c|}{\textbf{Year}} & \multicolumn{1}{c|}{\textbf{Live}} & \multicolumn{1}{c|}{\textbf{Highlight}}      & \multicolumn{1}{c|}{\textbf{Data Size}}    & \multicolumn{1}{c|}{\textbf{Metric}}  & \multicolumn{1}{c|}{\textbf{Measurement}}  & \multicolumn{1}{c}{\textbf{Link}}   \\ \hline
A3 \cite{chai2025a3androidagentarena} & Mobile Android & 2025 & Yes & Introduces a novel business-level LLM-based evaluation process, significantly reducing human labor and coding expertise requirements. & 201 tasks across 21 widely used apps. & Task Success Rate. & Element Match, Action Match. & \url{https://yuxiangchai.github.io/Android-Agent-Arena/}  \\ \hline
LlamaTouch \cite{zhang2024llamatouchfaithfulscalabletestbed} & Mobile Android & 2024 & Yes & Enables faithful and scalable evaluations for mobile UI task automation by matching task execution traces against annotated essential states & 496 tasks covering 57 unique Android applications & Task Success Rate, Step Success Rate, Efficiency Score & Text Match, Action Match, State Information Match & \url{https://github.com/LlamaTouch/LlamaTouch} \\ \hline
MobileAgentBench \cite{wang2024mobileagentbenchefficientuserfriendlybenchmark} & Mobile Android & 2024 & Yes & Provides a fully autonomous evaluation process on real Android devices and flexibility in judging success conditions across multiple paths to completion & 100 tasks across 10 open-source Android applications & Task Success Rate, Efficiency Score, Latency, Token Cost & State Information (UI State Matching) & \url{https://mobileagentbench.github.io/} \\ \hline
Mobile-Bench \cite{deng2024mobile} & Android & 2024 & Yes & Supports both UI and API-based actions in multi-app scenarios, testing agents on single and multi-task structures with a checkpoint-based evaluation approach. & 832 entries (200+ tasks) & Task Success Rate, Step Success Rate, Efficiency Score & Action Match, Path Length & \url{https://github.com/XiaoMi/MobileBench} \\ \hline
Mobile Safety Bench \cite{lee2024mobilesafetybench} & Android & 2024 & Yes & Prioritizes safety evaluation in mobile control tasks, with distinct tasks focused on helpfulness, privacy, and legal compliance. & 100 tasks & Task Success Rate, Risk Mitigation Success & Action Match with Safety Considered, Element Match, State Information & \url{https://mobilesafetybench.github.io/} \\ \hline
SPA-BENCH \cite{chen2024spa} & Android & 2024 & Yes & Extensive evaluation framework supporting single-app and cross-app tasks in English and Chinese, providing a plug-and-play structure for diverse task scenarios. & 340 tasks & Task Success Rate, Step Success Rate, Efficiency Score & Action Match, State Information, Time Spent, API Cost & \url{https://spa-bench.github.io} \\ \hline
SPHINX \cite{ran2025beyond} & Android & 2025 & Yes & Provides a fully automated benchmarking suite and introduces a multi-dimensional evaluation framework. & 284 common tasks. & Task Success Rate, Efficiency Score, Completion under Policy, Turn Success Rate. & Text Match, Image Match, Element Match, Action Match. & / \\ \hline
AEIA-MN \cite{chen2025aeia} & Mobile Android & 2025 & Yes & Introduces the Active Environment Injection Attack (AEIA) framework that actively manipulates environmental elements (\eg notifications) in mobile operating systems to mislead multimodal LLM-powered agents. & 61 tasks (AndroidWorld) + 45 tasks (AppAgent) & Task Success Rate, Risk Ratio, Efficiency Score & Text Match, State Information, Action Match & / \\ \hline
AutoEval \cite{sun2025autoeval} & Mobile Android & 2025 & Yes & Introduces a fully autonomous evaluation framework for mobile agents, eliminating the need for manual task reward signal definition and extensive evaluation code development. & 93 tasks & Task Success Rate & Action Match, State Information & / \\ \hline
LearnGUI \cite{liu2025learnactfewshotmobilegui} & Mobile Android & 2025 & Yes & The first benchmark to systematically study few-shot demonstration-based learning in mobile GUI agents, featuring both offline and online task environments & Offline: 2,252 tasks with k-shot variants across 44 apps; Online: 101 interactive tasks across 20 apps & Task Success Rate & Action Match & \url{https://lgy0404.github.io/LearnAct} \\ \hline
\end{tabular}}
\end{table*}

\subsection{Mobile Agent Benchmarks\label{sec:evaluation:mobile}}

Evaluating GUI agents on mobile platforms presents unique challenges due to the diversity of interactions and the complexity of mobile applications. Several benchmarks have been developed to address these challenges, each contributing to the advancement of mobile agent evaluation. We first provide an analysis for these mobile benchmarks in Tables~\ref{tab:mobile_benchmark1} and \ref{tab:mobile_benchmark2}.

An early effort in this domain is \textbf{PIXELHELP}~\cite{mappingnaturallanguageinstructions}, which focuses on grounding natural language instructions to actions on mobile user interfaces. Addressing the significant challenge of interpreting and executing complex, multi-step tasks, PIXELHELP provides a comprehensive dataset pairing English instructions with human-performed actions on a mobile UI emulator. It comprises 187 multi-step instructions across four task categories, offering a robust resource for evaluating models on task accuracy through metrics like Complete Match and Partial Match.

Building upon the need for systematic evaluation, \textbf{ANDROIDLAB}~\cite{xu2024androidlabtrainingsystematicbenchmarking} establishes a comprehensive framework for Android-based auto  mous agents. It introduces both an action space and operational modes that support consistent evaluations for text-only and multimodal models. By providing XML and SoM operation modes, ANDROIDLAB allows LLMs and LMMs to simulate real-world interactions in equivalent environments. The benchmark includes 138 tasks across nine apps, encompassing typical Android functionalities, and evaluates agents using metrics such as Success Rate and Reversed Redundancy.

To further challenge agents in handling both API and UI operations, \textbf{Mobile-Bench}~\cite{deng2024mobile} offers an in  vative approach by combining these elements within a realistic Android environment. Its multi-app setup and three distinct task categories test agents' capabilities in handling simple and complex mobile interactions, pushing beyond traditional single-app scenarios. The evaluation leverages CheckPoint metrics, assessing agents at each key action step, providing insights into planning and decision-making skills.

Emphasizing safety in mobile device control, \textbf{MobileSafetyBench}~\cite{lee2024mobilesafetybench} provides a structured evaluation framework that prioritizes both helpfulness and safety. It rigorously tests agents across common mobile tasks within an Android emulator, focusing on layered risk assessment, including legal compliance and privacy. A distinctive feature is its indirect prompt injection test to probe agent robustness. The evaluation ensures agents are scored on practical success while managing risks, advancing research in LLM reliability and secure auto  mous device control.

Expanding the scope to multiple languages and application scenarios, \textbf{SPA-BENCH}~\cite{chen2024spa} introduces an extensive benchmark for smartphone agents. It assesses both single-app and cross-app tasks in a plug-and-play framework that supports seamless agent integration. With a diverse task collection across Android apps, including system and third-party apps, SPA-BENCH offers a realistic testing environment measuring agent capabilities in understanding UIs and handling app navigation through metrics like success rate, efficiency, and resource usage.

Focusing on efficient and user-friendly evaluation, \textbf{MobileAgentBench}~\cite{wang2024mobileagentbenchefficientuserfriendlybenchmark} presents a benchmark tailored for agents on Android devices. It offers a fully auto  mous testing process, leveraging final UI state matching and real-time app event tracking. With 100 tasks across 10 open-source Android applications categorized by difficulty, it accommodates multiple paths to success, enhancing reliability and applicability. Comprehensive metrics, including task success rate, efficiency, latency, and token cost, provide insights into agent performance.

Complementing these efforts, \textbf{LlamaTouch}~\cite{zhang2024llamatouchfaithfulscalabletestbed} introduces a benchmark and testbed for mobile UI task automation in real-world Android environments. Emphasizing essential state an  tation, it enables precise evaluation of tasks regardless of execution path variability or dynamic UI elements. With 496 tasks spanning 57 unique applications, LlamaTouch demonstrates scalability and fidelity through advanced matching techniques, integrating pixel-level screenshots and textual screen hierarchies, reducing false negatives and supporting diverse task complexities.

Zhao \etal introduce \textbf{GTArena} \cite{zhao2024gui}, a formalized framework and benchmark designed to advance auto  mous GUI testing agents. GTArena provides a standardized evaluation environment tailored for multimodal large language models. Central to its design is the   vel Transition Tuple data structure, which systematically captures and analyzes GUI defects. The benchmark assesses three core tasks—test intention generation, task execution, and defect detection—using a diverse dataset comprising real-world, artificially injected, and synthetic defects, establishing GTArena as a pioneering benchmark for GUI testing agents.

Collectively, these benchmarks have significantly advanced the evaluation of mobile-based GUI agents, addressing challenges in task complexity, safety, efficiency, and scalability. Their contributions are instrumental in developing more capable and reliable agents for mobile platforms.

\begin{table*}[!h]
\centering
\caption{Overview of computer GUI agent benchmarks.\label{tab:computer_benchmark1}}
\resizebox{\textwidth}{!}{%
\begin{tabular}{p{2cm}|p{1.5cm}|p{1cm}|p{1cm}|p{3.5cm}|p{1.5cm}|p{2cm}|p{2cm}|p{2cm}}
\hline
\multicolumn{1}{c|}{\textbf{Benchmark}}     & \multicolumn{1}{c|}{\textbf{Platform}}  & \multicolumn{1}{c|}{\textbf{Year}} & \multicolumn{1}{c|}{\textbf{Live}} & \multicolumn{1}{c|}{\textbf{Highlight}}      & \multicolumn{1}{c|}{\textbf{Data Size}}    & \multicolumn{1}{c|}{\textbf{Metric}}  & \multicolumn{1}{c|}{\textbf{Measurement}}  & \multicolumn{1}{c}{\textbf{Link}}   \\ \hline
OSWorld \cite{xie2024osworldbenchmarkingmultimodalagents} & Linux, Windows, macOS & 2024 & Yes & Scalable, real computer environment for multimodal agents, supporting task setup, execution-based evaluation, and interactive learning across Ubuntu, Windows, and macOS. & 369 Ubuntu tasks, 43 Windows tasks & Task Success Rate & Execution-based State Information (such as internal file interpretation, permission management) & \url{https://os-world.github.io/} \\ \hline
Windows Agent Arena \cite{bonatti2024windowsagentarenaevaluating} & Windows & 2024 & Yes & Adaptation of OSWorld focusing exclusively on the Windows OS with diverse multi-step tasks, enabling agents to use a wide range of applications and tools. & 154 tasks & Task Success Rate & Same as OSWorld, scalable with cloud parallelization & \url{https://microsoft.github.io/WindowsAgentArena} \\ \hline
OmniACT \cite{kapoor2024omniactdatasetbenchmarkenabling} & macOS, Linux, Windows & 2024 & No & Assesses agents’ capability to generate executable programs for computer tasks across desktop and web applications in various OS environments, prioritizing multimodal challenges. & 9,802 data points & Task Success Rate, Step Success Rate & Action Match & \url{https://huggingface.co/datasets/Writer/omniact} \\ \hline
VideoGUI \cite{lin2024videoguibenchmarkguiautomation} & Windows & 2024 & No & Focuses on visual-centric tasks from instructional videos, emphasizing planning and action precision in applications like Adobe Photoshop and Premiere Pro. & 178 tasks, 463 subtasks & Task Success Rate & State Information, Action Match & \url{https://showlab.github.io/videogui} \\ \hline
Spider2-V \cite{cao2024spider2vfarmultimodalagents} & Linux & 2024 & Yes & Benchmarks agents across data science and engineering workflows in authentic enterprise software environments, covering tasks from data ingestion to visualization. & 494 tasks & Task Success Rate & Action Match, State Information & \url{https://spider2-v.github.io} \\ \hline
Act2Cap \cite{wu2024guiactionnarratordid} & Windows & 2024 & Yes & Emphasizes GUI action narration using cursor-based prompts in video format, covering a variety of GUI interactions like clicks, typing, and dragging. & 4,189 samples & Step Success Rate & Element Match & \url{https://showlab.github.io/GUI-Narrator} \\ \hline
OFFICEBENCH \cite{wang2024officebenchbenchmarkinglanguageagents} & Linux & 2024 & Yes & Tests cross-application automation in office workflows with complex multi-step tasks across applications like Word and Excel, assessing operational integration in realistic scenarios. & 300 tasks & Task Success Rate & Action Match, Text Match, State Information & \url{https://github.com/zlwang-cs/OfficeBench} \\ \hline
AssistGUI \cite{gao2024assistgui} & Windows & 2024 & Yes & The first benchmark focused on task-oriented desktop GUI automation & 100 tasks from 9 popular applications & Task Success Rate, Efficiency Score & Element Match, Action Match & \url{https://showlab.github.io/assistgui/} \\ \hline
WorldGUI \cite{zhao2025worldguidynamictestingcomprehensive} & Windows & 2025 & Yes & First GUI benchmark designed to evaluate dynamic GUI interactions by incorporating various initial states. & 315 total tasks from 10 Windows applications & Task Success Rate, Efficiency Score & Image Match, Element Match, Action Match & / \\ \hline
UI-Vision \cite{nayak2025ui} & Desktop (Windows, Linux) & 2025 & No & The first large-scale benchmark specifically designed for desktop GUI agents & 8,227 query–label pairs in total & Task Success Rate & Action Match, Text Match & \url{https://uivision.github.io} \\ \hline
Computer Agent Arena \cite{wang2025computer} & Windows, Ubuntu, macOS & 2025 & Yes & The first large-scale, open-ended evaluation platform for multimodal LLM-based agents in real desktop computing environments & User-proposed tasks & Task Success Rate & Human evaluators & \url{https://arena.xlang.ai/} \\ \hline
\end{tabular}}
\end{table*}

\subsection{Computer Agent Benchmarks\label{sec:evaluation:computer}}

Evaluating GUI agents on desktop computers involves diverse applications and complex workflows. Several benchmarks have been developed to assess agents' capabilities in these environments, each addressing specific challenges and advancing the field. We overview benchmarks for computer GUI agents in Table~\ref{tab:computer_benchmark1}.

An early benchmark in this domain is \textbf{Act2Cap}~\cite{wu2024guiactionnarratordid}, which focuses on capturing and narrating GUI actions in video formats using a cursor as a pivotal visual guide. Act2Cap emphasizes the detailed nuances of GUI interactions, particularly cursor-based actions like clicks and drags, essential for advancing automation capabilities in GUI-intensive tasks. It includes a substantial dataset of 4,189 samples across various Windows GUI environments, employing metrics based on element-wise Intersection over Union to evaluate semantic accuracy and temporal and spatial precision.

To provide a scalable and genuine computer environment for multimodal agents, \textbf{OSWorld}~\cite{xie2024osworldbenchmarkingmultimodalagents} introduces a pioneering framework that supports task setup, execution-based evaluation, and interactive learning across multiple operating systems, including Ubuntu, Windows, and macOS. OSWorld serves as a unified environment that mirrors the complexity and diversity of real-world computer use, accommodating arbitrary applications and open-ended computer tasks. It includes a comprehensive suite of 369 tasks on Ubuntu and 43 tasks on Windows, utilizing execution-based evaluation metrics like success rate for rigorous assessment.

Building on OSWorld, \textbf{WindowsArena}~\cite{bonatti2024windowsagentarenaevaluating} adapts the framework to create over 150 diverse tasks specifically for the Windows operating system. Focusing on multi-modal, multi-step tasks, it requires agents to demonstrate abilities in planning, screen understanding, and tool usage within a real Windows environment. Addressing the challenge of slow evaluation times, WindowsArena enables parallelized deployment in the Azure cloud, drastically reducing evaluation time and allowing for comprehensive testing across various applications and web domains.

Focusing on office automation tasks, \textbf{OFFICEBENCH}~\cite{wang2024officebenchbenchmarkinglanguageagents} introduces a groundbreaking framework for benchmarking LLM agents in realistic office workflows. Simulating intricate workflows across multiple office applications like Word, Excel, and Email within a Linux Docker environment, it evaluates agents' proficiency in cross-application automation. The benchmark challenges agents with complex tasks at varying difficulty levels, demanding adaptability to different complexities and use cases. Customized metrics assess operation accuracy and decision-making, providing critical insights into agents' capabilities in managing multi-application office scenarios.

Addressing the automation of data science and engineering workflows, \textbf{Spider2-V}~\cite{cao2024spider2vfarmultimodalagents} offers a distinctive benchmark. It features 494 real-world tasks across 20 enterprise-level applications, spanning the entire data science workflow from data warehousing to visualization. Assessing agents' abilities to handle both code generation and complex GUI interactions within authentic enterprise software environments on Ubuntu, it employs a multifaceted evaluation method that includes information-based validation, file-based comparison, and execution-based verification.

In the realm of productivity software, \textbf{AssistGUI}~\cite{gao2024assistgui} provides a pioneering framework for evaluating agents' capabilities. It introduces an Actor-Critic Embodied Agent framework capable of complex hierarchical task planning, GUI parsing, and action generation. The dataset includes diverse tasks across design, office work, and system settings, supported by project files for reproducibility. By emphasizing outcome-driven evaluation with pixel-level precision and procedural adherence, AssistGUI highlights the potential and limitations of current LLM-based agents in managing intricate desktop software workflows.

\textbf{WorldGUI \cite{zhao2025worldguidynamictestingcomprehensive}} is a benchmark designed to evaluate GUI agents under dynamic conditions on the Windows platform. Unlike previous static benchmarks, it introduces varied initial states to simulate real-world interactions across both desktop and web applications. Rather than always starting from a fixed default state, agents must adapt to changing UI layouts, user interactions, system settings, and pre-existing conditions, requiring robust adaptability to perform effectively. The benchmark comprises 315 tasks spanning 10 popular software applications and incorporates instructional videos, project files, and multiple pre-action scenarios, providing a comprehensive and realistic evaluation framework for assessing an agent's ability to handle complex task execution.

\textbf{Computer Agent Arena \cite{wang2025computer}} presents a new paradigm for benchmarking LLM-based GUI agents through live, user-configured desktop environments. Unlike traditional static datasets, it provides an interactive cloud-based infrastructure where agents are evaluated on tasks spanning web browsing, programming, and productivity using real applications like Google Docs, VSCode, and Slack. Its innovation lies in using head-to-head agent comparisons, human judgment, and Elo-based ranking to evaluate general-purpose digital agents in realistic settings. The benchmark supports Windows and Ubuntu, with MacOS support planned, and allows customized task scenarios with diverse software and website setups. By enabling crowdsourced evaluations and planning open-source releases, it fosters community-driven improvements and robust comparisons.

Collectively, these benchmarks provide comprehensive evaluation frameworks for GUI agents on desktop platforms, addressing challenges in task complexity, cross-application automation, scalability, and fidelity. Their contributions are instrumental in advancing the development of sophisticated agents capable of complex interactions in desktop environments.

\begin{table*}[!h]
\centering
\caption{Overview of cross-platform GUI agent benchmarks.\label{tab:cross_benchmark1}}
\resizebox{\textwidth}{!}{%
\begin{tabular}{p{2cm}|p{1.5cm}|p{1cm}|p{1cm}|p{3.5cm}|p{1.5cm}|p{2cm}|p{2cm}|p{2cm}}
\hline
\multicolumn{1}{c|}{\textbf{Benchmark}}     & \multicolumn{1}{c|}{\textbf{Platform}}  & \multicolumn{1}{c|}{\textbf{Year}} & \multicolumn{1}{c|}{\textbf{Live}} & \multicolumn{1}{c|}{\textbf{Highlight}}      & \multicolumn{1}{c|}{\textbf{Data Size}}    & \multicolumn{1}{c|}{\textbf{Metric}}  & \multicolumn{1}{c|}{\textbf{Measurement}}  & \multicolumn{1}{c}{\textbf{Link}}   \\ \hline
VisualAgent Bench \cite{liu2024visualagentbenchlargemultimodalmodels} & Web, Android, Game, Virtual Embodied & 2024 & Yes & First benchmark designed for visual foundation agents across GUI and multimodal tasks, focusing on vision-centric interactions in Android, web, and game environments. & 4,482 trajectories & Task Success Rate & Text Match & \url{https://github.com/THUDM/VisualAgentBench/} \\ \hline
SPR Benchmark \cite{fan2024readpointedlayoutawaregui} & Mobile, Web, Operating Systems & 2024 & Yes & Evaluates GUI screen readers' ability to describe both content and layout information & 650 screenshots annotated with 1,500 target points and regions & Task Success Rate, Efficiency Score & Text Match, Element Match & / \\ \hline
AgentStudio \cite{zheng2024agentstudiotoolkitbuildinggeneral} & Windows, Linux, macOS & 2024 & Yes & Open toolkit for creating and benchmarking general-purpose virtual agents, supporting complex interactions across diverse software applications. & NA & Step Success Rate & Action Match, State Information, Image Match & \url{https://computer-agents.github.io/agent-studio/} \\ \hline
CRAB \cite{xu2024crabcrossenvironmentagentbenchmark} & Linux, Android & 2024 & Yes & Cross-environment benchmark evaluating agents across mobile and desktop devices, using a graph-based evaluation method to handle multiple correct paths and task flexibility. & 120 tasks & Step Success Rate, Efficiency Score & Action Match & \url{https://github.com/crab-benchmark} \\ \hline
ScreenSpot \cite{cheng2024seeclickharnessingguigrounding} & iOS, Android, macOS, Windows, Web & 2024 & No & Vision-based GUI benchmark with pre-trained GUI grounding, assessing agents’ ability to interact with GUI elements across mobile, desktop, and web platforms using only screenshots. & 1,200 instructions & Step Success Rate & Action Match & \url{https://github.com/njucckevin/SeeClick} \\ \hline
\end{tabular}}
\end{table*}

\subsection{Cross-Platform Agent Benchmarks\label{sec:evaluation:cross}}

To develop GUI agents capable of operating across multiple platforms, cross-platform benchmarks are essential. These benchmarks challenge agents to adapt to different environments and interfaces, evaluating their versatility and robustness. We provide an overview of benchmarks for cross-platform GUI agents in Tables~\ref{tab:cross_benchmark1}.

Addressing this need, \textbf{VisualAgentBench} (VAB)~\cite{liu2024visualagentbenchlargemultimodalmodels} represents a pioneering benchmark for evaluating GUI and multimodal agents across a broad spectrum of realistic, interactive tasks. Encompassing platforms such as Web (WebArena-Lite \cite{webarenarealisticwebenvironment}), Android (VAB-Mobile \cite{xu2024androidlabtrainingsystematicbenchmarking}), and game environments, VAB focuses on vision-based interaction and high-level decision-making tasks. The benchmark employs a multi-level data collection strategy involving human demonstrations, program-based solvers, and model bootstrapping. Evaluation metrics concentrate on success rates, ensuring comprehensive performance assessments in tasks like navigation and content modification, thereby filling a significant gap in benchmarking standards for GUI-based LLM agents.

Complementing this, \textbf{CRAB}~\cite{xu2024crabcrossenvironmentagentbenchmark} introduces an in  vative benchmark by evaluating multimodal language model agents in cross-environment interactions. It uniquely supports seamless multi-device task execution, evaluating agents in scenarios where tasks span both Ubuntu Linux and Android environments. By introducing a graph-based evaluation method that breaks down tasks into sub-goals and accommodates multiple correct paths to completion, CRAB provides a nuanced assessment of planning, decision-making, and adaptability. Metrics such as Completion Ratio, Execution Efficiency, Cost Efficiency, and Success Rate offer comprehensive insights into agent performance.

Focusing on GUI grounding for cross-platform visual agents, \textbf{ScreenSpot}~\cite{cheng2024seeclickharnessingguigrounding} offers a comprehensive benchmark emphasizing tasks that rely on interpreting screenshots rather than structured data. ScreenSpot includes over 600 screenshots and 1,200 diverse instructions spanning mobile (iOS, Android), desktop (macOS, Windows), and web platforms. It evaluates click accuracy and localization precision by measuring how effectively agents can identify and interact with GUI elements through visual cues alone. By challenging models with a wide variety of UI elements, ScreenSpot addresses real-world complexities, making it an essential resource for evaluating visual GUI agents across varied environments.

These cross-platform benchmarks collectively advance the development of GUI agents capable of operating seamlessly across multiple platforms. By providing comprehensive evaluation frameworks, they are instrumental in assessing and enhancing the versatility and adaptability of agents in diverse environments.

\subsection{Takeaways\label{sec:evaluation:takeaways}}
  
The evolution of GUI agent benchmarks reflects a broader shift towards more realistic, interactive, and comprehensive evaluation environments. This section highlights key trends and future directions in the benchmarking of LLM-brained GUI agents.  

\begin{enumerate}
    \item \textbf{Towards More Interactive and Realistic Environments:} Recent advancements in GUI agent benchmarking emphasize the transition from synthetic scenarios to more interactive and realistic environments. This shift is evident in the use of simulators, Docker containers, and real-world applications to create "live" environments that better mimic genuine user interactions. Such environments   t only provide a more accurate assessment of agent capabilities but also pose new challenges in terms of performance and robustness.  
  
   \item \textbf{Cross-Platform Benchmarks:}  The emergence of cross-platform benchmarks that encompass mobile, web, and desktop environments represents a significant step towards evaluating the generalizability of GUI agents. However, these benchmarks introduce fundamental challenges unique to each platform. A unified interface for accessing platform-specific information, such as HTML and DOM structures, could substantially streamline the benchmarking process and reduce implementation efforts. Future work should focus on standardizing these interfaces to facilitate seamless agent evaluation across diverse environments.  
  
  \item \textbf{Increased Human Interaction and Realism:} There is a growing trend towards incorporating more human-like interactions in benchmarks, as seen in multi-turn and conversational scenarios. These setups mirror real-world use cases more closely, thereby providing a rigorous test of an agent's ability to handle dynamic, iterative interactions. As GUI agents become more sophisticated, benchmarks must continue to evolve to include these nuanced interaction patterns, ensuring agents can operate effectively in complex, human-centric environments.  
  
   \item \textbf{Scalability and Automation Challenges:} Scalability remains a significant concern in benchmarking GUI agents. The creation of realistic tasks and the development of evaluation methods for individual cases often require substantial human effort. Automation of these processes could alleviate some of the scalability issues, enabling more extensive and efficient benchmarking. Future research should explore automated task generation and evaluation techniques to enhance scalability.  
  
   \item \textbf{Emphasis on Safety, Privacy, and Compliance:}  There is a   table trend towards evaluating GUI agents on safety, privacy, and compliance metrics. These considerations are increasingly important as agents are integrated into sensitive and regulated domains. Encouraging this trend will help ensure that agents   t only perform tasks effectively but also adhere to necessary legal and ethical standards. Future benchmarks should continue to expand on these dimensions, incorporating evaluations that reflect real-world compliance and data security requirements. 
\end{enumerate}
The landscape of GUI agent benchmarking is rapidly evolving to meet the demands of increasingly complex and interactive environments. By embracing cross-platform evaluations, fostering human-like interactions, addressing scalability challenges, and prioritizing safety and compliance, the community can pave the way for the next generation of sophisticated GUI agents. Continued in  vation and collaboration will be essential in refining benchmarks to ensure they accurately capture the multifaceted capabilities of modern agents, ultimately leading to more intuitive and effective human-computer interactions.

\section{Applications of LLM-Brained GUI Agents\label{sec:applications}}
As LLM-brained GUI agents continue to mature, a growing number of applications leverage this concept to create more intelligent, user-friendly, and natural language-driven interfaces. These advancements are reflected in research papers, open-source projects, and industry solutions. Typical applications encompass \textit{(i)} \textbf{GUI testing}, which has transitioned from traditional script-based approaches to more intuitive, natural language-based interactions, and \textit{(ii)} \textbf{virtual assistants}, which automate users' daily tasks in a more adaptive and responsive manner through natural language interfaces.

\subsection{GUI Testing\label{sec:applications:testing}}

\begin{table*}[t]
\caption{Overview of GUI-testing with LLM-powered GUI agents (Part I).\label{tab:testing1}}
\resizebox{\textwidth}{!}{ % Resize the entire figure to fit \textwidth
\begin{tabular}{p{1.2cm}|p{1.3cm}|p{1.2cm}|p{1.5cm}|p{1.8cm}|p{1.8cm}|p{2.2cm}|p{3.6cm}|p{1.8cm}}
\hline
\multicolumn{1}{c|}{\textbf{Project}}                                & \multicolumn{1}{c|}{\textbf{Category}} & \multicolumn{1}{c|}{\textbf{Platform}} & \multicolumn{1}{c|}{\textbf{Model}} & \multicolumn{1}{c|}{\textbf{Perception}}                                       & \multicolumn{1}{c|}{\textbf{Action}}                                             & \multicolumn{1}{c|}{\textbf{Scenario}}                                                                                           & \multicolumn{1}{c|}{\textbf{Highlight}}                                                                                                                                        & \multicolumn{1}{c}{\textbf{Link}}                                     \\ \hline
Daniel and Anne \cite{10132233}                                      & General testing                        & General-purpose platforms              & GPT-3                               & GUI structure and state                                                        & Standard UI operations                                                           & Automates the software testing process using natural language test cases                                                         & Applies GPT-3’s language understanding capabilities to GUI-based software testing, enabling natural interaction through text-based test case descriptions.                     & \url{https://github.com/neuroevolution\%2Dai/SoftwareTestingLanguageModels} \\ \hline
Daniel and Anne \cite{zimmermann2023gui}                             & General testing                        & Web platforms                          & GPT-4                               & HTML DOM structure                                                             & Standard UI operations                                                           & Automated GUI testing to enhance branch coverage and efficiency                                                                  & Performs end-to-end GUI testing using GPT-4's natural language understanding and reasoning capabilities.                                                                       & \url{https://github.com/SoftwareTestingLLMs/WebtestingWithLLMs}          \\ \hline
GPTDroid \cite{liu2024make}                                          & General testing                        & Mobile Android                         & GPT-3.5                             & UI view hierarchy files                                                        & Standard UI operations and compound actions                                      & Automates GUI testing to improve testing coverage and detect bugs efficiently                                                    & Formulates GUI testing as a Q\& A task, utilizing LLM capabilities to provide human-like interaction.                                                                          & \url{https://github.com/franklinbill/GPTDroid}                           \\ \hline
DROID-AGENT \cite{yoon2024intent}                                    & General testing                        & Mobile Android                         & GPT-3.5, GPT-4                      & JSON representation of the GUI state                                           & Standard UI operations, higher-level APIs, and custom actions                    & Semantic, intent-driven automation of GUI testing                                                                                & Autonomously generates and executes high-level, realistic tasks for Android GUI testing based on app-specific functionalities.                                                 & \url{https://github.com/coinse/droidagent}                               \\ \hline
AUITest-Agent \cite{hu2024auitestagentautomaticrequirementsoriented} & General testing                        & Mobile Android                         & GPT-4                               & GUI screenshots, UI hierarchy files, and CV-enhanced techniques like Vision-UI & Standard UI operations                                                           & Automated functional testing of GUIs                                                                                             & Features dynamic agent organization for step-oriented testing and a multi-source data extraction strategy for precise function verification.                                   & \url{https://github.com/bz-lab/AUITestAgent}                             \\ \hline
VisionDroid \cite{liu2024visiondrivenautomatedmobilegui}             & General testing                        & Mobile Android                         & GPT-4                               & GUI screenshots with annotated bounding boxes, View hierarchy files            & Standard UI operations                                                           & Identifies non-crash bugs                                                                                                        & Integrates vision-driven prompts and GUI text alignment with vision-language models to enhance understanding of GUI contexts and app logic.                                    & \url{https://github.com/testtestA6/VisionDroid}                          \\ \hline
AXNav \cite{taeb2024axnav}                                           & Accessibility testing                  & iOS mobile devices                     & GPT-4                               & GUI screenshots, UI element detection model, and OCR                           & Gestures, capturing screenshots, and highlighting potential accessibility issues & Automates accessibility testing workflows, including testing features like VoiceOver, Dynamic Type, Bold Text, and Button Shapes & Adapts to natural language test instructions and generates annotated videos to visually and interactively review accessibility test results.                                   & /                                                                      \\ \hline
LLMigrate \cite{beyzaei2024automated} & General testing & Mobile Android & GPT-4o & DOM and screenshots & Standard UI operations & Automates the transfer of usage-based UI tests between Android apps & Leverages multimodal LLMs to perform UI test transfers without requiring source code access & / \\ \hline
\end{tabular}
}
\end{table*}

\begin{table*}[t]
\caption{Overview of GUI-testing with LLM-powered GUI agents (Part II).\label{tab:testing2}}
\resizebox{\textwidth}{!}{ % Resize the entire figure to fit \textwidth
\begin{tabular}{p{1.2cm}|p{1.3cm}|p{1.2cm}|p{1.5cm}|p{1.8cm}|p{1.8cm}|p{2.2cm}|p{3.6cm}|p{1.8cm}}
\hline
\multicolumn{1}{c|}{\textbf{Project}}                                & \multicolumn{1}{c|}{\textbf{Category}} & \multicolumn{1}{c|}{\textbf{Platform}} & \multicolumn{1}{c|}{\textbf{Model}} & \multicolumn{1}{c|}{\textbf{Perception}}                                       & \multicolumn{1}{c|}{\textbf{Action}}                                             & \multicolumn{1}{c|}{\textbf{Scenario}}                                                                                           & \multicolumn{1}{c|}{\textbf{Highlight}}                                                                                                                                        & \multicolumn{1}{c}{\textbf{Link}}                                     \\ \hline
Cui \etal \cite{cui2024large}                                        & Test input generation                  & Mobile Android                         & GPT-3.5, GPT-4                      & GUI structures and contextual information                                      & Entering text inputs                                                             & Generating and validating text inputs for Android applications                                                                   & Demonstrates the effectiveness of various LLMs in generating context-aware text inputs, improving UI test coverage, and identifying previously unreported bugs.                & /                                                                      \\ \hline
QTypist \cite{liu2023fill}                                           & Test input generation                  & Mobile Android                         & GPT-3                               & UI hierarchy files                                                             & Generates semantic text inputs                                                   & Automates mobile GUI testing by generating appropriate text inputs                                                               & Formulates text input generation as a cloze-style fill-in-the-blank language task.                                                                                             & /                                                                      \\ \hline
Crash-Translator \cite{huang2024crashtranslator}                     & Bug replay                             & Mobile Android                         & GPT-3                               & Crash-related stack trace information and GUI structure                        & Standard UI operations                                                           & Automates the reproduction of mobile application crashes                                                                         & Leverages LLMs for iterative GUI navigation and crash reproduction from stack traces, integrating a reinforcement learning-based scoring system to optimize exploration steps. & \url{https://github.com/wuchiuwong/CrashTranslator}                      \\ \hline
AdbGPT \cite{feng2024prompting}                                      & Bug replay                             & Mobile Android                         & GPT-3.5                             & GUI structure and hierarchy                                                    & Standard UI operations                                                           & Automates bug reproduction by extracting S2R (Steps to Reproduce) entities                                                       & Combines prompt engineering with few-shot learning and chain-of-thought reasoning to leverage LLMs for GUI-based tasks.                                                        & \url{https://github.com/sidongfeng/AdbGPT}                               \\ \hline
MagicWand \cite{ding2024improving}                                   & Verrification                          & Mobile Android                         & GPT-4V                              & UI screenshots and hierarchical UI control tree                                & Standard UI operations                                                           & Automates the verification of ``How-to'' instructions from a search engine                                                       & Features a three-stage process: extracting instructions, executing them in a simulated environment, and reranking search results based on execution outcomes.                  & /                                                                      \\ \hline
UXAgent \cite{lu2025uxagent} & Usability testing for web design & Web & Self-designed & Simplified HTML representations & Standard UI operations & Automated usability testing of web applications
& Enables LLM-powered automated usability testing by simulating thousands of user interactions, collecting both qualitative and quantitative data, and providing researchers with early feedback before real-user studies. & \url{https://uxagent.hailab.io} \\ \hline
Guardian \cite{ran2024guardian} & GUI Testing & Mobile Android & GPT-3.5 & GUI structure, Properties & Standard UI operations & Autonomously explores mobile applications, interacting with the UI to validate core functionalities. & Improves LLM-driven UI testing by offloading planning tasks to an external runtime system. & / \\ \hline
Test-Agent \cite{li2024test} & GUI Testing & Android, iOS, Harmony OS & Not Mentioned & GUI screenshots, UI structure information & Standard UI operations & Cross-platform mobile testing & Eliminates the need for pre-written test scripts by leveraging LLMs and multimodal perception to generate and execute test cases automatically. & / \\ \hline
VLM-Fuzz \cite{demissie2025vlm} & GUI Testing & Android (Mobile) & GPT-4o & GUI screenshots and UI structure information & Standard UI operations, system-level actions & Automated Android app testing for detection of crashes and bugs & Integrates vision-language reasoning with heuristic-based depth-first search (DFS) to systematically explore complex Android UIs, achieving significantly higher code coverage & / \\ \hline

\end{tabular}
}
\end{table*}

\begin{table*}[t]
\caption{Overview of GUI-testing with LLM-powered GUI agents (Part III).\label{tab:testing3}}
\resizebox{\textwidth}{!}{ % Resize the entire figure to fit \textwidth
\begin{tabular}{p{1.2cm}|p{1.3cm}|p{1.2cm}|p{1.5cm}|p{1.8cm}|p{1.8cm}|p{2.2cm}|p{3.6cm}|p{1.8cm}}
\hline
\multicolumn{1}{c|}{\textbf{Project}}                                & \multicolumn{1}{c|}{\textbf{Category}} & \multicolumn{1}{c|}{\textbf{Platform}} & \multicolumn{1}{c|}{\textbf{Model}} & \multicolumn{1}{c|}{\textbf{Perception}}                                       & \multicolumn{1}{c|}{\textbf{Action}}                                             & \multicolumn{1}{c|}{\textbf{Scenario}}                                                                                           & \multicolumn{1}{c|}{\textbf{Highlight}}                                                                                                                                        & \multicolumn{1}{c}{\textbf{Link}}                                     \\ \hline
BugCraft \cite{yapaugci2025bugcraft} & Bug Reproduction & Windows Computer & BugCraft based on GPT-4o & GUI screenshots & Standard UI operations & Automatically reproduces crash bugs in Minecraft by reading user-submitted bug reports, generating structured steps, and executing them to cause a crash & First end-to-end framework that automates crash bug reproduction in a complex open-world game (Minecraft) using LLM-driven agents, vision-based UI parsing, and structured action execution & \url{https://bugcraft2025.github.io/} \\ \hline

ReuseDroid \cite{li2025reusedroid} & GUI Testing & Mobile Android  & ReuseDroid based on GPT-4o & GUI screenshots and widget properties & Standard UI operations & Migrates GUI test cases between Android apps that share similar functionality but differ in operational logic & Leverages visual contexts and dynamic feedback mechanisms to significantly boost migration success rates compared to prior mapping- and LLM-based methods & / \\ \hline

SeeAct-ATA and PinATA \cite{chevrot2025autonomous} & GUI Testing & Web & SeeAct \cite{zheng2024gpt4visiongeneralistwebagent} & GUI structure and DOM & Standard UI operations & Automates manual end-to-end (E2E) web application testing & First open-source attempt to adapt LLM-powered Autonomous Web Agents into Autonomous Test Agents (ATA) for web testing & / \\ \hline

GERALLT \cite{rosenbach2025automated} & GUI Testing & Desktop (Windows/Linux) & GPT-4o & GUI screenshots and UI structure information & Standard UI operations & Finds unintuitive behavior, inconsistencies, and functional errors in GUIs without pre-defined test scripts & Pioneers LLM-driven testing on real-world desktop GUI applications (not web or mobile), combining structured GUI parsing with LLM-based control and evaluation & \url{https://github.com/DLR-SC/GERALLT} \\ \hline

ProphetAgent \cite{kongprophetagent} & GUI Testing & Android Mobile & SemanticAgent and GenerationAgent using foundation models (GPT-4o) & XML UI trees & Executable UI test scripts & Automates GUI test case generation from natural language for regression and compatibility testing in mobile apps & Innovatively combines LLM reasoning with a semantically enriched GUI graph (CUTG), significantly improving GUI test synthesis performance and efficiency over state-of-the-art tools & \url{https://github.com/prophetagent/Home}
\\ \hline

Agent for User \cite{feng2025agent} & GUI Testing & Android Mobile & GPT-4 & XML view hierarchy & Standard UI operations & Automated testing of multi-user interactive features & Introduces a multi-agent LLM framework where each agent simulates a user on a virtual device & / \\ \hline

\end{tabular}
}
\end{table*}

\begin{figure*}[t]
    \centering
    \includegraphics[width=\textwidth]{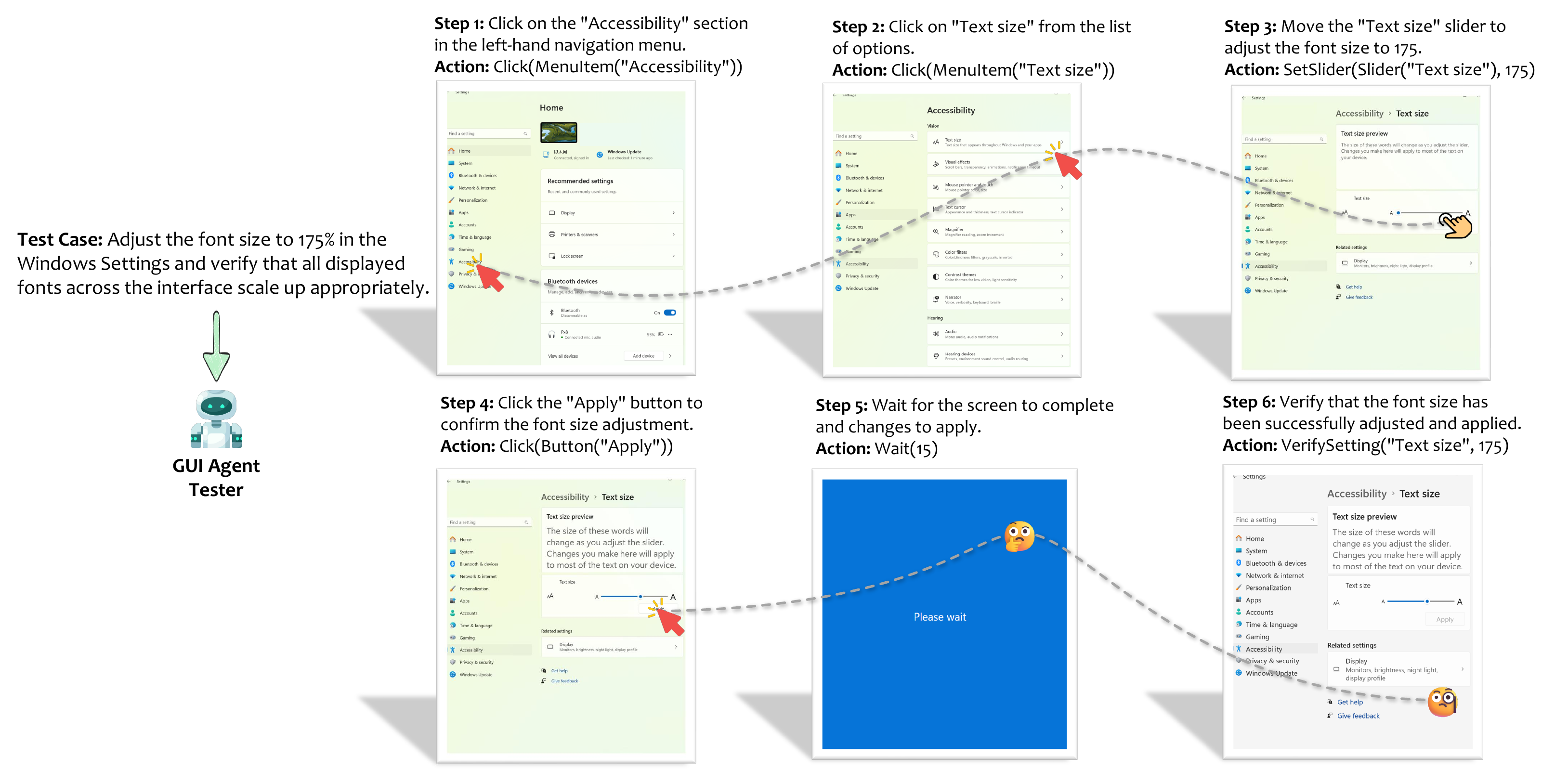}
    % \vspace{-2em}
    \caption{An example of testing font size adjustment using an LLM-powered GUI agent.}
    \label{fig:testing}
    % \vspace{-2em}
\end{figure*}

GUI testing evaluates a software application's graphical user interface to ensure compliance with specified requirements, functionality, and user experience standards. It verifies interface elements like buttons, menus, and windows, as well as their responses to user interactions. Initially conducted manually, GUI testing evolved with the advent of automation tools such as Selenium and Appium, enabling testers to automate repetitive tasks, increase coverage, and reduce testing time \cite{wang2024software, yu2023vision}. However, LLM-powered GUI agents have introduced a paradigm shift, allowing non-experts to test GUIs intuitively through natural language interfaces. 
These agents cover diverse scenarios, including general testing, input generation, and bug reproduction, without the need for traditional scripting \cite{wang2024software}. 

Figure~\ref{fig:testing} and illustrates the use of an LLM-powered GUI agent to test font size adjustment on Windows OS. With only a natural language test case description, the agent autonomously performs the testing by executing UI operations, navigating through the settings menu, and leveraging its screen understanding capabilities to verify the final outcome of font size adjustment. This approach dramatically reduces the effort required for human or script-based testing. Next, we detail the GUI testing works powered by GUI agents, and first provide an overview Tables~\ref{tab:testing1}, \ref{tab:testing2} and \ref{tab:testing3}.

\subsubsection{General Testing\label{sec:applications:testing:general}}
Early explorations demonstrated how LLMs like GPT-3 could automate GUI testing by interpreting natural language test cases and programmatically executing them. For example, one approach integrates GUI states with GPT-3 prompts, leveraging tools like Selenium and OpenCV to reduce manual scripting and enable black-box testing \cite{10132233}. Building on this, a subsequent study employed GPT-4 and Selenium WebDriver for web application testing, achieving superior branch coverage compared to traditional methods like monkey testing \cite{zimmermann2023gui}. These advances highlight how LLMs simplify GUI testing workflows while significantly enhancing coverage and efficiency.

Further pushing boundaries, \textbf{GPTDroid} reframed GUI testing as an interactive Q\&A task. By extracting structured semantic information from GUI pages and leveraging memory mechanisms for long-term exploration, it increased activity coverage by 32\%, uncovering critical bugs with remarkable precision \cite{liu2024make}. This approach underscores the potential of integrating conversational interfaces with memory for comprehensive app testing. For Android environments, \textbf{DROIDAGENT} introduced an intent-driven testing framework. It automates task generation and execution by perceiving GUI states in JSON format and using LLMs for realistic task planning. Its ability to set high-level goals and achieve superior feature coverage demonstrates how intent-based testing can transform functional verification in GUI applications \cite{yoon2024intent}.

ProphetAgent \cite{kongprophetagent} introduces a novel approach to LLM-powered GUI testing by automatically synthesizing Android application test scripts from natural language descriptions. Departing from previous methods that directly apply LLMs to GUI screenshots or app behaviors, ProphetAgent builds a Clustered UI Transition Graph (CUTG) enriched with semantic annotations. This structured representation enables more accurate mapping between natural language test steps and GUI operations, leading to significant improvements in completion rate (78.1\%) and action accuracy (83.3\%). The system employs a dual-agent architecture: SemanticAgent handles semantic annotation, while GenerationAgent generates executable scripts. ProphetAgent demonstrates strong scalability and real-world applicability—reducing tester workload by over 70\% at ByteDance. Its performance underscores the effectiveness of combining LLMs with explicit semantic knowledge graphs in GUI-based environments.

\textbf{AUITestAgent} extended the capabilities of LLM-powered GUI testing by bridging natural language-driven requirements and GUI functionality \cite{hu2024auitestagentautomaticrequirementsoriented}. Employing multi-modal analysis and dynamic agent organization, it efficiently executes both simple and complex testing instructions. This framework highlights the value of combining multi-source data extraction with robust language models to automate functional testing in commercial apps. Incorporating vision-based methods, \textbf{VisionDroid} redefined GUI testing by aligning screenshots with textual contexts to detect non-crash bugs \cite{liu2024visiondrivenautomatedmobilegui}. This innovation ensures application reliability by identifying logical inconsistencies and exploring app functionalities that conventional methods often overlook.

Accessibility testing has also benefited from LLM-powered agents. \textbf{AXNav} addresses challenges in iOS accessibility workflows, automating tests for features like VoiceOver and Dynamic Type using natural language instructions and pixel-based models. Its ability to generate annotated videos for interactive review positions AXNav as a scalable and user-friendly solution for accessibility testing \cite{taeb2024axnav}.

\subsubsection{Text Input generation\label{sec:applications:testing:text}}
In the realm of text input generation, Cui \etal demonstrated how GPT-3.5 and GPT-4 could enhance Android app testing by generating context-aware text inputs for UI fields \cite{cui2024large}. By systematically evaluating these inputs across multiple apps, they revealed the potential of LLMs in improving test coverage and detecting unique bugs with minimal manual intervention. Similarly, \textbf{QTypist} formulated text input generation as a fill-in-the-blank task, leveraging LLMs to improve activity and page coverage by up to 52\% \cite{liu2023fill}.

\subsubsection{Bug Replay\label{sec:applications:testing:bug}}
For bug reproduction, \textbf{CrashTranslator} automated the reproduction of crashes from stack traces by integrating reinforcement learning with LLMs. Its iterative navigation and crash prediction steps significantly reduced debugging time and outperformed state-of-the-art methods \cite{huang2024crashtranslator}. Meanwhile, \textbf{AdbGPT} demonstrated how few-shot learning and chain-of-thought reasoning could transform textual bug reports into actionable GUI operations. By dynamically inferring GUI actions, AdbGPT provided an efficient and lightweight solution for bug replay \cite{feng2024prompting}. 

\textbf{BugCraft} \cite{yapaugci2025bugcraft} leverages LLM-powered GUI agents to automate bug reproduction in games, specifically targeting the open-ended and complex environment of Minecraft. It employs GPT-4o as the inference engine, integrating textual bug reports, visual GUI understanding through OmniParser \cite{lu2024omniparserpurevisionbased}, and external knowledge from the Minecraft Wiki to generate and execute structured reproduction steps. Actions are carried out via a custom Macro API, enabling robust interaction with both the game's GUI and environment. BugCraft’s ability to translate unstructured bug descriptions into executable in-game behaviors highlights the strong potential of vision-enhanced LLM agents for advancing software testing and debugging.

\subsubsection{Verification\label{sec:applications:testing:verification}}
Finally, as a novel application in testing, \textbf{MagicWand} showcased the potential of LLMs in automating ``How-to'' verifications. By extracting, executing, and refining instructions from search engines, it addressed critical challenges in user-centric task automation, improving the reliability of GUI-driven workflows \cite{ding2024improving}.

In summary, LLM-powered GUI agents have revolutionized GUI testing by introducing natural language-driven methods, vision-based alignment, and automated crash reproduction. These innovations have enhanced test coverage, efficiency, and accessibility, setting new benchmarks for intelligent GUI testing frameworks.

\subsection{Virtual Assistants\label{sec:applications:assistants}}

\begin{table*}[h!]
\caption{Overview of virtual assistants with LLM-powered GUI agents (Part I).\label{tab:assistants1}}
\resizebox{\textwidth}{!}{ % Resize the entire figure to fit \textwidth
\begin{tabular}{p{1.3cm}|p{1.3cm}|p{1.2cm}|p{1.2cm}|p{1.8cm}|p{1.8cm}|p{2.2cm}|p{3.6cm}|p{1.8cm}}
\hline
\multicolumn{1}{c|}{\textbf{Project}}       & \multicolumn{1}{c|}{\textbf{Type}} & \multicolumn{1}{c|}{\textbf{Platform}} & \multicolumn{1}{c|}{\textbf{Model}} & \multicolumn{1}{c|}{\textbf{Perception}}                         & \multicolumn{1}{c|}{\textbf{Action}}                         & \multicolumn{1}{c|}{\textbf{Scenario}}                                                                       & \multicolumn{1}{c|}{\textbf{Highlight}}                                                                                                                          & \multicolumn{1}{c}{\textbf{Link}}                                                                \\ \hline
ProAgent \cite{ye2023proagent}              & Research                               & Web and Desktop                        & GPT-4                               & Task descriptions and structured application data                & Standard UI operations and dynamic branching                 & Automates business processes such as data analysis, report generation, and notifications via GUI-based tools & Introduces dynamic workflows where agents interpret and execute tasks flexibly, surpassing traditional RPA systems                                               & \url{https://github.com/OpenBMB/ProAgent}                                                          \\ \hline
LLMPA \cite{guan2024intelligent}            & Research                               & Mobile (Android)                       & AntLLM-10b                          & UI tree structures, visual modeling, and text extraction modules & Standard UI operations                                       & Automates user interactions within mobile apps, such as ticket booking                                       & Integrates LLM reasoning capabilities with a modular design that supports task decomposition, object detection, and robust action prediction in GUI environments & /                                                                                                \\ \hline
VizAbility \cite{gorniak2024vizability}     & Research                               & Desktop                                & GPT-4V                              & Keyboard-navigable tree views                                    & Navigates chart structures and generates answers             & Assists blind and low-vision users in exploring and understanding data visualizations                        & Integrates structured chart navigation with LLM-powered conversational capabilities, enabling visually impaired users to query in natural language               & \url{https://dwr.bc.edu/vizability/}                                                               \\ \hline
GPTVoice-Tasker \cite{vu2024gptvoicetasker}  & Research                               & Mobile (Android)                       & GPT-4                               & Android Accessibility Tree                                       & Standard UI operations                                       & Automates user interactions on mobile devices through voice commands                                         & Integrates LLMs for natural command interpretation and real-time GUI interactions, using a graph-based local database to record and replicate interactions       & \url{https://github.com/vuminhduc796/GPTVoiceTasker}                                               \\ \hline
AutoTask \cite{pan2023autotask}             & Research                               & Mobile (Android)                       & GPT-4                               & Android Accessibility Tree                                       & Standard UI operations                                       & Automates multi-step tasks on mobile devices                                                                 & Operates without predefined scripts or configurations, autonomously exploring GUI environments                                                                   & \url{https://github.com/BowenBryanWang/AutoTask}                                                   \\ \hline
AssistEditor \cite{gao2024assisteditor}     & Research                               & Windows                                & UniVTG \cite{lin2023univtg}         & GUI elements, user requirements, and video data                  & Standard UI operations                                       & Automates video editing workflows                                                                            & Employs a multi-agent collaboration framework where agents specialize in roles to integrate user requirements into video editing workflows                       & /                                                                                                \\ \hline
PromptRPA \cite{huang2024promptrpa}         & Research                               & Mobile (Android)                       & GPT-4 and GPT-3.5 Turbo             & Layout hierarchy and screenshots with OCR                        & Standard UI operations and application-level functionalities & Automates smartphone tasks and creates interactive tutorials                                                 & Integrates user feedback loops for continuous improvement, addressing interface evolution and task variability                                                   & /                                                                                                \\ \hline
EasyAsk \cite{gao2024easyask}               & Research                               & Mobile (Android)                       & GPT-4                               & Android Accessibility Tree                                       & Highlights specific UI elements for user interaction         & Assists older adults in learning and navigating smartphone functions through in-app interactive tutorials    & Combines voice and touch inputs, supplementing incomplete or ambiguous queries with in-app contextual information                                                & /                                                                                                \\ \hline
WebNav \cite{srinivasan2025webnav} & Research & Web  & Gemini 2.0 Flash Thinking & Standard UI operations
 & GUI screenshots and DOM & Assistive technology for visually impaired users, enabling voice-based navigation of complex websites & Combines a ReAct-style reasoning loop, real-time DOM labeling, and voice-driven interaction to support intelligent web navigation for visually impaired users & /  \\ \hline

\end{tabular}
}
\end{table*}

\begin{table*}[h!]
\caption{Overview of virtual assistants with LLM-powered GUI agents (Part II).\label{tab:assistants2}}
\resizebox{\textwidth}{!}{ % Resize the entire figure to fit \textwidth
\begin{tabular}{p{1.3cm}|p{1.3cm}|p{1.2cm}|p{1.2cm}|p{1.8cm}|p{1.8cm}|p{2.2cm}|p{3.6cm}|p{1.8cm}}
\hline
\multicolumn{1}{c|}{\textbf{Project}}       & \multicolumn{1}{c|}{\textbf{Type}} & \multicolumn{1}{c|}{\textbf{Platform}} & \multicolumn{1}{c|}{\textbf{Model}} & \multicolumn{1}{c|}{\textbf{Perception}}                         & \multicolumn{1}{c|}{\textbf{Action}}                         & \multicolumn{1}{c|}{\textbf{Scenario}}                                                                       & \multicolumn{1}{c|}{\textbf{Highlight}}                                                                                                                          & \multicolumn{1}{c}{\textbf{Link}}                                                                \\ \hline
OpenAdapt \cite{openadapt2024}              & Open-source                            & Desktop                                & LLM, VLM (e.g., GPT-4, ACT-1)       & Screenshots with CV tools for GUI parsing                        & Standard UI operations                                       & Automates repetitive tasks across industries                                                                 & Learns task automation by observing user interactions, eliminating manual scripting                                                                              & \url{https://github.com/OpenAdaptAI/OpenAdapt}                                                     \\ \hline
AgentSea \cite{agentsea2024}                & Open-source                            & Desktop and Web                        & LLM, VLM                            & Screenshots with CV tools for GUI parsing                        & Standard UI operations                                       & Automates tasks within GUI environments                                                                      & Offers a modular toolkit adhering to the UNIX philosophy, allowing developers to create custom AI agents for diverse GUI environments                            & \url{https://www.agentsea.ai/}                                                                     \\ \hline
Open Interpreter \cite{openinterpreter2024} & Open-source                            & Desktop, Web, Mobile (Android)         & LLM                                 & System perception via command-line                               & Shell commands, code, and native APIs                        & Automates tasks, conducts data analysis, manages files, and controls web browsers for research               & Executes code locally, providing full access to system resources and libraries, overcoming limitations of cloud-based services                                   & \url{https://github.com/OpenInterpreter/open-interpreter}                                          \\ \hline
MultiOn \cite{multion2024}                  & Production                             & Web                                    & LLM                                 & /                                                                & Standard UI operations                                       & Automates web-based tasks                                                                                    & Performs autonomous web actions via natural language commands                                                                                                    & \url{https://www.multion.ai/}                                                                      \\ \hline
YOYO Agent in MagicOS \cite{magicos}        & Production                             & Mobile (MagicOS 9.0)                   & MagicLM                             & GUI context                                                      & Executes in-app and cross-app operations                     & Automates daily tasks, enhancing productivity                                                                & Leverages MagicLM to understand and execute complex tasks across applications, learning user habits to provide personalized assistance                           & /                                                                                                \\ \hline
Power Automate \cite{powerautomate}         & Production                             & Windows                                & LLM, VLM                            & Records user interactions with the GUI                           & Standard UI operations                                       & Automates repetitive tasks and streamlines workflows                                                         & Translates natural language descriptions of desired automations into executable workflows                                                                        & \url{https://learn.microsoft.com/en-us/power-automate/desktop-flows/create\%2Dflow-using\%2Dai-recorder} \\ \hline
Eko \cite{eko_fellou_ai} & Production & Web browsers and computer environments & ChatGPT and Claude 3.5 & Visual-Interactive Element Perception (VIEP) technology for interacting with GUI elements. & Standard UI operations. & Automates tasks by handling diverse workflows. & Decomposes natural language task descriptions into executable workflows, enabling seamless integration of natural language and programming logic in agent design. & \url{https://eko.fellou.ai/} \\ \hline
\end{tabular}
}
\end{table*}

\begin{figure*}[t]
    \centering
    \includegraphics[width=\textwidth]{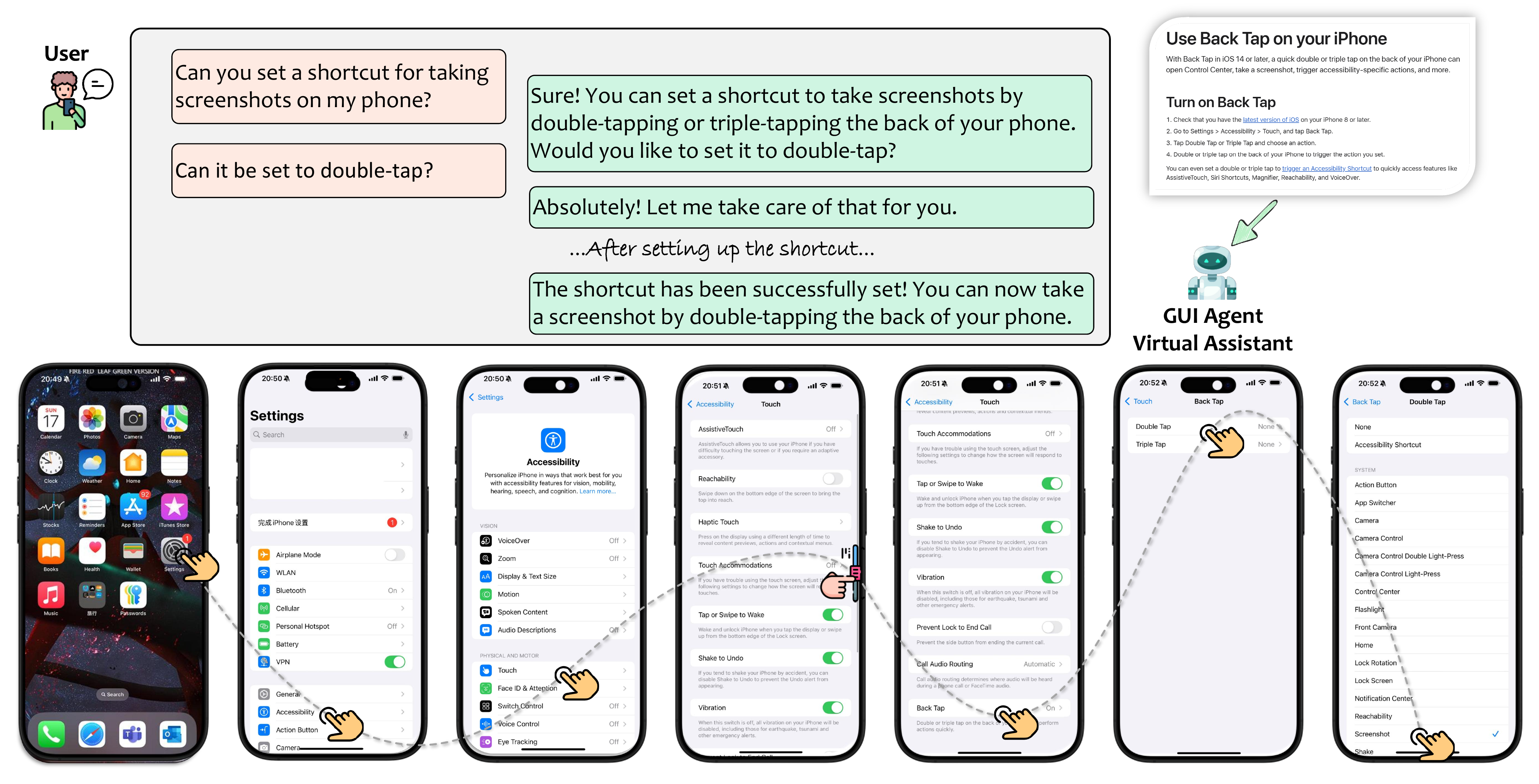}
    % \vspace{-2em}
    \caption{A conceptual example of a GUI agent-powered virtual assistant on a smartphone.}
    \label{fig:assistant}
    % \vspace{-2em}
\end{figure*}
Virtual assistants, such as Siri\footnote{\url{https://www.apple.com/siri/}}, are AI-driven applications that help users by performing tasks, answering questions, and executing commands across various platforms, including web browsers, mobile phones, and computers. Initially, these assistants were limited to handling simple commands via voice or text input, delivering rule-based responses or running fixed workflows similar to RPA. They focused on basic tasks, such as setting alarms or checking the weather.

With advancements in LLMs and agents, virtual assistants have evolved significantly. They now support more complex, context-aware interactions on device GUIs through textual or voice commands and provide personalized responses, catering to diverse applications and user needs on various platforms. This progression has transformed virtual assistants from basic utilities into intelligent, adaptive tools capable of managing intricate workflows and enhancing user productivity across platforms. Figure~\ref{fig:assistant} presents a conceptual example of a GUI agent-powered virtual assistant on a smartphone\footnote{The application and scenario depicted in the figure are conceptual and fabricated. They do not reflect the actual functionality of any specific smartphone. Readers should consult the phone manual or official guidance for accurate information on AI assistant capabilities.}. In this scenario, the agent enables users to interact through chat, handling tasks such as setting up a screenshot shortcut on their behalf. This feature is particularly beneficial for users unfamiliar with the phone's functionalities, simplifying complex tasks into conversational commands.

To explore more real-world applications of virtual assistants powered by GUI agents, we provide an overview of advancements across research, open-source initiatives, and production-level applications, as summarized in Table~\ref{tab:assistants1} and \ref{tab:assistants2}.

\subsubsection{Research\label{sec:applications:assistants:research}}

Recent research efforts have significantly advanced the capabilities of virtual assistants by integrating LLM-powered GUI agents, enabling more intelligent and adaptable interactions within various applications.

Firstly, the integration of LLMs into GUI-based automation has been explored to enhance business process automation. For instance, \cite{ye2023proagent} introduces Agentic Process Automation through the development of \textbf{ProAgent}, which automates both the creation and execution of workflows in GUI environments. By utilizing agents like ControlAgent and DataAgent, it supports complex actions such as dynamic branching and report generation in applications like Slack and Google Sheets. This approach transcends traditional RPA by enabling flexible, intelligent workflows, significantly reducing the need for manual intervention and highlighting the transformative potential of LLM-powered agents in virtual assistants.

Building upon the idea of integrating LLMs with GUI environments, researchers have focused on mobile platforms to automate complex tasks. \textbf{LLMPA}~\cite{guan2024intelligent} is a pioneering framework that leverages LLMs to automate multi-step tasks within mobile applications like Alipay. It interacts directly with app GUIs, mimicking human actions such as clicks and typing, and employs UI tree parsing and object detection for precise environment understanding. A unique controllable calibration module ensures logical action execution, demonstrating the potential of LLM-powered virtual assistants to handle intricate workflows and real-world impact in assisting users with diverse tasks.

Similarly, the automation of smartphone tasks through natural language prompts has been addressed by \textbf{PromptRPA}~\cite{huang2024promptrpa}. Utilizing a multi-agent framework, it automates tasks within smartphone GUI environments, tackling challenges like interface updates and user input variability. Advanced perception methods, including OCR and hierarchical GUI analysis, are employed to understand and interact with mobile interfaces. By supporting real-time feedback and iterative improvements, PromptRPA underscores the importance of user-centered design in LLM-driven virtual assistants.

In the realm of accessibility, LLM-powered GUI agents have been instrumental in enhancing user experience for individuals with disabilities. For example, \textbf{VizAbility}~\cite{gorniak2024vizability} enhances the accessibility of data visualizations for blind and low-vision users. By combining structured chart navigation with LLM-based conversational interactions, users can ask natural language queries and receive insights on chart content and trends. Leveraging frameworks like Olli\footnote{\url{https://mitvis.github.io/olli/}} and chart specifications such as Vega-Lite\footnote{\url{https://vega.github.io/}}, VizAbility allows exploration of visual data without direct visual perception, addressing real-world accessibility challenges in GUIs.

Furthermore, addressing the needs of older adults, \textbf{EasyAsk}~\cite{gao2024easyask} serves as a context-aware in-app assistant that enhances usability for non-technical users. By integrating multi-modal inputs, combining natural voice queries and touch interactions with GUI elements, it generates accurate and contextual tutorial searches. EasyAsk demonstrates how GUI agents can enhance accessibility by integrating contextual information and interactive tutorials, empowering users to navigate smartphone functions effectively.

Voice interaction has also been a focus area, with tools like \textbf{GPTVoiceTasker}~\cite{vu2024gptvoicetasker} facilitating hands-free interaction with Android GUIs through natural language commands. It bridges the gap between voice commands and GUI-based actions using real-time semantic extraction and a hierarchical representation of UI elements. By automating multi-step tasks and learning from user behavior, it enhances task efficiency and reduces cognitive load, highlighting the transformative potential of LLMs in improving accessibility and user experience in mobile environments.

Expanding on voice-powered interactions, \textbf{AutoTask}~\cite{pan2023autotask} enables virtual assistants to execute multi-step tasks in GUI environments without predefined scripts. It autonomously explores and learns from mobile GUIs, effectively combining voice command interfaces with dynamic action engines to interact with GUI elements. Utilizing trial-and-error and experience-driven learning, AutoTask adapts to unknown tasks and environments, showcasing its potential in enhancing voice-driven virtual assistants for hands-free interactions.

Finally, in the domain of creative workflows, \textbf{AssistEditor}~\cite{gao2024assisteditor} exemplifies a multi-agent framework for automating video editing tasks. By interacting with GUI environments, it autonomously performs complex workflows using dialogue systems and video understanding models to bridge user intent with professional editing tasks. The innovative use of specialized agents ensures efficient task distribution and execution, demonstrating the practical application of LLM-powered GUI agents in real-world scenarios and expanding automation into creative domains.

These research endeavors collectively showcase significant advancements in LLM-powered GUI agents, highlighting their potential to transform virtual assistants into intelligent, adaptable tools capable of handling complex tasks across various platforms and user needs.

\subsubsection{Open-Source Projects\label{sec:applications:assistants:projects}}

In addition to research prototypes, open-source projects have contributed substantially to the development and accessibility of LLM-brained GUI agents, enabling wider adoption and customization.

One such project is \textbf{OpenAdapt}~\cite{openadapt2024}, an open-source framework that utilizes large multimodal models to automate tasks by observing and replicating user interactions within GUI environments. It captures screenshots and records user inputs, employing computer vision techniques to understand and execute standard UI operations. Designed to streamline workflows across various industries, OpenAdapt learns from user demonstrations, thereby reducing the need for manual scripting and showcasing adaptability in GUI-based task automation.

Similarly, \textbf{AgentSea}~\cite{agentsea2024} offers a comprehensive and modular toolkit for creating intelligent agents that can navigate and interact with various GUI environments across multiple platforms. Its flexibility is particularly beneficial for developing virtual assistants capable of automating complex tasks within applications, enhancing user productivity. By adhering to the UNIX philosophy, AgentSea ensures that each tool is specialized, promoting ease of use and extensibility. Its open-source nature fosters community collaboration and innovation in AI-driven GUI automation.

\textbf{Open Interpreter}~\cite{openinterpreter2024} further exemplifies the potential of open-source contributions by leveraging large language models to execute code locally. Users can interact with their computer's GUI through natural language commands, supporting multiple programming languages and operating across various platforms. By facilitating tasks such as data analysis, web automation, and system management, Open Interpreter provides unrestricted access to system resources and libraries, enhancing flexibility and control. Its customization capabilities make it a valuable asset for users aiming to streamline operations through AI-powered virtual assistance.

These open-source projects not only advance the state of LLM-powered GUI agents but also democratize access to intelligent virtual assistants, enabling developers and users to tailor solutions to specific needs and applications.

\subsubsection{Production\label{sec:applications:assistants:production}}

The integration of LLM-brained GUI agents into production environments demonstrates their practical viability and impact on enhancing user experiences in commercial applications.

\textbf{Power Automate}~\cite{powerautomate} exemplifies an AI-powered GUI agent that enhances user interaction with desktop applications. By allowing users to describe tasks in natural language while recording actions, it translates these descriptions into automated workflows, effectively bridging the gap between user intent and execution. Its ability to record and replicate user actions within the GUI streamlines the automation of repetitive tasks, making it a valuable tool for increasing efficiency and highlighting advancements in user-friendly automation solutions.

In the realm of web interactions, \textbf{MultiOn}~\cite{multion2024} serves as a personal AI agent that autonomously interacts with web-based GUIs to execute user-defined tasks. Leveraging large language models, it interprets natural language commands and translates them into precise web actions, effectively automating complex or repetitive tasks. MultiOn's approach to perceiving and manipulating web elements enables seamless functioning across various web platforms, enhancing user productivity and streamlining web interactions.

On mobile platforms, the \textbf{YOYO Agent} in \emph{MagicOS}~\cite{magicos} exemplifies an LLM-powered GUI agent operating within the MagicOS 9.0 interface. Utilizing Honor's MagicLM, it comprehends and executes user commands across various applications, learning from user behavior to offer personalized assistance. This integration demonstrates how large language models can enhance virtual assistants, enabling them to perform complex tasks within GUI environments and improving user experience and productivity on mobile devices.

Eko \cite{eko_fellou_ai} serves as a prime example of a versatile and efficient tool for developing intelligent agents capable of interacting with GUIs across various platforms. Its integration with multiple LLMs and the innovative Visual-Interactive Element Perception (VIEP) technology highlight its capability to perform complex tasks through natural language instructions. Eko's  comprehensive tool support make it a valuable resource for developers aiming to create customizable and production-ready agent-based workflows. By facilitating seamless interaction within GUI environments, Eko exemplifies the advancements in virtual assistants powered by LLMs.

These production-level implementations highlight the practical applications and benefits of LLM-brained GUI agents in enhancing automation, productivity, and user engagement across different platforms and industries.

\subsection{Takeaways\label{sec:applications:takeaways}}

The application of LLM-brained GUI agents has ushered in new capabilities and interfaces for tasks such as GUI testing and virtual assistance, introducing natural language interactions, enhanced automation, and improved accessibility across platforms. These agents are transforming the way users interact with software applications by simplifying complex tasks and making technology more accessible. However, despite these advancements, LLM-brained GUI agents are still in their infancy, and several challenges need to be addressed for them to reach maturity. Key insights from recent developments include:
\begin{enumerate}
    \item \textbf{Natural Language-Driven Interactions:} LLM-powered GUI agents have enabled users to interact with applications using natural language, significantly lowering the barrier to entry for non-expert users. In GUI testing, tools like GPTDroid~\cite{liu2024make} and AUITestAgent~\cite{hu2024auitestagentautomaticrequirementsoriented} allow testers to specify test cases and requirements in plain language, automating the execution and verification processes. Similarly, virtual assistants like {LLMPA}~\cite{guan2024intelligent} and {ProAgent}~\cite{ye2023proagent} interpret user commands to perform complex tasks, showcasing the potential of natural language interfaces in simplifying user interactions across platforms.

    \item \textbf{Enhanced Automation of Complex Tasks:} These agents have demonstrated the ability to automate multi-step and intricate workflows without the need for manual scripting. Projects like {AutoTask}~\cite{pan2023autotask} and {GPTVoiceTasker}~\cite{vu2024gptvoicetasker} autonomously explore and interact with GUI environments, executing tasks based on high-level goals or voice commands. In GUI testing, agents have improved coverage and efficiency by automating the generation of test inputs and reproducing bugs from textual descriptions, as seen in {CrashTranslator}~\cite{huang2024crashtranslator} and {AdbGPT}~\cite{feng2024prompting}.

    \item \textbf{Multimodal Perception and Interaction:} Integrating visual and textual inputs has enhanced the agents' understanding of GUI contexts, leading to better decision-making and interaction accuracy. Agents like {VizAbility}~\cite{gorniak2024vizability} and {OpenAdapt}~\cite{openadapt2024} utilize screenshots, UI trees, and OCR to perceive the environment more comprehensively. This multimodal approach is crucial for applications that require precise identification and manipulation of GUI elements, especially in dynamic or visually complex interfaces.

    \item \textbf{Improved Accessibility and User Experience:} LLM-brained GUI agents have contributed to making technology more accessible to users with disabilities or limited technical proficiency. Tools like {VizAbility}~\cite{gorniak2024vizability} aid blind and low-vision users in understanding data visualizations, while {EasyAsk}~\cite{gao2024easyask} assists older adults in navigating smartphone functions. By tailoring interactions to the needs of diverse user groups, these agents enhance inclusivity and user experience.
\end{enumerate}
LLM-brained GUI agents are transforming the landscape of GUI interaction and automation by introducing natural language understanding, enhanced automation capabilities, and improved accessibility. While they are still in the early stages of development, the ongoing advancements and emerging applications hold great promise for the future. Continued research and innovation are essential to overcome current challenges and fully realize the potential of these intelligent agents across diverse domains and platforms.
\section{Limitations, Challenges and Future Roadmap\label{sec:limitation}}

Despite significant advancements in the development of LLM-brained GUI agents, it is important to acknowledge that this field is still in its infancy. Several technical challenges and limitations hinder their widespread adoption in real-world applications. Addressing these issues is crucial to enhance the agents' effectiveness, safety, and user acceptance. In this section, we outline key limitations and propose future research directions to overcome these challenges, providing concrete examples to illustrate each point.

\subsection{Privacy Concerns\label{sec:limitation:privacy}}

Privacy is a critical concern uniquely intensified in the context of LLM-powered GUI agents. These agents often require access to sensitive user data—such as screenshots, interaction histories, personal credentials, and confidential documents—to effectively perceive and interact with the GUI environment. In many cases, this data must be transmitted to remote servers for model inference, especially when relying on cloud-based LLMs \cite{liao2024eia, he2024emerged, gan2024navigatingriskssurveysecurity}. Such deployments raise significant privacy risks, including data breaches, unauthorized access, and misuse of personal information. These concerns are further amplified when sensitive inputs are routed through third-party APIs or processed off-device, creating compliance and security vulnerabilities that can deter real-world adoption. 

For instance, a GUI agent tasked with managing a user's email inbox may need to read, classify, and respond to messages containing highly personal or confidential content. Offloading this processing to the cloud introduces risks of exposure, prompting hesitation among users and organizations due to potential privacy violations \cite{zharmagambetov2025agentdam, yang2024security, zhang2024privacyasst}. Compared to traditional LLM applications, GUI agents operate at a finer granularity of user activity and often require broader system access, making privacy-preserving deployment strategies a critical and domain-specific challenge.

\textbf{Potential Solutions:} To mitigate privacy concerns, future research should focus on enabling \emph{on-device inference}, where the language model operates directly on the user's device without uploading personal data \cite{xu2024device, qu2024mobile}. Achieving this requires advancements in model compression techniques \cite{lin2024awq}, on-device optimization \cite{liu2024mobilellm}, and efficient inference algorithms \cite{zhou2024survey} to accommodate the computational limitations of user devices. In addition, frameworks must incorporate data redaction, secure communication channels, and explicit scoping of data usage within the agent's context. Furthermore, integration with system-level privacy controls and user consent mechanisms (\eg runtime permission dialogs or sandboxed execution) is essential for deployment in regulated domains.

From the technical perspective,  implementing privacy-preserving techniques like federated learning \cite{kuang2024federatedscope}, differential privacy \cite{mai2023split}, and homomorphic encryption \cite{de2024privacy} can enhance data security while allowing the model to learn from user data. Furthermore, developers of GUI agents should collaborate with privacy policymakers to ensure that user data and privacy are appropriately protected \cite{wolff2024lessons}. They should make the data handling processes transparent to users, clearly informing them about what data are being transmitted and how they are used, and obtain explicit user consent \cite{zhang2024s}.

\subsection{Latency, Performance, and Resource Constraints\label{sec:limitation:latency}}
One challenge that is particularly salient for GUI agents—distinct from general LLM applications is the issue of latency in interactive, multi-step execution environments. Since GUI agents rely on large language models to plan and issue actions, their computational demands can lead to high latency and slow response times, which directly impact user experience \cite{li2024llm}. This is especially critical in time-sensitive or interactive scenarios, where delays in action execution can cause user frustration or even trigger unintended system behavior. Unlike single-shot LLM tasks, GUI agents typically operate over extended sequences of steps, making latency cumulative and more disruptive over time. The problem is further amplified in on-device deployments, where computational resources are limited. For example, running an LLM-powered agent within a mobile app may result in sluggish performance or rapid battery depletion, significantly undermining usability on resource-constrained platforms \cite{xu2024survey, chen2024llm, krupp2025towards}. These concerns are uniquely pronounced in GUI agents due to their need for real-time perception, decision-making, and UI control in dynamic environments \cite{chen2024llm}.

\textbf{Potential Solutions:} Future work should aim to reduce inference latency by optimizing model architectures for speed and efficiency \cite{wan2023efficient}. Techniques such as model distillation can create smaller, faster models without substantially compromising performance \cite{xu2024survey2}. Leveraging hardware accelerators like GPUs, TPUs, or specialized AI chips, and exploring parallel processing methods can enhance computational efficiency \cite{kachris2024survey}. Implementing incremental inference and caching mechanisms may also improve responsiveness by reusing computations where applicable \cite{lee2024infinigen}. Additionally, research into model optimization and compression techniques, such as pruning \cite{wang2019structured} and quantization \cite{lin2024awq} can produce lightweight models suitable for deployment on resource-constrained devices. Exploring edge computing \cite{qu2024mobile} and distributed inference \cite{wu2023fast} can help distribute the computational load effectively.

Moreover, GUI agents should collaborate with application developers to encourage them to expose high-level native APIs for different functionalities \cite{song2024beyond, lu2024turn}, which combine several UI operations into single API calls. By integrating these APIs into the GUI agent, tasks can be completed with fewer steps, making the process much faster and reducing cumulative latency.

\subsection{Safety and Reliability\label{sec:limitation:safety}}
The real-world actuation capabilities of GUI agents introduce unique and significant safety and reliability risks beyond those faced by general-purpose LLMs. Because GUI agents can directly manipulate user interfaces—clicking buttons, deleting files, submitting forms, or initiating system-level operations—errors in action generation can have irreversible consequences \cite{anwar2024foundational, gan2024navigatingriskssurveysecurity}. These may include data corruption, accidental message dispatches, application crashes, or unauthorized access to sensitive system components \cite{zhong2023study, yuan2024r}. Such risks are compounded by the inherent uncertainty and non-determinism in LLM outputs: agents may hallucinate actions, misinterpret UI contexts, or behave inconsistently across sessions \cite{zhang2024respond, zhao2025robustness, zhang2023sirenssongaiocean, chiang2025web, chen2025obvious}. For example, an agent automating financial transactions could mistakenly execute the wrong transfer, leading to material losses. Furthermore, GUI agents expose a broader attack surface than traditional LLM applications—they are susceptible to black-box adversarial attacks that could manipulate their inputs or exploit their decision policies \cite{xu2024advweb}.

Unlike passive language models, GUI agents operate within dynamic software ecosystems where incorrect actions can propagate across applications or escalate into system-wide disruptions. Integration challenges also arise, including compatibility with evolving UI frameworks, user permission boundaries, and software-specific safety constraints, and malicious attacks \cite{yang2025systematiccategorizationconstructionevaluation, aichberger2025attacking}. These concerns, coupled with the lack of interpretability and formal guarantees, contribute to skepticism and reluctance from users and developers alike. Addressing safety and reliability in GUI agents thus requires not only robust model behavior but also runtime safeguards \cite{lee2025safeguarding}, rollback mechanisms, and interface-aware verification techniques tailored specifically to this interaction paradigm.

\textbf{Potential Solutions:} Ensuring safety and reliability necessitates robust error detection and handling mechanisms \cite{Pan2023AutomaticallyCL}. Future research should focus on integrating validation steps that verify the correctness of inferred actions before execution \cite{Huang2023ASO}. Developing formal verification methods \cite{Jha2023DehallucinatingLL}, implementing exception handling routines \cite{Zhang2023ACR}, and establishing rollback procedures \cite{Koo1986CheckpointingAR} are essential for preventing and mitigating the impact of errors. Additionally, incorporating permission management \cite{Luo2017AndroidMS, hao2013effectiveness, felt2011android, lutaaya2018rethinking} to limit the agent's access rights can prevent unauthorized or harmful operations.

Furthermore, creating standardized interaction protocols can facilitate smoother and safer integration with various applications and systems \cite{xiang2024guardagent}. Ensuring that agents comply with security best practices, such as secure authentication and authorization protocols \cite{Berkovits1998AuthenticationFM}, is essential.

\subsection{Human-Agent Interaction\label{sec:limitation:hai}}
\begin{figure*}[t]
    \centering
    \includegraphics[width=\textwidth]{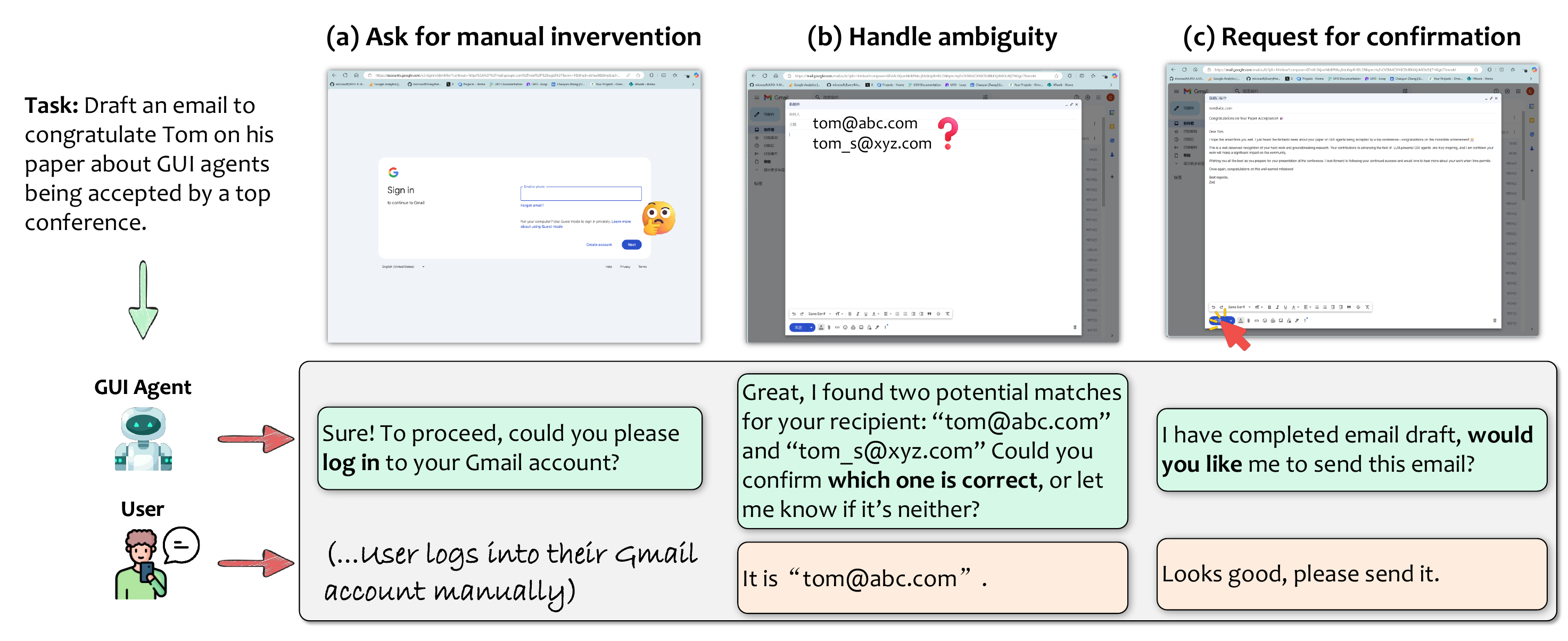}
    % \vspace{-2em}
    \caption{An illustrative example of human-agent interaction for completing an email sending request.}
    \label{fig:hai}
    % \vspace{-2em}
\end{figure*}

Human-agent interaction introduces distinct challenges in the context of GUI agents, where the agent and user operate within the same dynamic interface. Any user intervention—such as moving the mouse, altering window states, or modifying inputs—can inadvertently interfere with the agent's ongoing execution, potentially causing conflicts, unintended actions, or breakdowns in task flow \cite{gao2024taxonomy, bradshaw2017human}. Designing robust collaboration protocols that govern when the agent should yield control, pause execution, or defer to the user is a non-trivial problem specific to GUI-based automation.

Further complicating this interaction is the ambiguity of user instructions. Natural language commands may be vague, under-specified, or context-dependent, leading to misinterpretations or incomplete task plans. GUI agents may also encounter runtime uncertainties—such as unexpected popups, missing inputs, or conflicting UI states—that require them to seek user clarification or feedback \cite{zhang2024ufouifocusedagentwindows, feng2024large}. Determining when and how an agent should request user input—whether for disambiguation, permission, or verification—is critical for ensuring both reliability and user trust \cite{amayuelas2023knowledge, shi2025towards, chen2025toward}.

This challenge is exemplified in the fabricated scenario shown in Figure~\ref{fig:hai}, where a GUI agent is instructed to send an email to ``Tom.'' The agent must first prompt the user to log in securely, protecting credentials by avoiding automated input. It then encounters ambiguity when multiple contacts named ``Tom'' are found, and resolves it by prompting the user to select the intended recipient. Finally, before dispatching the email, the agent requests explicit confirmation, recognizing that email-sending is a non-reversible action with privacy implications \cite{zhang2024ufouifocusedagentwindows}. Although the task appears simple, it reflects the complexity of real-world human-GUI agent collaboration, involving privacy preservation, ambiguity resolution, and intentionality confirmation \cite{kim2024understanding}. These are not generic LLM issues, but domain-specific challenges rooted in shared interaction with software interfaces—underscoring the need for new design paradigms around shared control, interruption handling, and proactive clarification in GUI agent systems.

\textbf{Potential Solutions:} Emphasizing \emph{user-centered design} \cite{lu2024ai} principles can address user needs and concerns, providing options for customization and control over the agent's behavior \cite{feng2024large}. Equipping agents with the ability to engage in \emph{clarification dialogues} when user instructions are unclear can enhance task accuracy \cite{wester2024ai}. Natural language understanding components can detect ambiguity and prompt users for additional information. For instance, the agent could ask, ``There are two contacts named John. Do you mean John Smith or John Doe?'' Incorporating \emph{human-in-the-loop} systems allows for human intervention during task execution, enabling users to guide or correct the agent's decisions when necessary \cite{wang2024understanding}. Developing adaptive interaction models that facilitate seamless collaboration between humans and agents is essential. Additionally, providing transparency and explainability in the agent's reasoning processes can build user trust and improve cooperation \cite{cambria2024xaimeetsllmssurvey, wu2024usable, shi2025towards}. 

\begin{figure*}[t]
    \centering
    \includegraphics[width=0.95\textwidth]{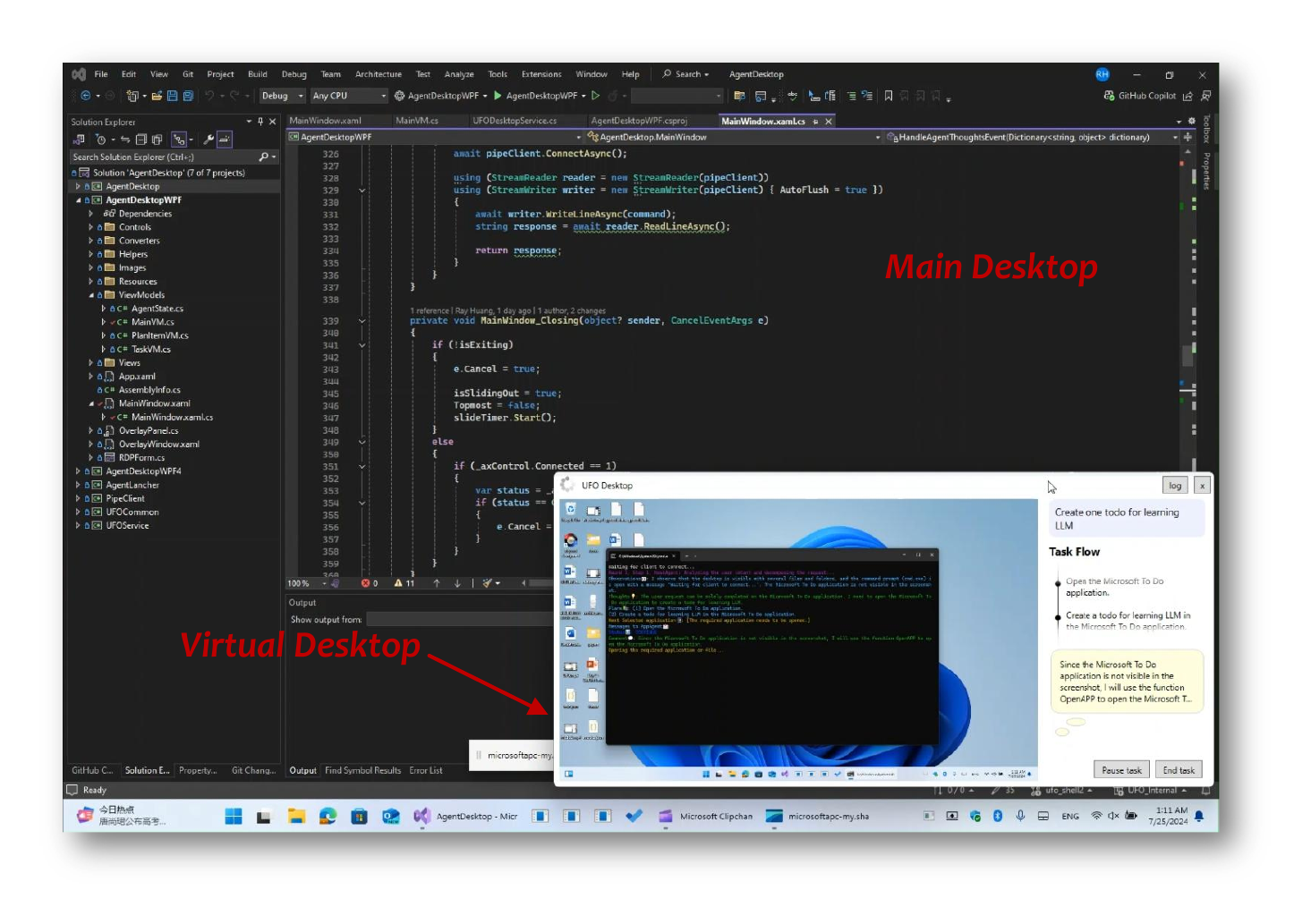}
    \vspace{-3.em}
    \caption{The Picture-in-Picture interface in UFO\textsuperscript{2}: a virtual desktop window enabling non-disruptive automation. Figure adapted from \cite{zhang2025ufo2}.}
    \label{fig:pip}
    % \vspace{-2em}
\end{figure*}

Lastly, developing a virtual desktop environment for the agent to operate in—one that connects to the user's main desktop session without disrupting their workflow, can significantly enhance the user experience (UX) in human-agent interaction. The picture-in-picture mode implemented in UFO\textsuperscript{2} \cite{zhang2025ufo2} demonstrates this concept in practice, as illustrated in Figure~\ref{fig:pip}. By allowing the agent to run within a resizable and movable virtualized desktop, users can easily minimize or reposition the agent window as needed. This flexibility improves both the usability and the overall UX of interacting with GUI-based agents.

\subsection{Customization and Personalization\label{sec:limitation:customization}}

Effective GUI agents must go beyond generic task completion and provide experiences that are personalized to individual users, adapting to their unique workflows, preferences, and behavioral patterns \cite{li2024personal, cai2024large}. Unlike general LLM applications that operate in isolated prompts or conversations, GUI agents work across software environments where user interaction styles can vary significantly. A one-size-fits-all agent may fail to align with how a particular user edits documents, navigates interfaces, or organizes tasks—resulting in friction, inefficiency, or user frustration \cite{li2024hello}.

For instance, a GUI agent assisting with document editing must learn the user's preferred tone, formatting conventions, and vocabulary. Without this contextual understanding, the agent may offer irrelevant suggestions or enforce formatting inconsistent with the user's intent. Personalization in GUI agents thus requires longitudinal learning, where the agent continually adapts based on prior interactions, fine-tunes its behavior to match user expectations, and preserves consistency across sessions \cite{li2023traineragent}.

However, this introduces new challenges. The high variability in user preferences—especially in free-form GUI environments—makes it difficult to define universal personalization strategies. Moreover, collecting and leveraging user-specific data must be done responsibly, raising critical concerns around privacy, data retention, and on-device learning. Striking a balance between effective customization and user trust is particularly important for GUI agents, which often operate over sensitive documents, personal applications, or system-level interfaces.

\textbf{Potential Solutions:} Future research should focus on developing mechanisms for \emph{user modeling} \cite{tan2023user}  and \emph{preference learning} \cite{gao2024aligning}, enabling agents to tailor their actions to individual users. Techniques such as reinforcement learning from user feedback \cite{kaufmann2023survey}, collaborative filtering \cite{kim2024large}, and context-aware computing \cite{talukdar2024improving} can help agents learn user preferences over time. Ensuring that personalization is achieved without compromising privacy is essential \cite{xiao2006personalized}, potentially through on-device learning and anonymized data processing. In a more futuristic, cyberpunk-inspired scenario, agents may inversely generate GUIs tailored to users' needs, enabling greater customization and personalization \cite{hojo2025generativegui}.

\subsection{Ethical and Regulatory Challenges\label{sec:limitation:ethical}}

LLM-powered GUI agents raise distinct ethical and regulatory concerns due to their ability to perform real-world actions across software interfaces. Unlike traditional LLMs, these agents can autonomously trigger operations, manipulate data, and interact with sensitive applications—amplifying risks around accountability, fairness, and user consent \cite{gan2024navigatingriskssurveysecurity, sarker2024llm, biswas2023guardrails, li2023survey222, zhang2025interaction}.

A key concern is bias inherited from training data, which can lead to unfair behavior in sensitive workflows. For example, a GUI agent assisting in hiring may unknowingly exhibit gender or racial bias \cite{ferrara2023should, yu2024large}. These risks are harder to audit at the GUI level due to limited traceability across multi-application actions. Regulatory compliance adds further complexity. GUI agents often operate across domains with strict data protection laws, but lack standardized mechanisms for logging actions or securing user consent. This makes it challenging to meet legal and ethical standards, especially when agents act in opaque or background contexts. Addressing these issues requires tailored solutions for GUI agents, including permission controls, runtime confirmations, and transparent activity logs—ensuring safe, fair, and compliant deployment across diverse environments.

\textbf{Potential Solutions:} Addressing these concerns requires establishing clear ethical guidelines and regulatory frameworks for the development and use of GUI agents \cite{pineiro2023ethical}. Future work should focus on creating mechanisms for auditing and monitoring agent behavior \cite{zhengagentmonitor2} to ensure compliance with ethical standards and legal requirements \cite{chan2024agentmonitor}. Incorporating bias detection and mitigation strategies in language models can help prevent discriminatory or unfair actions \cite{lin2024investigating}. Providing users with control over data usage and clear information about the agent's capabilities can enhance transparency and trust.

\subsection{Scalability and Generalization\label{sec:limitation:scalability}}

GUI agents often struggle to scale beyond specific applications or environments, limiting their generalization. Each software interface features unique layouts, styles, and interaction patterns—even common UI elements like pop-up windows can vary widely \cite{zhang2024attackingvisionlanguagecomputeragents}. These variations make it difficult to design agents that operate robustly across platforms without retraining or fine-tuning.

A further challenge is the dynamic nature of real-world GUIs. Frequent changes due to software updates, A/B testing, or interface redesigns—such as repositioned buttons or modified widget hierarchies—can easily break previously functional agents. For example, an agent trained on one version of a word processor may fail when the layout changes, or when deployed on a different program with similar functionality but a different interface structure. Even when GUIs share visual similarities, agents often fail to generalize without additional exploration or adaptation \cite{shekkizhar2025agicomingrightai}. This lack of robustness restricts deployment in practical settings and increases the cost of maintenance, requiring frequent updates or retraining to stay aligned with evolving environments \cite{grosse2023studying, zhang2024out, li2024websuite}. Overcoming this challenge remains critical for developing truly scalable and adaptable GUI agents.

\textbf{Potential Solutions:} To enhance scalability and generalization, one solution from the dataset perspective is to create comprehensive GUI agent datasets that cover a wide range of environments, user requests, GUI designs, platforms, and interaction patterns. By exposing the LLM to diverse data sources during training, the model can learn common patterns and develop a more generalized understanding, enabling it to adapt to infer the functionality of new interfaces based on learned similarities \cite{song2024agentbank}.

To further enhance adaptability, research can focus on techniques such as \emph{transfer learning} \cite{weiss2016survey} and \emph{meta-learning} \cite{chen2021meta}. \emph{Transfer learning} involves pre-training a model on a large, diverse dataset and then fine-tuning it on a smaller, task-specific dataset. In the context of GUI agents, this means training the LLM on a wide array of GUI interactions before customizing it for a particular application or domain. \emph{Meta-learning}, enables the model to rapidly adapt to new tasks with minimal data by identifying underlying structures and patterns across different tasks. These approaches enable agents to generalize from limited data and adapt to new environments with minimal retraining.

However, even with these measures, the agent may still encounter difficulties in unfamiliar environments. To address this, we advocate for developers to provide helpful knowledge bases, such as guidance documents, application documentation, searchable FAQs, and even human demonstrations on how to use the application \cite{zhu2024knowagent, guan2024explainable, hsieh2023tool}. Techniques like RAG \cite{gao2023retrieval} can be employed, where the agent retrieves relevant information from a knowledge base at runtime to inform its decisions \cite{kagaya2024rap}. For instance, if the agent encounters an unknown interface element, it can query the documentation to understand its purpose and how to interact with it. This approach enhances the agent's capabilities without requiring extensive retraining. Implementing these solutions requires collaborative efforts not only from agent developers but also from application or environment providers. 

\subsection{Summary\label{sec:limitation:summary}}
LLM-brained GUI agents hold significant promise for automating complex tasks and enhancing user productivity across various applications. However, realizing this potential requires addressing the outlined limitations through dedicated research and development efforts. By addressing these challenges, the community can develop more robust and widely adopted GUI agents.

Collaboration among researchers, industry practitioners, policymakers, and users is essential to navigate these challenges successfully. Establishing interdisciplinary teams can foster innovation and ensure that GUI agents are developed responsibly, with a clear understanding of technical, ethical, and societal implications. As the field progresses, continuous evaluation and adaptation will be crucial to align technological advancements with user needs and expectations, ultimately leading to more intelligent, safe, and user-friendly GUI agents.
\section{Conclusion\label{sec:conclusion}}
The combination of LLMs and GUI automation marks a transformative moment in human-computer interaction. LLMs provide the ``brain'' for natural language processing, comprehension, and GUI understanding, while GUI automation tools serve as the ``hands'', translating the agent's cognitive abilities into actionable commands within software environments. Together, they form LLM-powered GUI agents that introduce a new paradigm in user interaction, allowing users to control applications through straightforward natural language commands instead of complex, platform-specific UI operations. This synergy has shown remarkable potential, with applications flourishing in both research and industry.

In this survey, we provide a comprehensive, systematic, and timely overview of the field of LLM-powered GUI agents. Our work introduces the core components and advanced techniques that underpin these agents, while also examining critical elements such as data collection, model development, frameworks, evaluation methodologies, and real-world applications. Additionally, we explore the current limitations and challenges faced by these agents and outline a roadmap for future research directions. We hope this survey serves as a valuable handbook for those learning about LLM-powered GUI agents and as a reference point for researchers aiming to stay at the forefront of developments in this field.

As we look to the future, the concept of LLM-brained GUI agents promises to become increasingly tangible, fundamentally enhancing productivity and accessibility in daily life. With ongoing research and development, this technology stands poised to reshape how we interact with digital systems, transforming complex workflows into seamless, natural interactions.

% or


% \appendices

% \section*{Acknowledgment}


\ifCLASSOPTIONcaptionsoff
  \newpage
\fi
% \bibliographystyle{IEEETrans}
% \bibliography{reference}



\bibliographystyle{IEEEtran}
\bibliography{reference}
\end{document}